\pdfoutput=1

\documentclass[11pt]{article}

\usepackage[preprint]{acl}

\usepackage{times}
\usepackage{latexsym}

\usepackage[T1]{fontenc}

\usepackage[utf8]{inputenc}

\usepackage{microtype}

\usepackage{inconsolata}
\usepackage{tabularx}
\usepackage{graphicx}
\usepackage{subcaption}
\usepackage{amsmath}
\usepackage{authblk}
\usepackage{amssymb}
\usepackage{booktabs}
\usepackage{xspace}
\usepackage{xcolor}
\definecolor{lightthistle}{rgb}{129,184,223}  
\definecolor{cultureEastAsia}{RGB}{254,129,125}
\definecolor{cultureSouthAsia}{RGB}{0,76,153} 
\definecolor{cultureMiddleEast}{RGB}{200,160,90}    
\definecolor{cultureEurope}{rgb}{0.65, 0.45, 0.65}  
\definecolor{cultureNorthAmerica}{RGB}{100,200,200}
\definecolor{cultureLatinAmerica}{RGB}{120,180,120}   
\definecolor{cultureAfrica}{rgb}{1, 0.8, 0.35}
\definecolor{cultureAustrlia}{rgb}{1.0, 0.55, 0.2}
\newcommand{\framework}{\textsc{MakiEval}\xspace}

\title{\includegraphics[width=0.6cm]{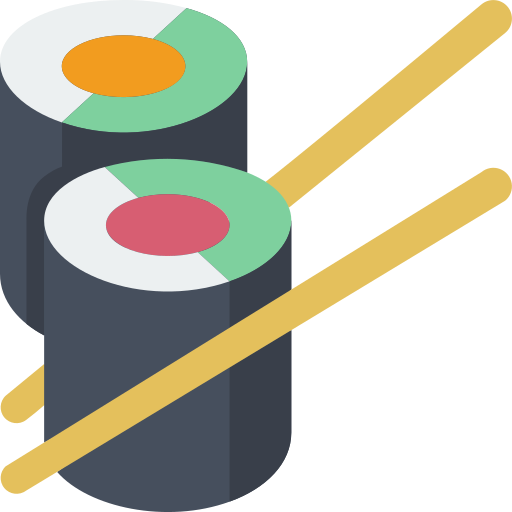}\framework: A \textsc{M}ultilingual \textsc{A}utomatic Wi\textsc{Ki}data-based Framework for Cultural Awareness Evaluation for LLMs}

\makeatletter
\renewcommand\AB@affilsepx{\hspace{1em} \protect\Affilfont}
\makeatother

\author{\textbf{Raoyuan Zhao, Beiduo Chen, Barbara Plank, Michael A. Hedderich}}
\affil{MaiNLP, Center for Information and Language Processing, LMU Munich \\
Munich Center for Machine Learning (MCML) \\
\texttt{\{rzhao,beiduochen,bplank,hedderich\}@cis.lmu.de}}

\begin{document}
\maketitle
\begin{abstract}
Large language models (LLMs) are used globally across many languages, but their English-centric pretraining raises concerns about cross-lingual disparities for cultural awareness, often resulting in biased outputs. However, comprehensive multilingual evaluation remains challenging due to limited benchmarks and questionable translation quality. To better assess these disparities, we introduce \framework, an automatic multilingual framework for evaluating cultural awareness in LLMs across languages, regions, and topics. \framework evaluates open-ended text generation, capturing how models express culturally grounded knowledge in natural language. Leveraging Wikidata’s multilingual structure as a cross-lingual anchor, it automatically identifies cultural entities in model outputs and links them to structured knowledge, enabling scalable, language-agnostic evaluation without manual annotation or translation.  We then introduce four metrics that capture complementary dimensions of cultural awareness: granularity, diversity, cultural specificity, and consensus across languages. We assess 7 LLMs developed from different parts of the world, encompassing both open-source and proprietary systems, across 13 languages, 19 countries and regions, and 6 culturally salient topics (e.g., food, clothing). Notably, we find that models tend to exhibit stronger cultural awareness in English, suggesting that English prompts more effectively activate culturally grounded knowledge.

\end{abstract}

\begin{figure}[t]
    \centering
    \includegraphics[width=0.95\linewidth]{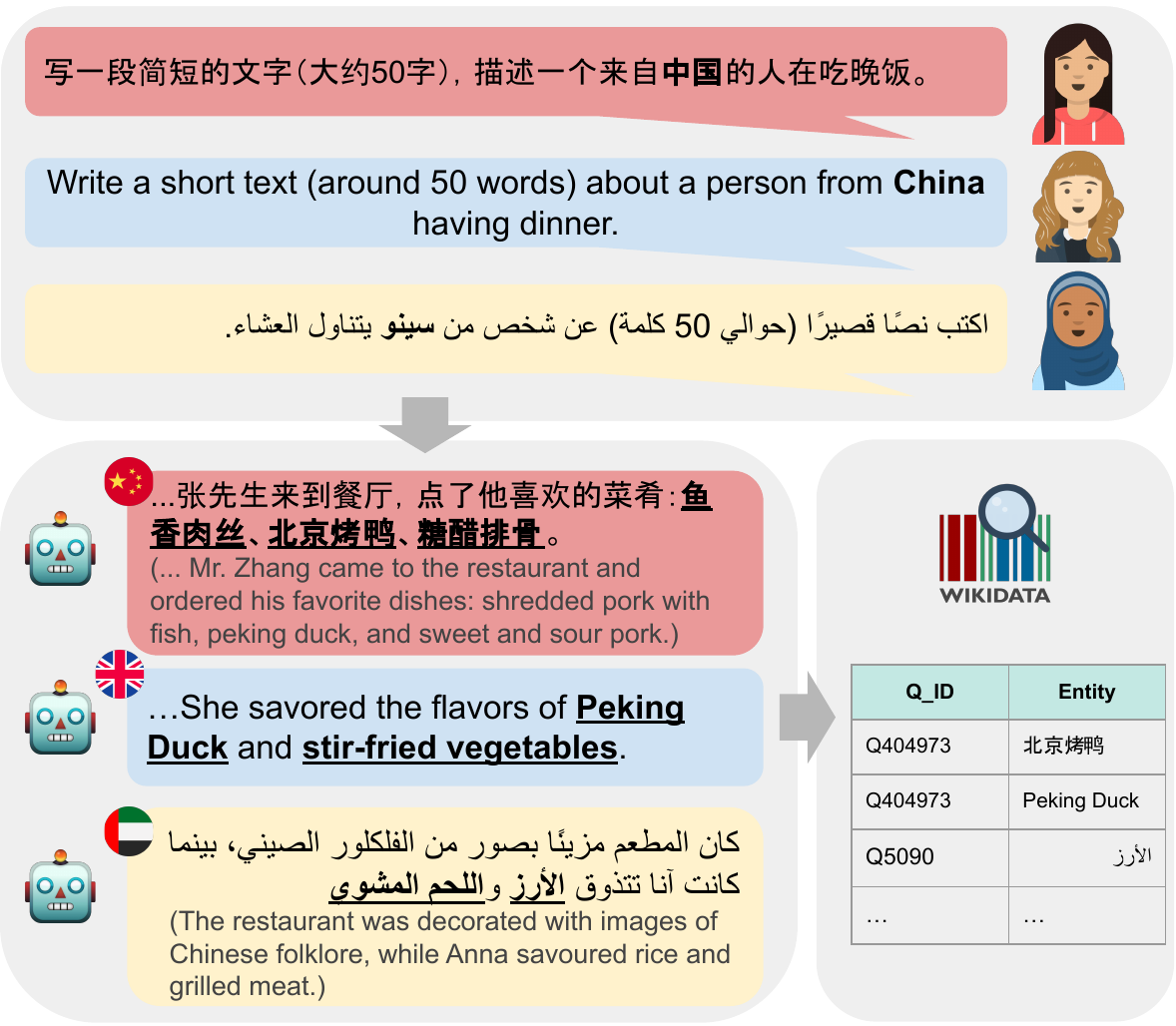}
    \caption{
        Multilingual prompts and corresponding model outputs for a culturally grounded generation task.
        Our culture-aware evaluation focuses on searching and assessing entities within these outputs using Wikidata.
    }
    \vspace{-8pt}
    \label{fig:culture_prompt_intro}
\end{figure}

\section{Introduction}

The global deployment of LLMs in applications ranging from conversational agents to content creation makes their cultural awareness increasingly critical~\cite{pawar2024surveyculturalawarenesslanguage}. Without such awareness, models risk producing biased or culturally insensitive outputs, which can perpetuate stereotypes and harm user trust~\cite{bender2021dangers, hutchinson-etal-2020-social,blodgett-etal-2020-language}. 
Yet despite its importance, we ask fundamentally: %
\textit{How can we evaluate a text generation model's cultural awareness in a flexible and generalizable manner?}

Figure~\ref{fig:culture_prompt_intro} presents a realistic scenario: when prompted in different languages to describe a Chinese person having dinner, the same LLM produces outputs that differ in food entity specificity (i.e., whether the food items are culturally specific to China),  diversity, and cross-lingual consistency. 
Capturing these differences across languages in evaluation, especially under the flexible nature of text generation models, remains a major challenge. %
Particularly, we identify three key limitations:

$\mathrm{i}$) \textit{test format}: dominant evaluation methods often rely on simplified setups such as cloze tests, single-token prediction, or predefined cultural entity lists~\cite{naous-etal-2024-beer,wang-etal-2024-countries}. While computationally efficient, these formats fail to capture how cultural knowledge naturally manifests in language generation~\cite{jones1995evaluating,bhatt-diaz-2024-extrinsic}.

$\mathrm{ii}$) \textit{language and cultural coverage}: most existing studies focus on English, adopt monolingual setups~\cite{onohara-etal-2025-jmmmu,seveso-etal-2025-italic,naous-etal-2024-beer,nikandrou-etal-2025-crope}, or depend on translation-based pipelines for multilingual evaluation~\cite{berger-ponti-2025-cross}. Although translation offers convenience, it introduces errors or meaning loss—particularly with culturally specific terms~\cite{yao-etal-2024-benchmarking}. 

$\mathrm{iii}$) \textit{evaluation benchmark}: many existing benchmarks rely on manually curated datasets with fixed cultural entities~\cite{schneider-sitaram-2024-m5,nayak-etal-2024-benchmarking}. This static design lacks scalability and fails to adapt to emerging or less commonly documented cultural expressions, leading to incomplete assessments.

To address these limitations, we introduce \framework, an automatic approach for  $\mathrm{i}$) open-ended text evaluation,  $\mathrm{ii}$) leveraging Wikidata as a language bridge and  $\mathrm{iii}$) enabling structured evaluation metrics and scalable, up-to-date evaluation.
Instead of static benchmarking, \framework is inspired from a human perspective and prior work: %
when humans describe daily life in their own region, they tend to be more detailed (high granularity), mention a wider range of items (high diversity), refer to culturally specific concepts (high culture specificity), and maintain consistent representations across languages (high cultural consensus)~\cite{chafe1994discourse,dewaele2003productivity,peters2018vocabulary}. For example, when asked about Chinese breakfast, an English speaker unfamiliar with the culture might respond simply “A Chinese eats noodles in the morning,” whereas a Chinese speaker would provide a richer description such as “youtiao (fried dough sticks) with soy milk, jianbing, or millet porridge in winter.” This contrast highlights all four dimensions. These dimensions have long been discussed in psycholinguistics and linguistics, and they also resonate with recent NLP research on semantic abstraction, coverage, cultural awareness, and cross-lingual consistency. In what follows, we draw on this line of work to design four operational metrics that capture cultural awareness in multilingual generation.

Building on these metrics, we conduct comprehensive experiments to investigate how model origin, prompt language, and the mentioned country affect LLMs’ cultural awareness. We found models align more with culturally related languages and their regions of origin. We publicly release
our code and data.\footnote{\href{https://github.com/mainlp/MAKIEval}{https://github.com/mainlp/MAKIEval}}
Our contributions are:

\noindent $\bullet$ We propose \framework, an automatic framework for evaluating cultural awareness in multilingual text generation,
accompanied by four distinct cultural-aware metrics: granularity, diversity, culture specificity, and culture consensus.
    
\noindent $\bullet$  We enable direct cross-linguistic comparison without relying on translation and conduct a large-scale evaluation in 13 languages across 19 countries and 7 models to reveal patterns of cultural awareness.
    
\noindent $\bullet$  We observe differences in cultural awareness may arise from a mix of factors, including model-, country- and language-specific characteristics.

\section{Related Work}
\paragraph{Dimensions of Cultural Evaluation} While ``culture'' has become a prominent focus in natural language processing, existing work evaluates cultural awareness from widely varying dimensions. \citet{liu2024culturally} address this fragmentation by proposing a taxonomy of cultural elements, organized into ideational, linguistic, and social categories, spanning both concrete and abstract phenomena. 

Some investigations examine abstract or high-level aspects of culture, such as value systems~\cite{ma-etal-2024-potential,agarwal-etal-2024-ethical,shi-etal-2024-culturebank}, political ideology~\cite{bang-etal-2024-measuring}, gender norms~\cite{wan-etal-2023-kelly}, or expressions of stereotypes or hate~\cite{cheng-etal-2023-marked,deshpande-etal-2023-toxicity}. 
Others focus on concrete, content-level evaluations, assessing how LLMs engage with specific cultural knowledge or practices~\cite{naous-etal-2024-beer,wang-etal-2024-countries,jiang-joshi-2024-cpopqa,zhou-etal-2025-mapo,hu-etal-2024-bridging,ghaboura-etal-2025-time}. In this work, we also focus on the ideational branch, particularly on concrete cultural artifacts such as food, music, and clothing, which allow for a more structured, multilingual evaluation.

\paragraph{Benchmarks for Cultural Awareness And Bias} To systematically evaluate cultural bias and awareness in LLMs, researchers have proposed a range of benchmarks by pre-collecting data and constructing dedicated benchmarks~\cite{yuksel-etal-2024-turkishmmlu,jin-etal-2024-kobbq,chiu2024culturalbenchrobustdiversechallenging,singh-etal-2025-global}. Here, we highlight several recent works. JMMMU~\cite{onohara-etal-2025-jmmmu} focuses on Japanese culture awareness. ITALIC~\cite{seveso-etal-2025-italic} targets cultural and linguistic understanding in Italian through a large-scale multiple-choice benchmark. CROPE~\cite{nikandrou-etal-2025-crope} constructs a culturally grounded multimodal benchmark by collecting concepts from Wikipedia and generating 1,060 binary-choice English questions that ask whether an image corresponds to a culturally associated entity. \citet{naous-etal-2024-beer} analyze cultural bias in Arabic prompts by examining how language models balance Arab and Western representations across three tasks: stereotype generation (via adjective sentiment), fairness in NER and sentiment classification, and cultural entity prediction in masked tweets. Predicted entities are compared with the CAMeL Cultural Entities dataset, where overuse of Western entities signals bias.

However, these benchmarks are mostly single-language, multiple-choice in format, and often cover only one specific country or region. Their reliance on pre-defined questions also limits generalization to open-ended settings.

\begin{figure*}[htb]
	\centering
	\includegraphics[width=1\textwidth]{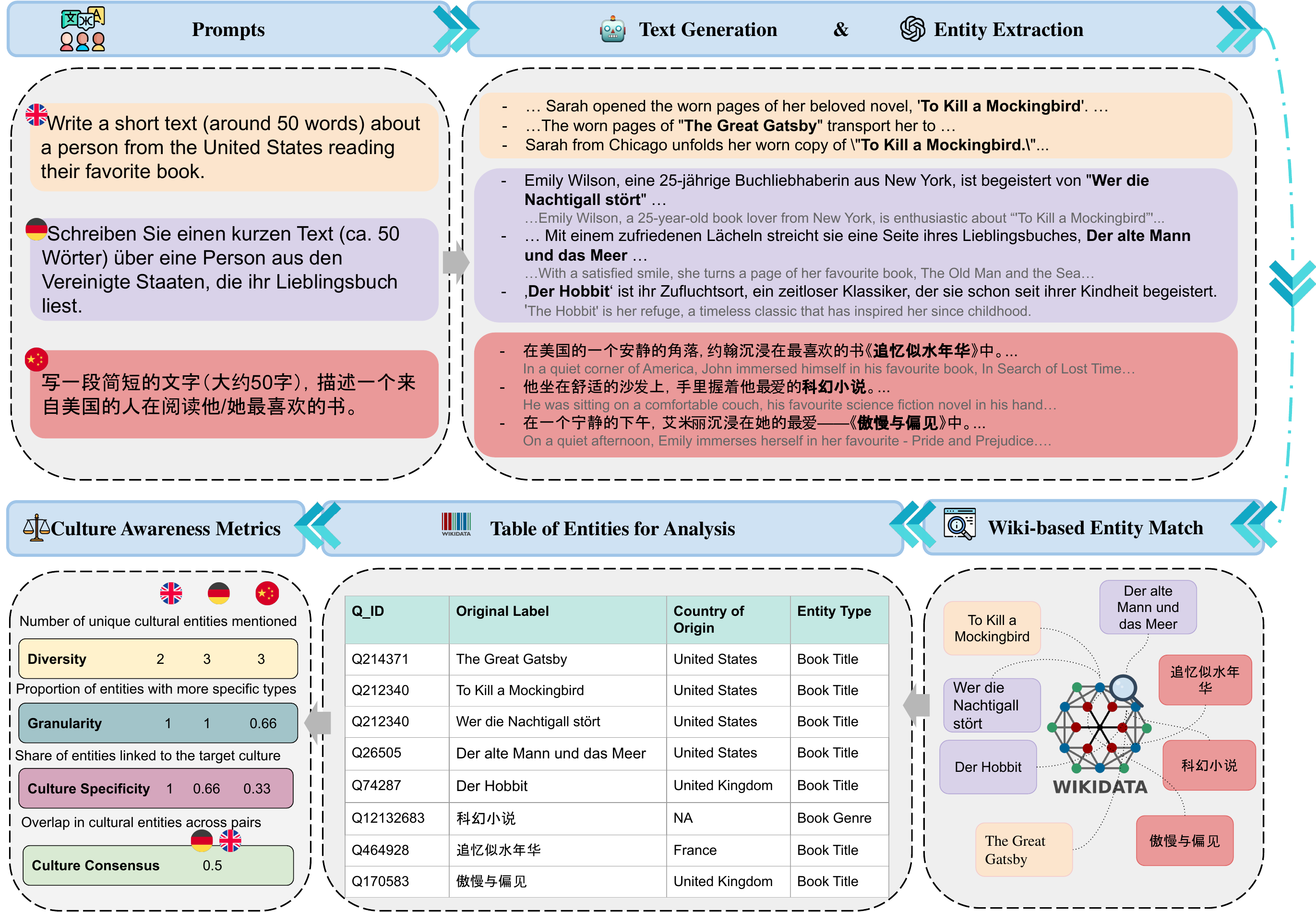}
	\caption{Overview of the \framework, a multilingual, generation-based framework for evaluating cultural awareness in LLMs. Given prompts in different languages to LLM, \framework collects open-ended generations, extracts named cultural entities and matched to Wikidata entries. Matched entities are augmented with metadata for downstream metric computation and analysis.}
    \vspace{-8pt}
	\label{fig:framework}
\end{figure*}
\paragraph{Cross-Lingual Bias Comparison}
Some recent works extend the evaluation to multilingual settings. \citet{myung2024blend} introduced a multilingual benchmark primarily consisting of short-answer and multiple-choice questions. They found that for resource-rich languages, LLMs tend to perform better in their native language. \citet{berger-ponti-2025-cross} compared textual descriptions of the same images produced by human speakers of different languages, using tools like Google Translate and WordNet to facilitate alignment. Their analysis revealed cross-cultural differences in what aspects people focus on when describing the same visual content. \citet{wang-etal-2024-countries} prompted LLMs in different languages to list cultural artifacts like holidays, then identify the country attributes with the help of Wikipedia. Rather than constraining output formats to entity listing, our framework utilizes natural, open-ended prompts to guide the model in generating coherent scene descriptions.
We presented the most relevant works to our setting here. For a more comprehensive overview, we refer to the survey by~\citet{10.1162/COLI.a.14}.

\section{\framework}
We propose \framework, an automatic and multilingual evaluation framework for evaluating cultural awareness.
In Figure \ref{fig:framework}, we illustrate our pipeline, which
consists of four steps: text generation, cultural entity extraction, wikidata-based entity match and metric-based analysis.

\subsection{Text Generation}
We prompt LLMs 
to generate culturally grounded texts in multiple languages. Unlike prior evaluations that rely on constraint output formats such as cloze tasks or entity listing, our setup encourages open-ended generation, allowing models to produce more realistic outputs that reflect typical usage scenarios and cultural entities embedded in natural language text.

To enable systematic comparison across languages, topics, and countries/regions, we introduce a set of template structures for the input prompts. Each template contains a placeholder for a country or region name, which we instantiate dynamically to create culturally contextualized prompts. Importantly, the templates only serve to initialize the generation context—models are otherwise unconstrained in their output.

For each \texttt{(language, topic, country)} combination, we generate 500 responses per model to capture generation variance. Section~\ref{subsec:prompts} details the prompt design, including template examples and strategies for ensuring cross-lingual alignment.

\subsection{Cultural Entity Extraction}

We extract topic-relevant cultural entities from each generated text using GPT-4o-mini~\cite{openai2025gpt4o}. For each topic (e.g., food, books, or music), GPT-4o-mini is prompted to identify mentions of relevant entities in the text (e.g., dish names for the food topic or author names for the book topic). In addition to extraction, the model classifies each entity by its semantic granularity. For example, for the food topic, it distinguishes whether the mention refers to a specific dish (e.g., Kung Pao Chicken), a dish type (e.g., stir-fry), an individual ingredient (e.g., onion), or an ingredient class (e.g., vegetable).

This process allows us to capture not only what cultural elements the model generates, but also how specific or abstract those references are—an essential signal for evaluating the depth and precision of cultural awareness.

\subsection{Wikidata-Based Entity Match}

To normalize entity mentions across languages and surface forms, we match each extracted entity to its corresponding entry in Wikidata~\cite{vrandevcic2014wikidata}. Our retrieval process matches not only entity labels but also aliases (\textsc{``Also Known As''}) in the prompt's language, falling back to English if no match is found. Each Wikidata entry is identified through its unique QID.

Since Wikidata contains ambiguous labels (e.g., ``lemon'' may refer to a fruit or a film, each with a different QID), we implement a disambiguation procedure that leverages hierarchical semantic properties. For each matched label, we collect all candidate QIDs and perform a recursive traversal over their \texttt{instance of}, \texttt{subclass of}, and \texttt{part of} relations. Each candidate may be linked to multiple such properties, resulting in a tree of semantic paths. We traverse all reachable paths upward in the ontology graph and retain the entity if any path leads to a culturally relevant category (e.g. \texttt{food}, \texttt{literary work}). If no valid semantic path is found, the disambiguation is marked as unresolved, and the corresponding entity is discarded from evaluation.

Once the correct entity is identified, we retrieve a set of metadata from Wikidata. This includes the entity’s description, country of origin, and, where applicable, the nationality of the author or performer (e.g., for books or music). 

All matched entities and their corresponding metadata are compiled into a structured table that supports both quantitative and qualitative analysis. Each entry includes the entity’s QID, surface labels in multiple languages, granularity tags, and country/region-level information. A manual evaluation of the performance indicated strong results, with full details presented in Appendix~\ref{sec:ner}.

\section{Evaluation Metrics}

We propose four metrics to evaluate different aspects of cultural awareness in multilingual language models: \textbf{Granularity}, \textbf{Diversity}, \textbf{Culture Specificity}, and \textbf{Culture Consensus}. All metrics operate over a set of cultural entities extracted from the evaluated model's responses.

We denote the union of all predicted entities across responses of a specific evaluated LLM and a specific prompt as \( E \).

Each entity \( e \in E \) is linked to a Wikidata QID, from which we obtain relevant metadata (e.g., country of origin, entity type). The following sections define each metric in detail.

\subsection{Granularity}

This metric captures the model’s ability to reference culturally specific concepts. We define \textbf{granularity} as the level of cultural detail reflected in the predicted entities. Inspired by a prior work on semantic abstraction in human descriptions~\cite{berger-ponti-2025-cross}, we assign a granularity score to each entity based on a predefined cultural scale (e.g. music genre vs.\ song title, see Appendix~\ref{sec:ner_prompt}), with higher values indicating finer granularity. Let \( \text{Gran}(e) \) be the granularity level of entity \( e \). We compute:
\begin{equation}
    \text{Granularity} = \frac{1}{|E|} \sum_{e \in E} \text{Gran}(e),
\end{equation}

\subsection{Diversity}

Diversity measures the breadth of cultural coverage by counting the number of unique cultural entities generated and is inspired by and adapted from work by~\citet{berger-ponti-2025-cross}. As shown in Figure~\ref{fig:framework}, the prompt asks about books, and the metric calculates the number of distinct book-related entities mentioned. Although the English response contains three extracted entities, one is repeated, so the resulting diversity score is 2.
Higher diversity indicates that the model is capable of expressing a wider range of cultural knowledge for a given language and topic.

\subsection{Culture Specificity}

{Culture specificity evaluates whether the entities in the generated text are aligned with the cultural context specified in the prompt. Figure \ref{fig:framework} illustrates a case where the prompt inquires about the US. The culture specificity score is calculated by computing the proportion of entities associated with the United States among all extracted entities.
This metric reflects the model’s ability to generate culturally appropriate and contextually aligned content. Its design is motivated by earlier studies on cultural awareness and bias \cite{wang-etal-2024-countries, liu2024culturally}.

\subsection{Culture Consensus}

This metric is motivated by the observations of \citet{hedderich-etal-2025-whats}, who showed that LLMs prompted with the same query in different languages often yield inconsistent outputs, highlighting the need to evaluate whether models preserve shared cultural understanding across languages. \textbf{Culture consensus} quantifies the agreement between model outputs for the same topic and cultural context setting across different languages. Let \( Q_A \) and \( Q_B \) denote the sets of Wikidata QIDs extracted from responses to the same prompt in languages \( A \) and \( B \), respectively. We compute their overlap using Jaccard similarity:

\begin{equation}
    \text{Consensus}(A, B) = \frac{|Q_A \cap Q_B|}{|Q_A \cup Q_B|}.
\end{equation}

Higher consensus implies more stable cultural grounding across languages, suggesting that the model maintains consistent cultural knowledge despite surface linguistic variation.

\section{Experimental Setups}

\subsection{Models}

\begin{table*}[t]
\centering
{\small
\begin{tabular}{@{}p{0.15\textwidth} p{0.8\textwidth}@{}}
\toprule
\textbf{Type} & \textbf{Prompt} \\
\midrule
Neutral & Write a short text (around 50 words) about a person having dinner. \\
Explicit & Write a short text (around 50 words) about a person from \{country\} having dinner. \\
Implicit & Write a short text (around 50 words) about \{name\} having dinner. \\
\bottomrule
\end{tabular}}
\caption{English prompt templates for the \textit{food} category under three conditions: \textbf{Neutral}, which makes no cultural reference; \textbf{Explicit} and \textbf{Implicit} include a hint for the referred cultural context. Full prompt templates for all categories are provided in  Appendix~\ref{sec:prompt_generation}.}
\label{tab:food_prompts}
\end{table*}

We evaluate our framework on seven multilingual language models, chosen to represent a diverse range of capabilities, resource requirements, and geographic origins. Our selection includes three open-source models developed across different regions: \href{https://huggingface.co/meta-llama/Llama-3.1-8B-Instruct}{Llama-3.1-8B-Instruct} (US)~\cite{grattafiori2024llama}, \href{https://huggingface.co/mistralai/Mistral-7B-Instruct-v0.1}{Mistral-7B-Instruct-v0.1} (France)~\cite{jiang2023mistral}, and \href{https://huggingface.co/Qwen/Qwen2.5-7B-Instruct}{Qwen2.5-7B-Instruct} (China)~\cite{yang2024qwen2}. To complement these, we also include two proprietary large-scale models: ChatGPT-4o-mini (US) and DeepSeek-V3 (China)~\cite{liu2024deepseek}, which are widely used but not openly available or easily deployable on local infrastructure. In addition, we evaluate \href{https://huggingface.co/meta-llama/Llama-3.3-70B-Instruct}{Llama-3.3-70B-Instruct}~\cite{grattafiori2024llama} and \href{https://huggingface.co/CohereLabs/aya-expanse-8b}{aya-expanse-8b} (Canada)~\cite{ustun-etal-2024-aya}. The inclusion of both Llama3 variants enables a controlled comparison of model size, while Aya allows us to assess cross-lingual generalization in a model explicitly trained for multilinguality.

\subsection{Languages}
\begin{table}[t]
\centering
\resizebox{0.9\columnwidth}{!}{\small
\begin{tabular}{lp{4.3cm}}
\toprule
\textbf{Lang} & \textbf{Countries/Regions} \\
\midrule
ar     & \textcolor{cultureMiddleEast}{United Arab Emirates} \\
en     & \textcolor{cultureNorthAmerica}{United States}, \textcolor{cultureEurope}{United Kingdom}, \textcolor{cultureNorthAmerica}{Canada}, \textcolor{cultureAustrlia}{Australia}, \textcolor{cultureAfrica}{Nigeria} \\
de     & \textcolor{cultureEurope}{Germany} \\
es     & \textcolor{cultureLatinAmerica}{Mexico}, \textcolor{cultureEurope}{Spain}, \textcolor{cultureLatinAmerica}{Argentina} \\
fa     & \textcolor{cultureMiddleEast}{Iran} \\
hi     & \textcolor{cultureSouthAsia}{India} \\
it     & \textcolor{cultureEurope}{Italy} \\
ja     & \textcolor{cultureEastAsia}{Japan} \\
ko     & \textcolor{cultureEastAsia}{South Korea} \\
th     & \textcolor{cultureEastAsia}{Thailand} \\
tr     & \textcolor{cultureMiddleEast}{Turkey} \\
zh  & \textcolor{cultureEastAsia}{China} \\
zh-tw  & \textcolor{cultureEastAsia}{Taiwan} \\
\bottomrule
\end{tabular}
}
\caption{
Prompt languages and country/region mentioned countries/regions in our experiments. Country/region names are color-coded by geographical region: \textcolor{cultureEastAsia}{East Asia}, \textcolor{cultureSouthAsia}{South Asia}, \textcolor{cultureMiddleEast}{Middle East}, \textcolor{cultureEurope}{Europe}, \textcolor{cultureNorthAmerica}{North America}, \textcolor{cultureLatinAmerica}{Latin America}, \textcolor{cultureAustrlia}{Australia} and \textcolor{cultureAfrica}{Africa}.
}
\vspace{-8pt}
\label{tab:lang_country_region_colored}
\end{table}

We aim at a balanced and diverse coverage of major cultural regions and continents. Therefore, we select 13 languages as the primary focus of our study: Arabic, English, German, Hindi, Italian, Japanese, Korean, Persian, Simplified Chinese, Spanish, Thai, Traditional Chinese, and Turkish. The chosen languages span multiple language families (e.g., Indo-European, Sino-Tibetan), writing systems (e.g., Kanji, Devanagari), and geographic regions (for example, Middle East). Table~\ref{tab:lang_country_region_colored} provides an overview of the selected languages and their corresponding regions\!\footnote{In addition, we initially explored two low-resource African languages: Hausa and Yoruba. However, during experimentation, we found that open-source models produced text of very poor quality in these two languages, often entirely unrelated to the prompts (see Appendix \ref{sec:prompt}). As a result, we set aside experiments involving native African languages as future work. }

\subsection{Prompts}
\label{subsec:prompts}

We design three categories of prompts to elicit culturally grounded responses from language models: \textbf{explicit contextualized prompts}, where the cultural background is explicitly mentioned (e.g. ``about a person from Spain''), \textbf{implicit contextualized prompts}, where a culturally associated name is mentioned (e.g. ``about Juan'', where ``Juan'' is a popular name in Spain) and the \textbf{neutral prompts} with no mention of nationality, ethnicity, or location, allowing us to assess cultural preferences expressed by the model without direct cues. Table~\ref{tab:food_prompts} provides examples of contextualized and neutral prompt templates.

All prompts are constructed across six topics: \textit{food}, \textit{beverages}, \textit{clothing}, \textit{books}, \textit{music}, and \textit{transportation}. These categories were chosen because they reflect culturally meaningful yet concrete aspects of daily life that are comparable across languages.

To ensure linguistic accuracy, all prompt templates are reviewed by native speakers to validate both meaning and fluency. For Spanish, the prompts were checked by native speakers from both Latin America and Europe to account for regional variation. For morphologically-rich languages like Turkish, we design templates that keep the placeholder position grammatically neutral to allow automatic template filling.

During pilot experiments, we observed notable inconsistencies in language alignment for certain models. Specifically, Mistral frequently responded in English even when prompted in other languages, while Qwen exhibited a strong tendency to reply in Chinese, reaching up to 80\% of responses. Such behavior was not observed in other models under the same conditions. To mitigate this, we appended an additional instruction to the prompts used with these models to explicitly enforce language consistency. Details of prompts are provided in Appendix~\ref{sec:prompt}.
\begin{figure*}[t]
    \centering
    \includegraphics[width=1\linewidth]{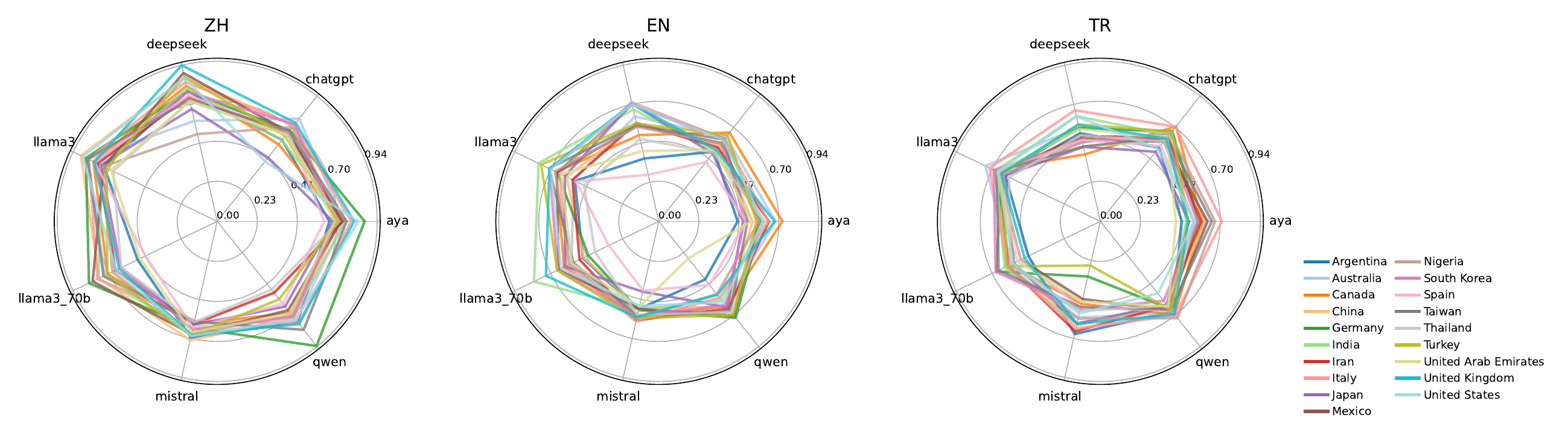}
    \caption{
        Average granularity ratio for Chinese, English, and Turkish prompts across all models. Each point represents the average granularity score aggregated over six cultural topics. 
        Full results are provided in Appendix~\ref{app:granularity}.
    }
    \label{fig:granularity_pattern}
\end{figure*}

\section{Results and Analysis}
Our evaluation is based on a total of 1,716 unique prompts, covering 13 languages, 6 cultural topics, 19 countries, and both contextualized and neutral settings. Each prompt is used to generate 500 responses per model, resulting in over 85.8 million generated texts. \framework identifies a total of 24,042 matched cultural entities (details in Appendix \ref{sec:entity}), which form the basis for computing our four proposed metrics.

In the remainder of this work, we explore the factors that influence cultural granularity, diversity, specificity and consensus, highlighting the complex interplay between model, language as well as cultural context.

\begin{figure}[t]
    \centering
    \includegraphics[width=1\linewidth]{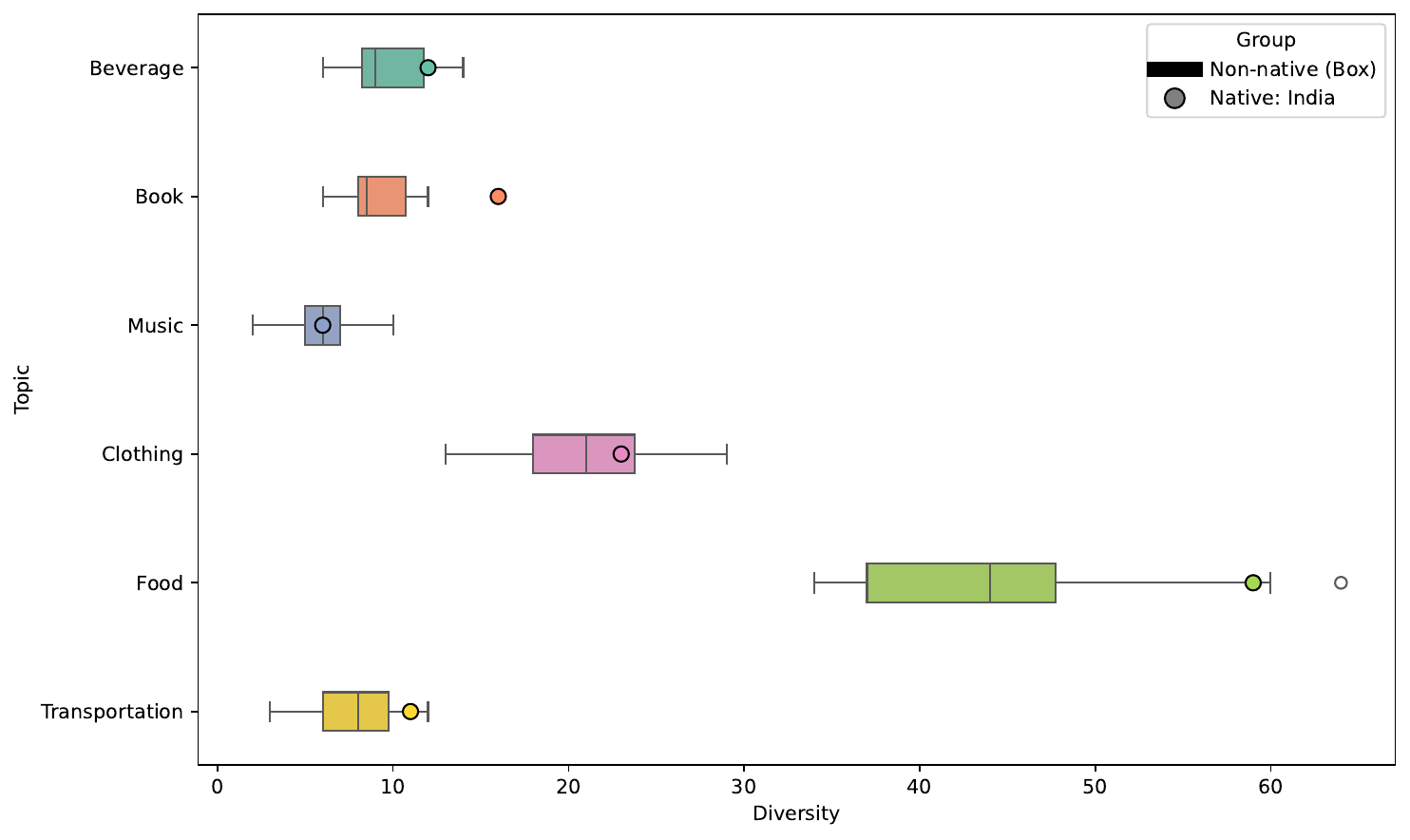}
    \caption{Diversity metric comparing prompts with culture context India in the native language Hindi vs. other languages for Llama3. Each row shows a separate cultural topic.}
    \label{fig:diversity-native}
\end{figure}

\begin{figure*}[t]
    \centering
    \includegraphics[width=0.49\linewidth]{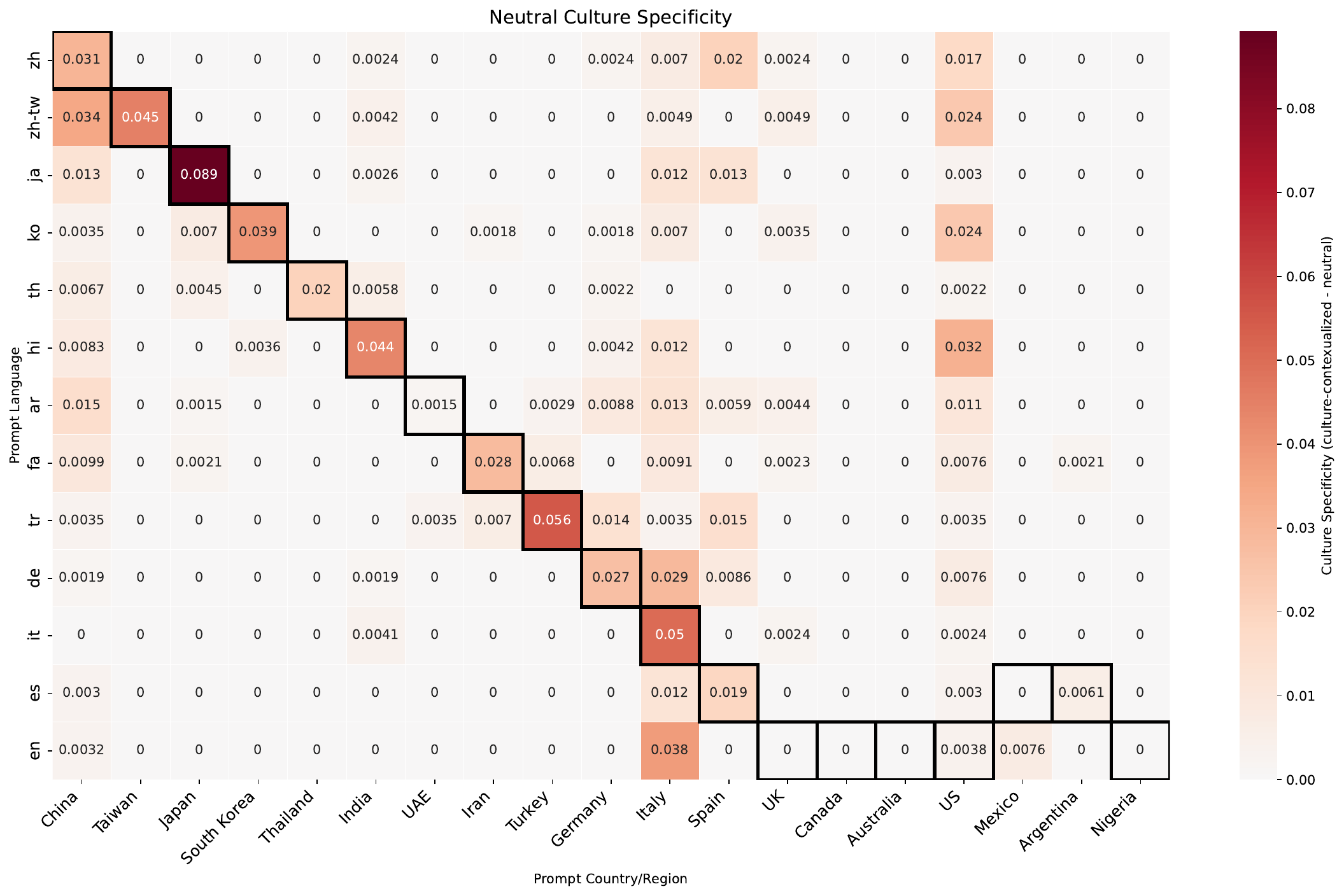}
    \includegraphics[width=0.49\linewidth]{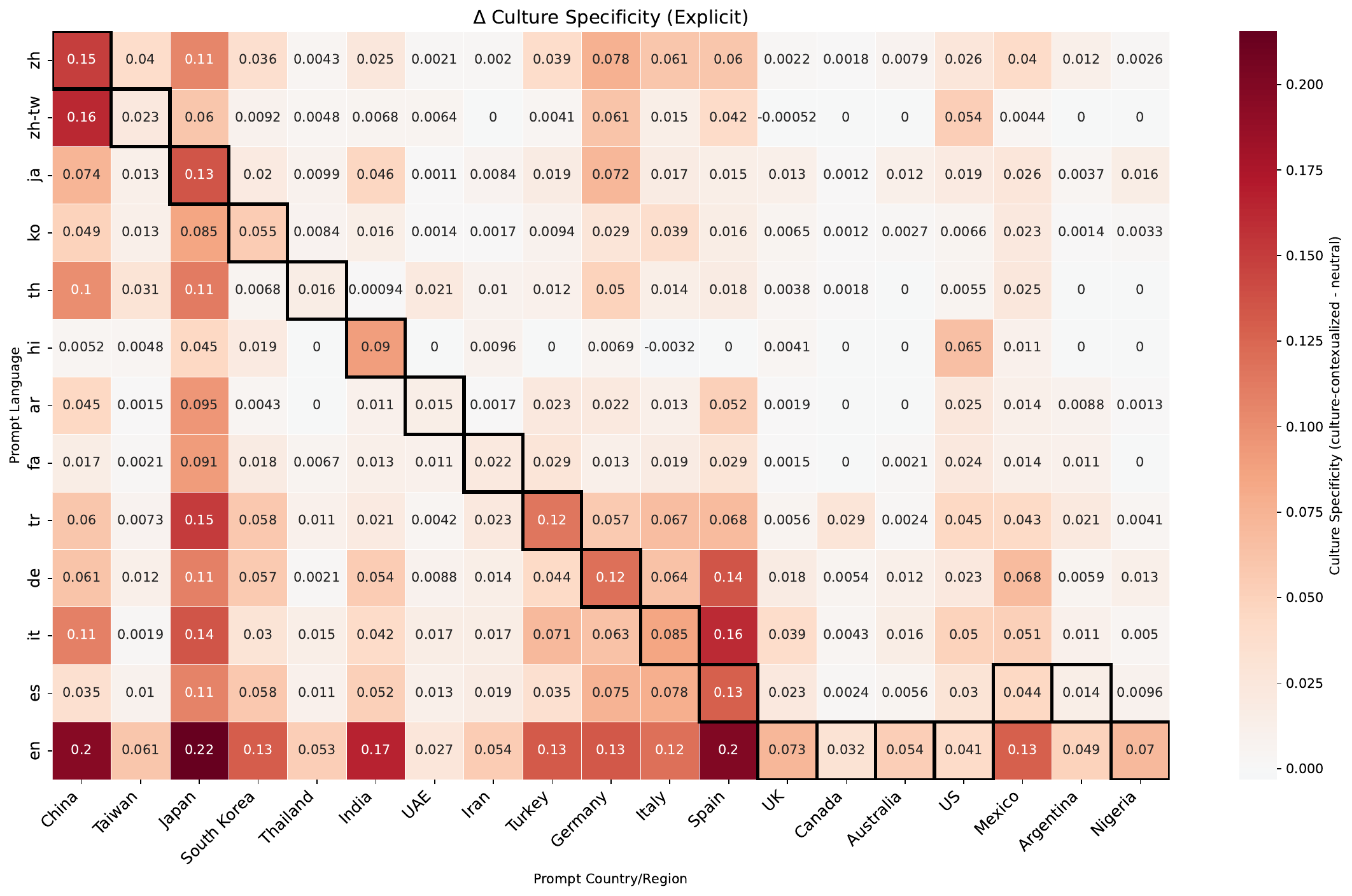}
    \caption{Culture specificity across country-language pairs for Llama3 without cultural context (neutral prompt, left) and with explicit context (right). Black boxes mark matching country and native language.}
    \label{fig:culture-specificity}
\end{figure*}

\subsection{Granularity}
 Our analysis reveals that, for each model, the average granularity is mostly the same given the same prompt language, independent of the referenced country/cultural context. We show the results for Chinese, English and Turkish as typical representatives in Figure~\ref{fig:granularity_pattern}, with all results given in Appendix~\ref{app:granularity}. Individual outliers exist (e.g., Mistral prompted in Turkish with Germany as context), but no context-specific pattern is visible. This suggests that each model maintains a language-specific baseline for cultural granularity that remains stable across cultural contexts.

Differences exist when moving with the same model to a different prompt language. E.g., English and Turkish have lower granularity levels for most models compared to Chinese. The observed patterns may stem from inherent linguistic differences in lexical specificity when describing cultural concepts, where Chinese, e.g., might employ more fine-grained expressions. Or alternatively, these differences might reflect limitations in the models’ multilingual capabilities. 

\subsection{Diversity}

The diversity metric varies widely depending on the setting, ranging from a handful of distinct entities per topic to over a hundred. The differences depend on both the prompt language and the model, with Llama3 having the highest mean diversity of 36.65 and DeepSeek the lowest with 14.62. 
For DeepSeek, when prompted in English on the book topic for the United States, we observe a diversity score close to 1, with the model producing the entity \textit{To Kill a Mockingbird} in 499 out of 500 generations.

Online texts in the native language of a country, used to train models, likely contain a diverse and maybe even less stereotypical representation of a country's culture compared to texts in other languages. We, therefore, explore whether language models exhibit stronger cultural diversity when prompted in the native language of the country being described.  On average, the diversity is higher for a country when prompted in the country's native language (26.518 vs. 23.617). Figure~\ref{fig:diversity-native} shows India and Hindi as examples. There is, however, no guarantee that using the native language increases diversity, with Appendix~\ref{sec:box_diverse} also showing many counterexamples across models, languages and topics. For English and Spanish, languages spoken natively in different countries, the diversity also differs by country. This shows that cultural diversity representation goes beyond simply the prompt language.

\subsection{Cultural Specificity}

Taking Llama3 as an example, shown in Figure~\ref{fig:culture-specificity} left, we can see that the neutral prompt (without mention of a cultural context or country) has a high cultural specificity when matching countries with their native language (e.g., Chinese-China and Turkish-Turkey). This suggests that language inherently encodes cultural signals. The model's outputs are, however, not limited to this language-country connection with entities from other countries also appearing in the outputs, especially from China, Italy and the United States.

Prompting with the explicit cultural context increases the cultural specificity in most cases (Figure~\ref{fig:culture-specificity} right). Implicit prompts (that do not mention the country explicitly) in LLaMA3 produce weaker gains than explicit ones (Figure~\ref{fig:culturespec_llama3} in Appendix), indicating reliance on overt cultural cues. Crucially, English prompts consistently achieve the highest specificity gains when adding cultural context across nearly all countries and models (see Appendix~\ref{sec:specificity}). This suggests that LLMs are more sensitive to cultural signals in English, likely due to the disproportionate amount of English data in their pretraining corpora. This further underscores the importance of multilingual evaluation, as cultural awareness is not only model-dependent but also deeply shaped by language.

\begin{table}[t]
\centering
\resizebox{0.8\linewidth}{!}{
\begin{tabular}{lc}
\toprule
\textbf{Model} & \textbf{Culture Consensus} \\
\midrule
Aya            & 0.1839 $\pm$ 0.0030 \\
ChatGPT        & 0.2036 $\pm$ 0.0015 \\
DeepSeek       & 0.2013 $\pm$ 0.0015 \\
Llama3         & 0.1813 $\pm$ 0.0010 \\
Llama3-70B     & 0.2097 $\pm$ 0.0011 \\
Mistral        & 0.1106 $\pm$ 0.0041 \\
Qwen           & 0.1623 $\pm$ 0.0018 \\
\bottomrule
\end{tabular}
}
\caption{Culture consensus scores for different models averaged over all topics and languages (mean $\pm$ variance).}
\label{tab:culture-consensus-alpha}
\end{table}

\subsection{Cultural Consensus}

Cultural consensus measures for a model how similar the cultural entities are across different prompt languages. Note that this is computed for all cultural contexts, not only that of the native language. A high consensus might thus be an indicator that the model has a representation of the cultures that is independent of the specific prompt language. 

As seen in Table~\ref{tab:culture-consensus-alpha}, model size seems to affect consensus, with the group of three larger models having a similar consensus value of around $0.2$. Model size is, however, not the only factor, with the difference in consensus being high among the smaller models of similar size. The figures in Appendix~\ref{sec:all_consensus}, provide a more detailed picture with large differences of consensus across language pairs. Taking ChatGPT on the food topic as an example in Figure~\ref{fig:cultural-consensus_heatmap_gpt-food}, we see that there is a high consensus between pairs of European languages as well as between Japanese and Korean. We hypothesize that this intra-regional alignment could be caused by shared perspectives on their own countries and those abroad, which then manifests in large-scale web or multilingual corpora through food blogs, reviews, or regional discussions. This effect suggests that cultural awareness in language models may be influenced not only by language itself but also by the latent cultural connections among the communities represented by those languages.

\begin{figure}
    \centering
    \includegraphics[width=\linewidth]{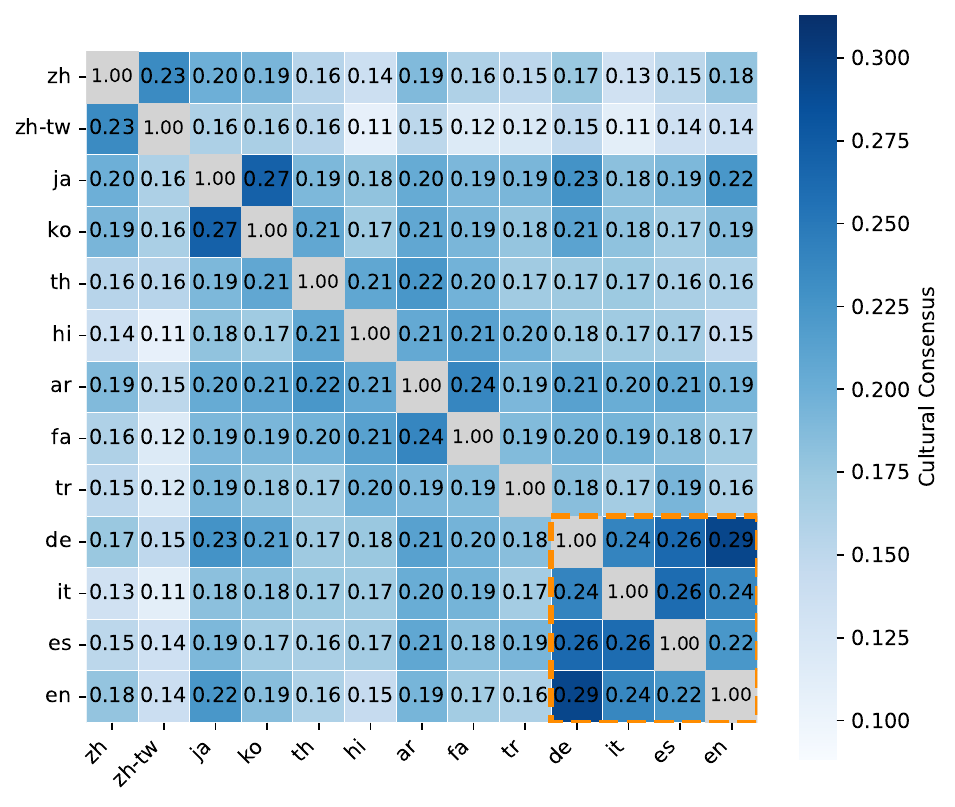}
    \caption{Cultural consensus between language pairs (over all countries) for the food topic for ChatGPT.}
    \label{fig:cultural-consensus_heatmap_gpt-food}
\end{figure}

\section{Discussion}
The design of the metrics in \framework is informed by insights from psycholinguistics and linguistics, as well as prior work in NLP, and our experiments demonstrate systematic differences across languages and models. However, how these differences should be interpreted and applied in practice remains an open question that merits further discussion. \framework is not intended as a prescriptive benchmark in which higher values are inherently preferable across all dimensions. Instead, it provides a set of complementary measures to capture different aspects of how LLMs represent and express cultural knowledge. 

The relevance and desirability of these measures vary with the characteristics of the downstream task and the interpretation of the proposed metrics requires careful consideration. For example, high specificity and granularity may be beneficial in settings where interlocutors share a cultural background, while lower values may be more effective in intercultural communication where common ground cannot be assumed. Similarly, cross-lingual consensus may indicate robustness in knowledge-intensive applications such as multilingual question answering and cross-lingual understanding~\cite{DBLP:conf/icpr/ChenGLT22,qi-etal-2023-cross,wang-etal-2025-lost-multilinguality}, whereas divergence can be advantageous in tasks that benefit from creativity or cultural variation, including storytelling or adaptation to local contexts. Given the diversity of generation tasks, no single value or direction of change should be considered universally optimal. Instead, the task-specific needs should be considered along the the proposed cultural dimensions.

\section{Conclusion}

We present \framework, an automatic, scalable and multilingual framework for evaluating cultural awareness in LLMs. By combining prompt-controlled generation, structured entity extraction, Wikidata grounding, and four culture metrics, our framework enables analysis of how models encode and express cultural knowledge across different languages, countries/regions, and topics.

Our results highlight that cultural awareness in LLMs is more complex than simple descriptions like a general bias to Western culture. Rather, it manifests in different forms, like the granularity, specificity and diversity of cultural entities. Besides general model capabilities, we show that the prompting language plays a crucial role in evoking these cultural differences, even when no explicit cultural context is given. The sensitivity to cultural signals in English, the connection between countries and their local languages, and the cultural consensus that links regional areas emphasize the need for multilingual and multicontextual evaluation of cultural awareness.

To foster further analysis of the cultural awareness of LLMs, we will release the full dataset of generated outputs and cultural entities for all model, language, country and topic combinations. Our work also provides guidance for future research in identifying how these factors have been picked up by the models during training and alignment, and how culture is represented internally in the models. More broadly, our framework offers a generalizable paradigm for multilingual evaluation, our findings reveal systematic disparities that call for rethinking prompting and training beyond English, and our metrics provide actionable signals for improving fairness and inclusivity in global LLM applications. We hope this work contributes to more equitable evaluation and deployment of language technologies worldwide.

\section*{Limitations}

While \framework enables scalable multilingual evaluation of cultural awareness, it has several limitations. 

First, our experiments are limited to a subset of high- and mid-resource languages. Although Wikidata provides labels in many low-resource languages, we excluded low-resource languages due to the low generation quality from current language models in these settings.

Second, our method relies on metadata retrieved from Wikidata to match and contextualize cultural entities. As an open, community-curated knowledge base, Wikidata may contain incomplete or inaccurate information. In practice, we observe that such cases are rare and have limited impact on our overall evaluation, though they may introduce noise in entity-level metrics such as specificity or consensus.

Finally, \framework primarily targets concrete cultural expression—entities such as food, books, and music that are grounded and observable. It is not designed to evaluate abstract cultural constructs, such as moral reasoning, political ideology, or values, which remain important directions for future work.

\paragraph{Ethical Concern} We do not foresee any ethical concerns associated with this work. All analyses were conducted using publicly available models. No private or sensitive information was used. The texts we studied were model-generated and do not necessarily represent our views. Additionally, we will release our code, produced data and documentations to support transparency and reproducibility.

\section*{Acknowledgments}
We thank the members of MaiNLP for their valuable feedback on this project, especially Verena Blaschke, Yupei Du, Florian Echin, Rob van der Goot, Bolei Ma, Elena Senger, and Soh-Eun Shim. We are further grateful to a number of these colleagues, as well as to Fatima Abdulla Mohammed Obaid Almatrooshi, Tianshi Feng, Paravee Jungbauer, Neysa Kaustubh, Abdullatif Köksal, Yuchen Li, Ali Modarressi, and Jing'an Qian for their assistance in verifying the language-specific texts studied in this paper.

This research was supported by the ERC Consolidator Grant DIALECT 101043235. We gratefully acknowledge that experiments involving API calls to GPT-4o-mini were supported by a compute grant from OpenAI.

\appendix
\section{Computational Resources and AI Assistance}

We checked the licenses of all the models and data used, which are publicly available resources.

\paragraph{Computational Resources}
We generate the texts with temperature $= 0.7$, top-$p = 0.9$, and top-$k = 10$ for all LLMs that run locally. All models were executed on NVIDIA A100 GPUs, except for the LLaMA3-70B model, which was run on an H200 GPU. The DeepSeek and ChatGPT models were accessed via their respective public APIs.
\paragraph{AI Assistance}
The authors acknowledge the use of ChatGPT solely for correcting grammatical errors, enhancing the coherence of the final manuscript, and providing assistance with coding.
\section{Prompt Design}
\label{sec:prompt}
\subsection{Prompt Template for Text Generation}
\label{sec:prompt_generation}
We use structured templates to generate prompts across six topics and 13 languages. To ensure linguistic accuracy and cultural appropriateness, we invited native speakers to review and validate the prompts. A total of 15 native speakers participated in this validation process, with at least one annotator per language. Their language backgrounds include German, Spanish (Spain), Spanish (Mexico), English (American), English (British), Turkish, Persian, Arabic, Hindi, Japanese, Korean, Traditional Chinese (Taiwan), Thai, Chinese(Mainland) and Italian.

To avoid redundancy, we present prompt templates for the \texttt{food} topic in Table~\ref{tab:food-prompts-nonbias-bias} here only. Templates for other topics share the same structural format and are available in the released code.

During our experiments, we observed a notable mismatch between the prompt language and the generated response language in the Mistral and Qwen models. To ensure accurate evaluation, we appended a language-specific instruction as listed in Table~\ref{tab:language-prompts} to the end of the original prompt when querying these models. This addition effectively resolved the misalignment issue, enabling the models to generate responses in the intended language.

\begin{table}[h]
\centering
\includegraphics[width=\linewidth]{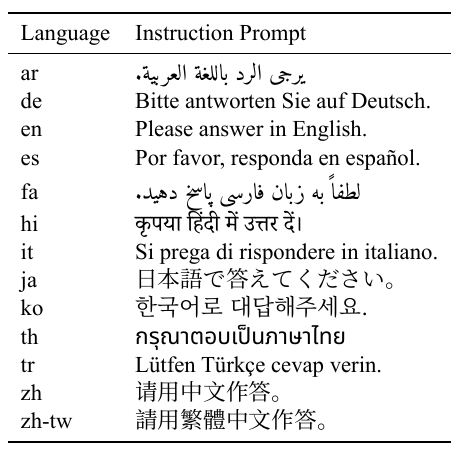}
\caption{Language-specific answer instructions.}
\label{tab:language-prompts}
\end{table}

\begin{figure*}[htbp]
\centering
\includegraphics[width=\textwidth]{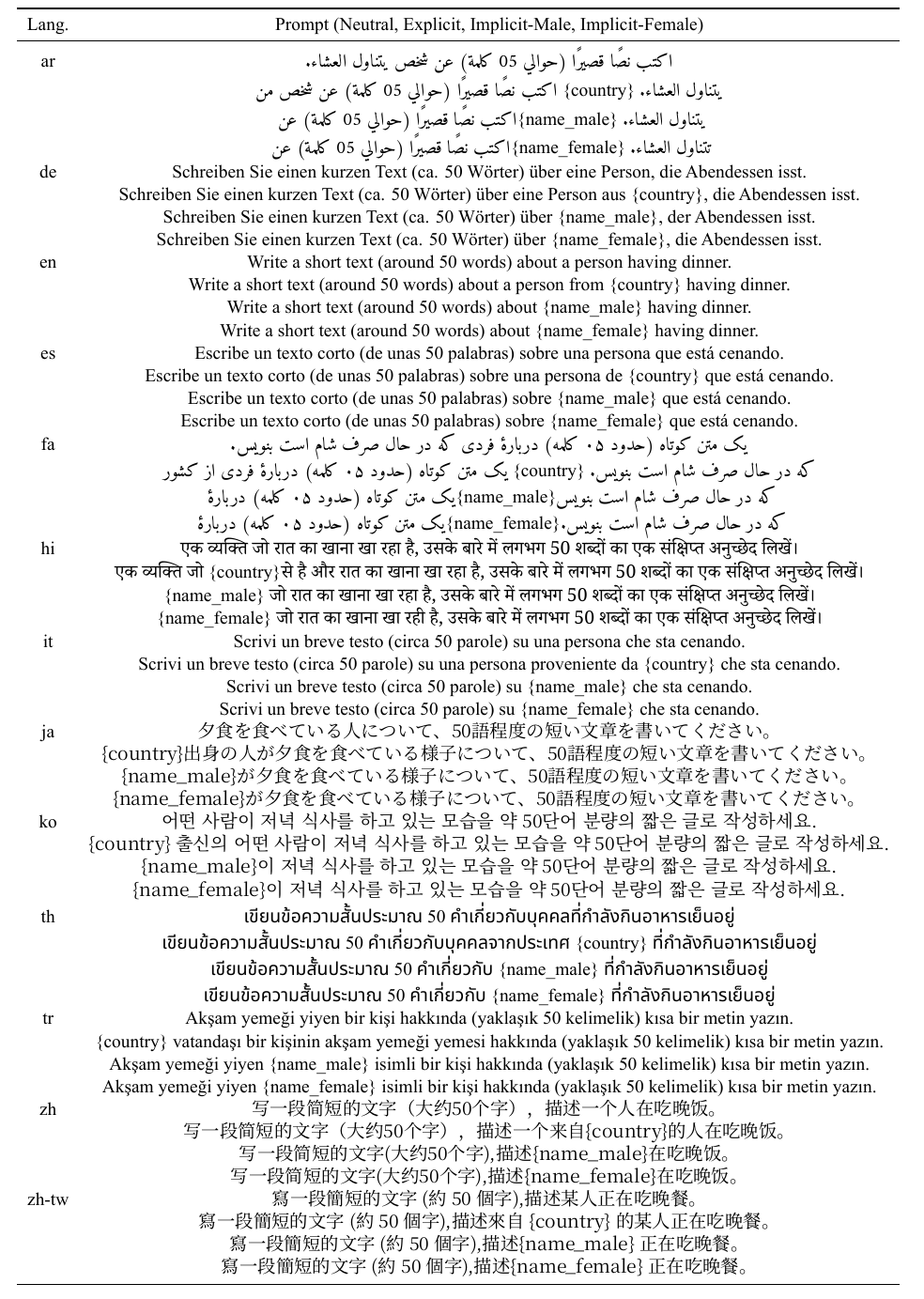}
\caption{Neutral, explicit, and implicit prompts (food topic) across 13 languages.}
\label{tab:food-prompts-nonbias-bias}
\end{figure*}

\subsection{Prompt for Entity Extraction}
\label{sec:ner_prompt}
\noindent
All prompts for entity extraction follow this format:

\begin{quote}
Help me extract the words or phrases in the given text which are \{topic-specific types\}.\\
Answer in a dictionary format with the type of the extracted text labeled.\\
Do not extract redundant entities or provide explanations.\\
Only output the dictionary in this format: \\
\texttt{\{"<extracted\_text>": "type", ...\}}\\
\textbf{Text:} \{response\}
\end{quote}
And based on the topic, as shown in Table~\ref{tab:prompt-extraction}, the system dynamically switches to different extraction branches. In addition to extracting topic-specific entities, we also include named entities such as person names and geographic locations to support more comprehensive downstream analyses.

Furthermore, we require that each extracted entity be assigned a granularity label. While we initially considered leveraging Wikidata to define these granular levels, the hierarchy was often inconsistent and difficult to generalize across languages and cultures. As a result, we employed ChatGPT-4-o-mini to infer the granularity of each entity. For example, food-related entities were categorized into labels such as dish, dish category, specific ingredient, and ingredient category. To quantify overall granularity, we assign a score of 1 to entities with specific references (e.g., spaghetti, coriander) and a score of 0 to more general categories (e.g., pasta dishes, herbs). The final granularity score for a given set of entities is computed as the average of these binary scores. More details can be found in the prompt design in Table~\ref{tab:prompt-extraction}.

\begin{table*}[h]
\centering
\begin{tabularx}{\textwidth}{lX}
\toprule
\textbf{Topic} & \textbf{Prompt} \\
\midrule

\textbf{book} & Extract book titles, book genres, author names, places, or reader names. For example: \texttt{\{"<book\_title>": "book\_title", "<author\_name>": "author\_name"\}}. \\

\textbf{food} & Extract dish names, dish categories, ingredient categories, specific ingredients, places, or person names. For example: \texttt{\{"<dish\_name>": "dish\_name", "<specific\_ingredient>": "specific\_ingredient"\}}. \\

\textbf{music} & Extract song names, music genres, artist names, places, or listener names. For example: \texttt{\{"<song\_name>": "song\_name", "<artist\_name>": "artist\_name"\}}. \\

\textbf{clothing} & Extract clothing types, fashion styles, places, or person names. For example: \texttt{\{"<clothing\_type>": "clothing\_type", "<fashion\_style>": "fashion\_style"\}}. \\

\textbf{transportation} & Extract modes of transport, vehicle types, places, or person names. For example: \texttt{\{"<mode\_of\_transport>": "mode\_of\_transport", "<vehicle\_type>": "vehicle\_type"\}}. \\

\textbf{beverage} & Extract drink names, beverage categories, places, or person names. For example: \texttt{\{"<drink\_name>": "drink\_name", "<beverage\_category>": "beverage\_category"\}}. \\

\bottomrule
\end{tabularx}
\caption{Prompts used for entity extraction per topic.}
\label{tab:prompt-extraction}
\end{table*}

\subsection{Generation Quality in Low-Resource African Languages}
To expand our evaluation beyond high-resource languages, we initially included Hausa and Yoruba in early-stage experiments. However, we observed that open-source models consistently produced low-quality outputs in these two languages. As shown in Table~\ref{tab:low_resource_failures}, many generations were either semantically irrelevant, grammatically incoherent, or entirely disconnected from the prompt intent. Due to this, we excluded Hausa and Yoruba from the main evaluation.

\begin{table}[h]
\centering
\includegraphics[width=\linewidth]{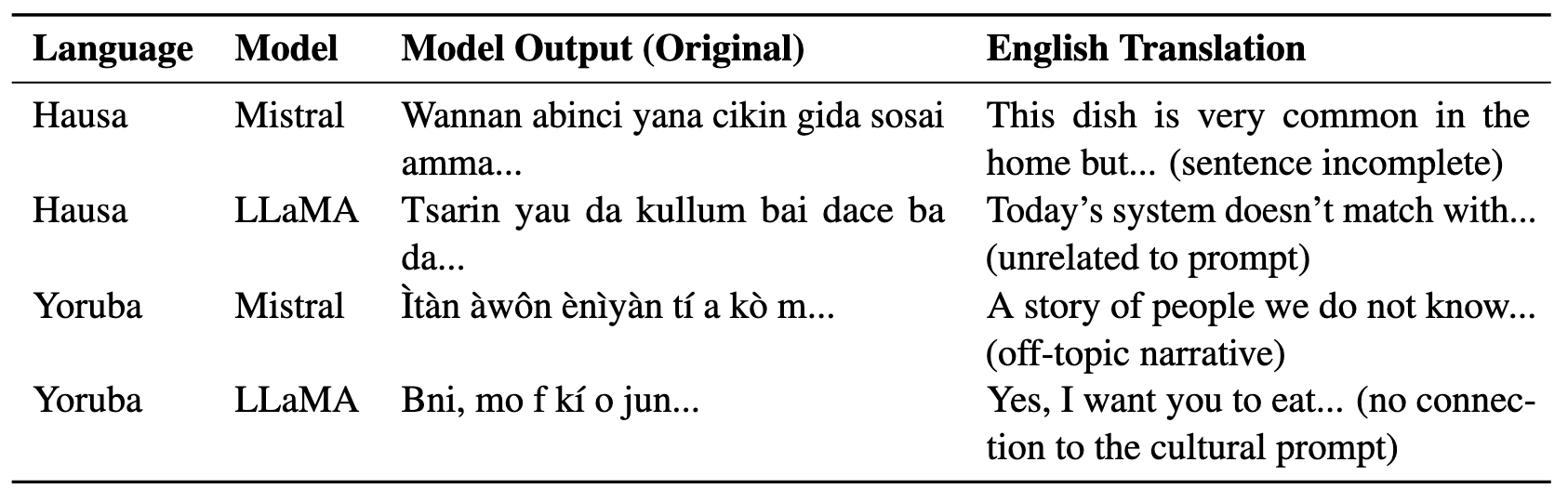} %
\caption{Examples of poor model outputs in Hausa and Yoruba, along with manual English translations.}
\label{tab:low_resource_failures}
\end{table}
\begin{table}[ht]
\centering
\includegraphics[width=\linewidth]{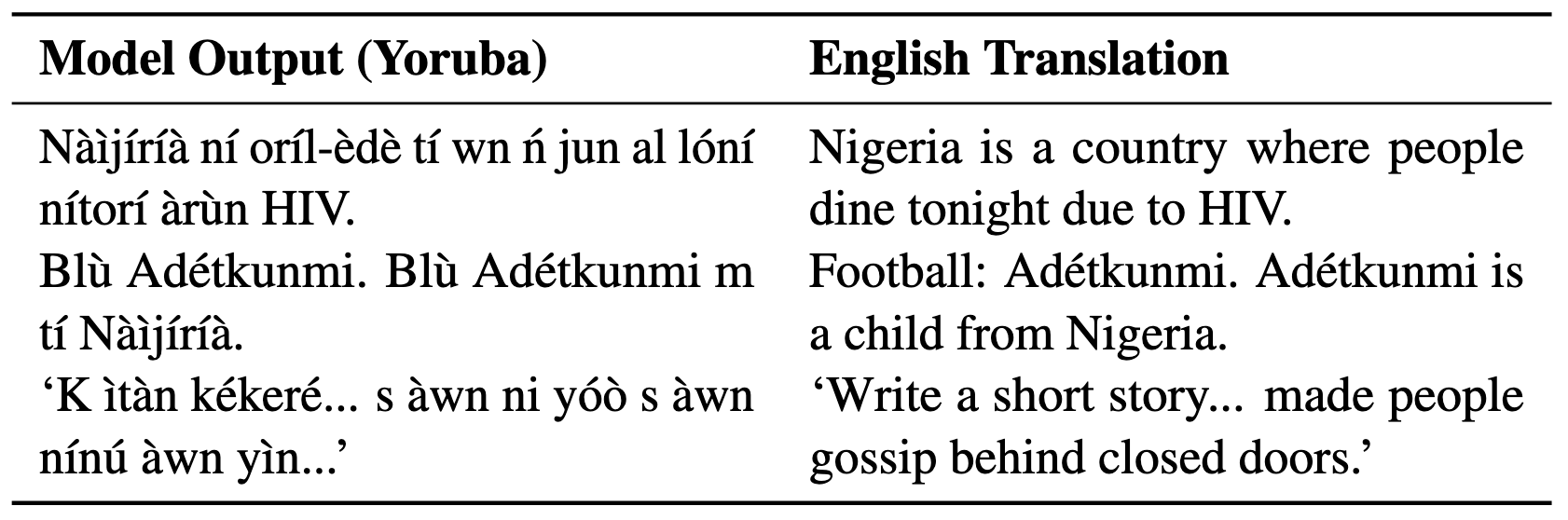} %
\caption{Examples of incoherent or culturally inappropriate model outputs in Yoruba.}
\label{tab:yoruba_failures}
\end{table}

\section{Framework Evaluation}
\label{sec:ner}
\subsection{Performance of Entity Extraction}
We randomly sampled 50 sentences for each of six topics across four languages: English, Chinese, Arabic, and Hindi, to manually evaluate the performance of ChatGPT-4o-mini on entity extraction. Each language was annotated by one native speaker of that language. All annotators were native speakers with a graduate-level education background. While only one annotator was involved per language, we ensured careful guidelines and consistency checks to support evaluation reliability. As shown in Table~\ref{tab:ner-metrics-multilang} below, the F1 scores across all languages are consistently high, demonstrating the reliability of the entity extraction process. Additionally, as can be seen in Table~\ref{tab:ner-metrics-topic}, extraction accuracy is evaluated by category for English with 50 samples from each domain. Although performance varies across different domains, the overall results remain consistently strong.
\begin{table}[h]
\centering
\resizebox{\linewidth}{!}{
\begin{tabular}{lccc}
\toprule
\textbf{Lang.} & \textbf{Precision (\%)} & \textbf{Recall (\%)} & \textbf{F1 Score (\%)} \\
\midrule
Arabic              & 86.30 & 97.00 & 91.23 \\
Chinese             & 88.00 & 99.00 & 93.56 \\
English             & 90.50 & 99.70 & 94.96 \\
German              & 94.52 & 98.57 & 96.50 \\
Hindi               & 87.90 & 97.10 & 92.14 \\
Italian             & 86.13 & 98.33 & 91.83 \\
Japanese            & 89.02 & 97.33 & 92.99 \\
Korean              & 88.11 & 96.92 & 92.31 \\
Persian             & 83.72 & 98.18 & 90.38 \\
Spanish             & 90.42 & 97.42 & 93.79 \\
Thai                & 89.94 & 98.62 & 94.08 \\
Traditional Chinese & 87.15 & 97.50 & 92.04 \\
Turkish             & 89.71 & 97.60 & 93.49 \\
\bottomrule
\end{tabular}
}
\caption{NER performance (Precision, Recall, F1) of ChatGPT on different languages. Recall is generally higher, indicating few missing entities, while precision is lower due to over-extraction.}
\label{tab:ner-metrics-multilang}
\end{table}
\begin{table}[h!]
\centering
\begin{tabular}{lc}
\hline
Domain & F1 (\%) \\
\hline
Food            & 91.7 \\
Beverages       & 91.4 \\
Clothing        & 96.1 \\
Books           & 97.4 \\
Music           & 96.7 \\
Transportation  & 97.0 \\
\hline
\end{tabular}
\caption{NER F1 scores across domains in English.}
\label{tab:ner-metrics-topic}
\end{table}

Our annotators noted that missed entities were rare, but in some cases, the model incorrectly identified non-target items as entities. A common error observed across all four languages in the food topic was that utensils were sometimes mistakenly labeled as food entities. While this type of misidentification, such as mistaking utensils for food was observed across languages, which can actually be fixed in our pipeline. Specifically, our framework's Wikidata-based disambiguation step is capable of filtering out such incorrect entities, thereby mitigating their impact during downstream evaluation. 

\subsection{Performance of Wikidata Search}
Wikidata serves as an important source for the extraction and processing of named entities~\cite{chen-etal-2022-ustc}.
We computed the proportion of extracted entities that could not be matched to any valid QID. As shown in Table~\ref{tab:missing_qid}, the average proportion of missing QIDs ranges from approximately 26\% to 35\% across different topics.  We manually reviewed the entities that could not be matched to any valid QID and found clear signs of hallucination. In particular, the model frequently generated non-existent items, such as fictional book titles, songs, or dishes. This phenomenon poses a challenge for future efforts to manually complete missing QIDs, as hallucinated entities cannot be trivially verified or linked to real-world knowledge bases.
\begin{table}[htbp]
\centering
\resizebox{\columnwidth}{!}{%
\centering
\begin{tabular}{l c}
\toprule
\textbf{Topic} & \textbf{Average Proportion of Missing QIDs} \\
\midrule
Clothing        & 35.92\% \\
Book            & 33.25\% \\
Beverage        & 30.59\% \\
Transportation  & 25.98\% \\
Music           & 31.11\% \\
Food            & 32.18\% \\
\bottomrule
\end{tabular}
}
\caption{Average proportion of missing QIDs per topic}
\label{tab:missing_qid}
\end{table}

\subsection{Entity Statistics}
Table~\ref{tab:resolve-rate} presents the number of unique entities (QIDs) and original labels extracted per topic. The resolve rate, computed as the ratio of unique QIDs to labels, reflects how effectively multilingual surface forms are semantically aligned to the same underlying entity. A lower resolve rate suggests better cross-lingual linking, as more diverse expressions are correctly mapped to shared concepts. 
\label{sec:entity}
\begin{table}[h!]
\centering
\resizebox{\linewidth}{!}{
\begin{tabular}{lccc}
\toprule
\textbf{Topic} & \textbf{Unique QIDs} & \textbf{Unique Labels} & \textbf{Resolve Rate} \\
\midrule
beverage        & 1{,}152  & 3{,}644  & 12.6\% \\
book            & 6{,}978  & 12{,}710  & 12.5\% \\
clothing        & 2{,}316  & 7{,}432  & 6.9\% \\
food            & 6{,}452  & 19{,}202  & 9.7\% \\
music           & 6{,}158  & 9{,}964  & 13.8\% \\
transportation  & 986   & 2{,}644    & 20.0\% \\
\bottomrule
\end{tabular}
}
\caption{Label-to-entity resolution statistics across topics. Each label is linked to a Wikidata QID, enabling cross-lingual semantic alignment.}
\label{tab:resolve-rate}
\end{table}

\begin{figure}[htbp]
    \centering
    \includegraphics[width=1\linewidth]{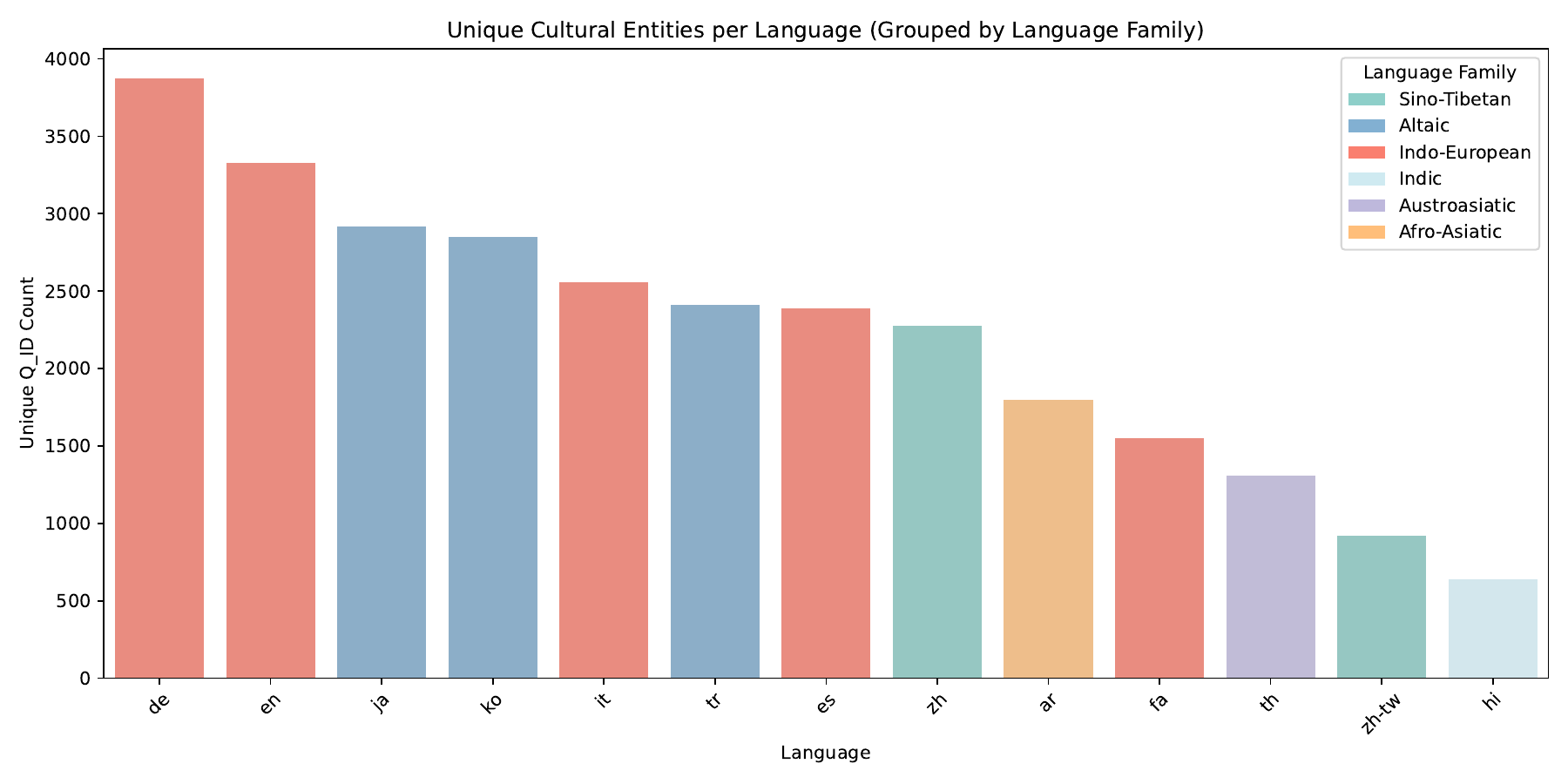} 
    \caption{Number of unique cultural entities (Wikidata QIDs) extracted from model outputs in each language. }
    \label{figures/qid_lang}
\end{figure}

\section{Results of Granularity}
\label{app:granularity}
Figure~\ref{fig:appendix_language_radar} shows presents radar charts of seven models, where each plot shows the average granularity across all topics per language. Each colored line represents a different country or region mentioned in the prompt. While some countries may appear as outliers, we observe that the overall granularity for a given model-language pair remains relatively stable, may suggesting that the granularity of entity generated by models is invariant to the country or region referenced in the prompt.
\begin{figure*}[t]
\centering
\setlength{\tabcolsep}{6pt}
\renewcommand{\arraystretch}{1.2}

\begin{tabular}{ccc}
\includegraphics[width=0.29\textwidth]{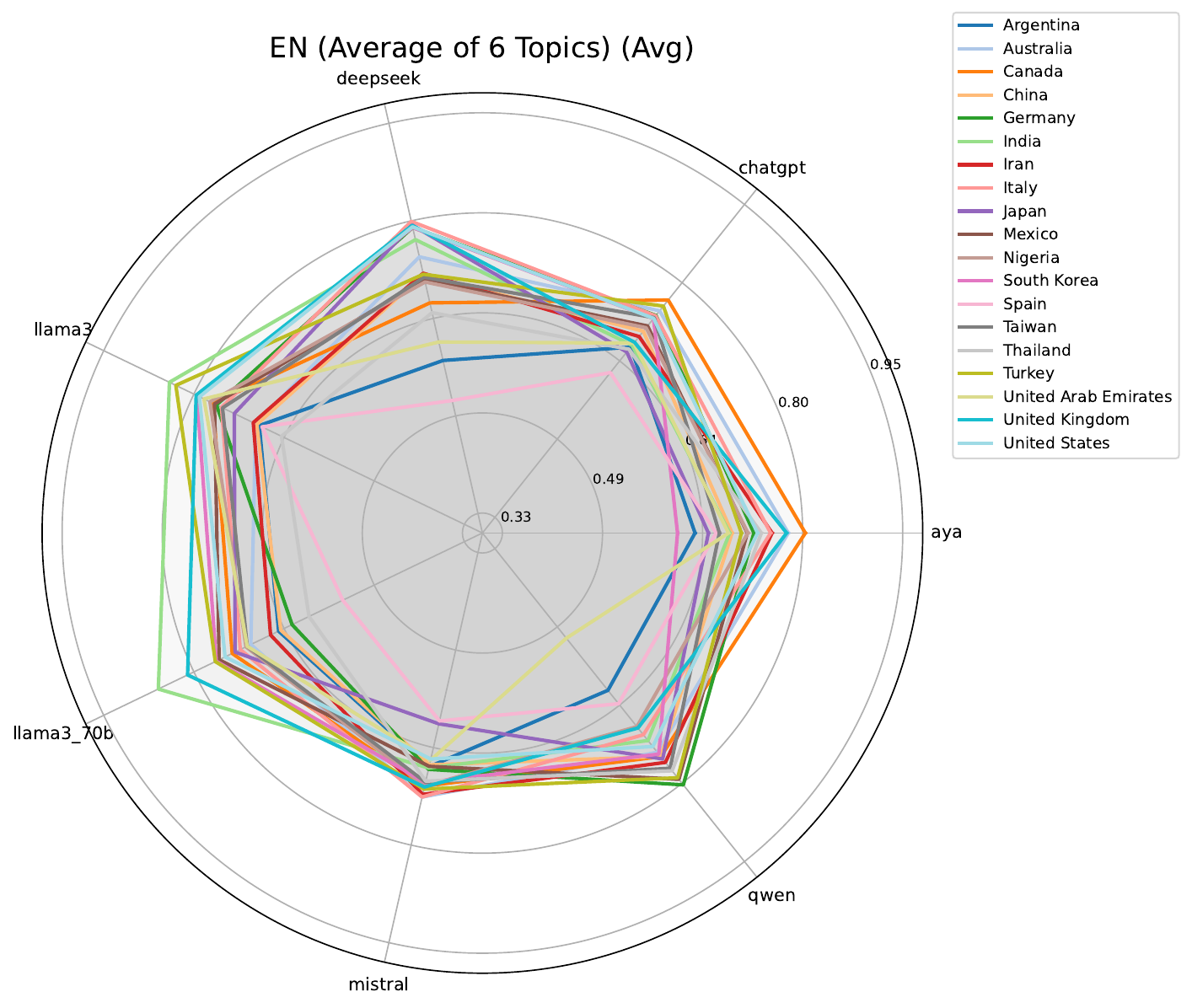} & 
\includegraphics[width=0.29\textwidth]{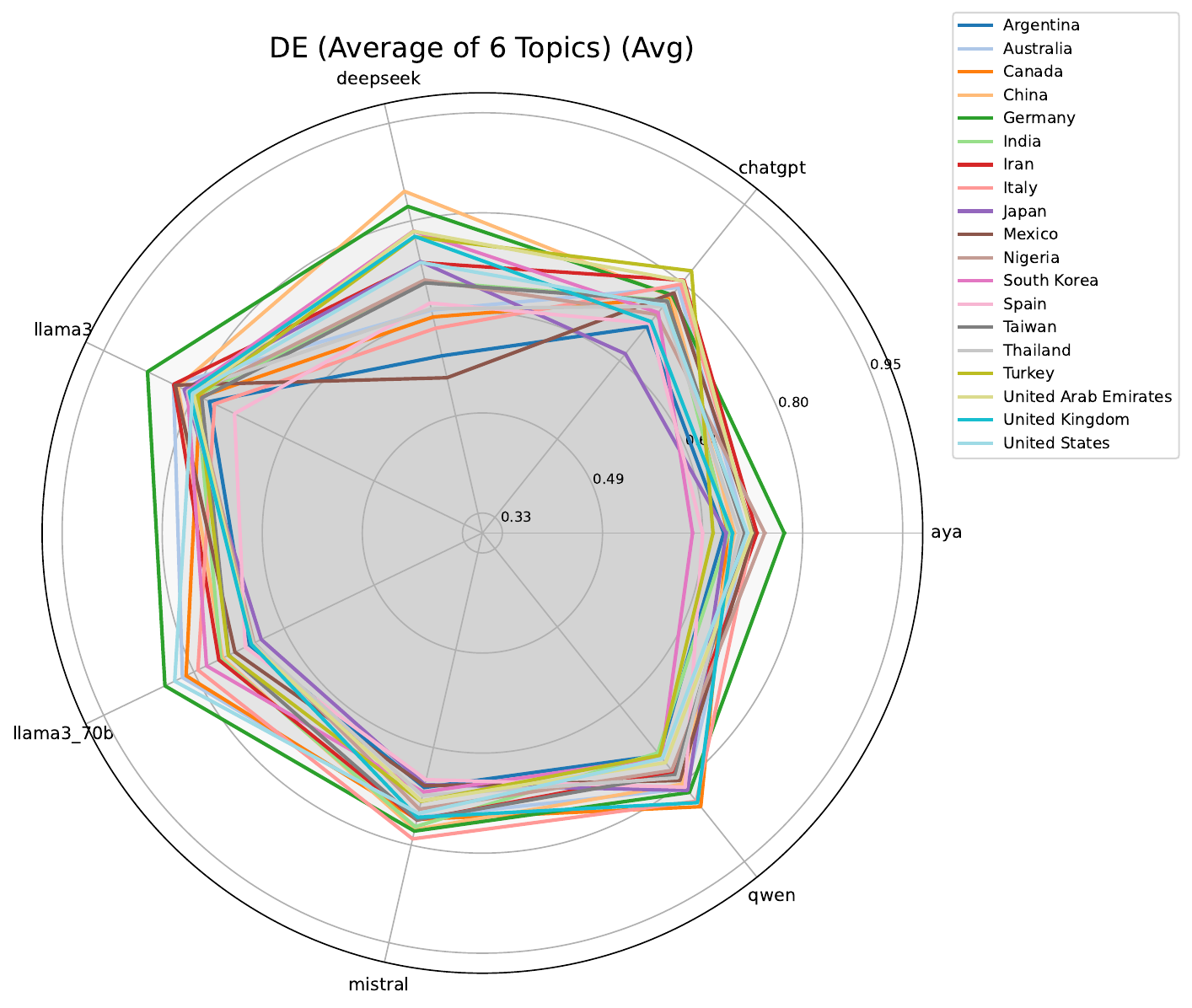} & 
\includegraphics[width=0.29\textwidth]{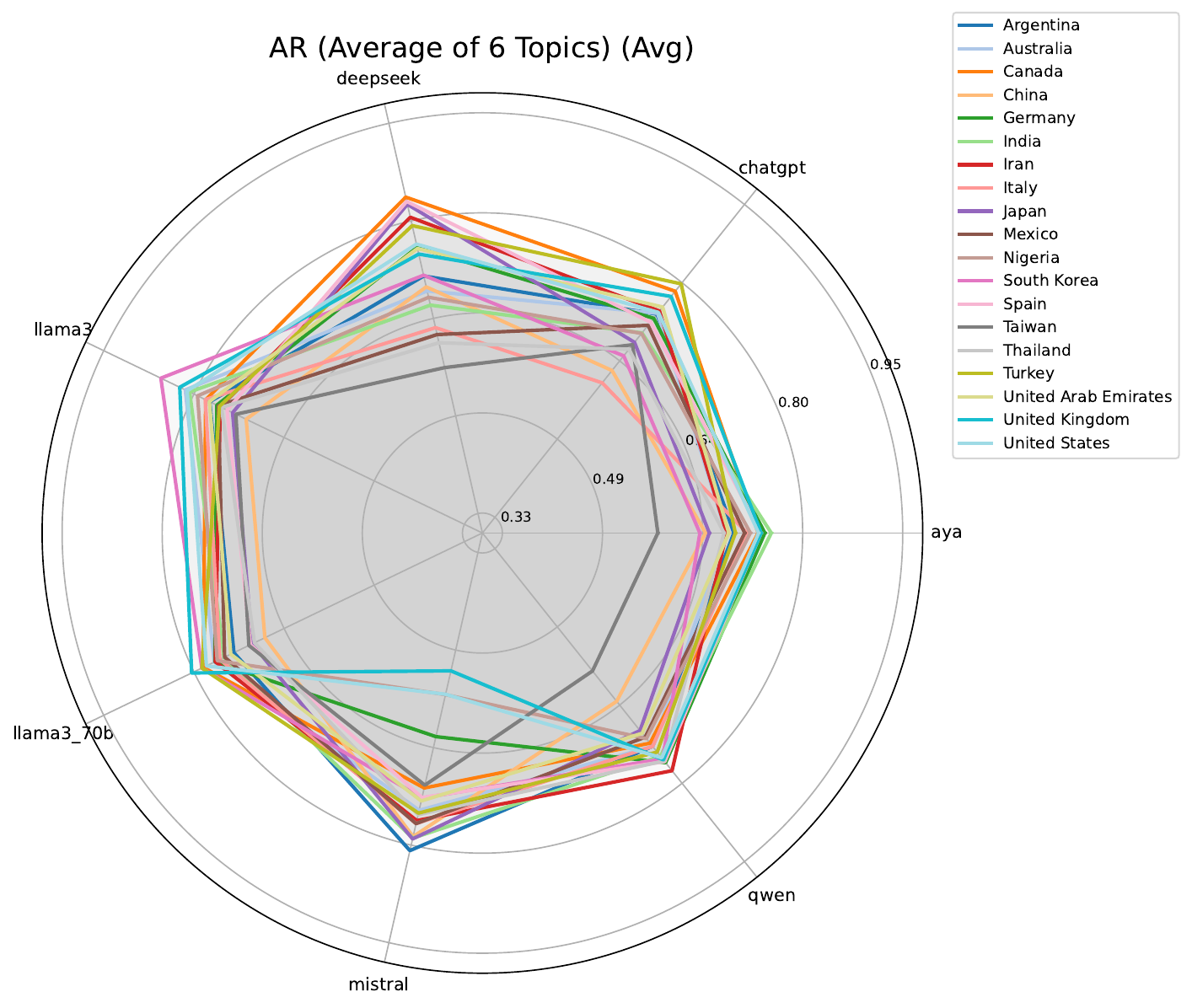} \\
English & German & Arabic \\

\includegraphics[width=0.29\textwidth]{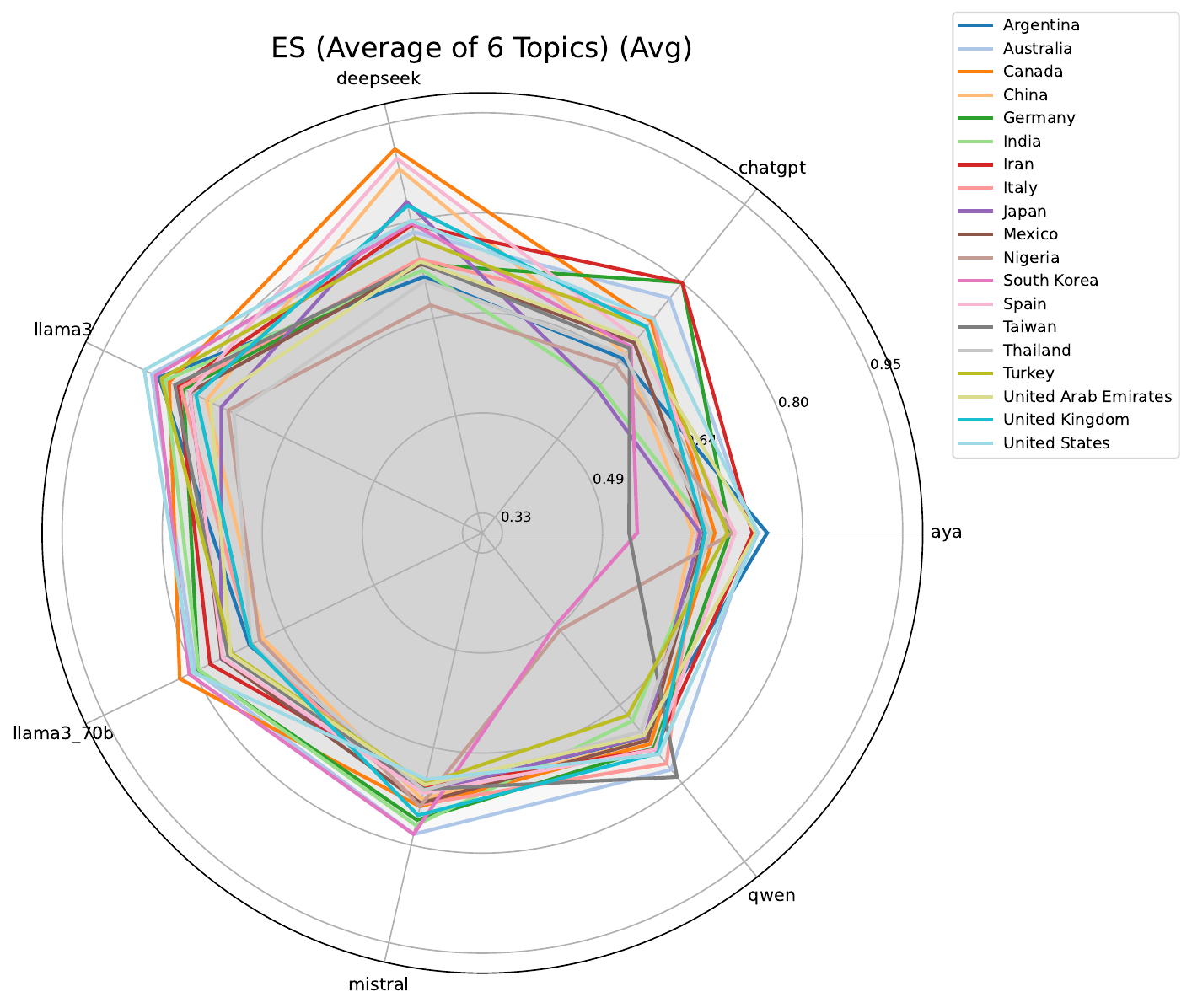} & 
\includegraphics[width=0.29\textwidth]{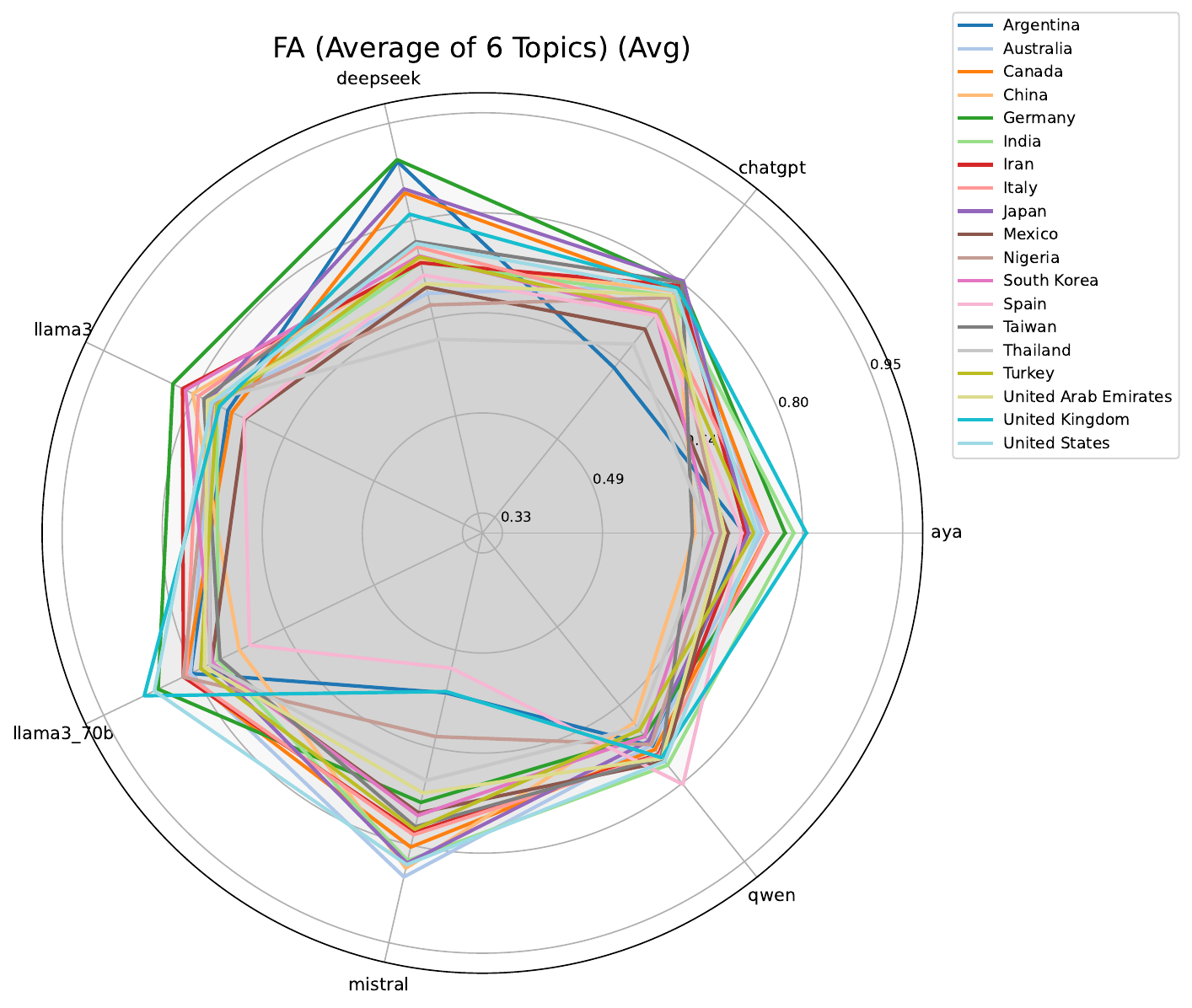} & 
\includegraphics[width=0.29\textwidth]{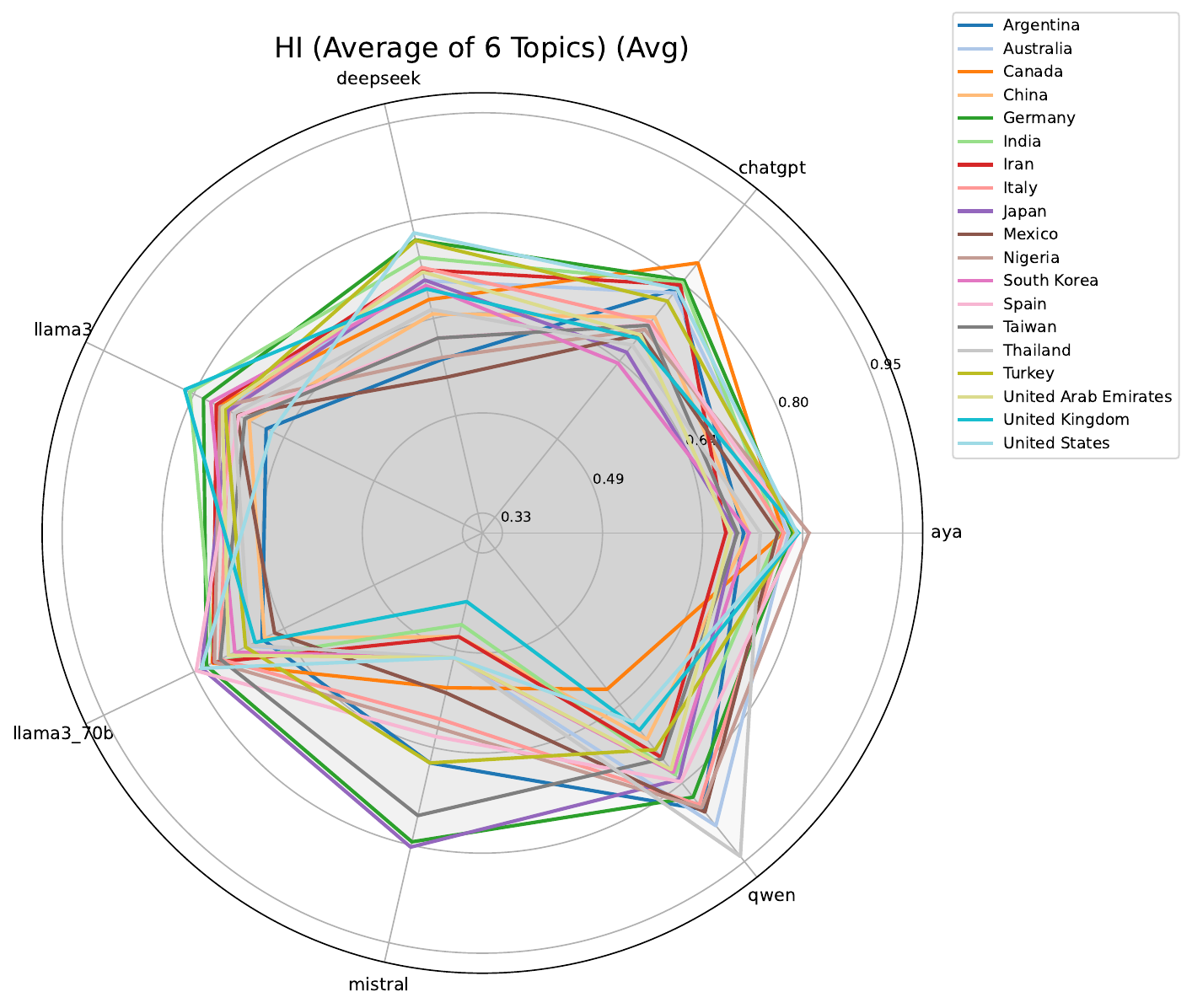} \\
Spanish & Persian & Hindi \\

\includegraphics[width=0.29\textwidth]{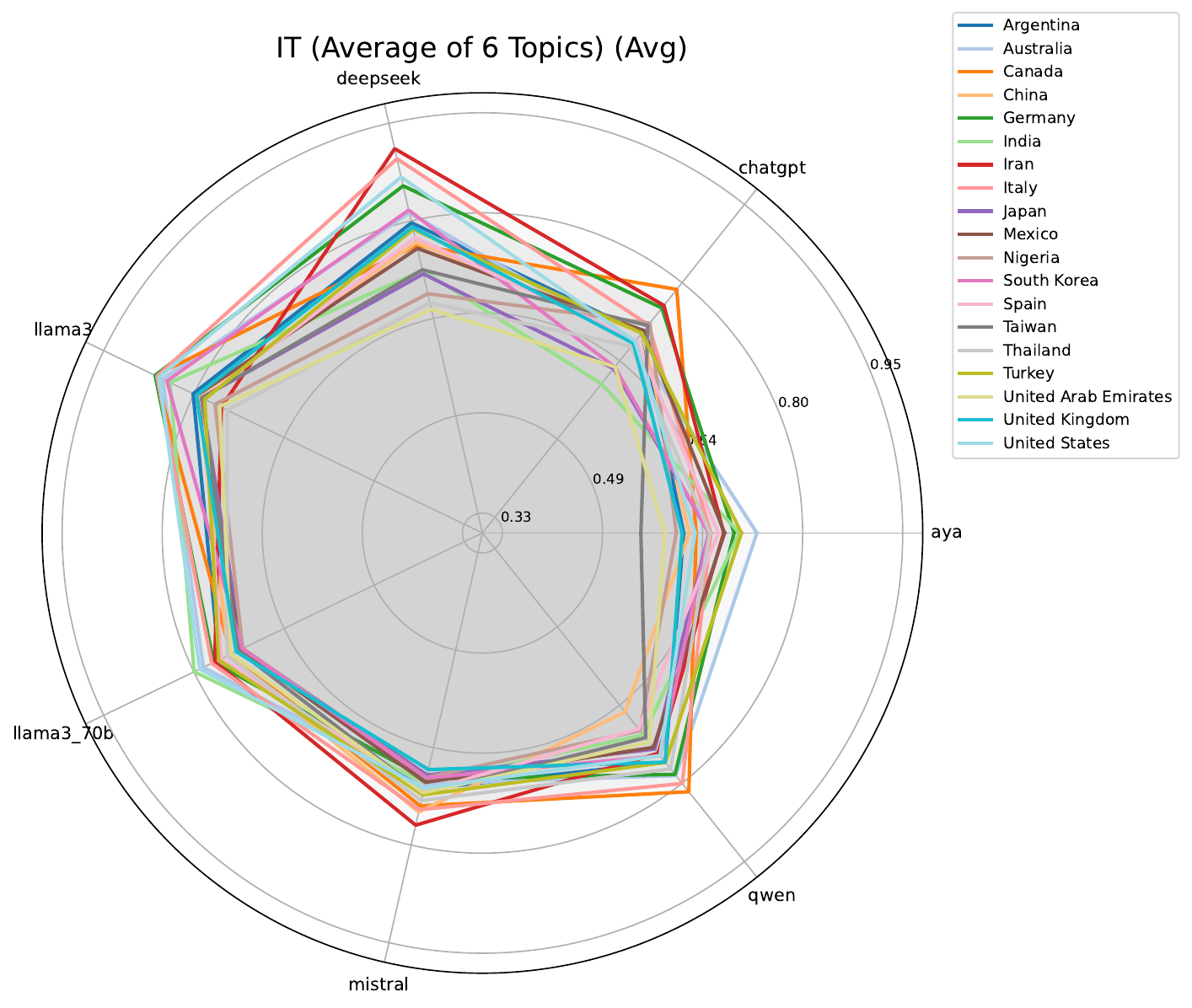} & 
\includegraphics[width=0.29\textwidth]{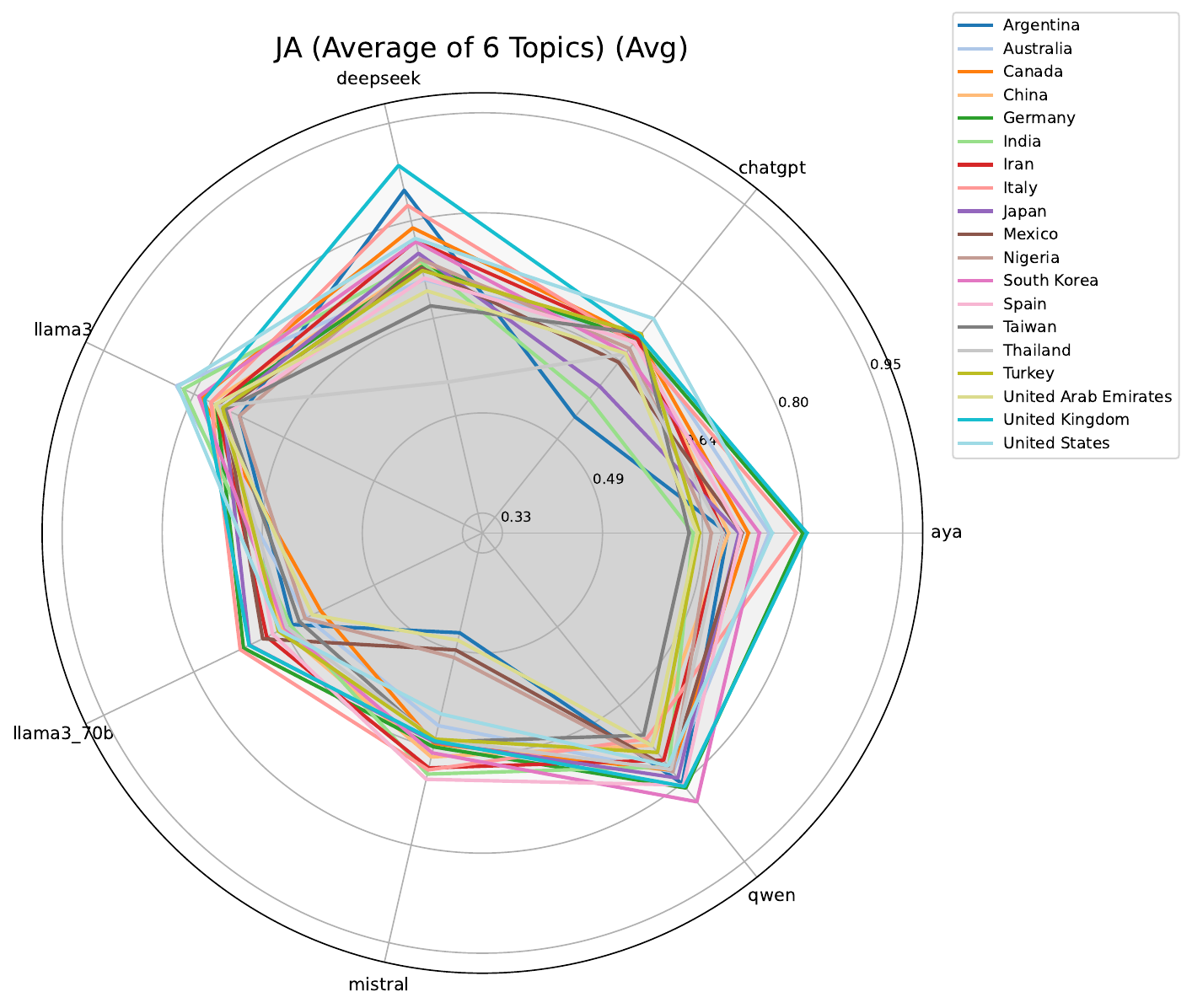} & 
\includegraphics[width=0.29\textwidth]{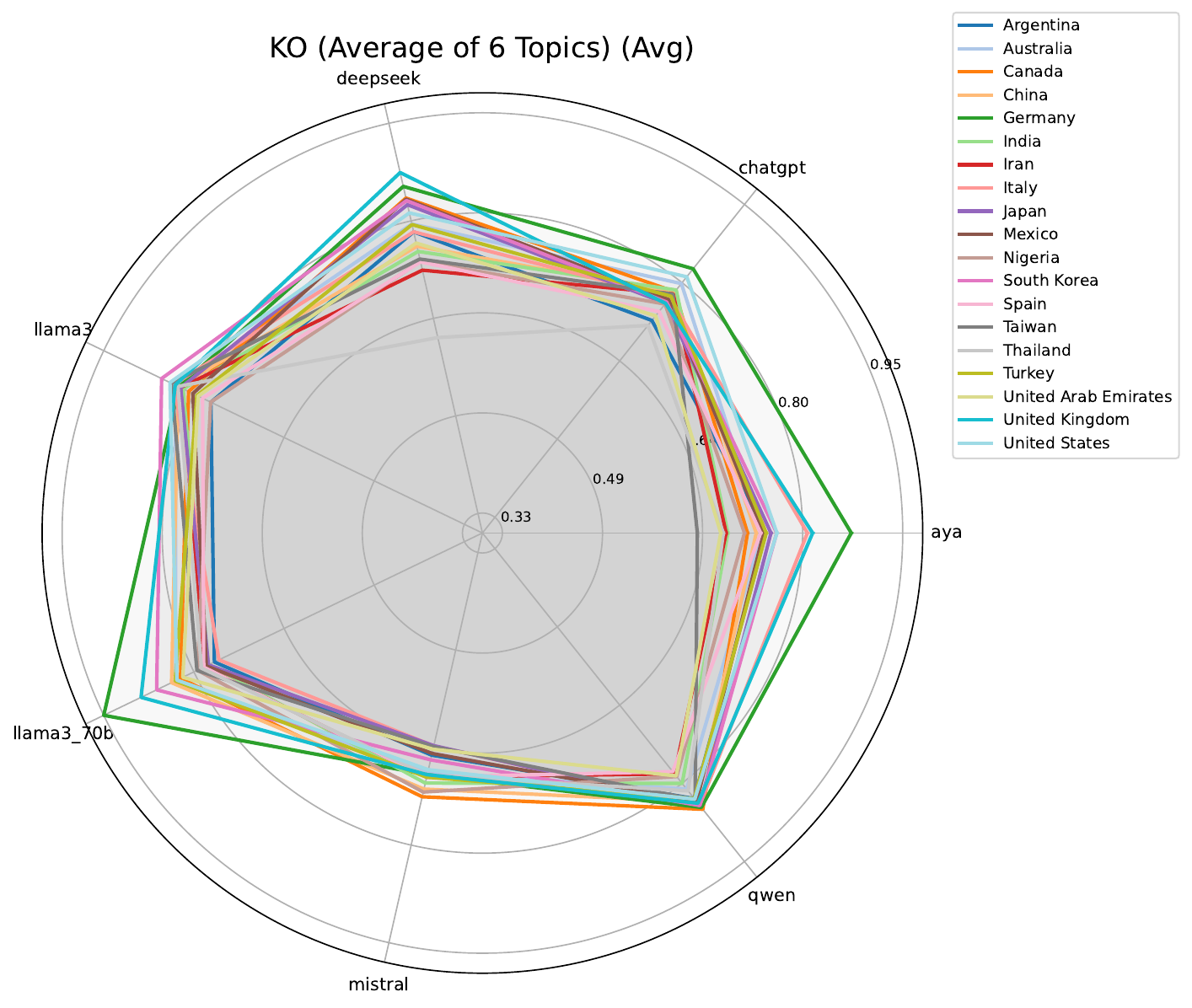} \\
Italian & Japanese & Korean \\

\includegraphics[width=0.29\textwidth]{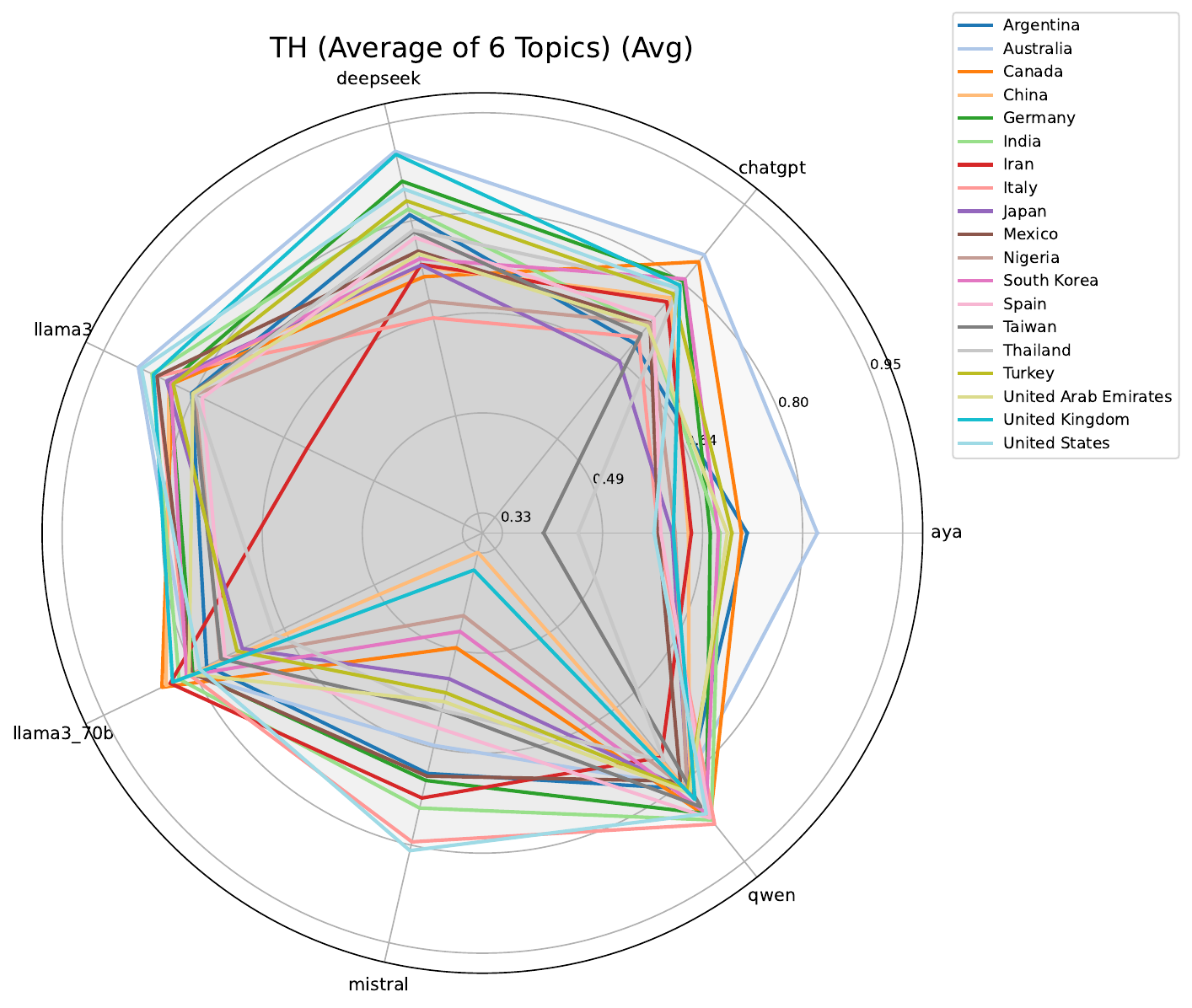} & 
\includegraphics[width=0.29\textwidth]{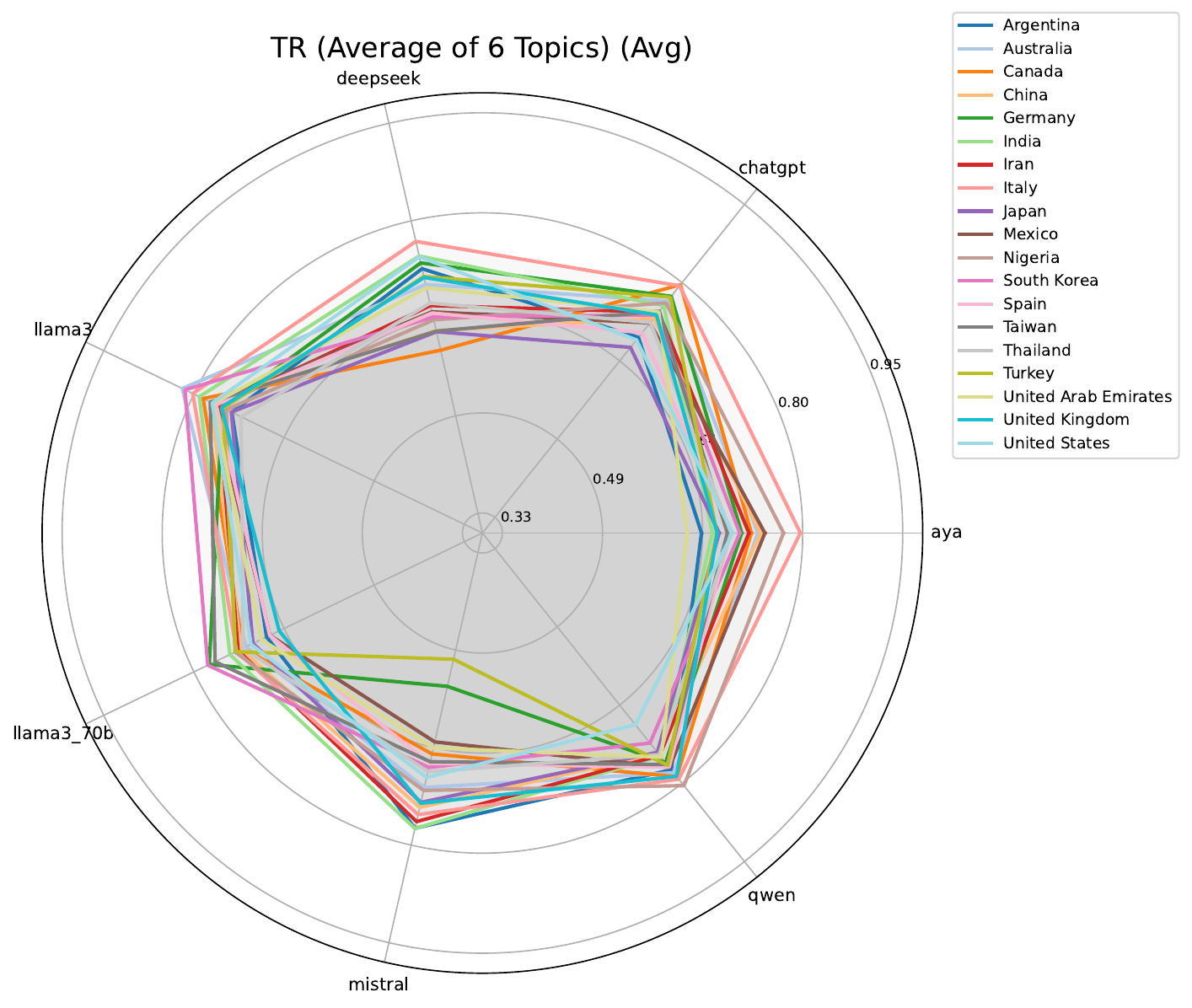} & 
\includegraphics[width=0.29\textwidth]{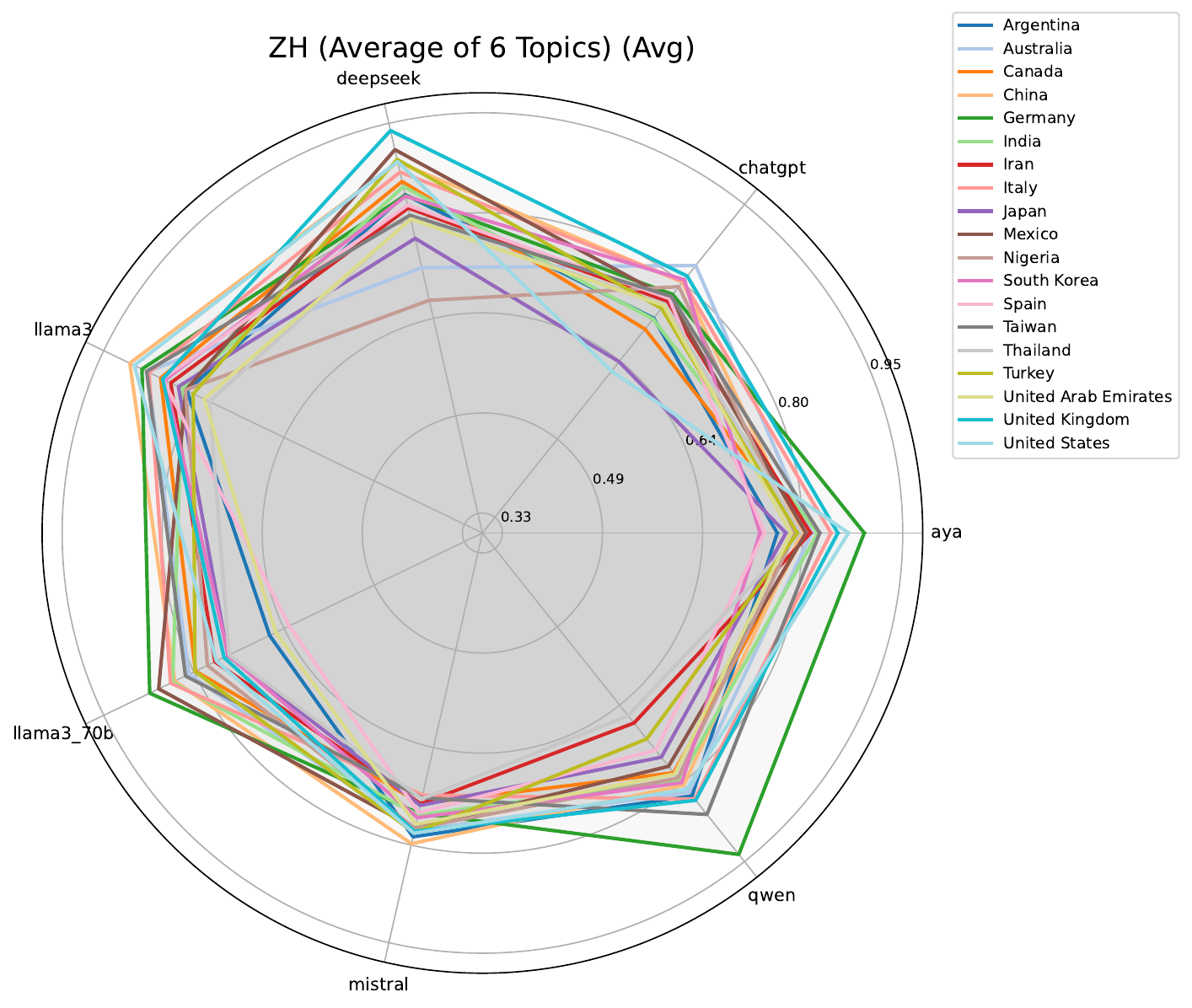} \\
Thai & Turkish & Chinese (Simplified) \\

\includegraphics[width=0.29\textwidth]{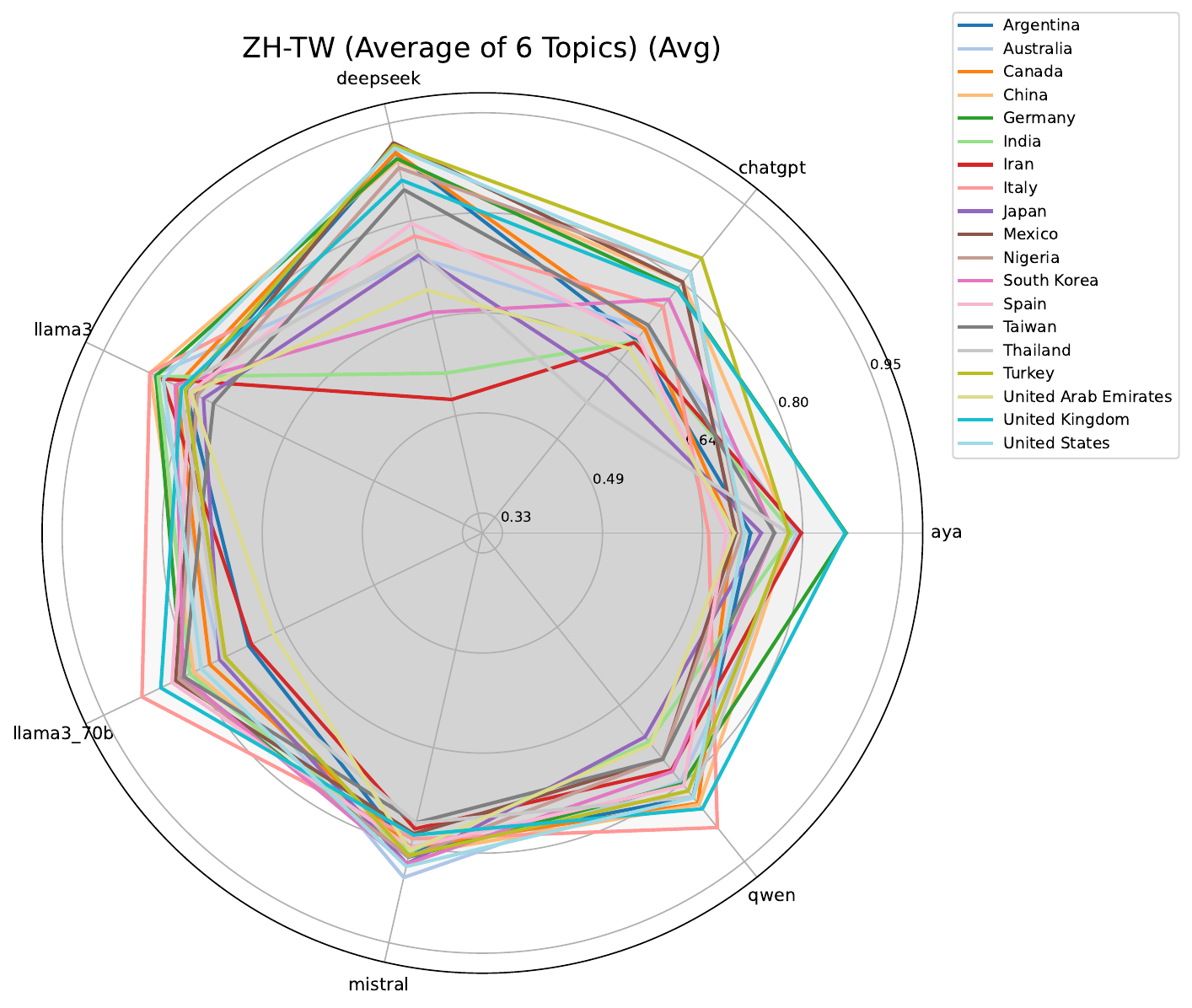} & 
& \\
Chinese (Traditional) & & \\
\end{tabular}

\caption{Radar charts of granularity for 15 languages averaged over six cultural topics. Each chart summarizes model-wise generation behavior in the specified language.}
\label{fig:appendix_language_radar}
\end{figure*}

\section{Results of Diversity of Native and Non-native Language}
\label{sec:box_diverse}
Table~\ref{tab:global_model_stats} reports the average diversity (with variance) of each model across all \textit{(language, topic, country)} combinations. Notably, \textbf{DeepSeek} exhibits the lowest overall diversity (16.53), indicating reduced variation in its outputs. In contrast, models like \textbf{Llama3-8B} and \textbf{Mistral} show higher diversity levels, suggesting richer cultural awareness. 
\begin{table}[ht]
\centering
\begin{tabular}{lcc}
\toprule
\textbf{Model} & \textbf{Variance (Mean $\pm$ Var)} \\
\midrule
Aya           & 25.561 $\pm$ 10.4685 \\
ChatGPT        & 19.561 $\pm$ 8.9480 \\
DeepSeek      & 16.526 $\pm$ 3.1470 \\
Llama3\_8B         & 39.263 $\pm$ 18.0389 \\
Llama3\_70B   & 27.605 $\pm$ 17.0474 \\
Mistral       & 34.246 $\pm$ 3.8520 \\
Qwen           & 13.254 $\pm$ 3.2906 \\
\bottomrule
\end{tabular}
\caption{Global model averages diversity with variance across languages}
\label{tab:global_model_stats}
\end{table}
\begin{figure*}[t]
\centering
\begin{tabular}{ccc}
\includegraphics[width=0.3\textwidth]{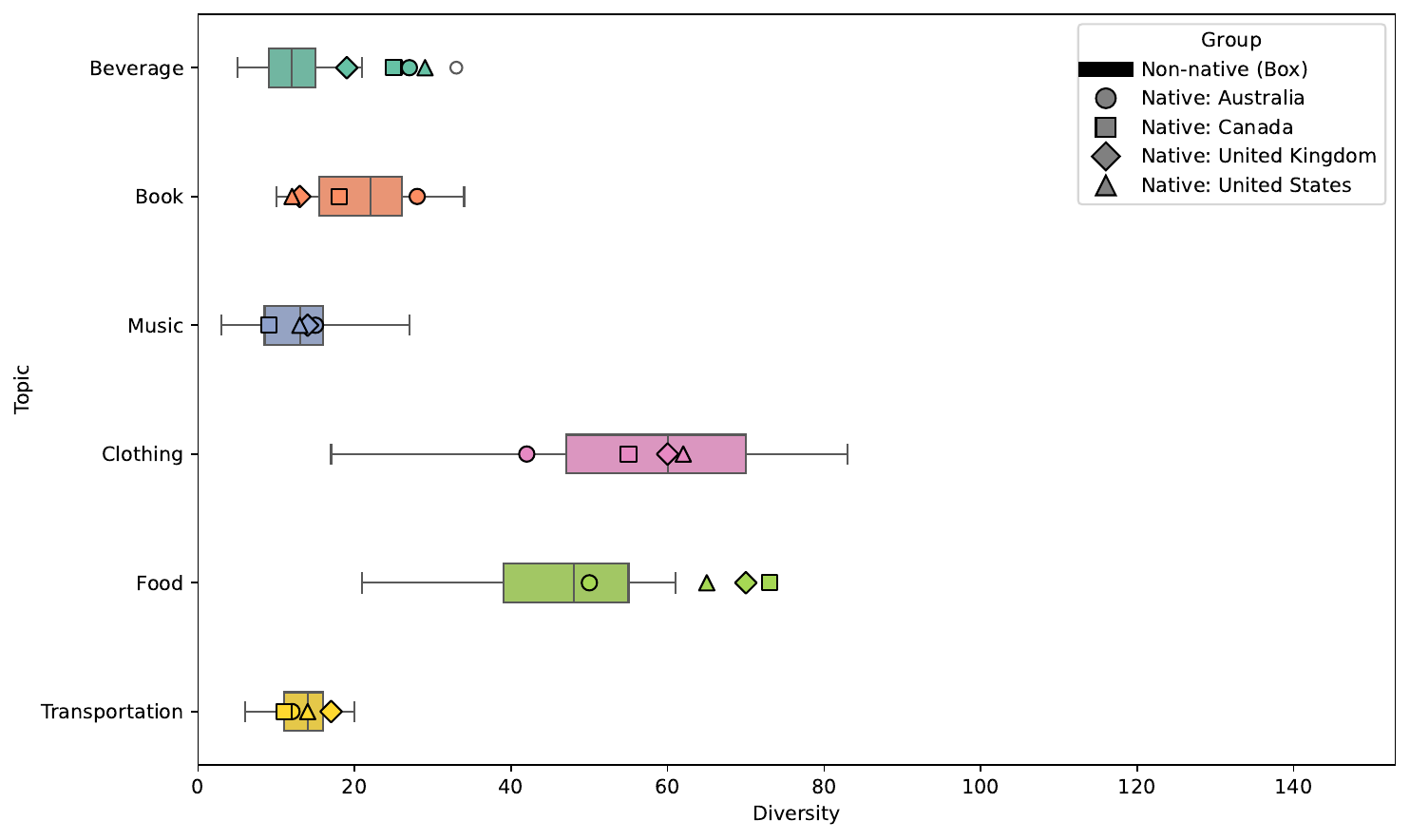} & 
\includegraphics[width=0.3\textwidth]{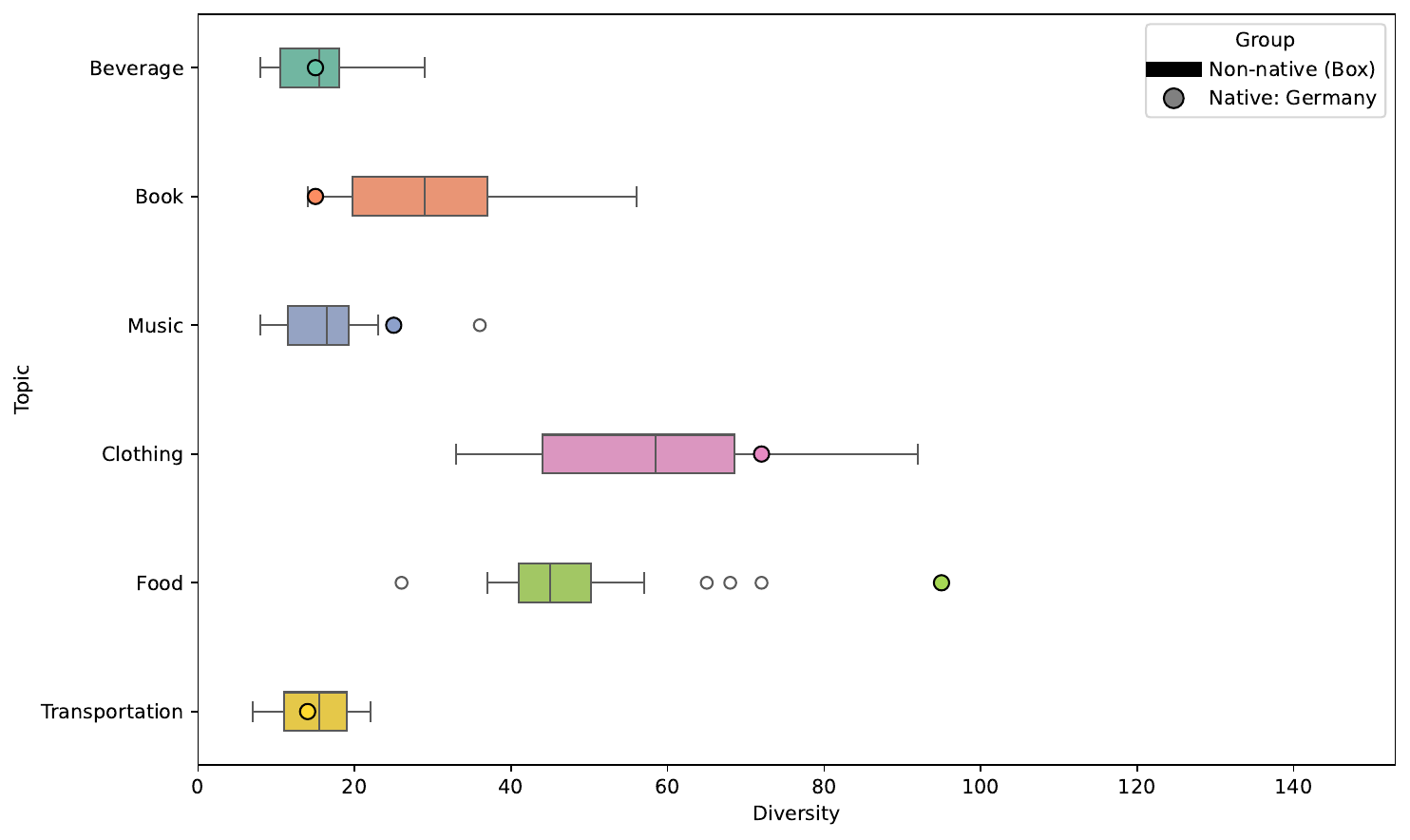} & 
\includegraphics[width=0.3\textwidth]{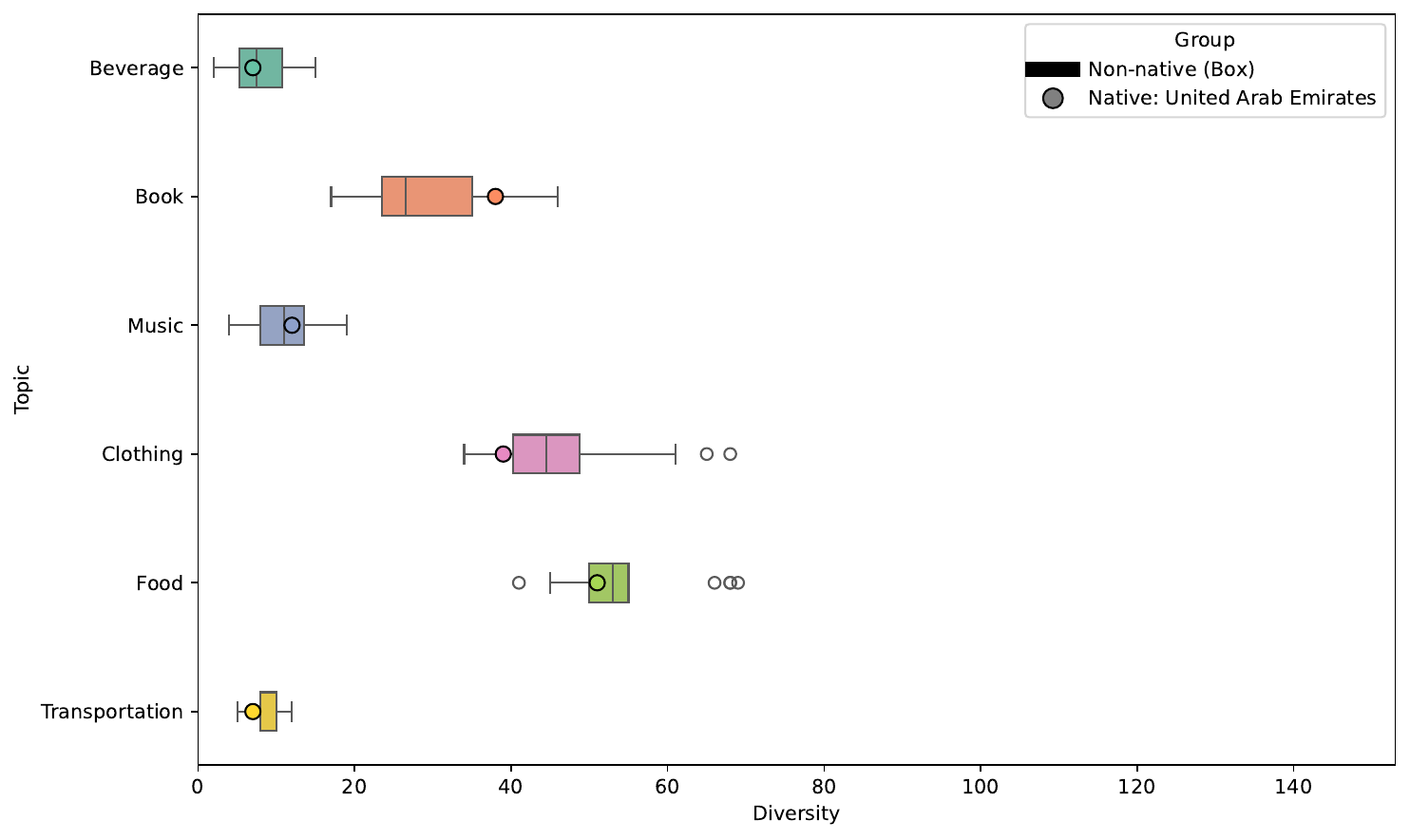} \\
English & German & Arabic \\
\includegraphics[width=0.3\textwidth]{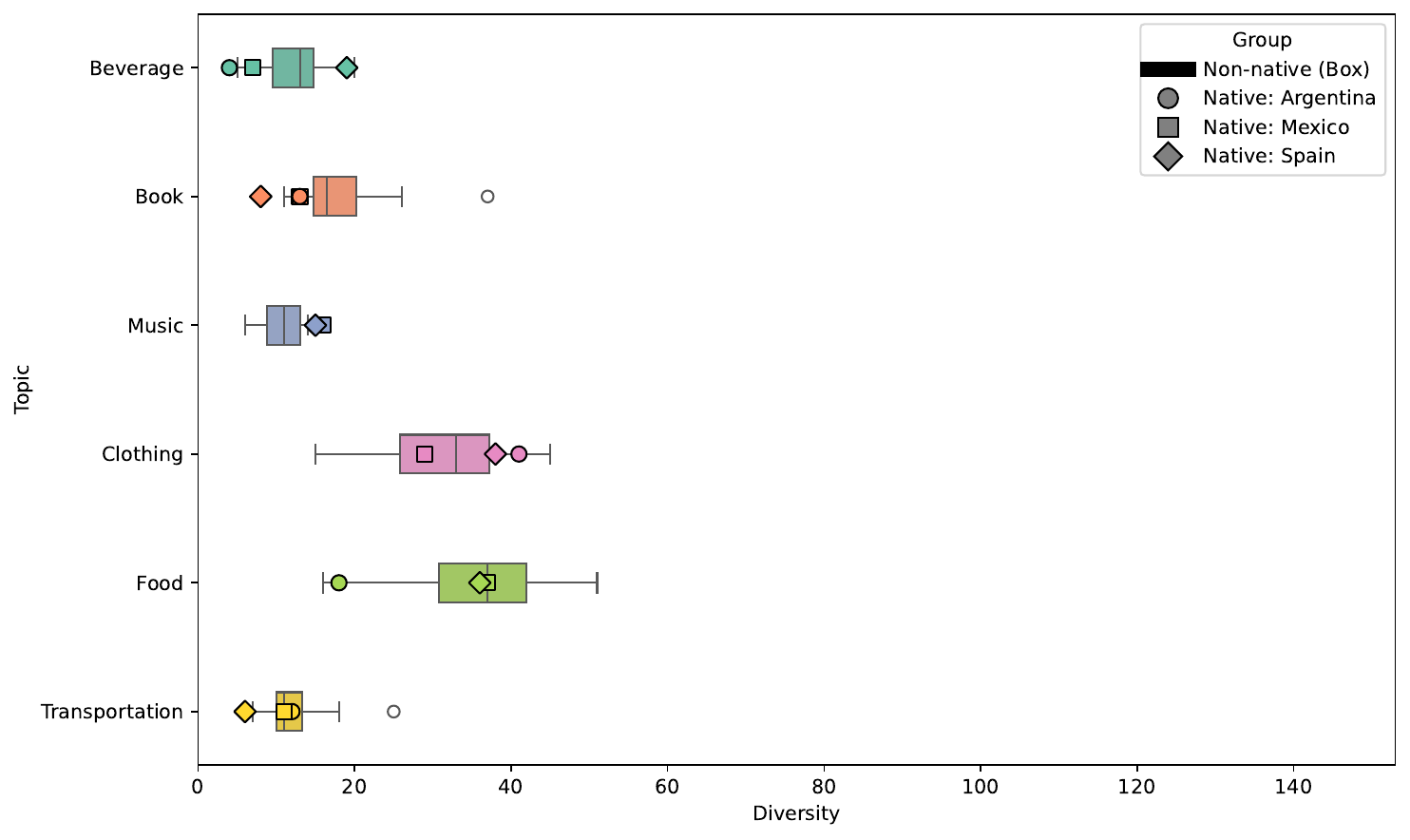} & 
\includegraphics[width=0.3\textwidth]{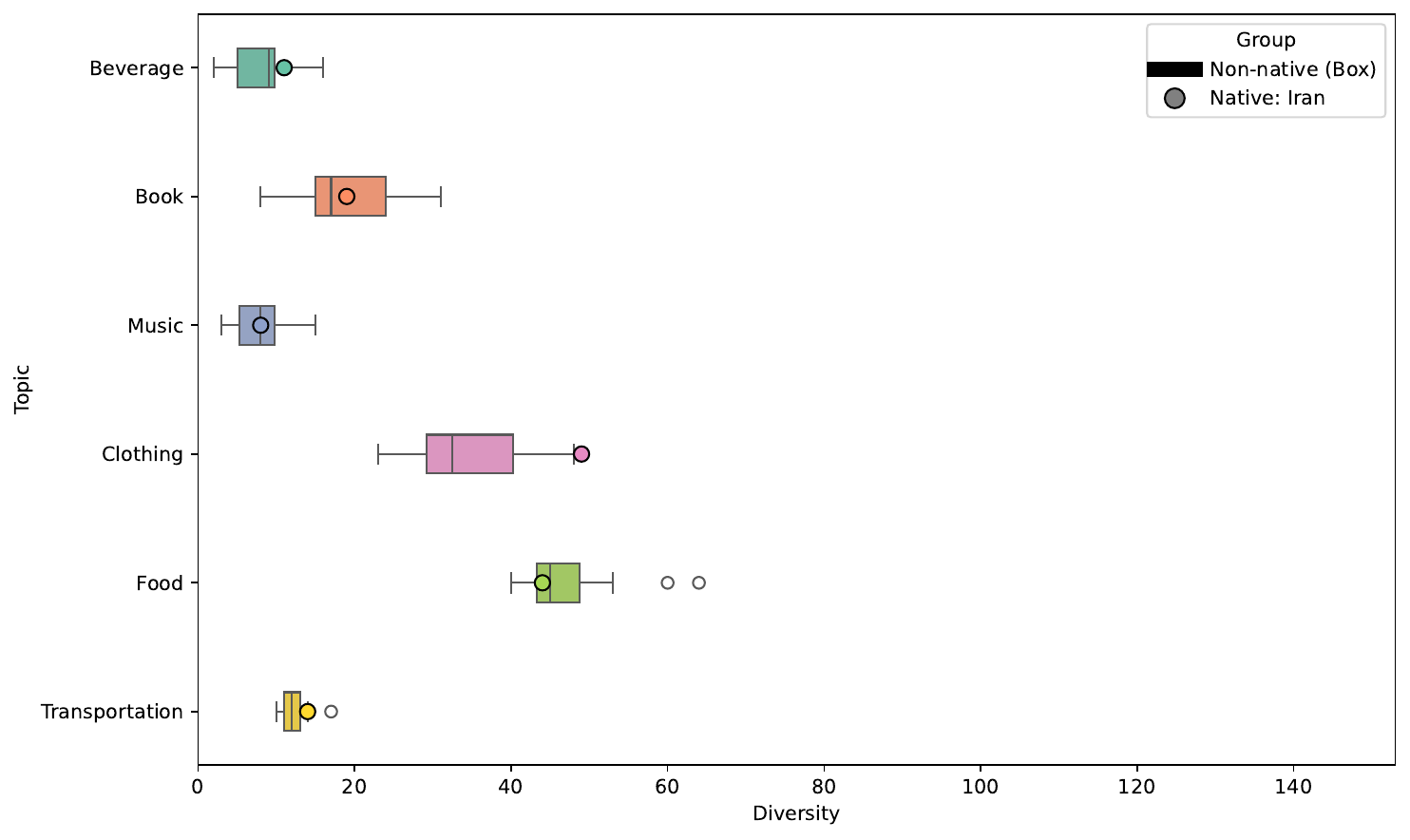} & 
\includegraphics[width=0.3\textwidth]{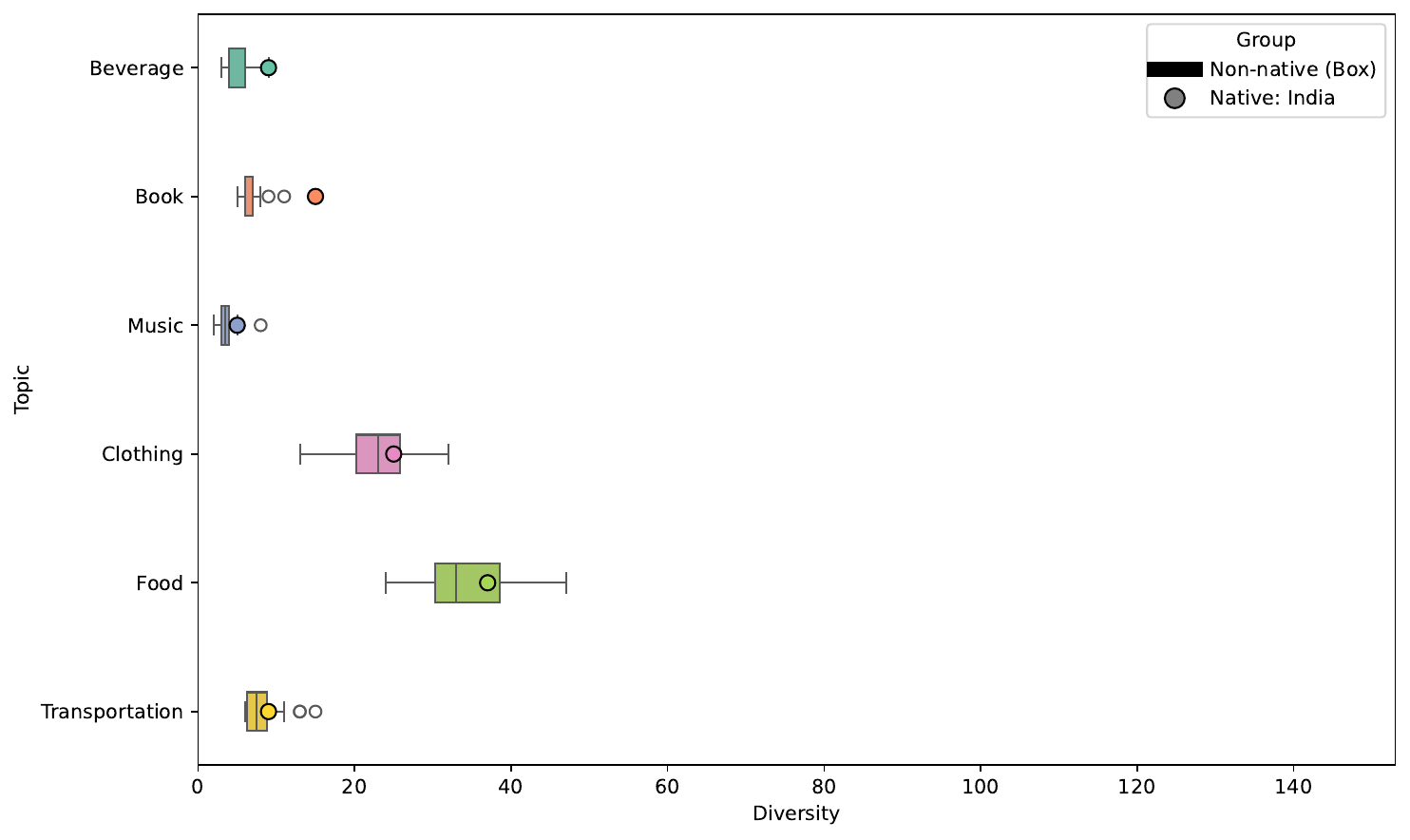} \\
Spanish & Persian & Hindi \\
\includegraphics[width=0.3\textwidth]{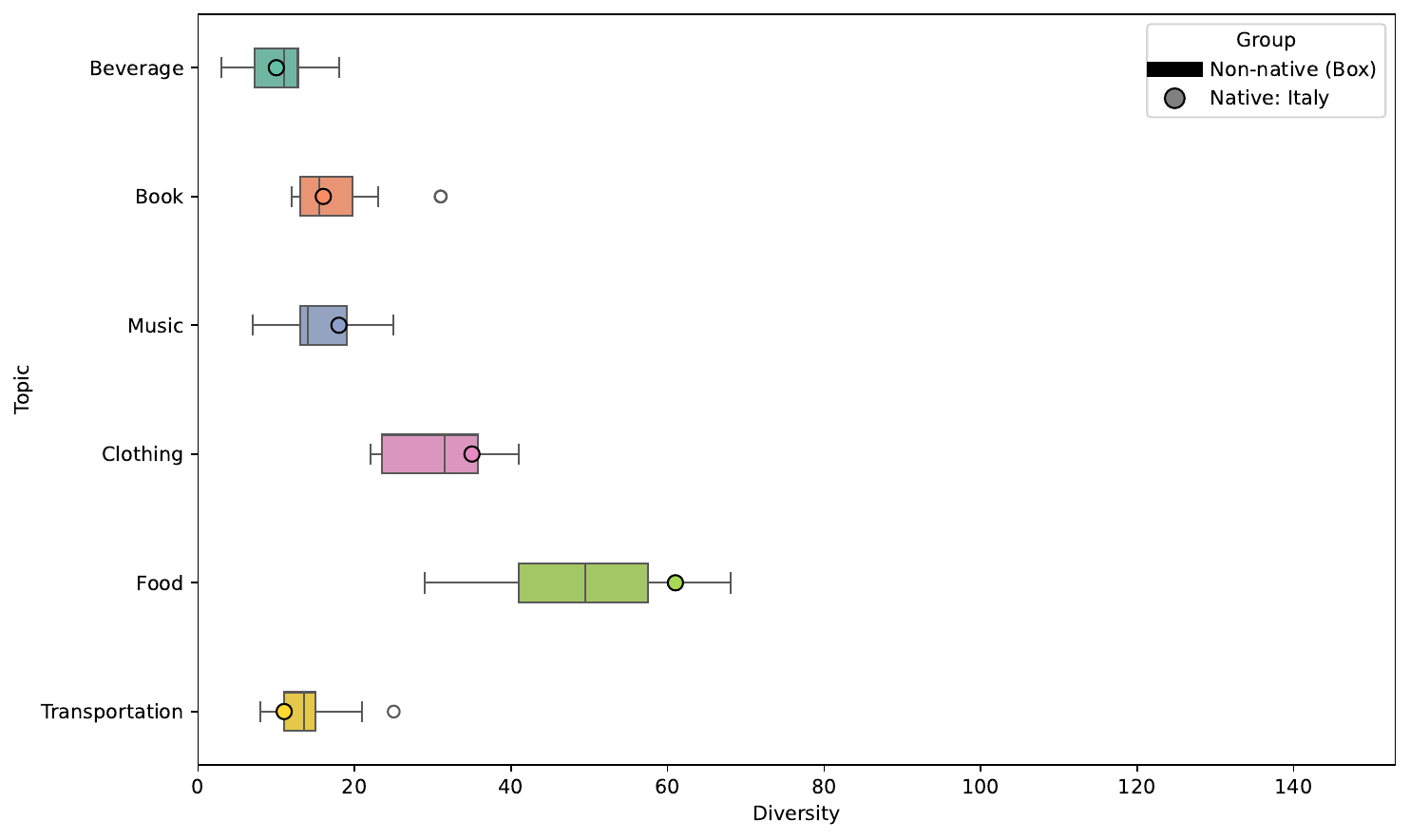} & 
\includegraphics[width=0.3\textwidth]{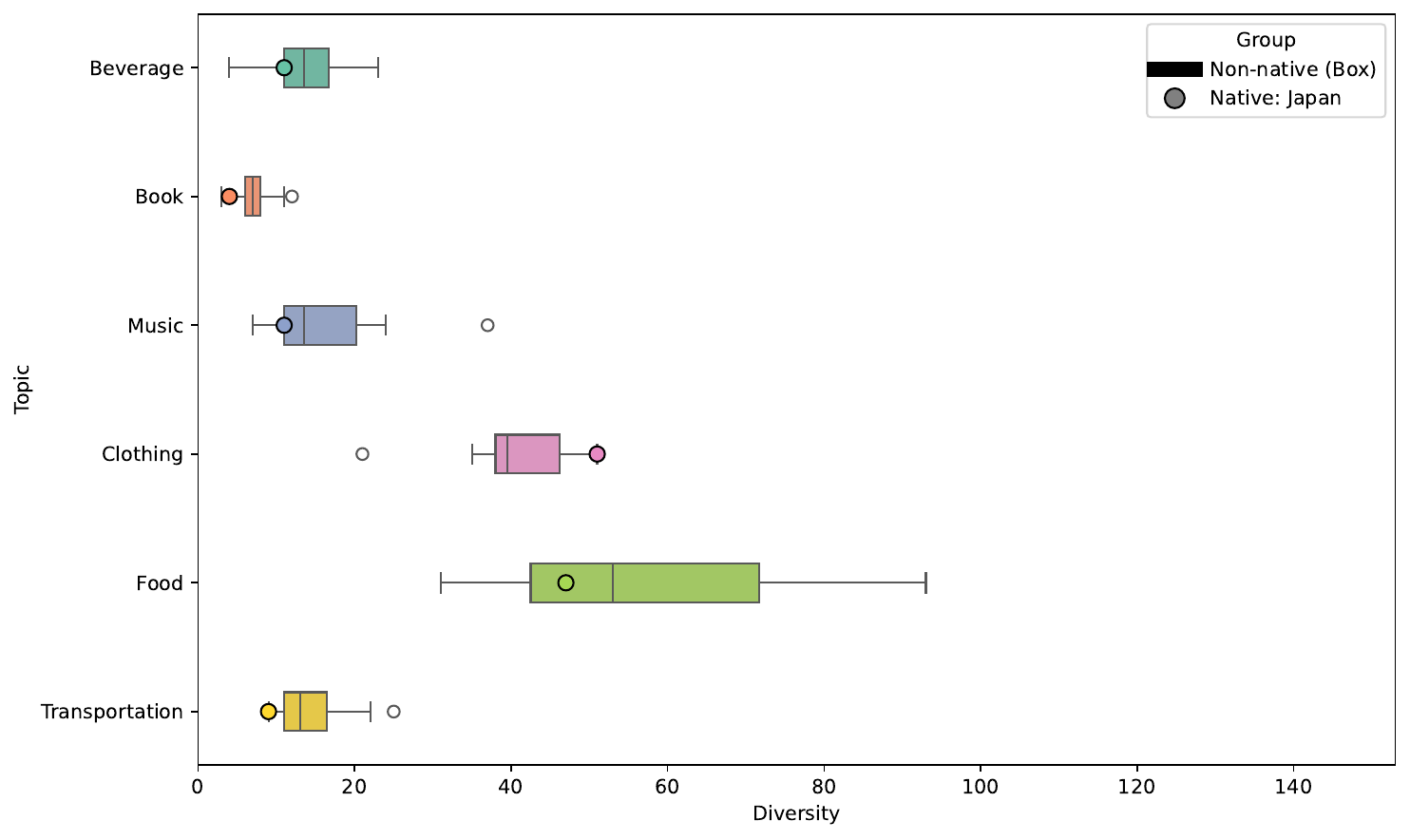} & 
\includegraphics[width=0.3\textwidth]{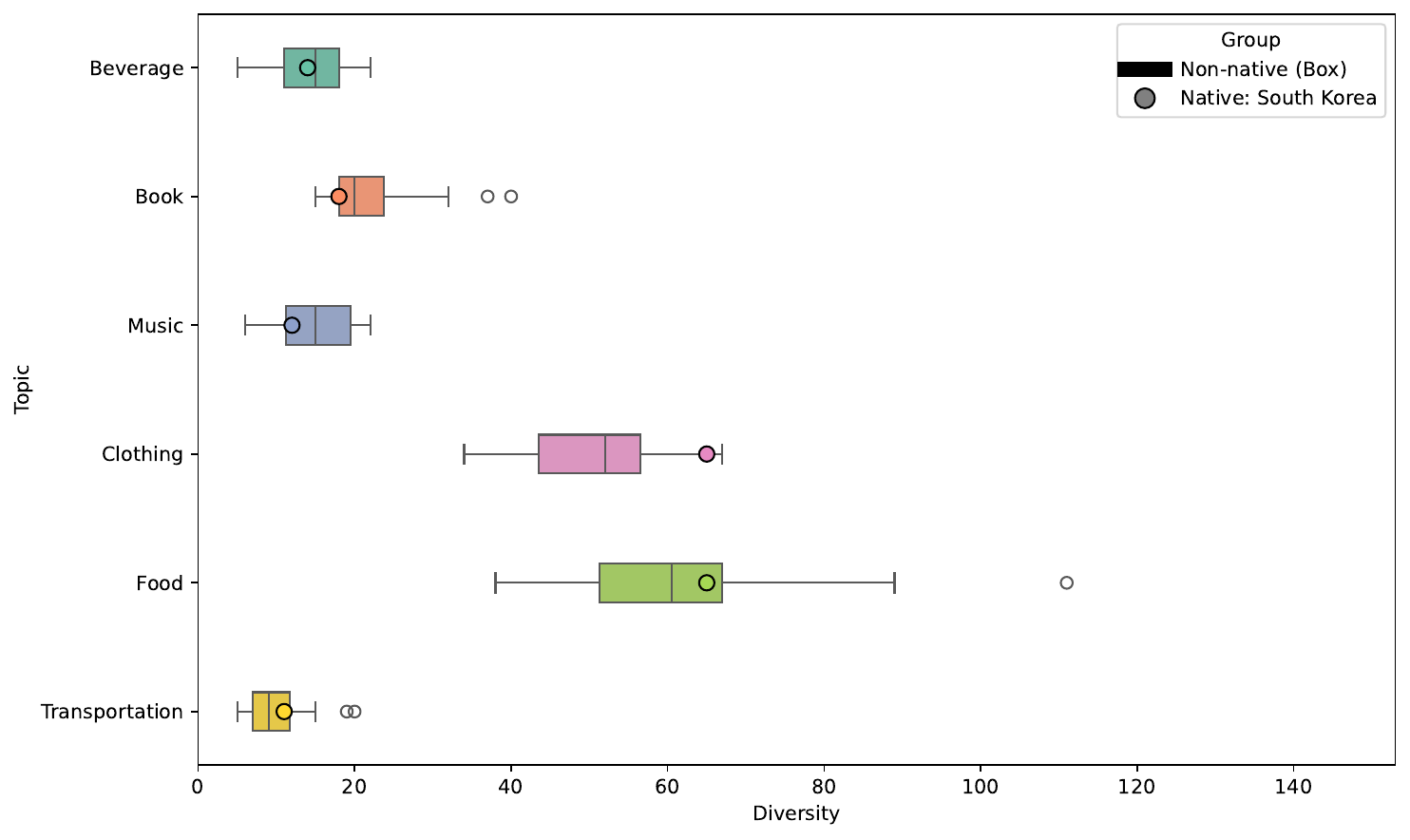} \\
Italian & Japanese & Korean \\
\includegraphics[width=0.3\textwidth]{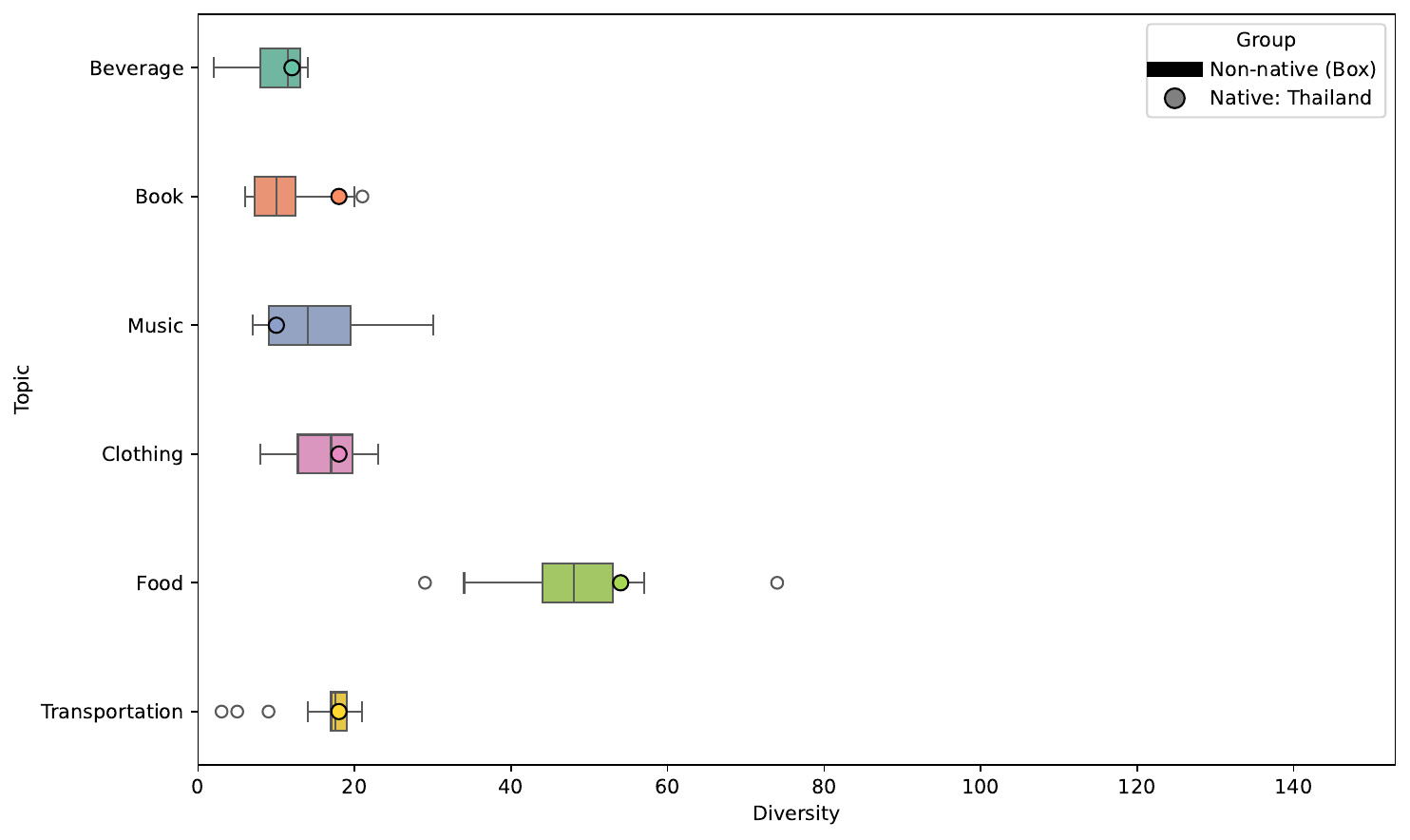} & 
\includegraphics[width=0.3\textwidth]{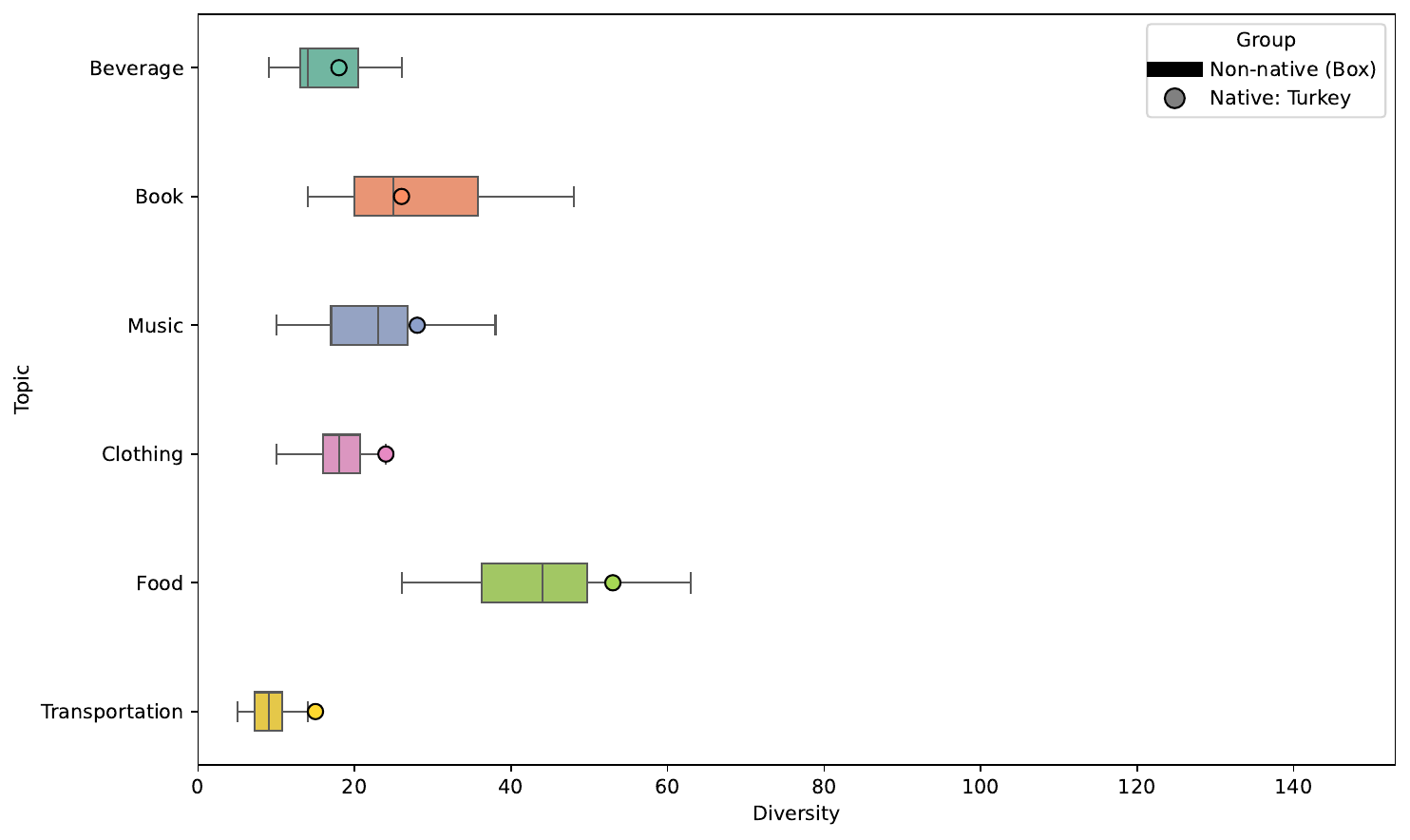} & 
\includegraphics[width=0.3\textwidth]{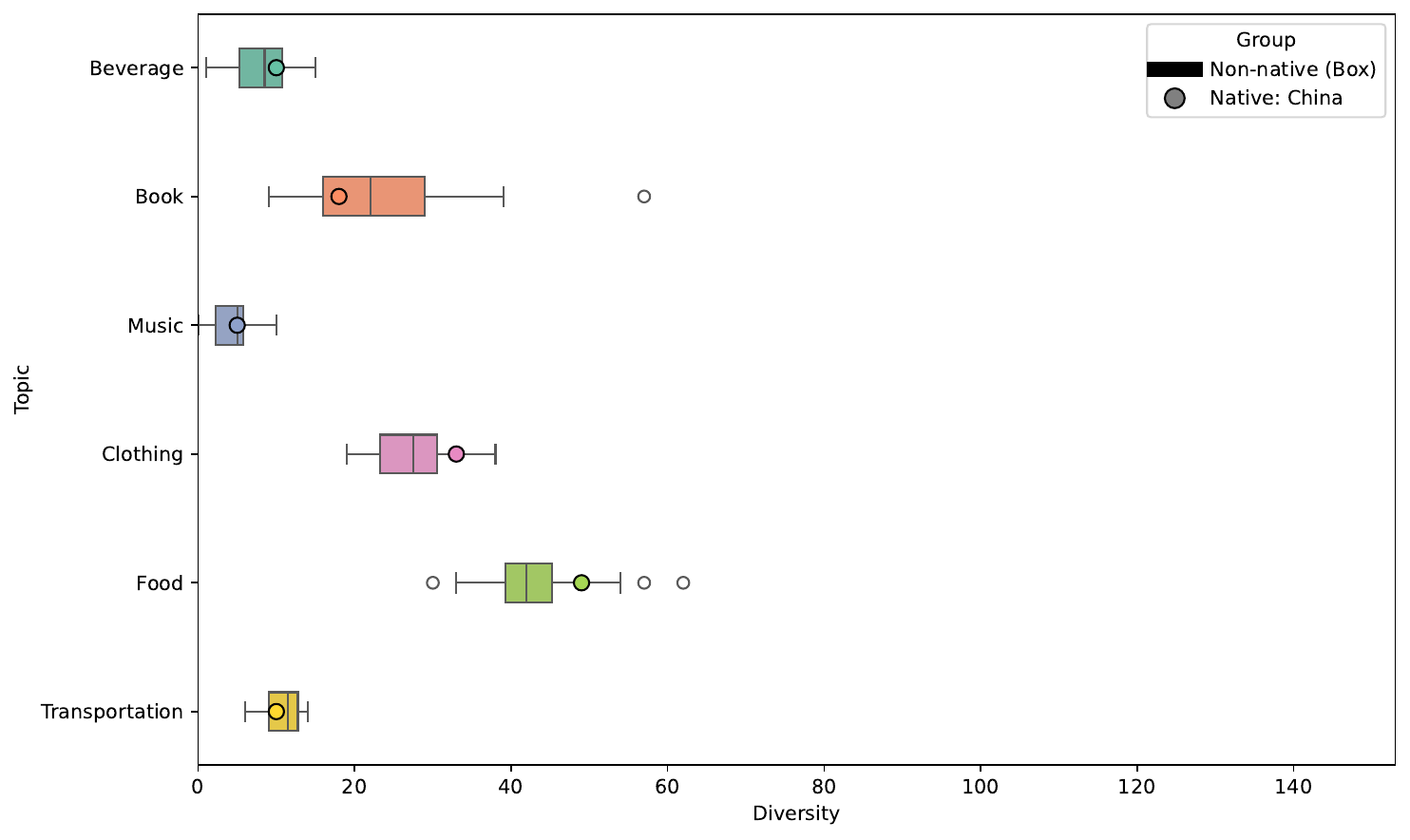} \\
Thai & Turkish & Chinese (Simplified) \\
\includegraphics[width=0.3\textwidth]{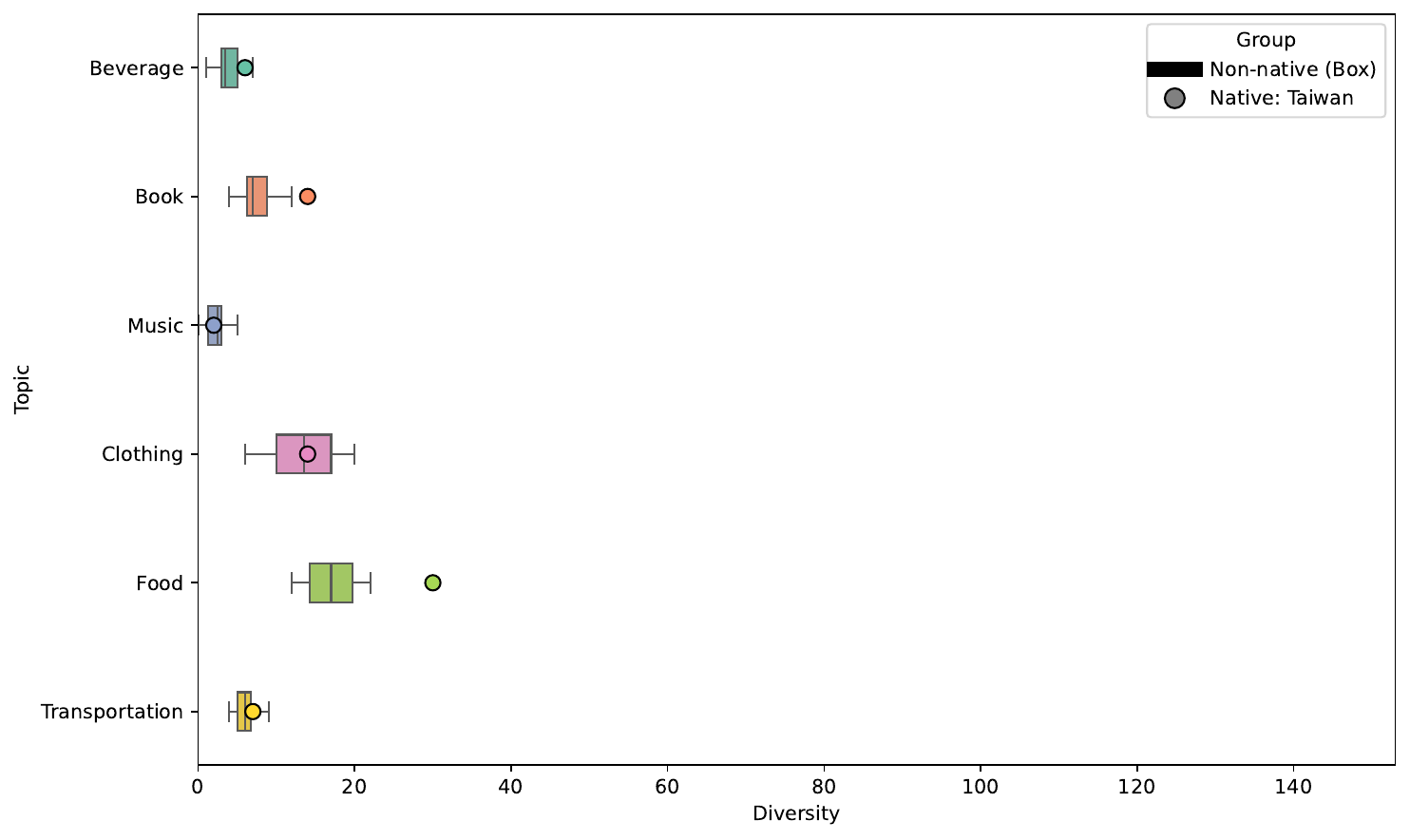} & & \\
Chinese (Traditional) & &
\end{tabular}
\caption{Box plots for diversity comparison between native and non-native languages for \textsc{ChatGPT} across 13 languages.}
\label{fig:box_chatgpt}
\end{figure*}

\begin{figure*}[t]
\centering
\begin{tabular}{ccc}
\includegraphics[width=0.3\textwidth]{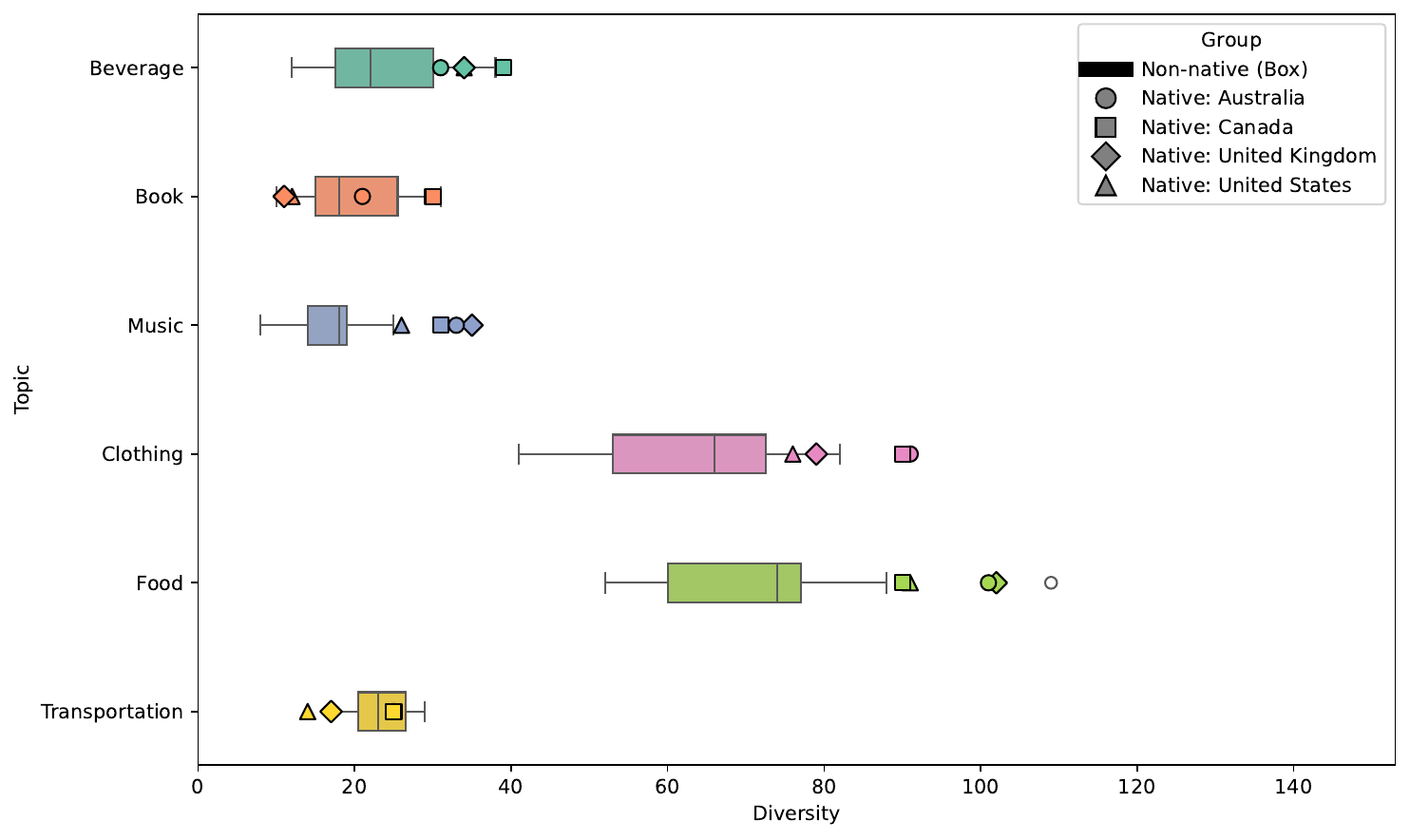} & 
\includegraphics[width=0.3\textwidth]{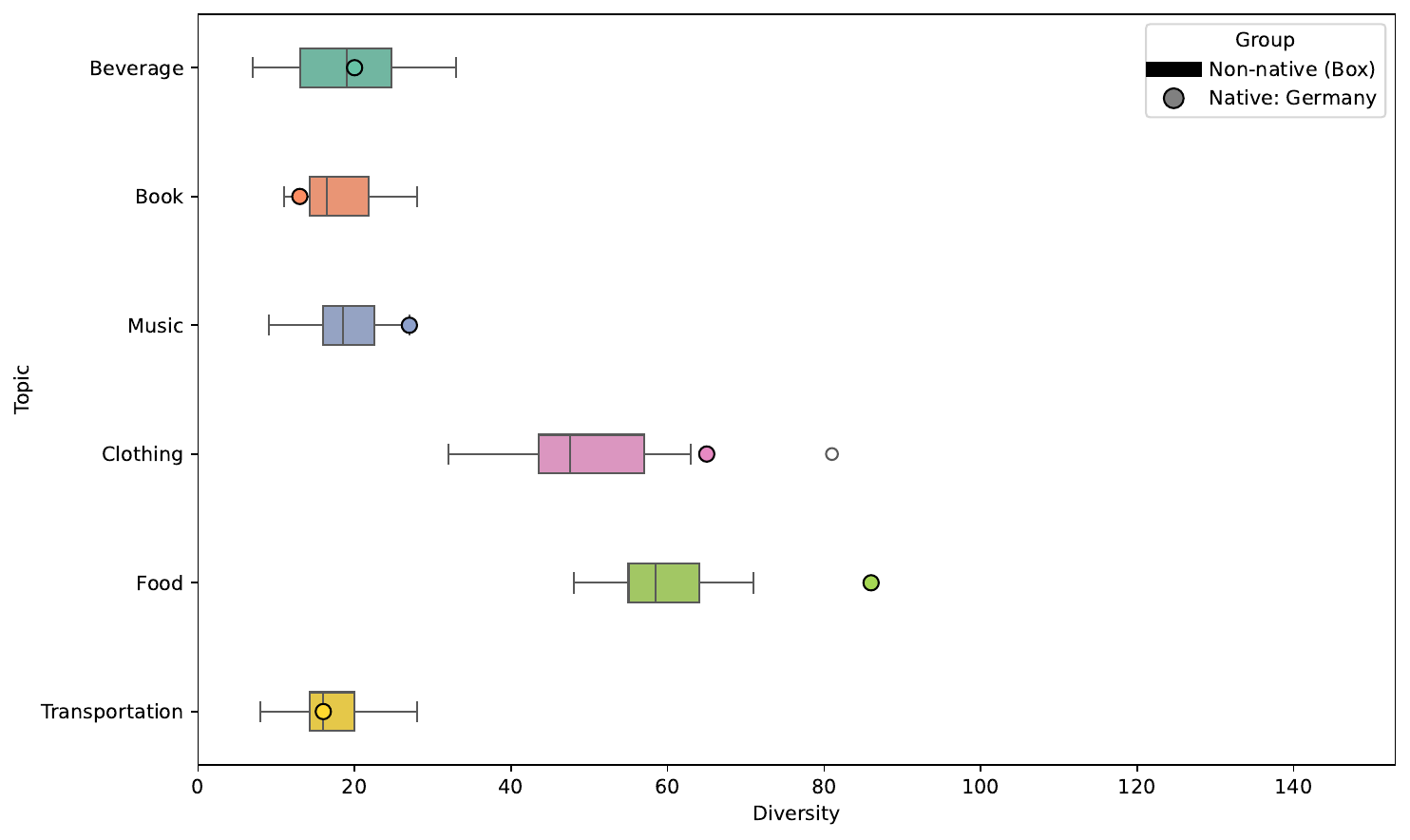} & 
\includegraphics[width=0.3\textwidth]{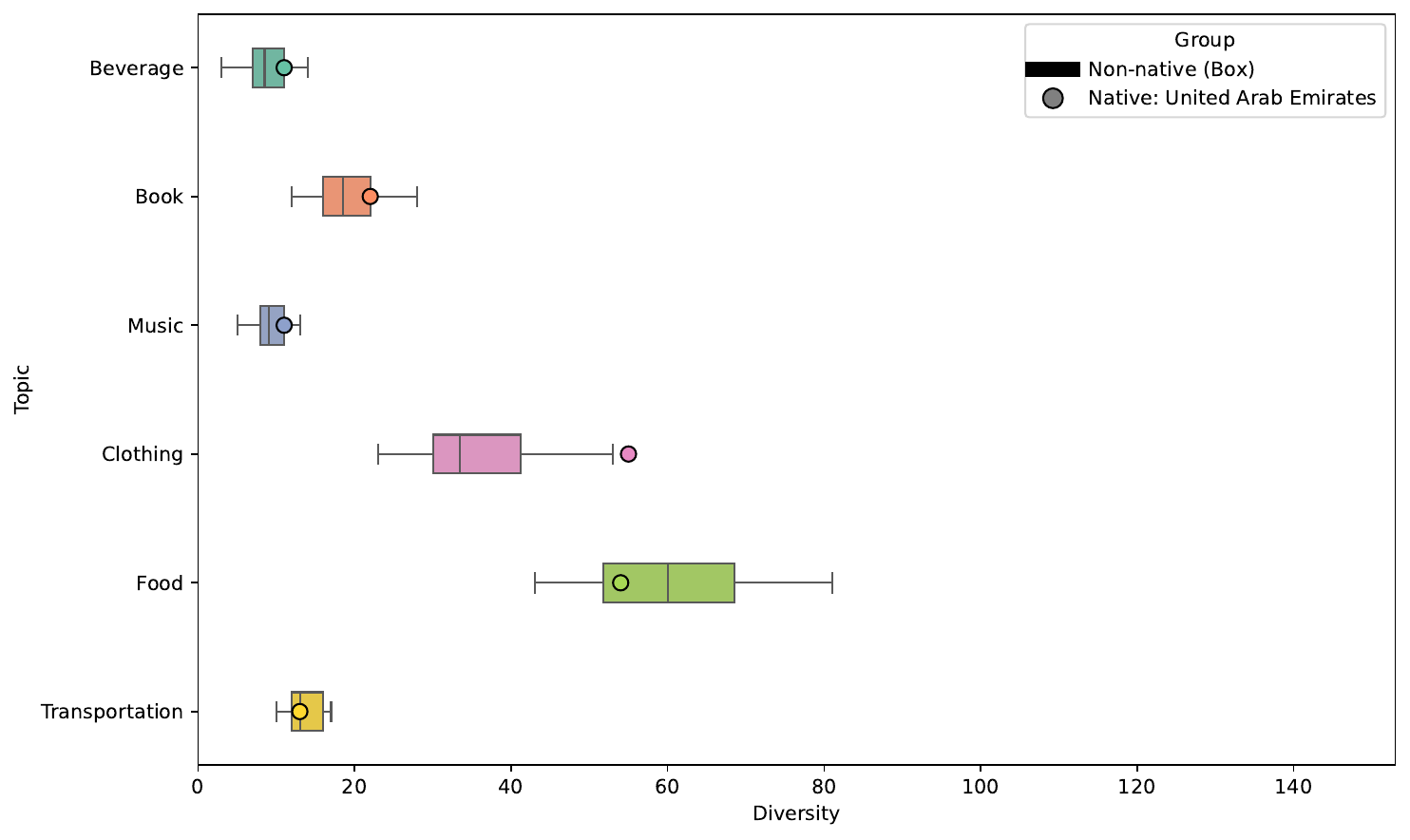} \\
English & German & Arabic \\
\includegraphics[width=0.3\textwidth]{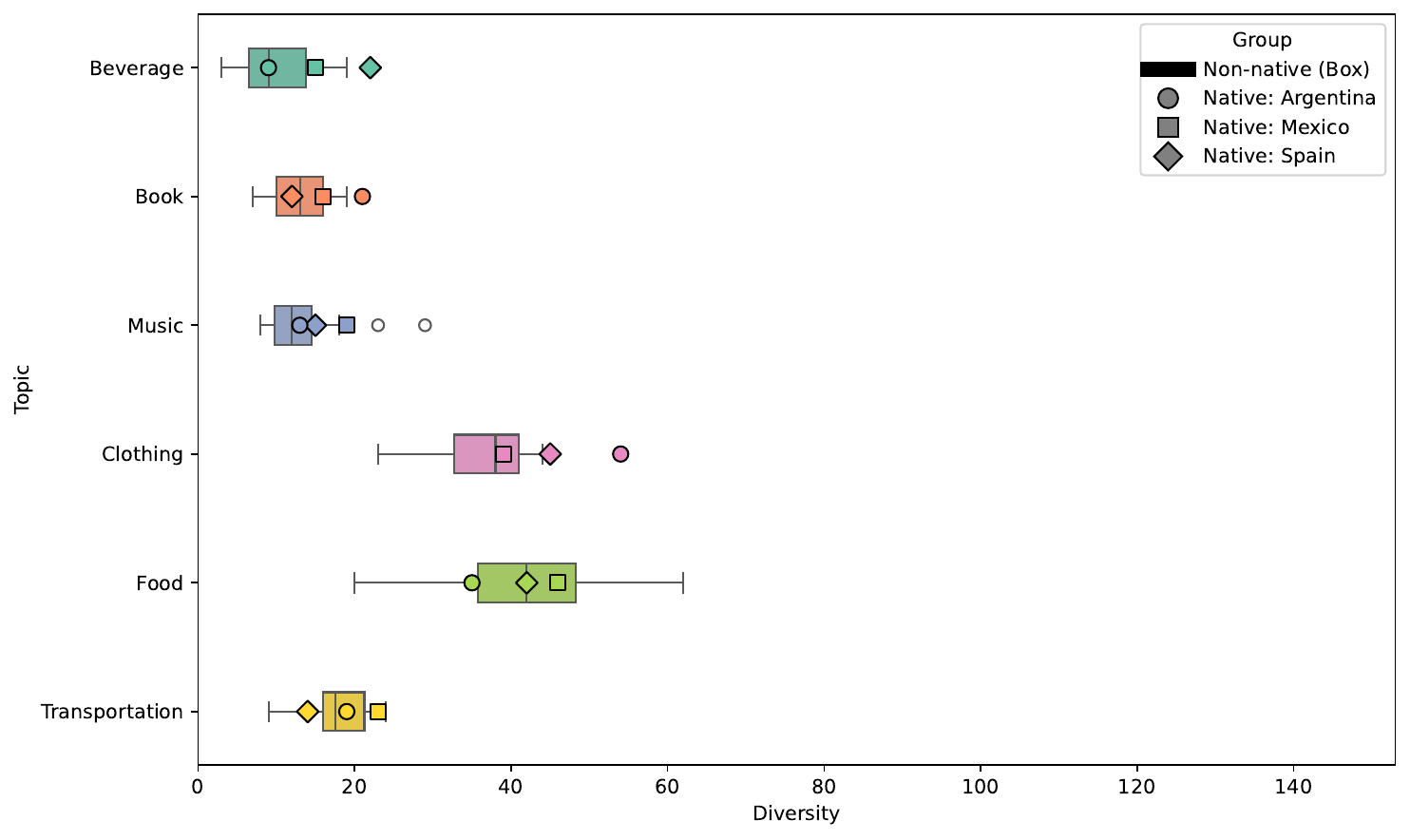} & 
\includegraphics[width=0.3\textwidth]{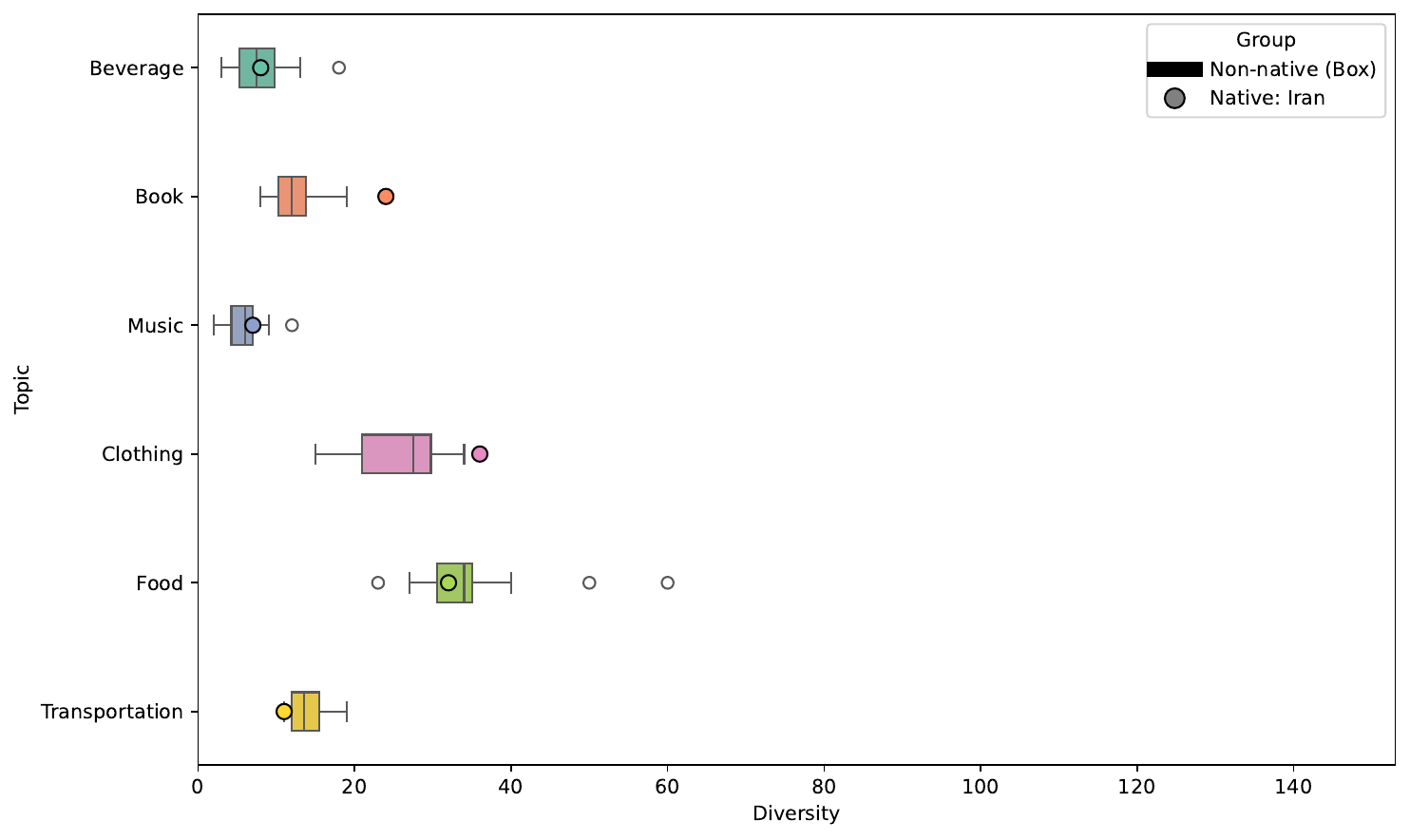} & 
\includegraphics[width=0.3\textwidth]{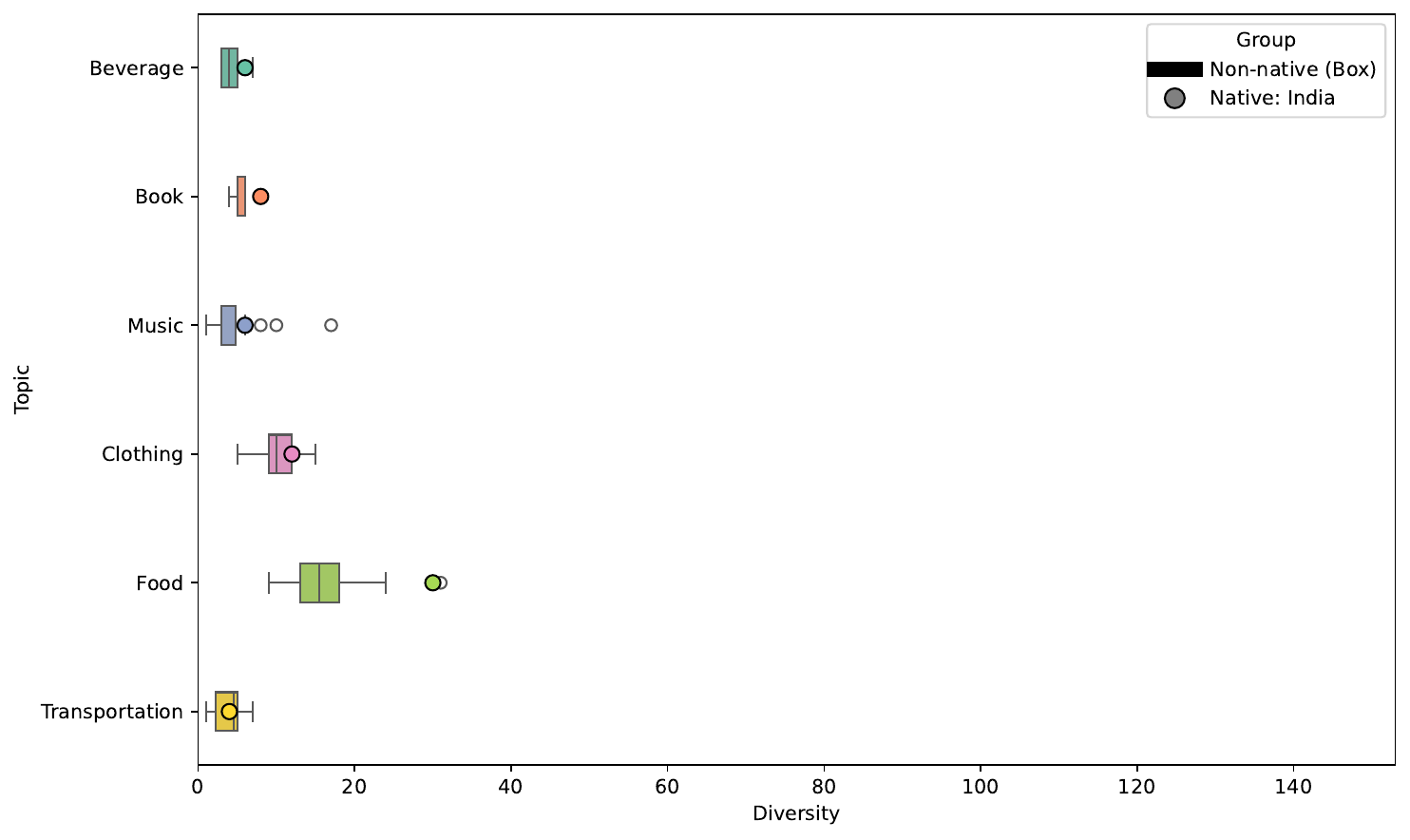} \\
Spanish & Persian & Hindi \\
\includegraphics[width=0.3\textwidth]{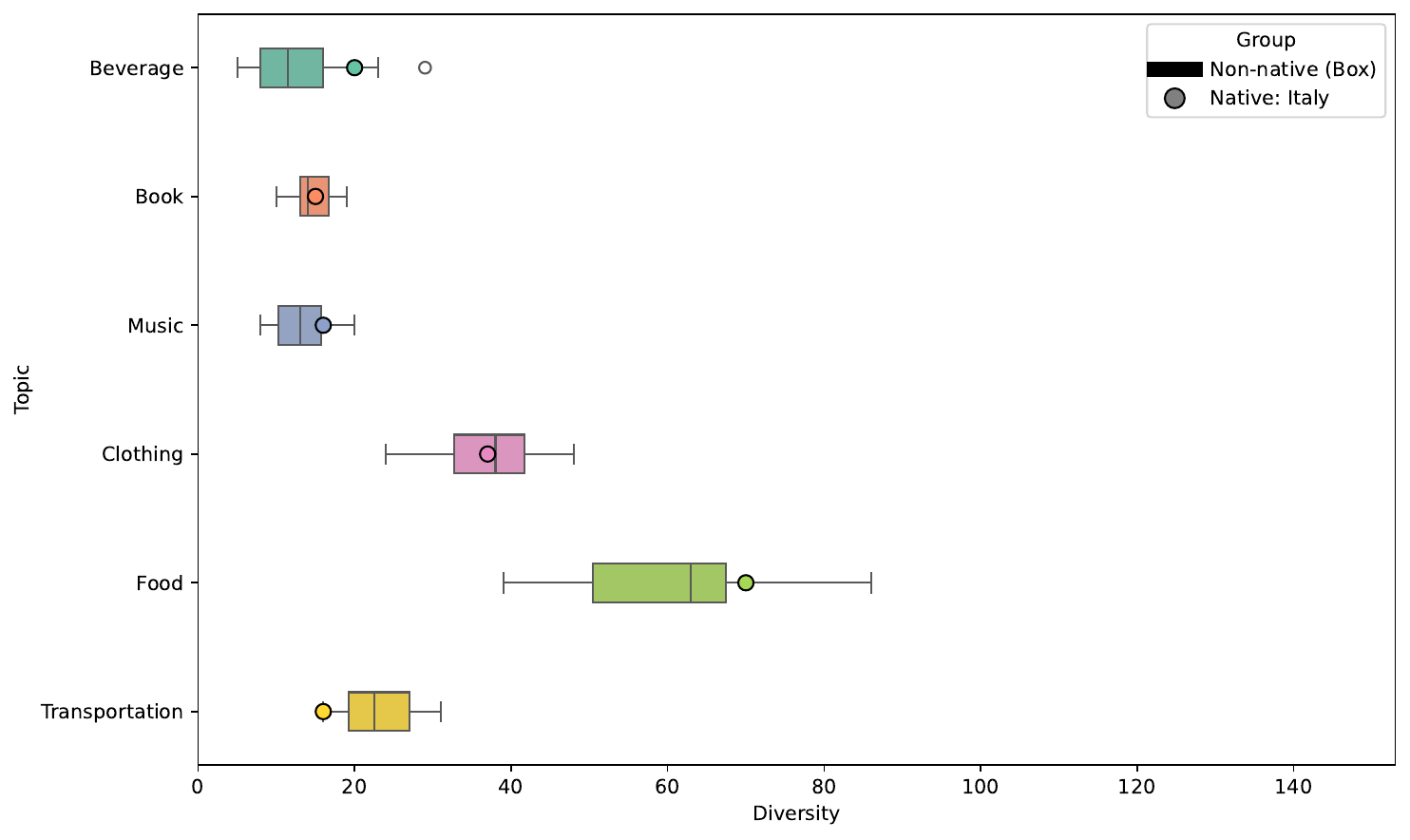} & 
\includegraphics[width=0.3\textwidth]{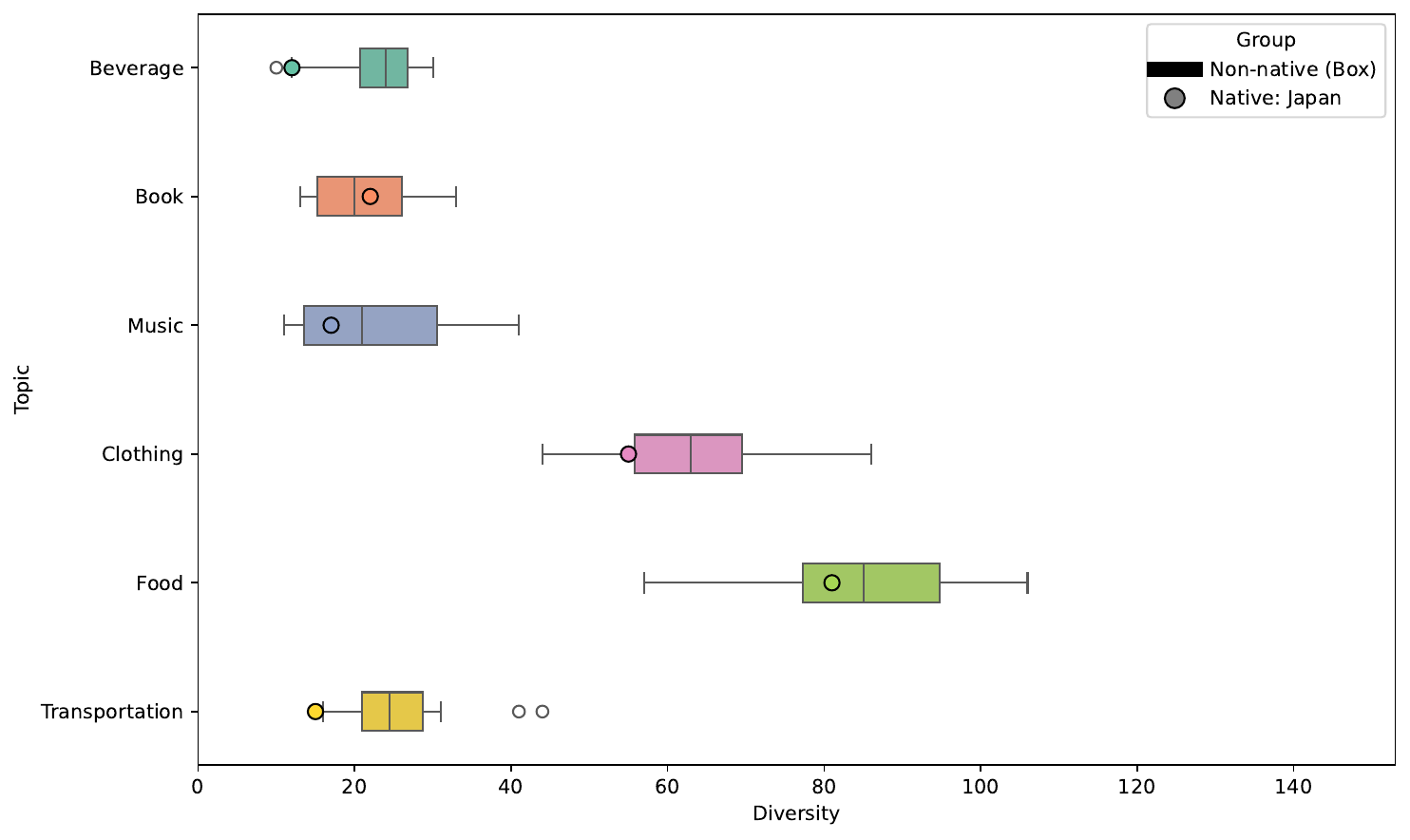} & 
\includegraphics[width=0.3\textwidth]{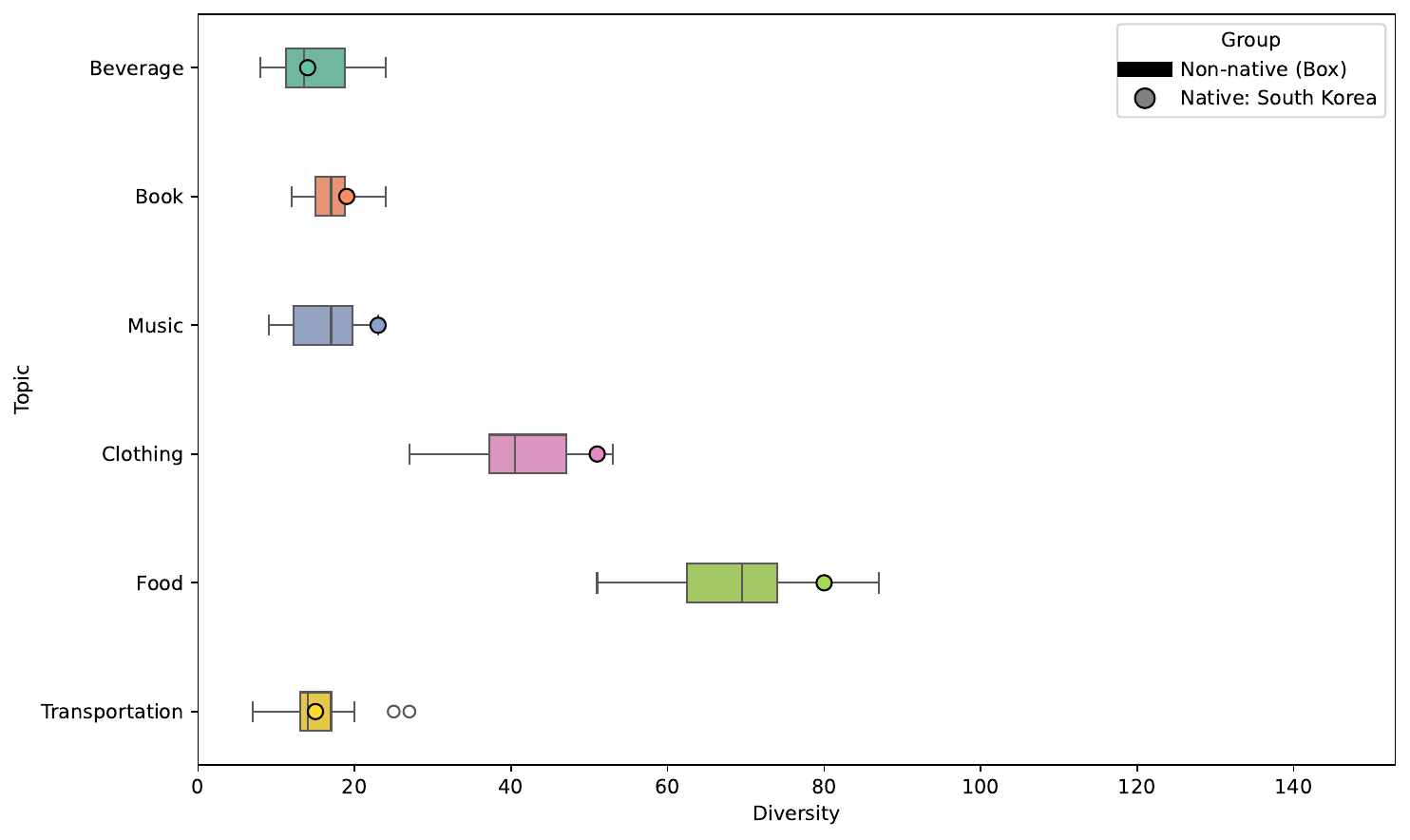} \\
Italian & Japanese & Korean \\
\includegraphics[width=0.3\textwidth]{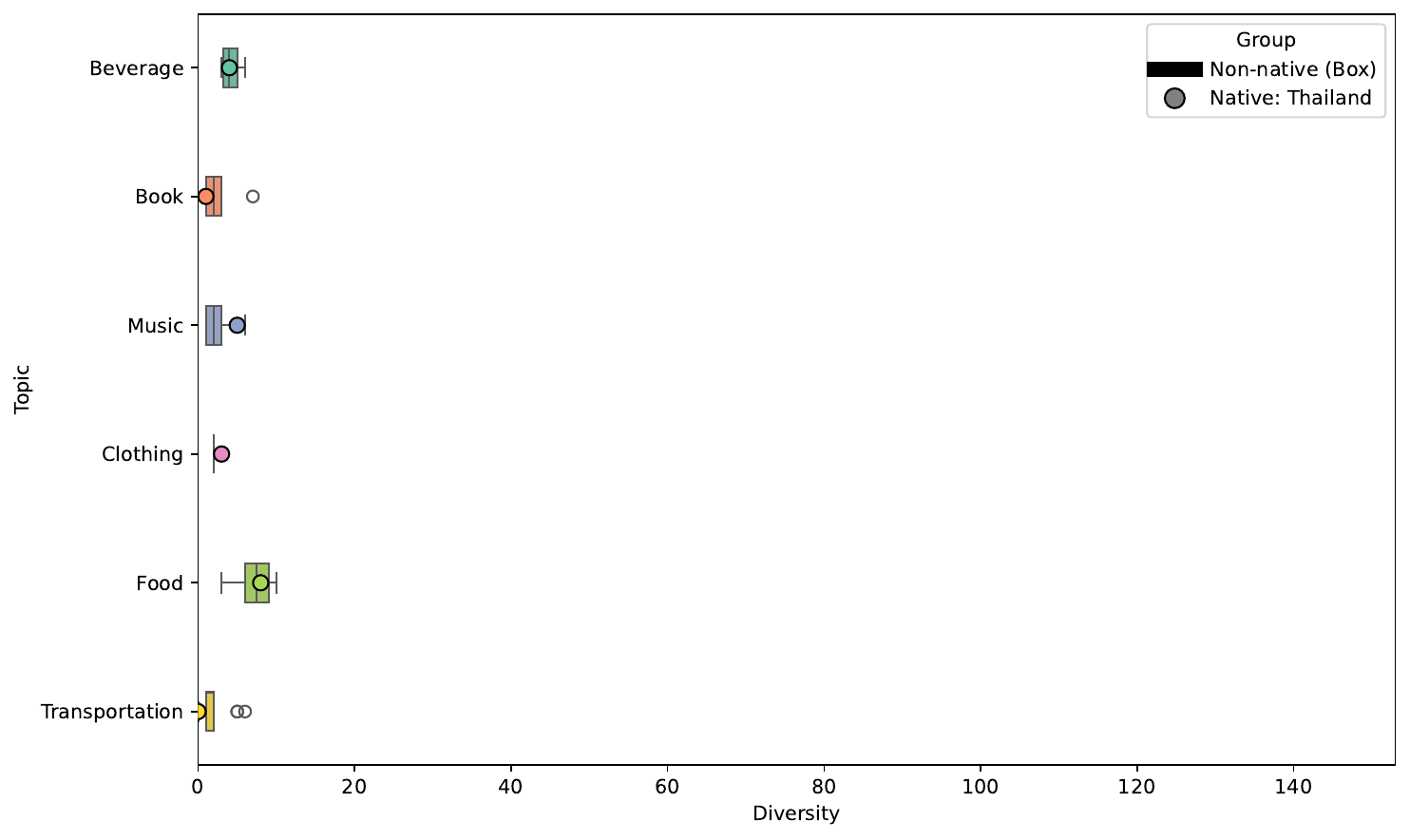} & 
\includegraphics[width=0.3\textwidth]{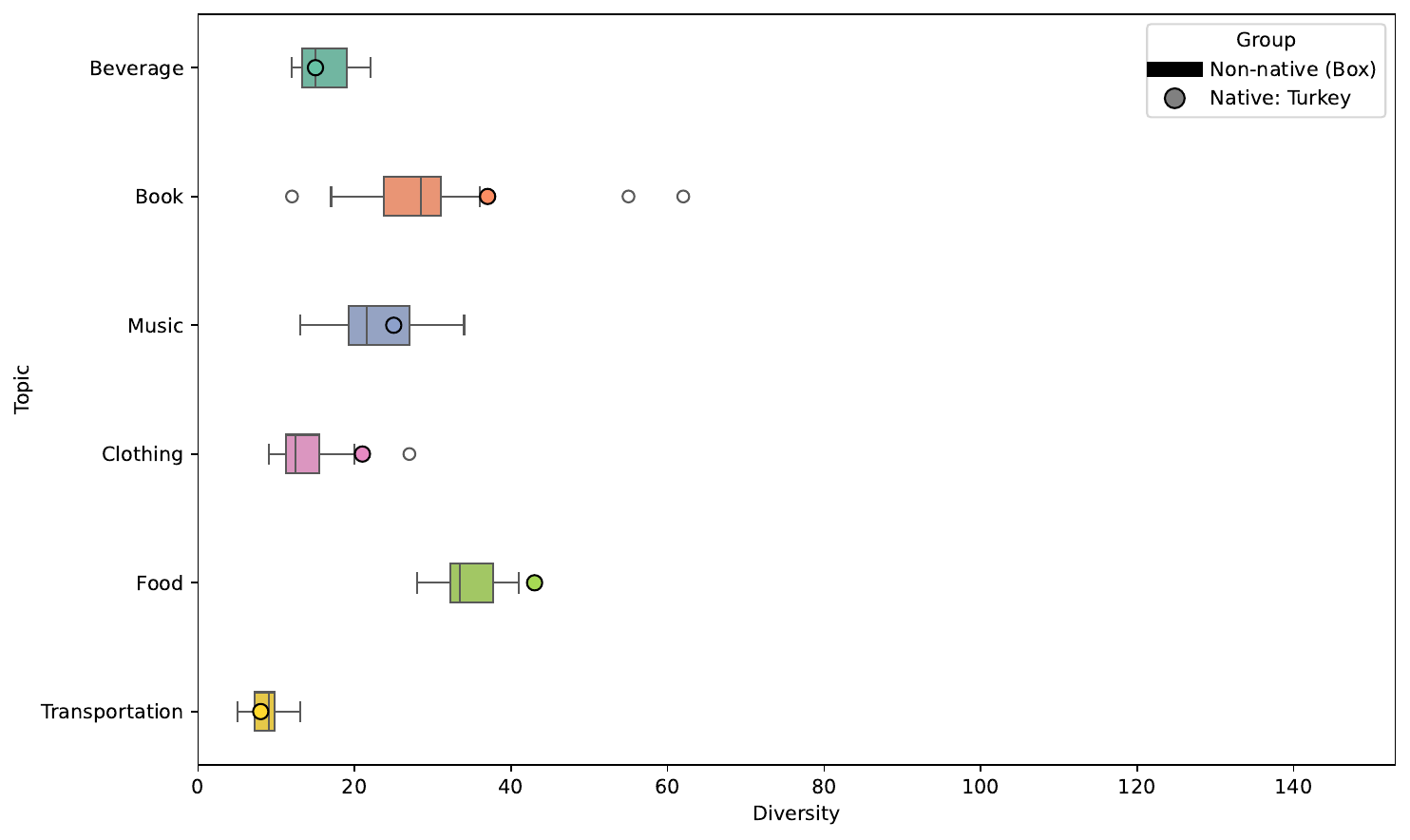} & 
\includegraphics[width=0.3\textwidth]{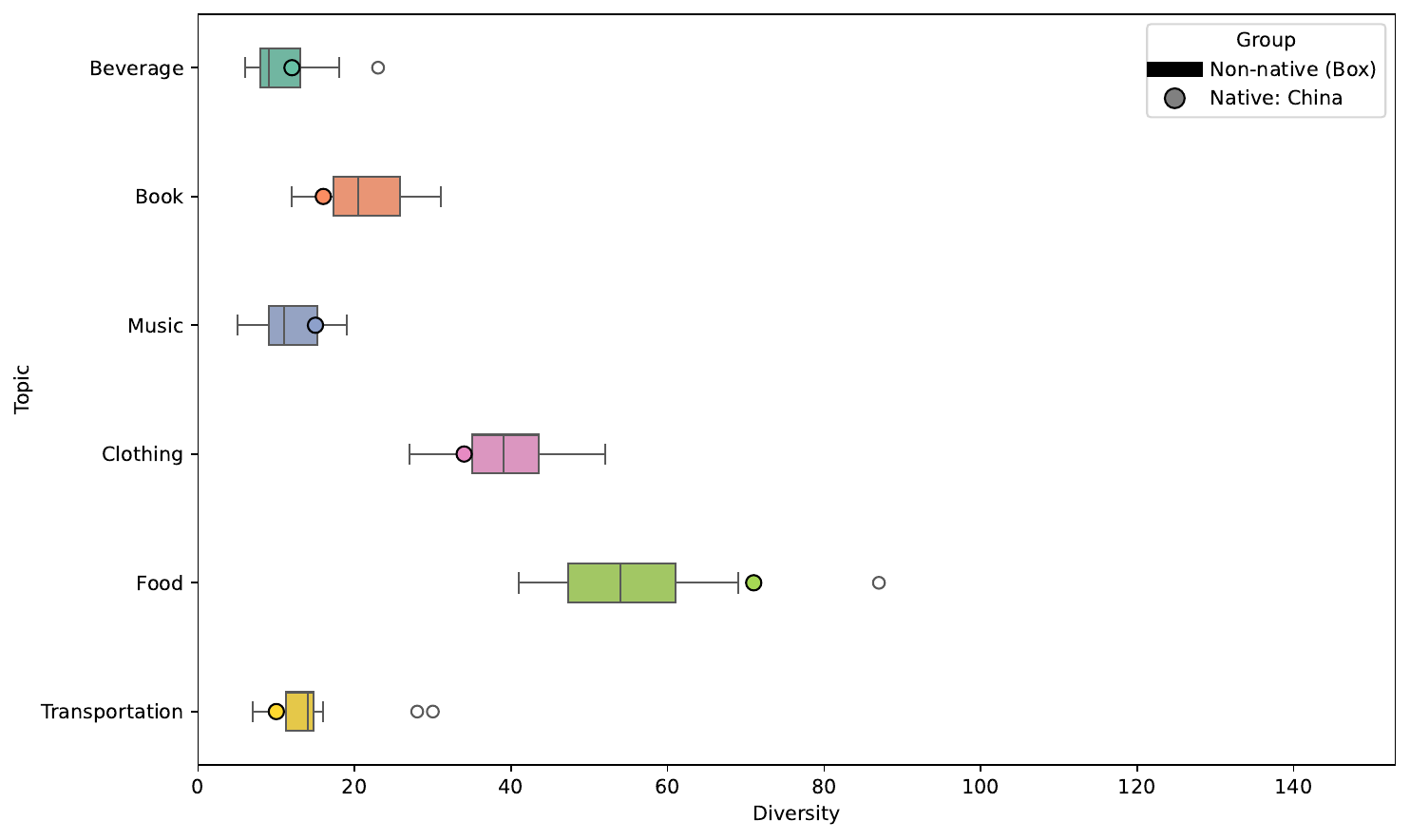} \\
Thai & Turkish & Chinese (Simplified) \\
\includegraphics[width=0.3\textwidth]{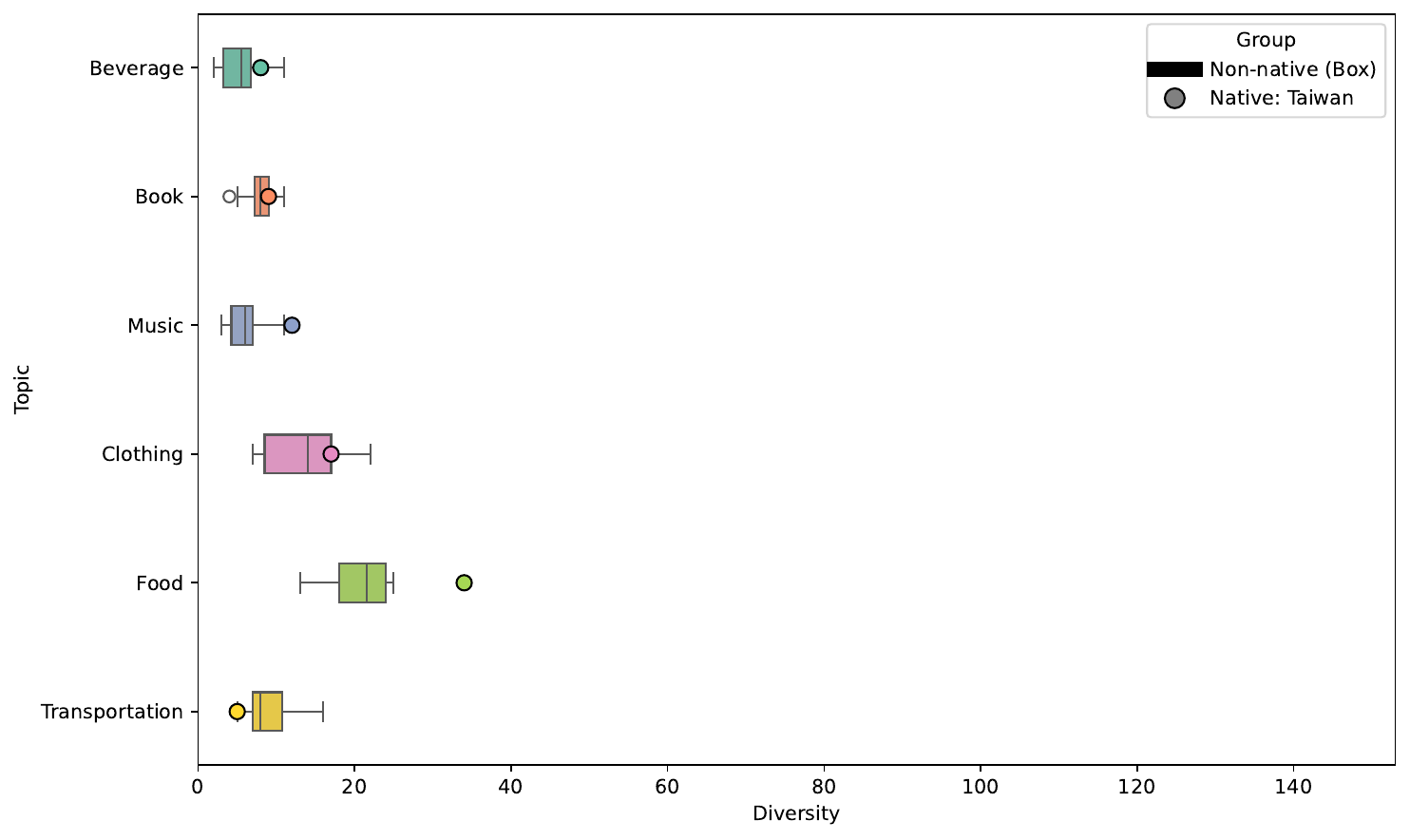} & & \\
Chinese (Traditional) & &
\end{tabular}
\caption{Box plots for diversity comparison between native and non-native languages for \textsc{Aya} across 13 languages.}
\label{fig:box_aya}
\end{figure*}

\begin{figure*}[t]
\centering
\begin{tabular}{ccc}
\includegraphics[width=0.3\textwidth]{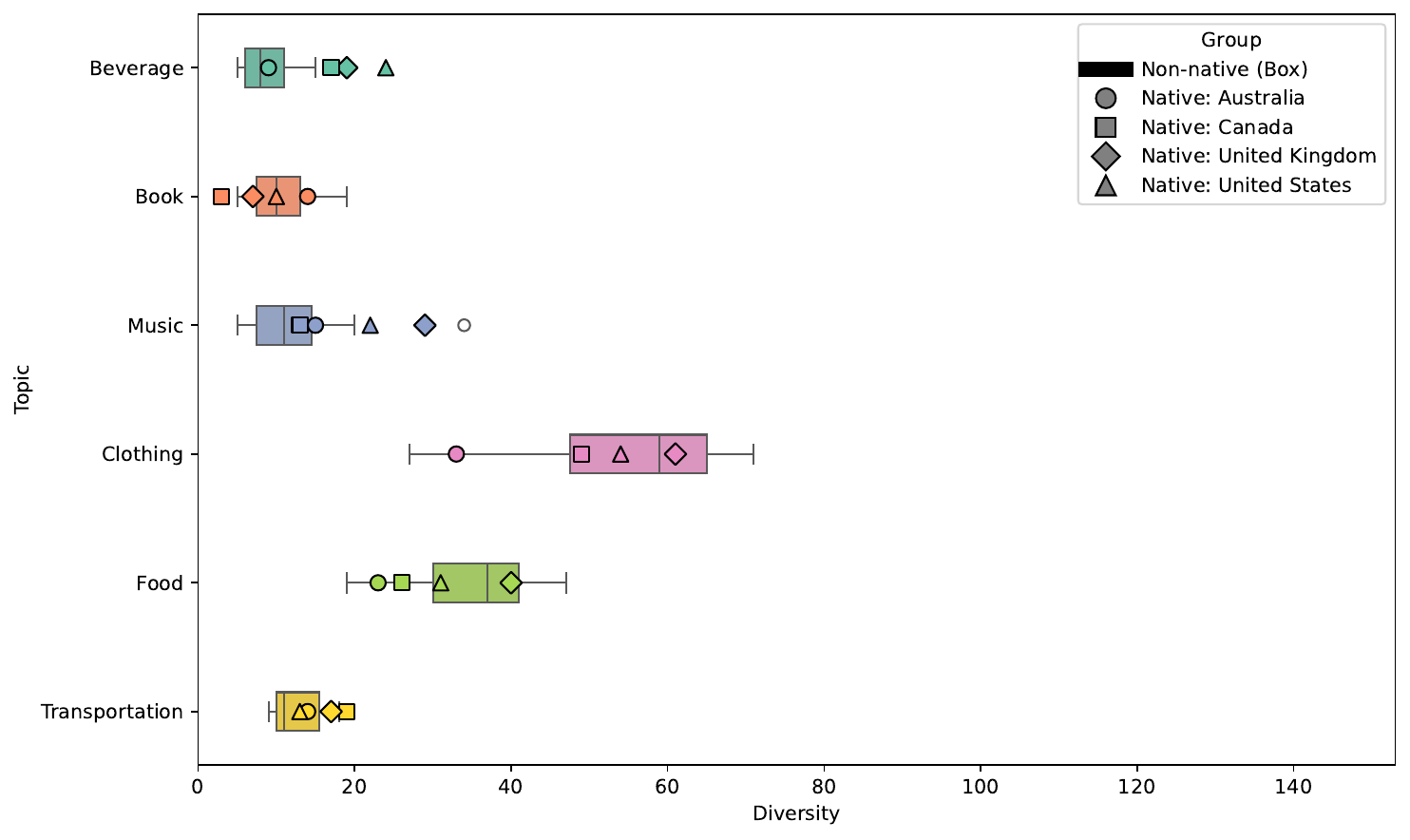} & 
\includegraphics[width=0.3\textwidth]{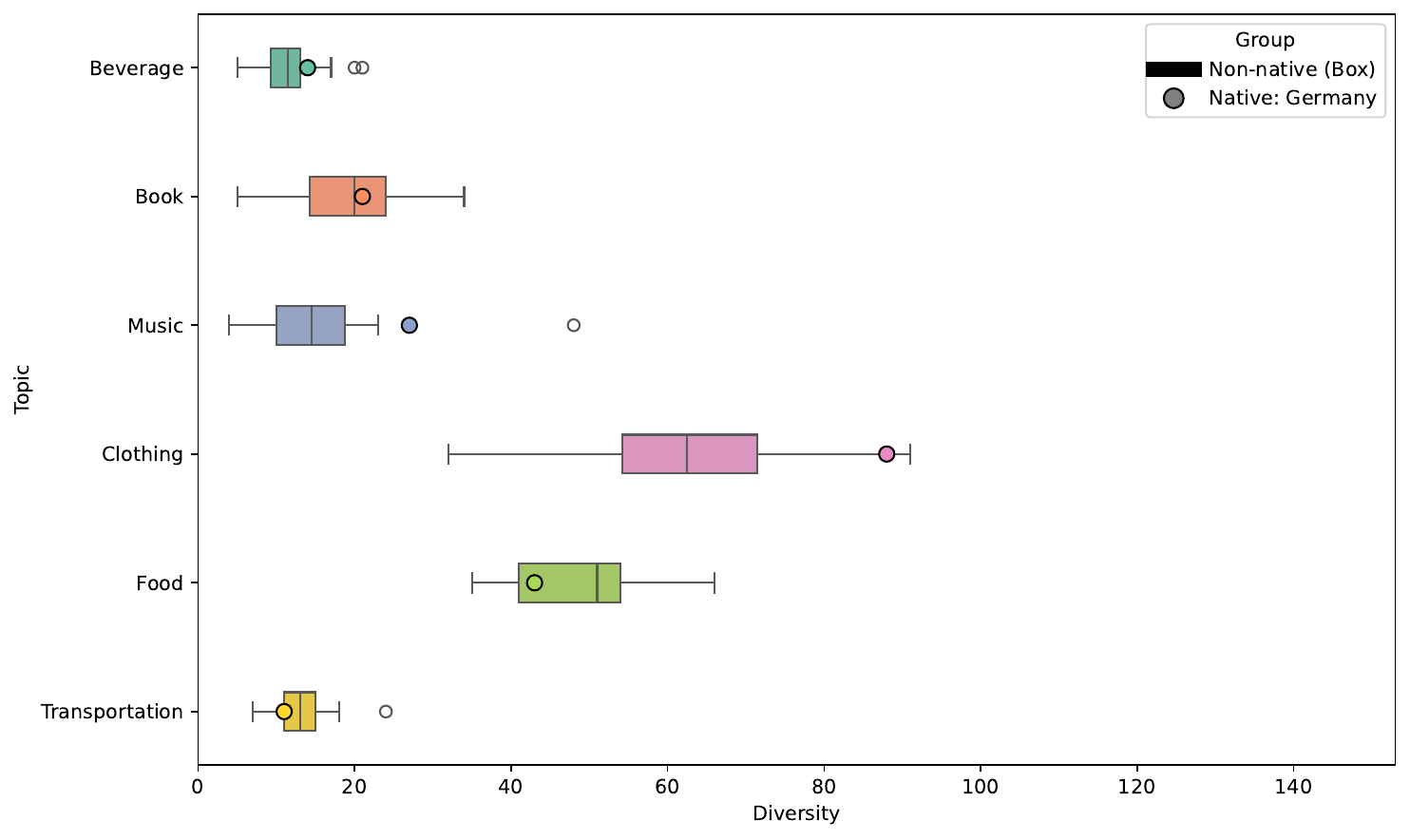} & 
\includegraphics[width=0.3\textwidth]{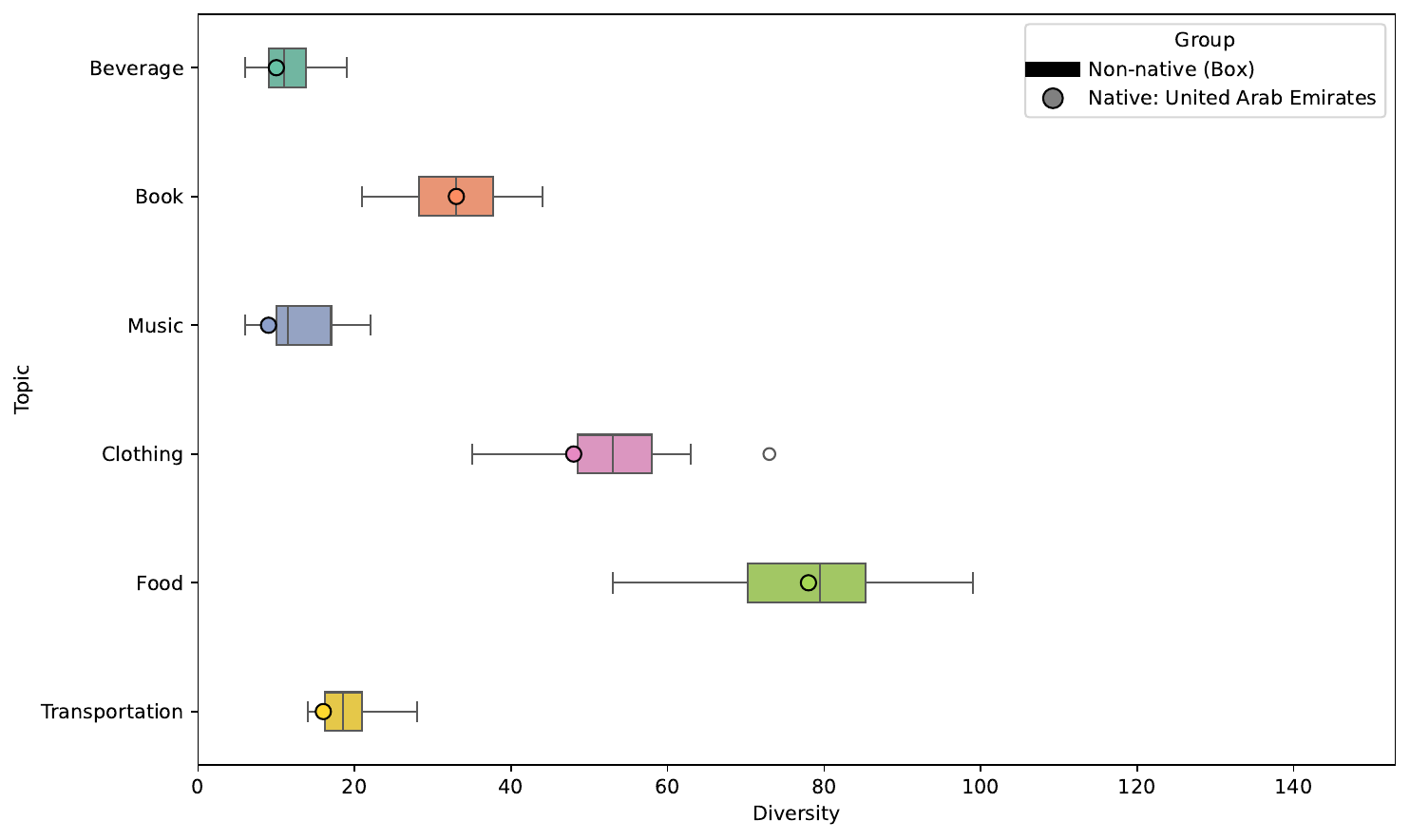} \\
English & German & Arabic \\
\includegraphics[width=0.3\textwidth]{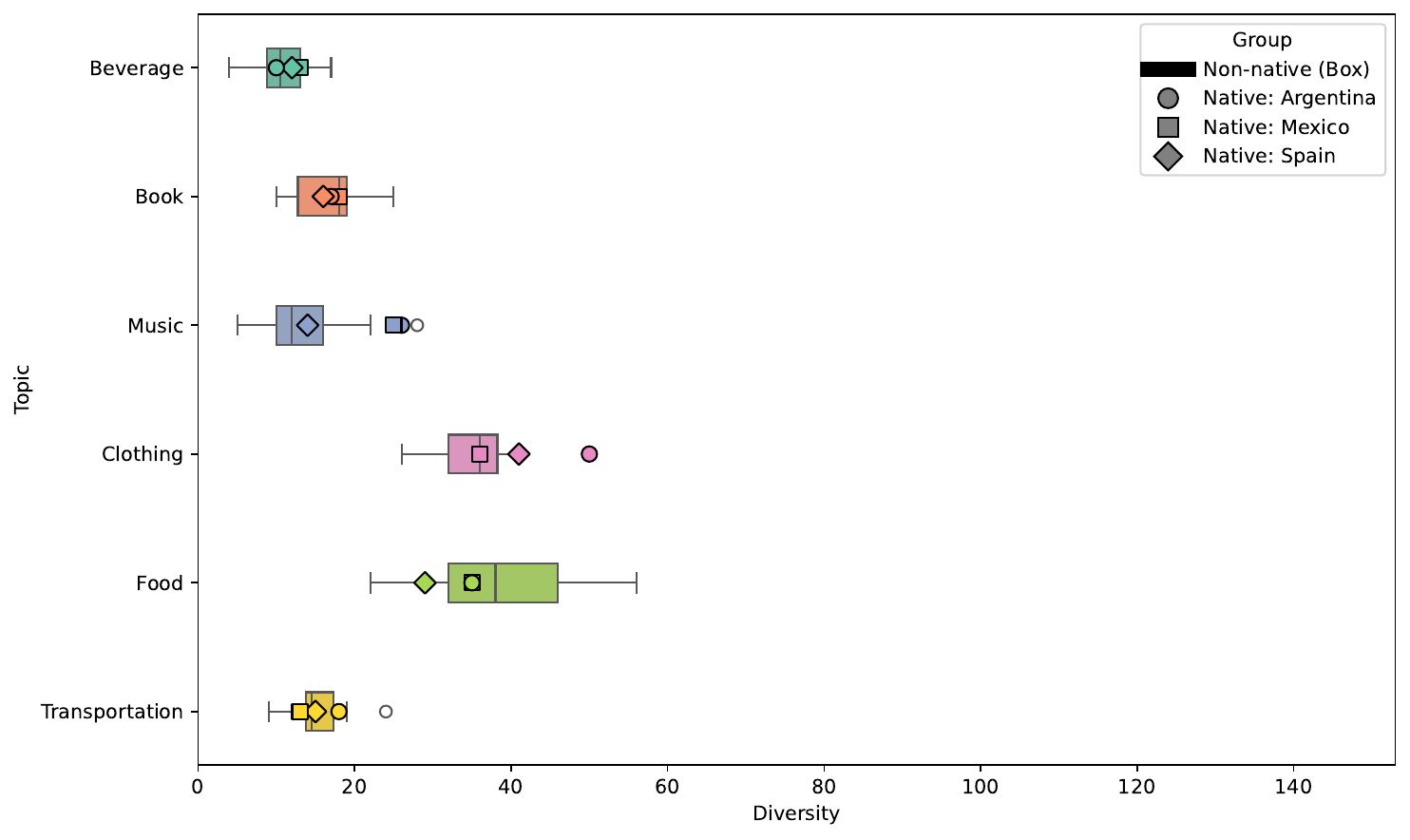} & 
\includegraphics[width=0.3\textwidth]{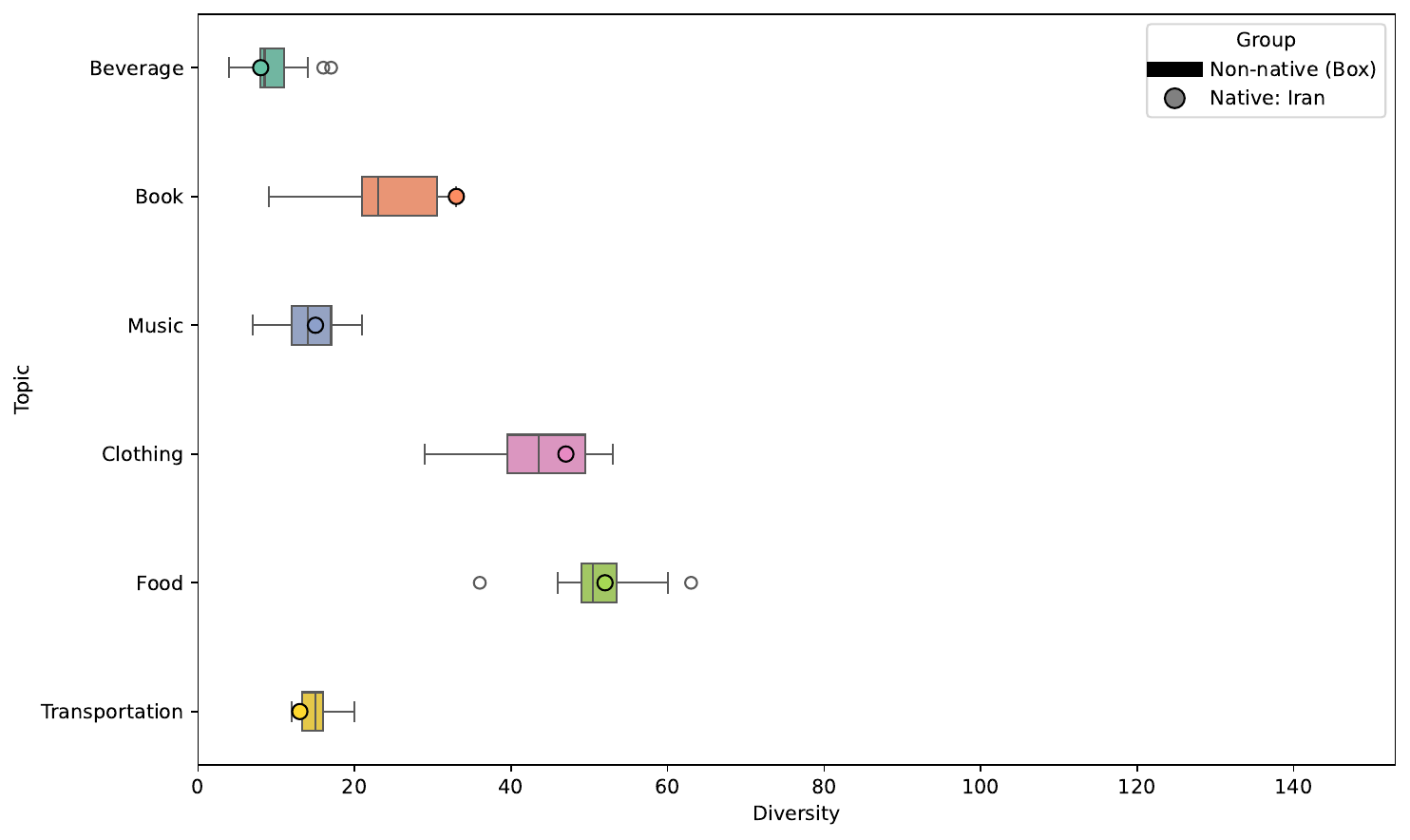} & 
\includegraphics[width=0.3\textwidth]{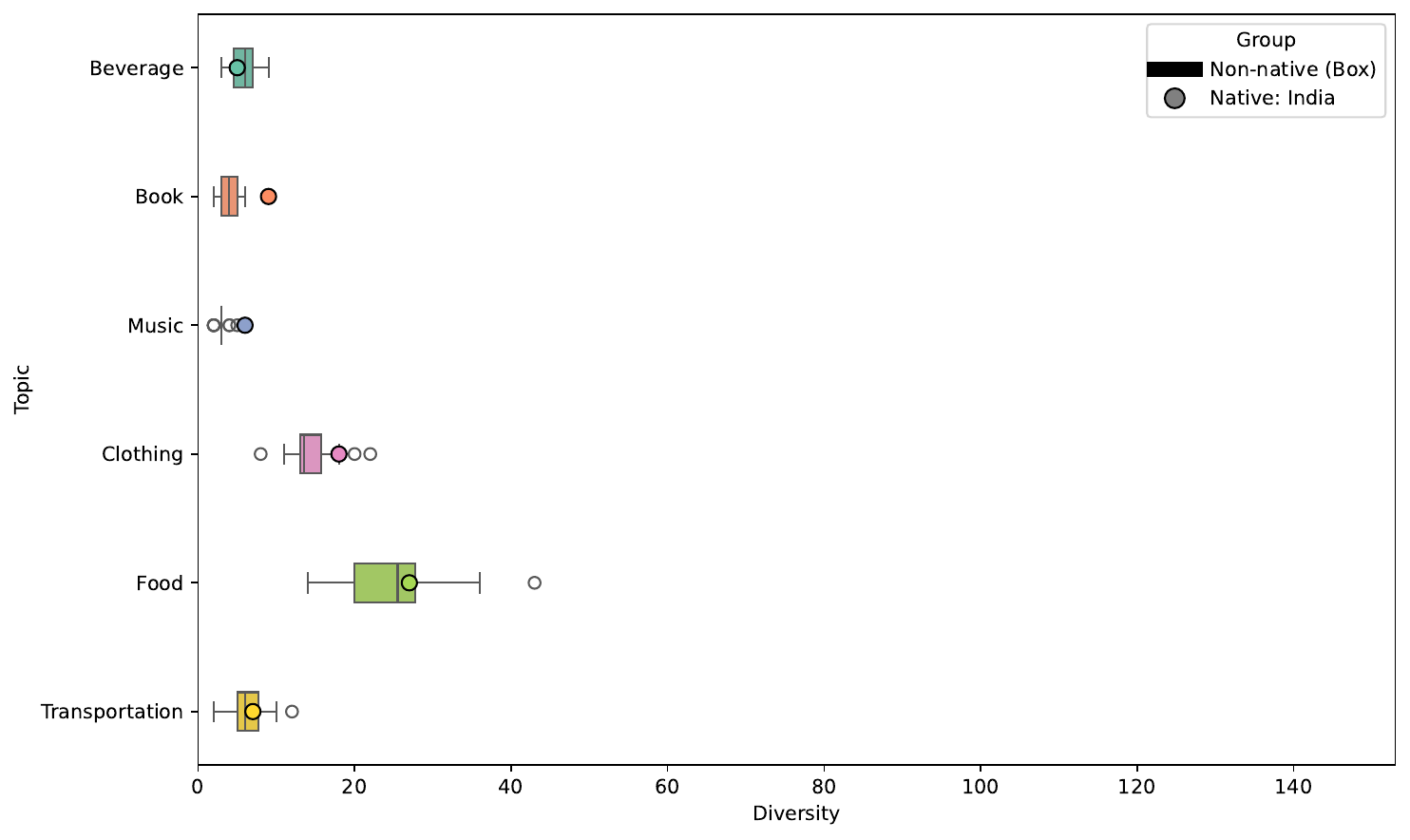} \\
Spanish & Persian & Hindi \\
\includegraphics[width=0.3\textwidth]{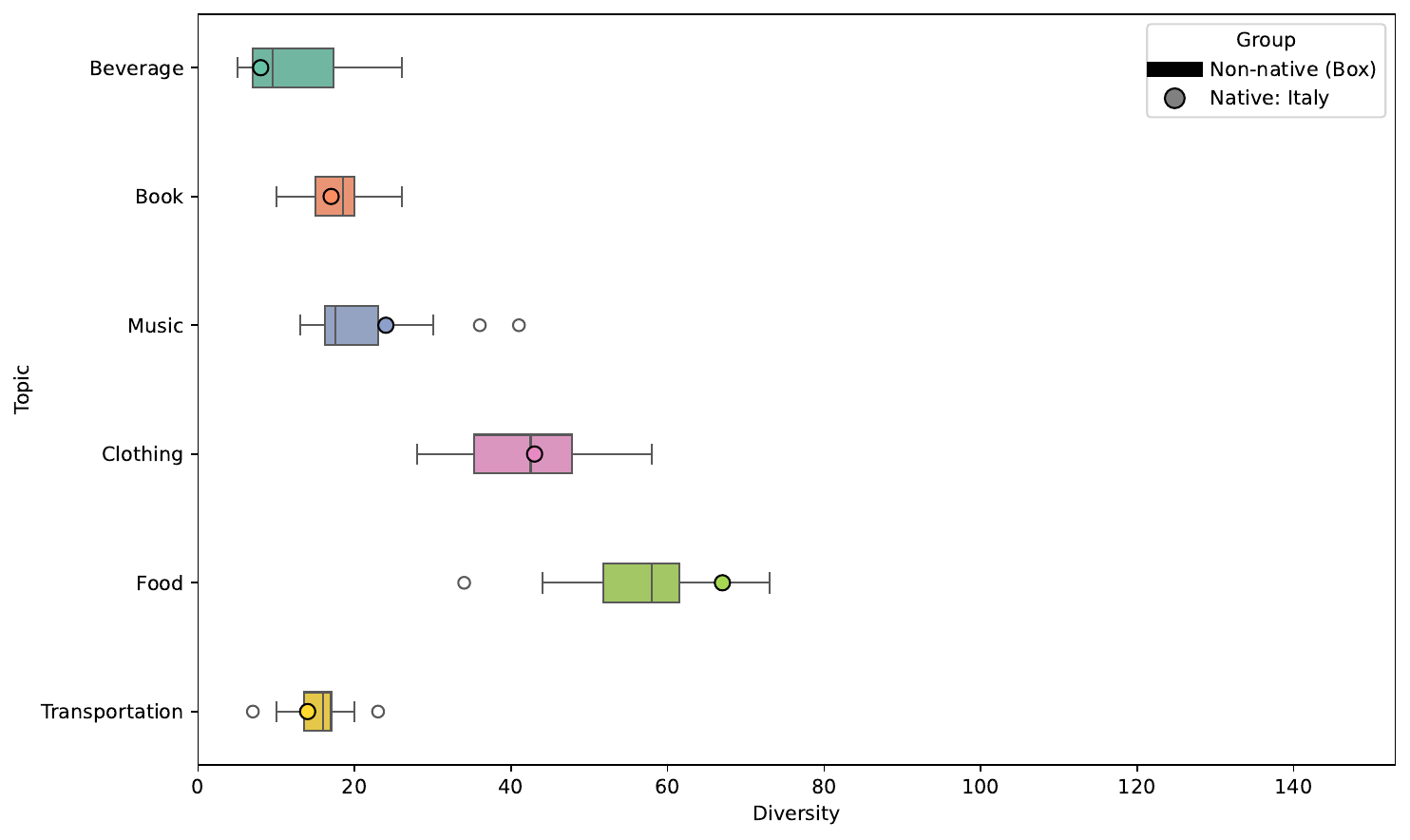} & 
\includegraphics[width=0.3\textwidth]{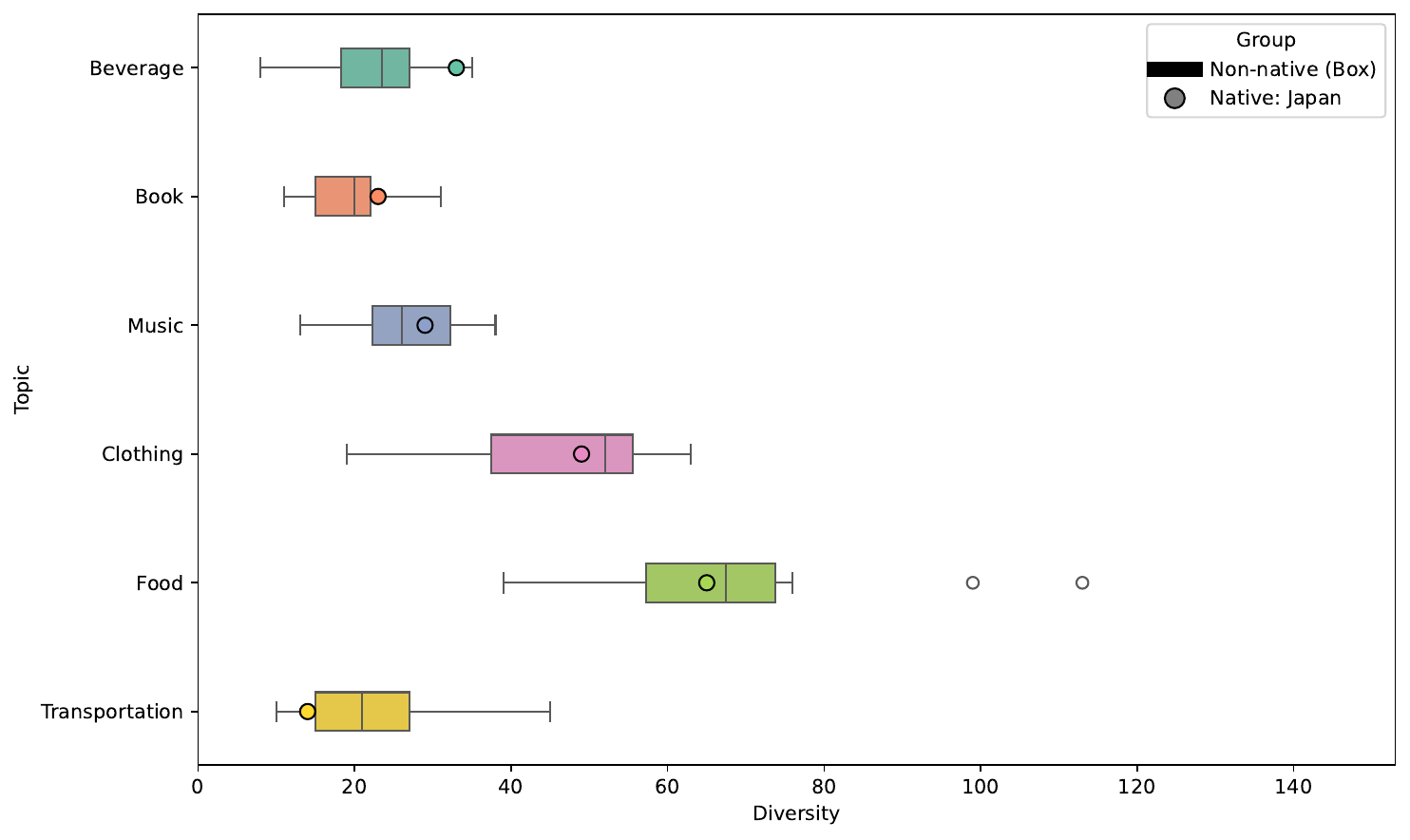} & 
\includegraphics[width=0.3\textwidth]{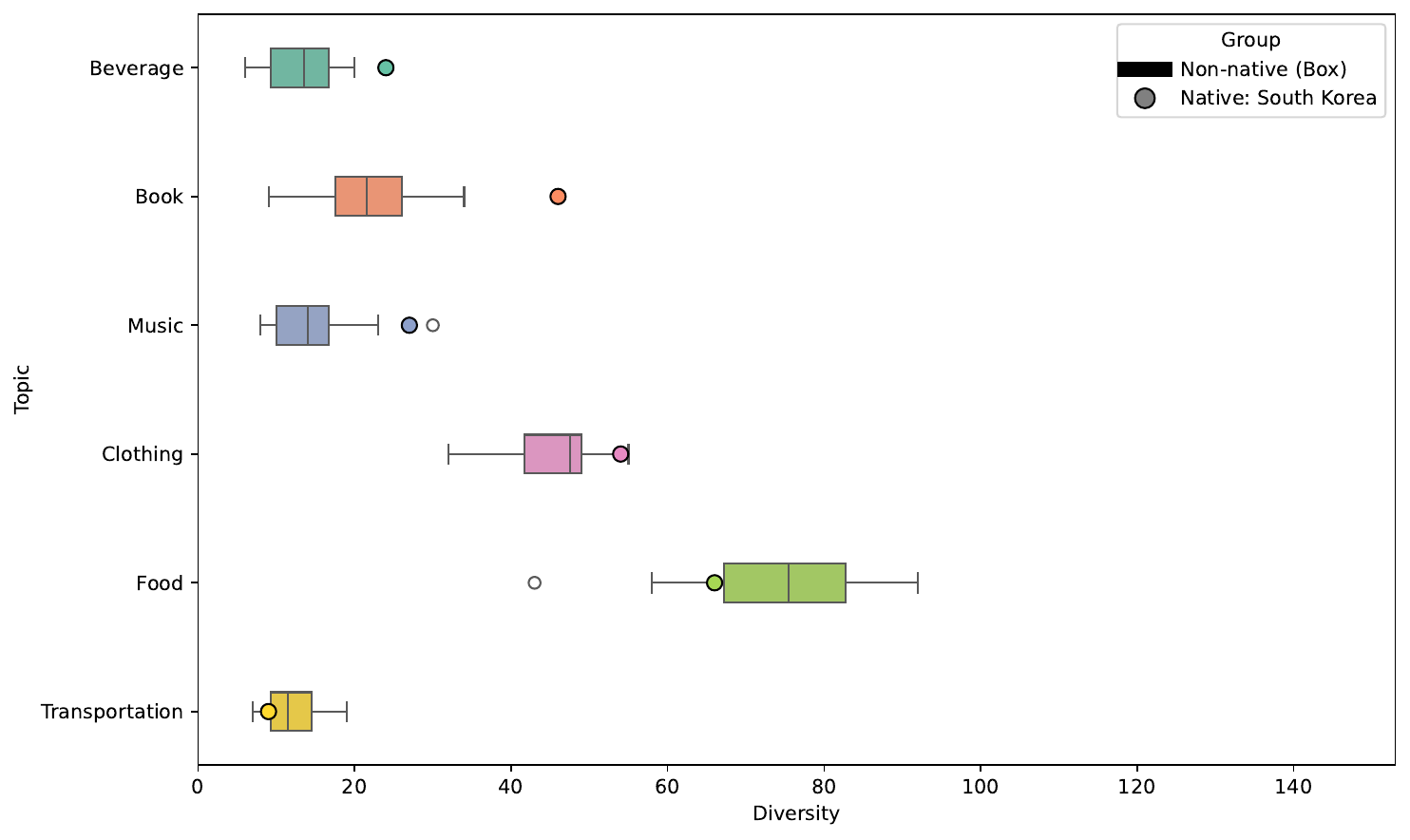} \\
Italian & Japanese & Korean \\
\includegraphics[width=0.3\textwidth]{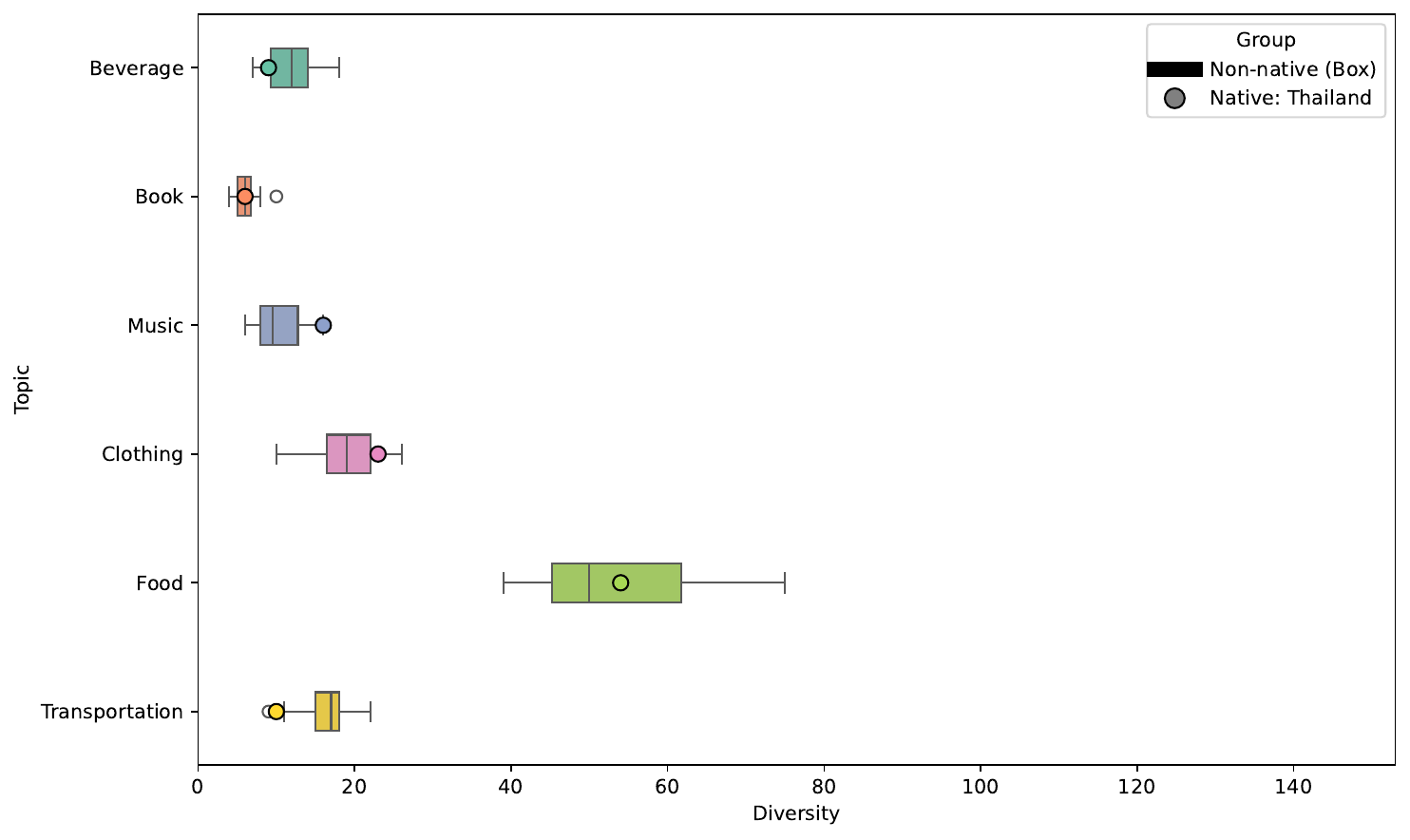} & 
\includegraphics[width=0.3\textwidth]{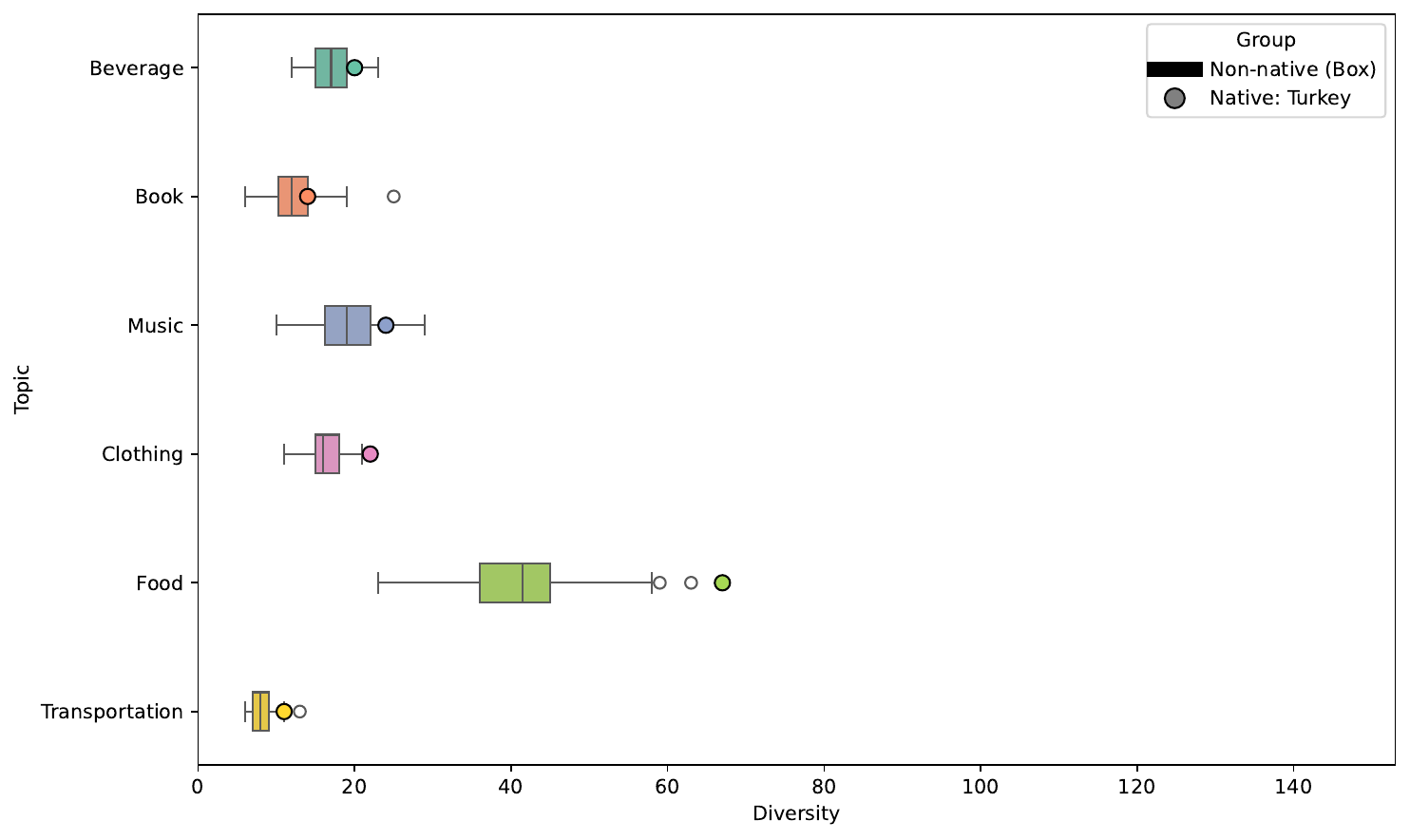} & 
\includegraphics[width=0.3\textwidth]{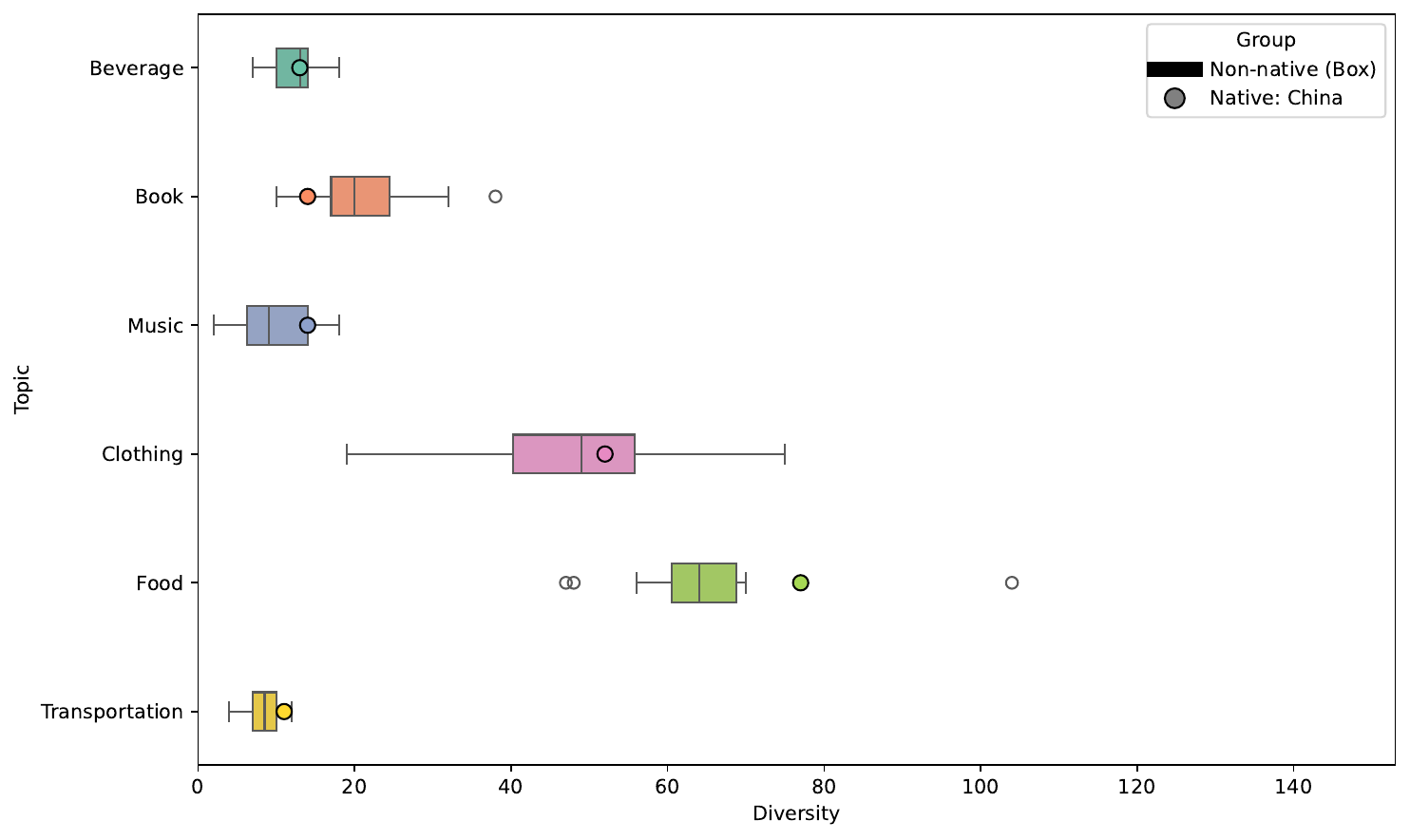} \\
Thai & Turkish & Chinese (Simplified) \\
\includegraphics[width=0.3\textwidth]{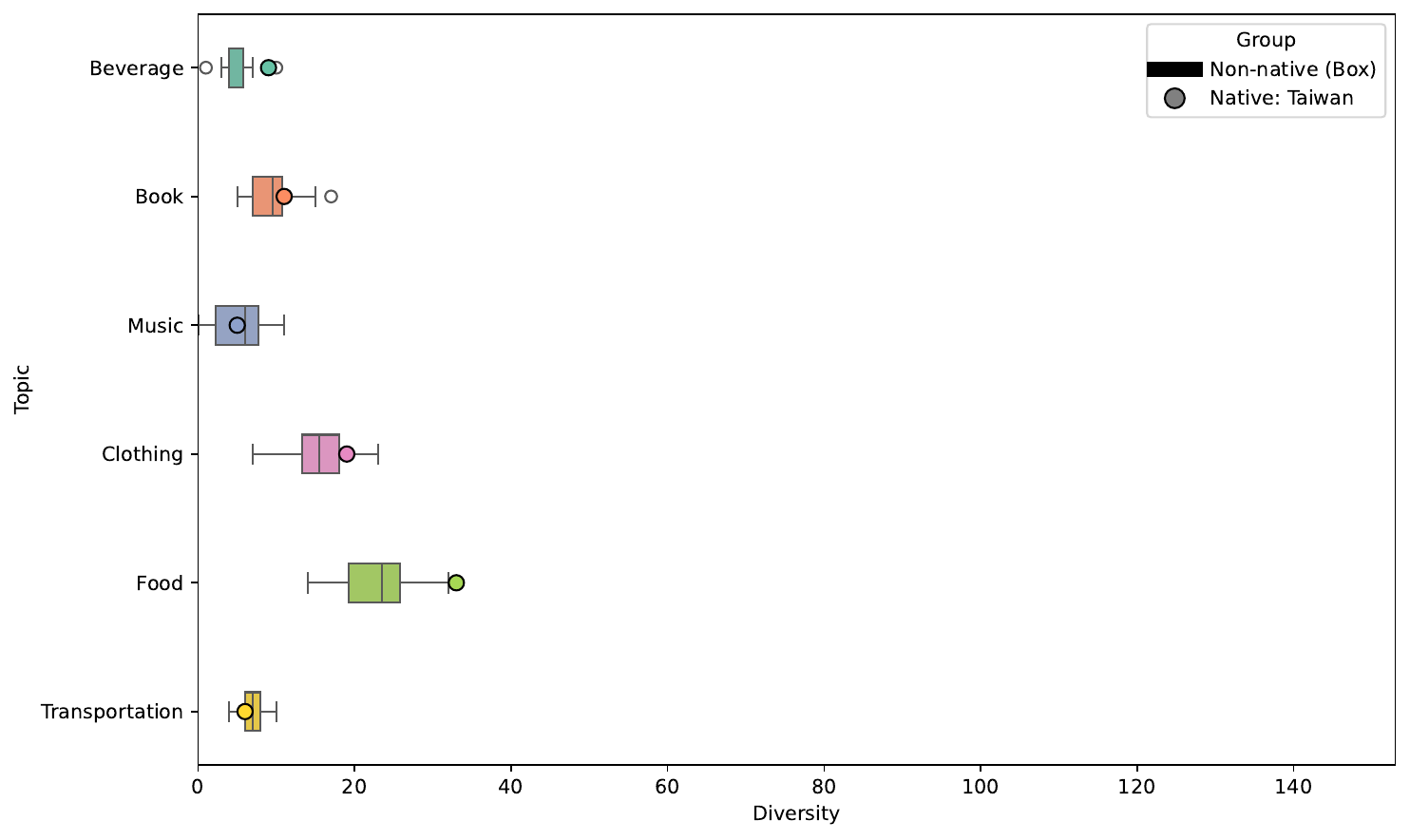} & & \\
Chinese (Traditional) & &
\end{tabular}
\caption{Box plots for diversity comparison between native and non-native languages for \textsc{Llama3-70B} across 13 languages.}
\label{fig:box_llama3_70b}
\end{figure*}

\begin{figure*}[t]
\centering
\begin{tabular}{ccc}
\includegraphics[width=0.3\textwidth]{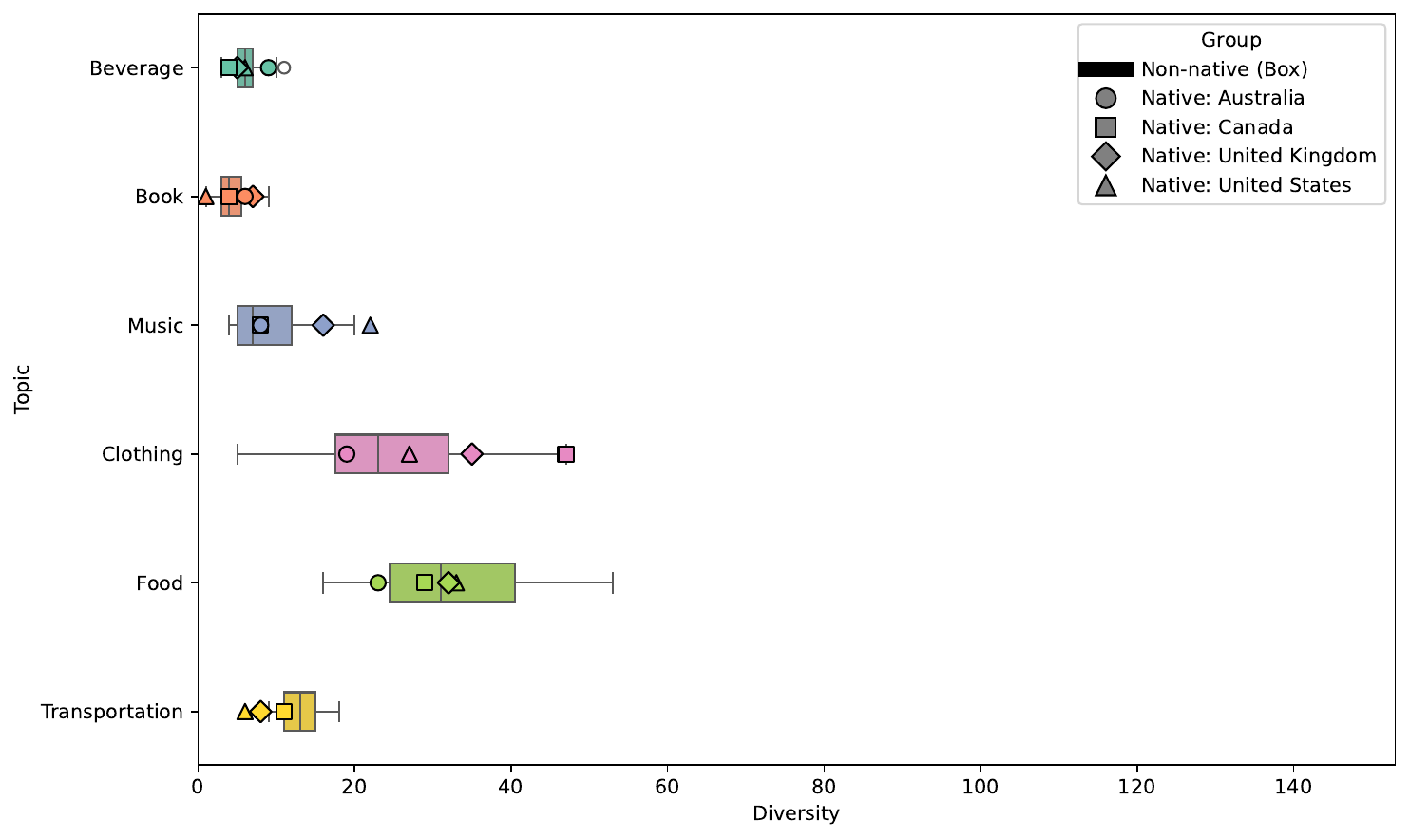} & 
\includegraphics[width=0.3\textwidth]{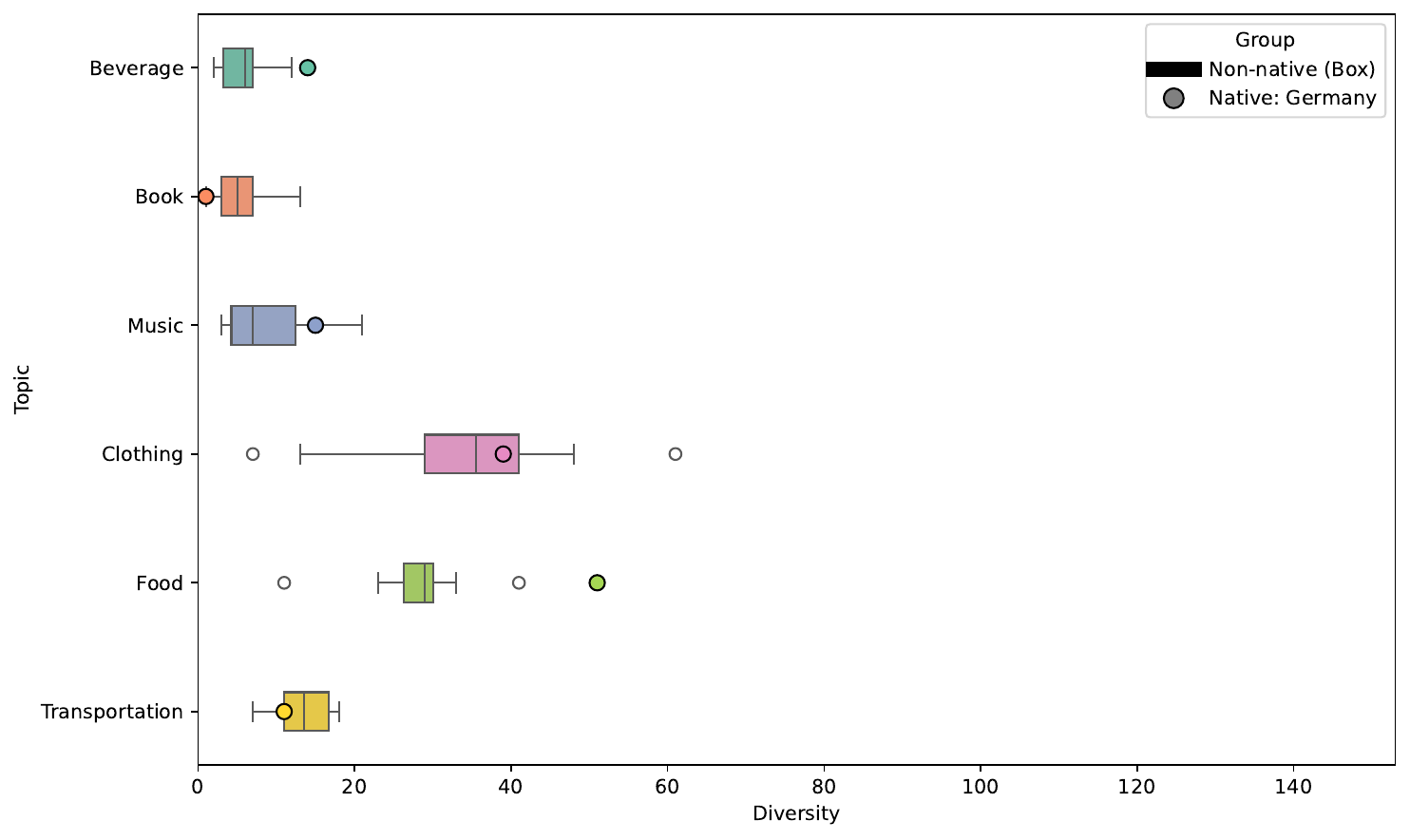} & 
\includegraphics[width=0.3\textwidth]{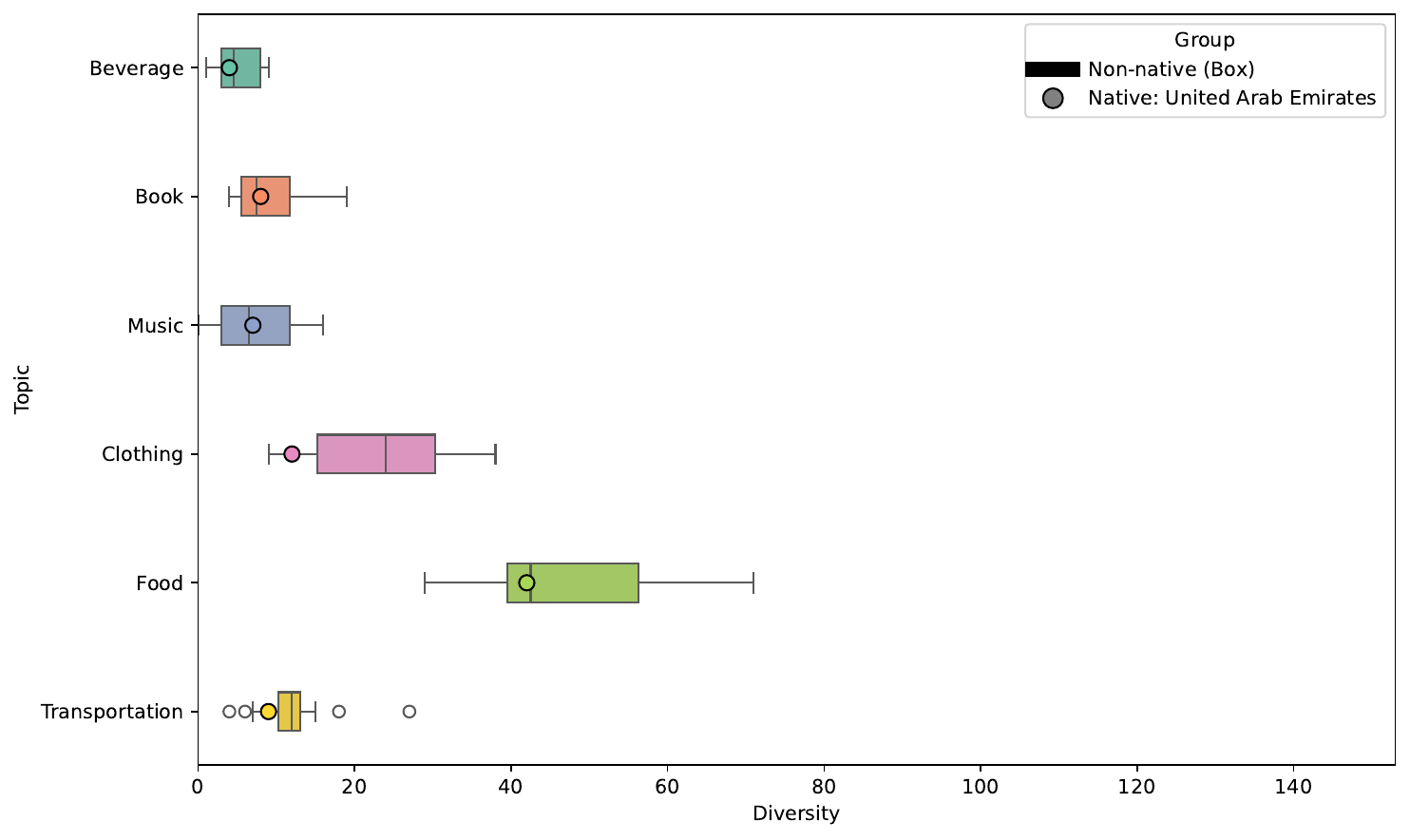} \\
English & German & Arabic \\
\includegraphics[width=0.3\textwidth]{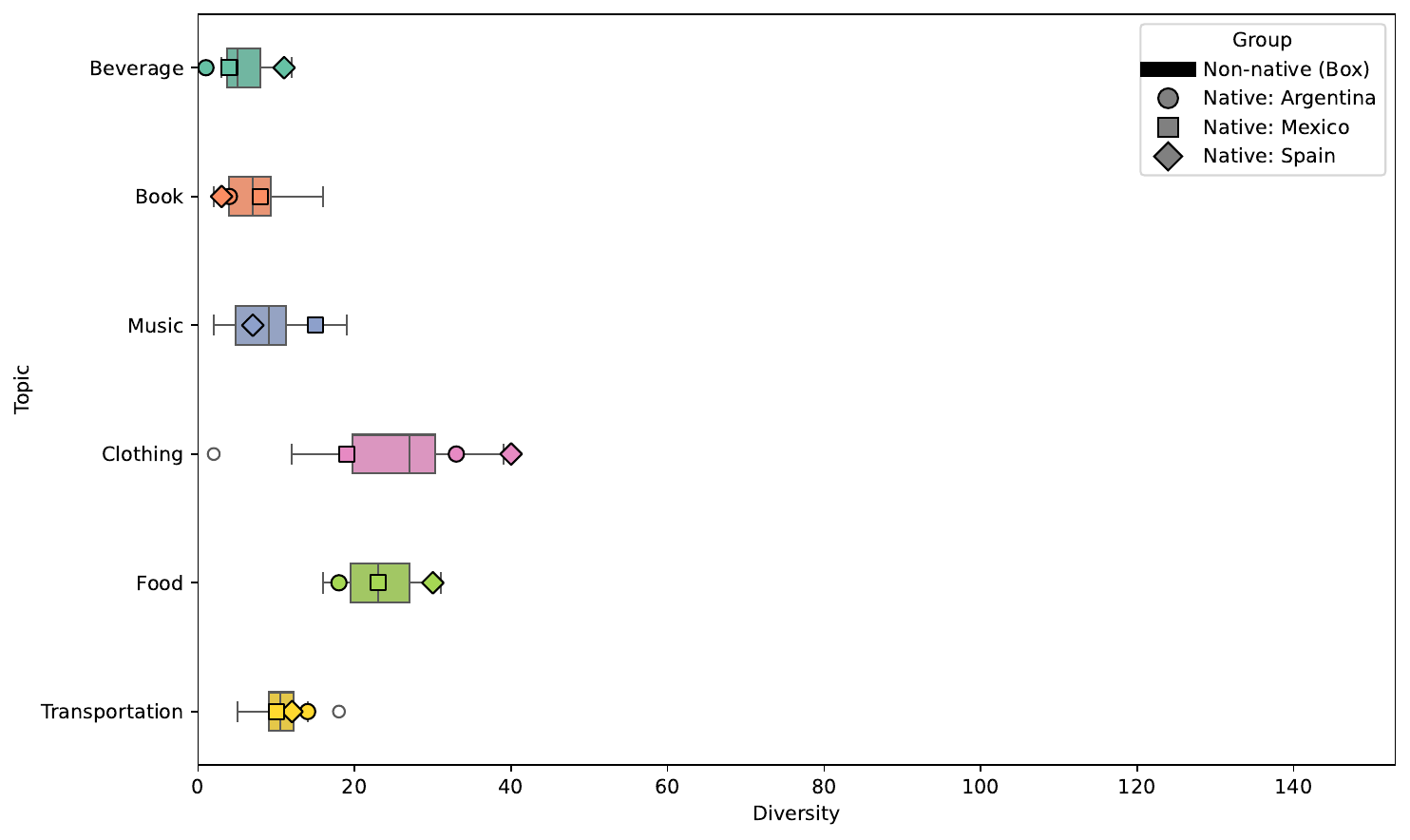} & 
\includegraphics[width=0.3\textwidth]{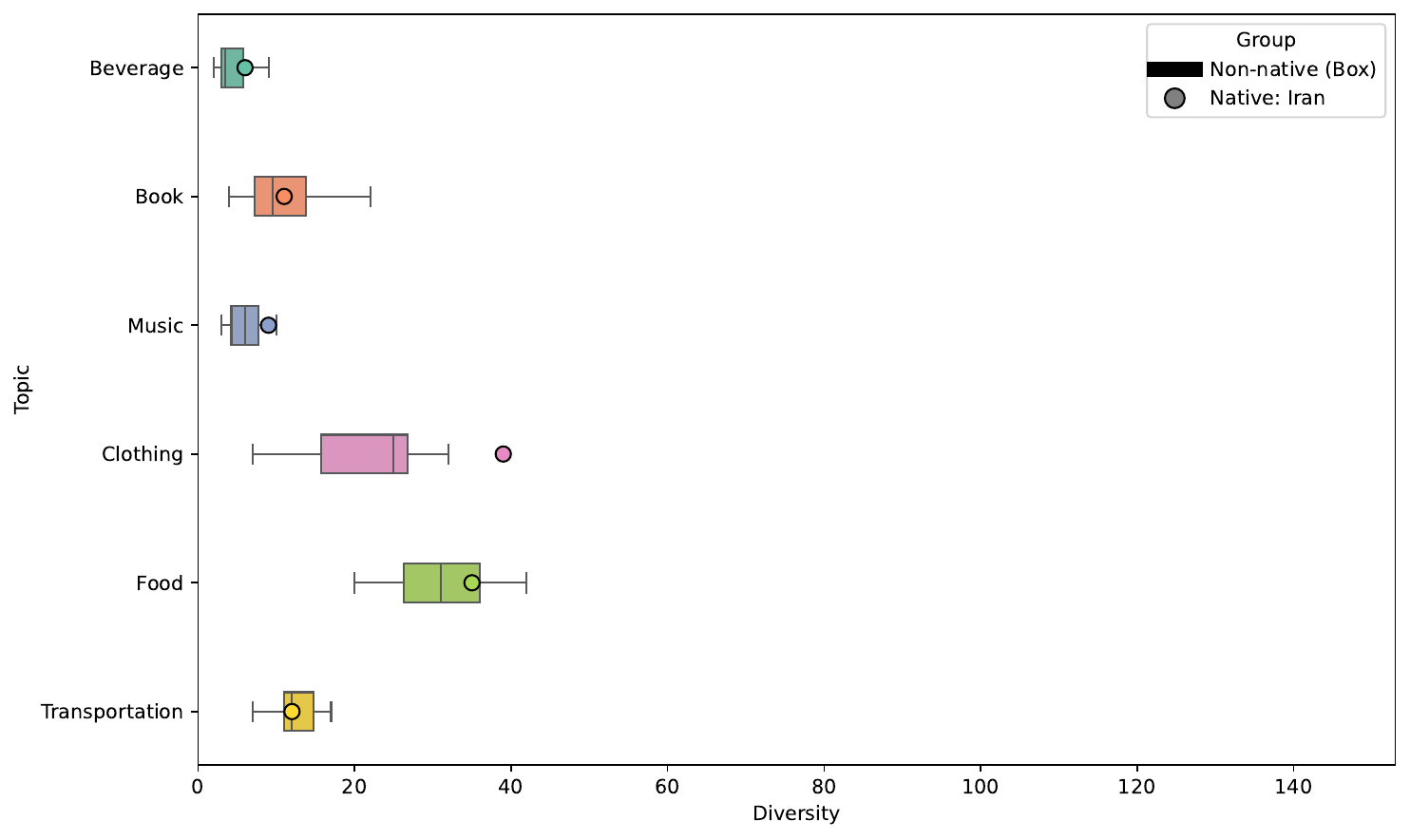} & 
\includegraphics[width=0.3\textwidth]{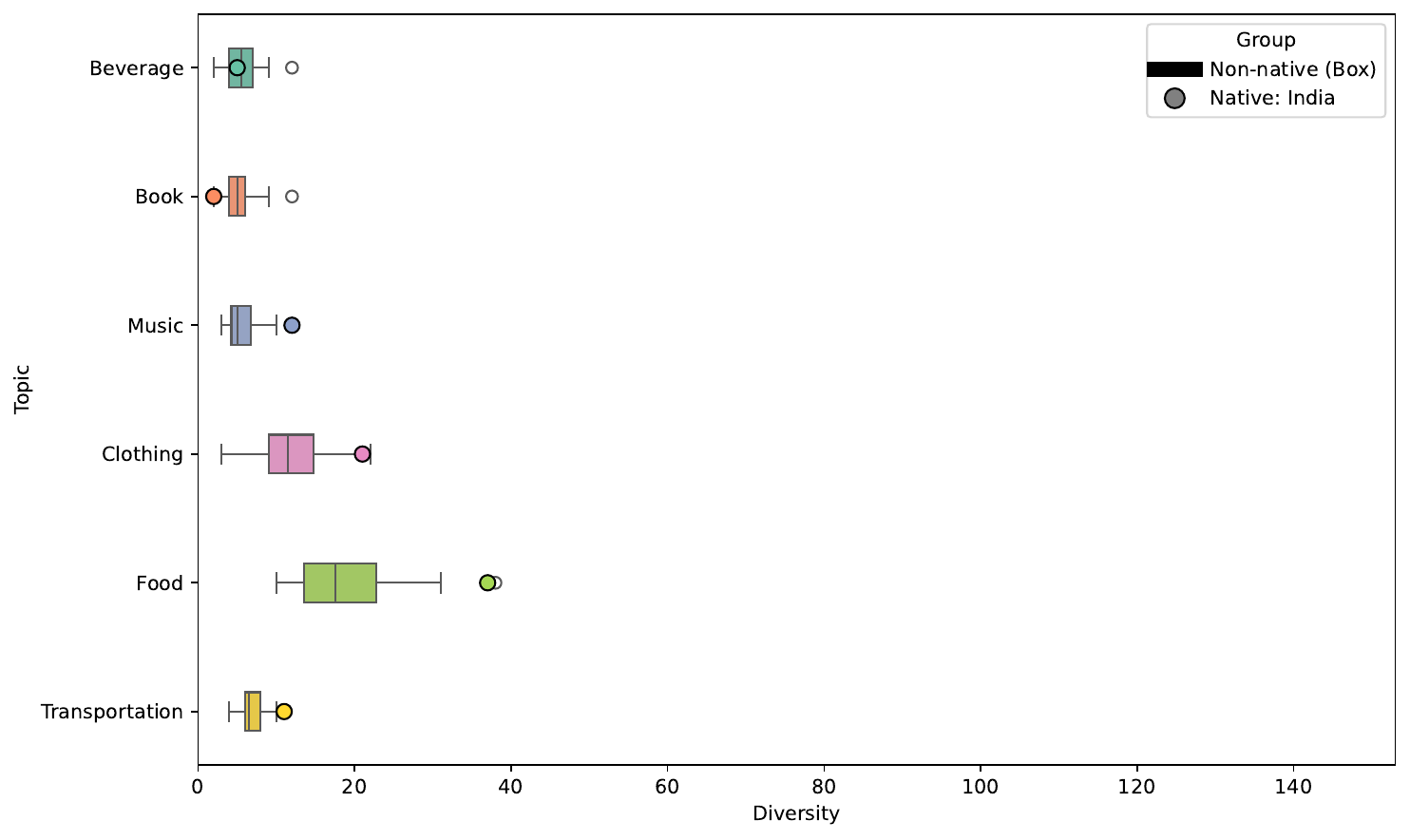} \\
Spanish & Persian & Hindi \\
\includegraphics[width=0.3\textwidth]{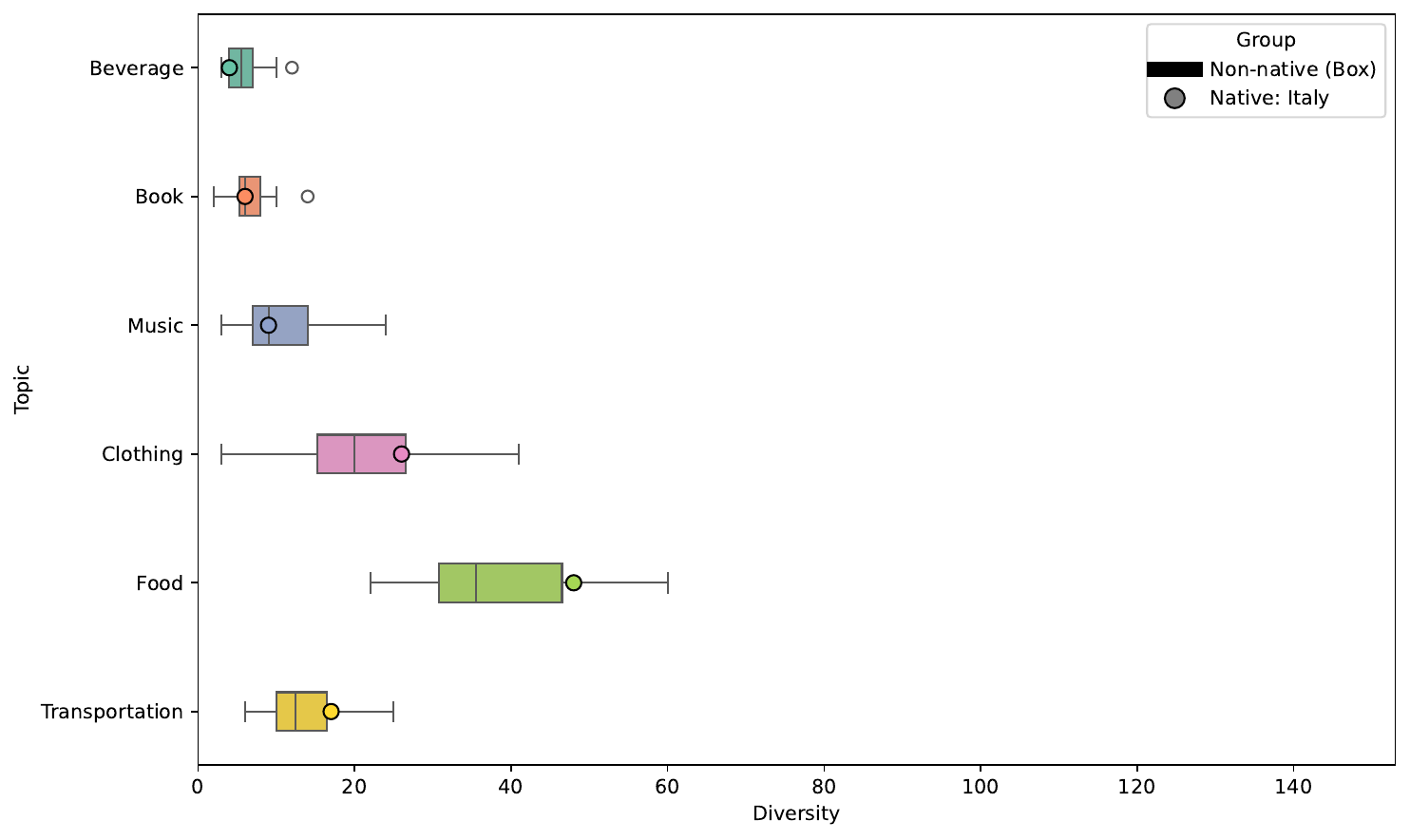} & 
\includegraphics[width=0.3\textwidth]{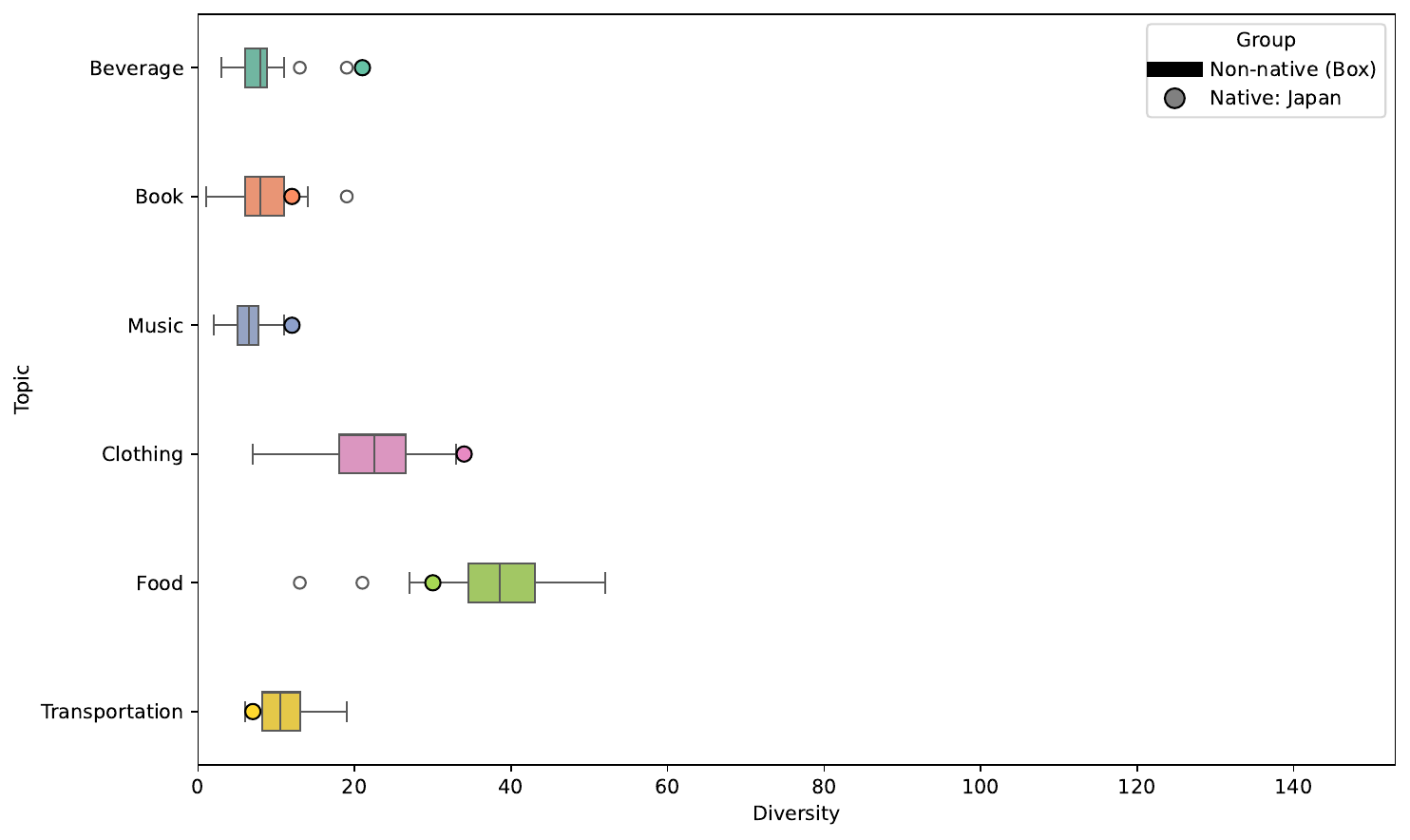} & 
\includegraphics[width=0.3\textwidth]{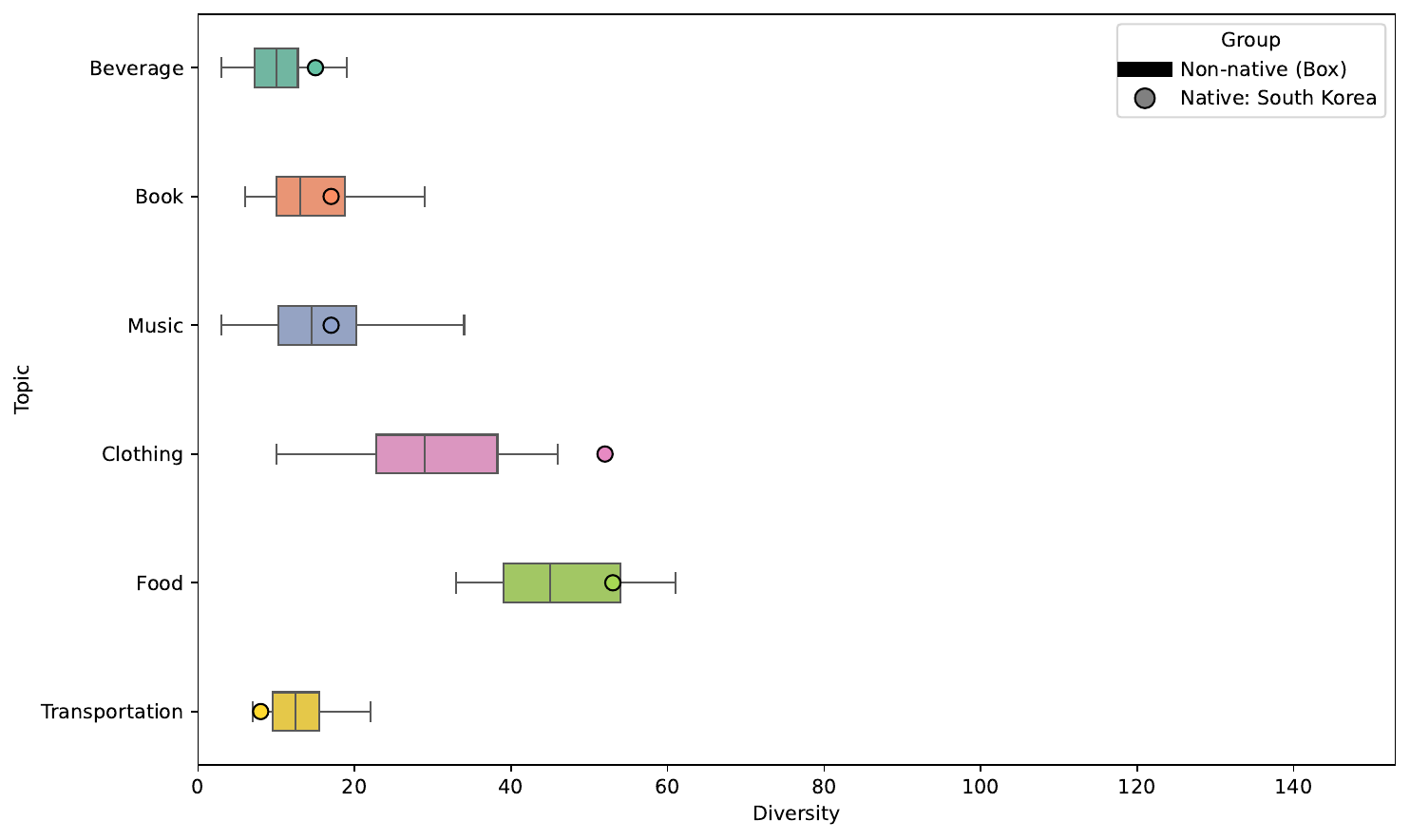} \\
Italian & Japanese & Korean \\
\includegraphics[width=0.3\textwidth]{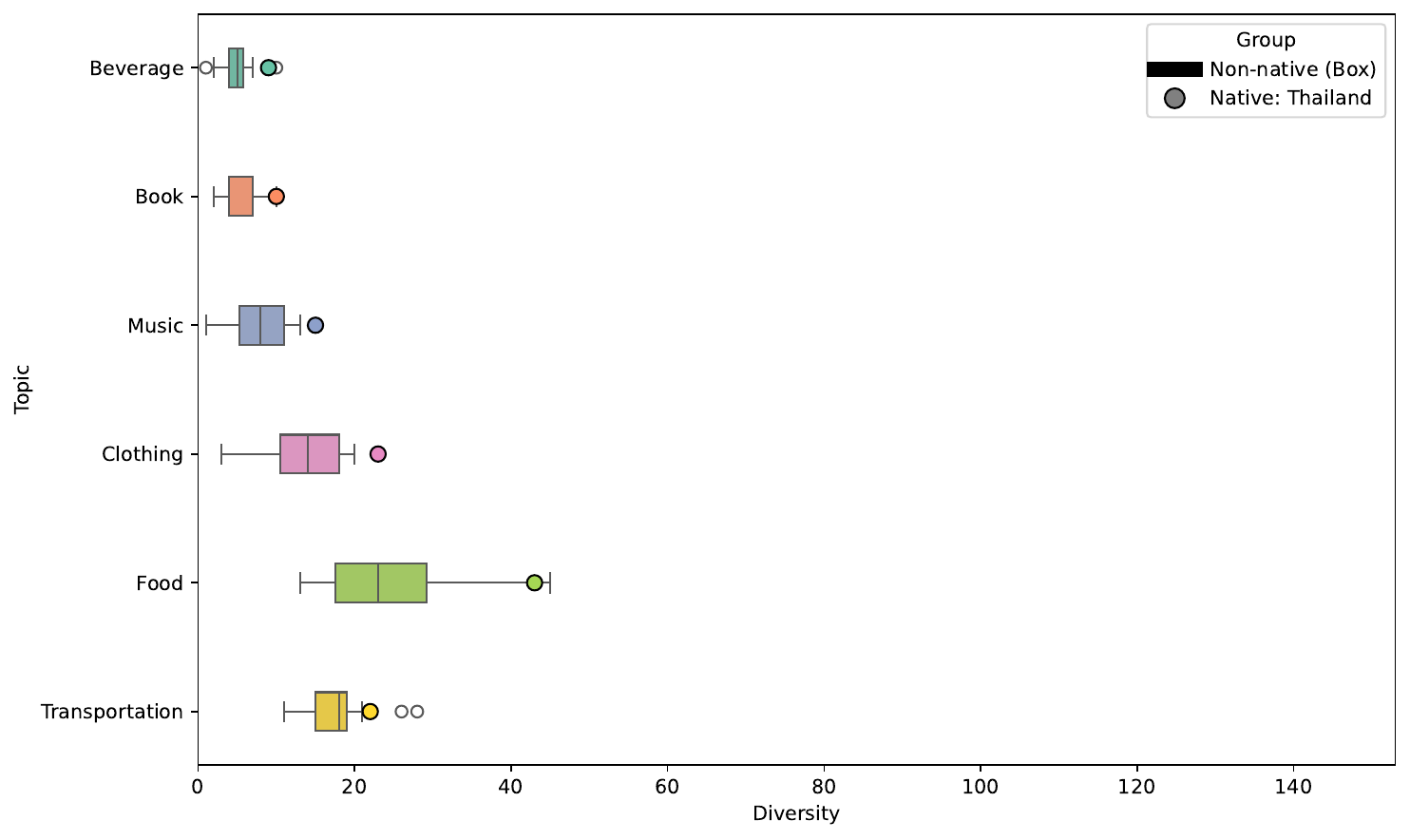} & 
\includegraphics[width=0.3\textwidth]{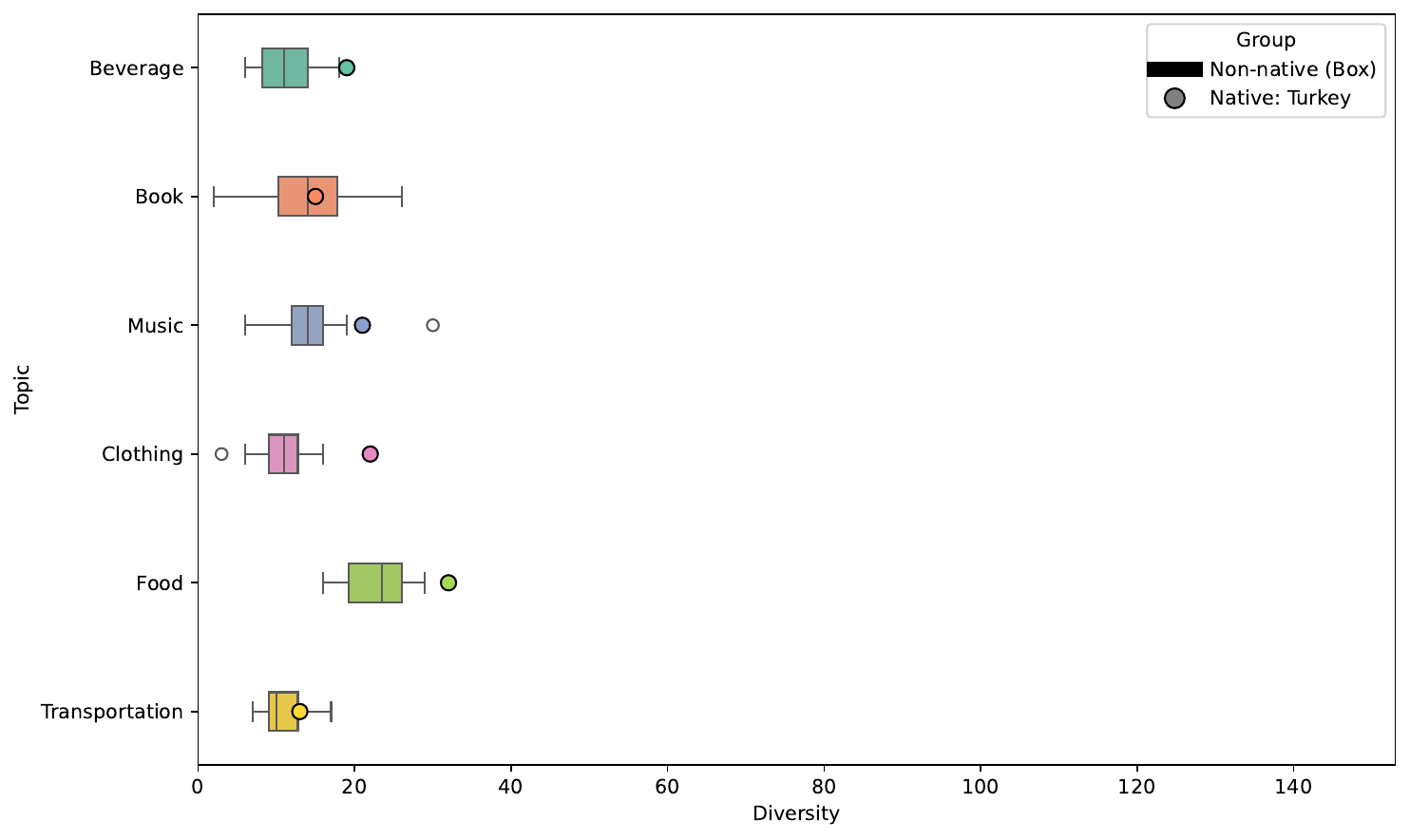} & 
\includegraphics[width=0.3\textwidth]{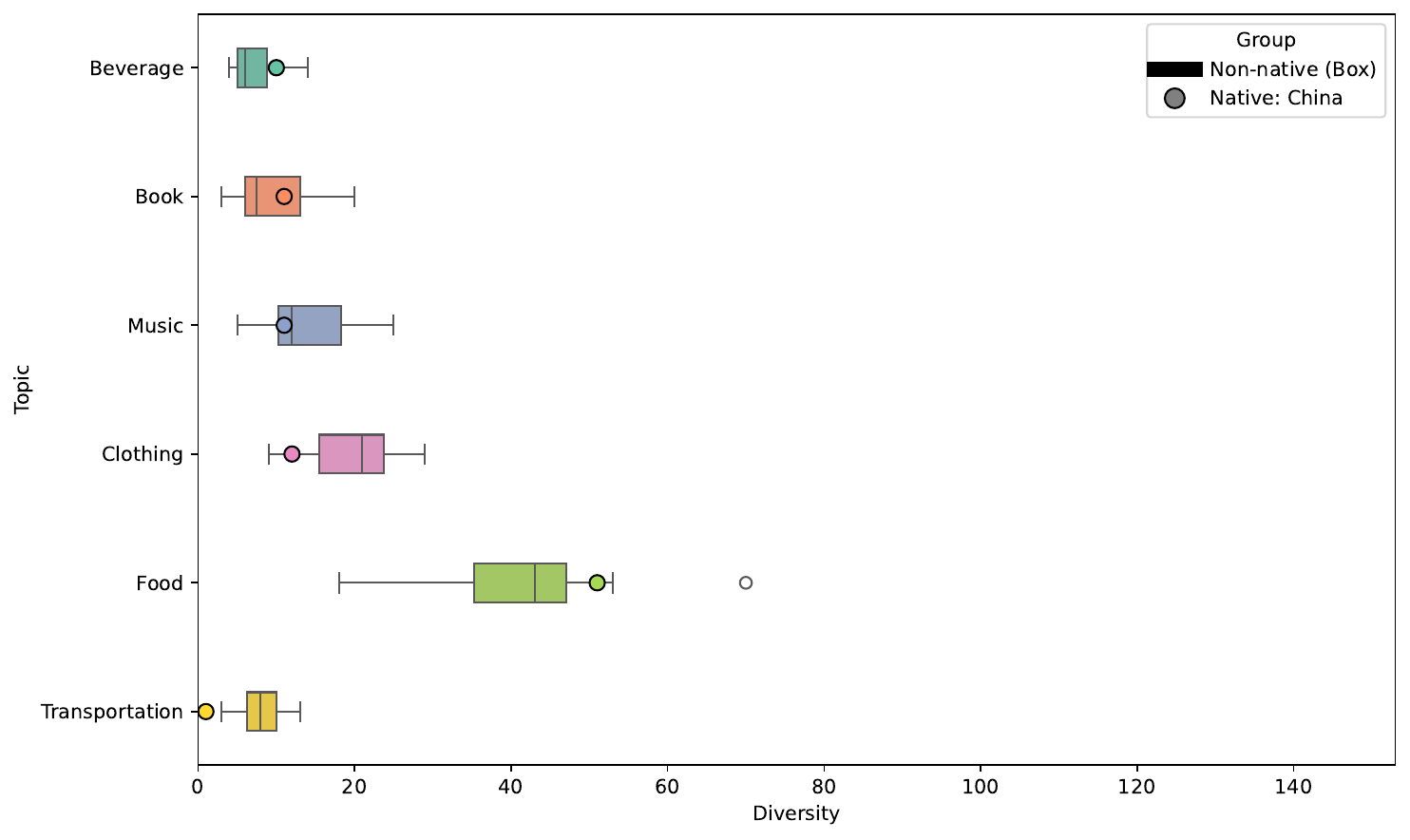} \\
Thai & Turkish & Chinese (Simplified) \\
\includegraphics[width=0.3\textwidth]{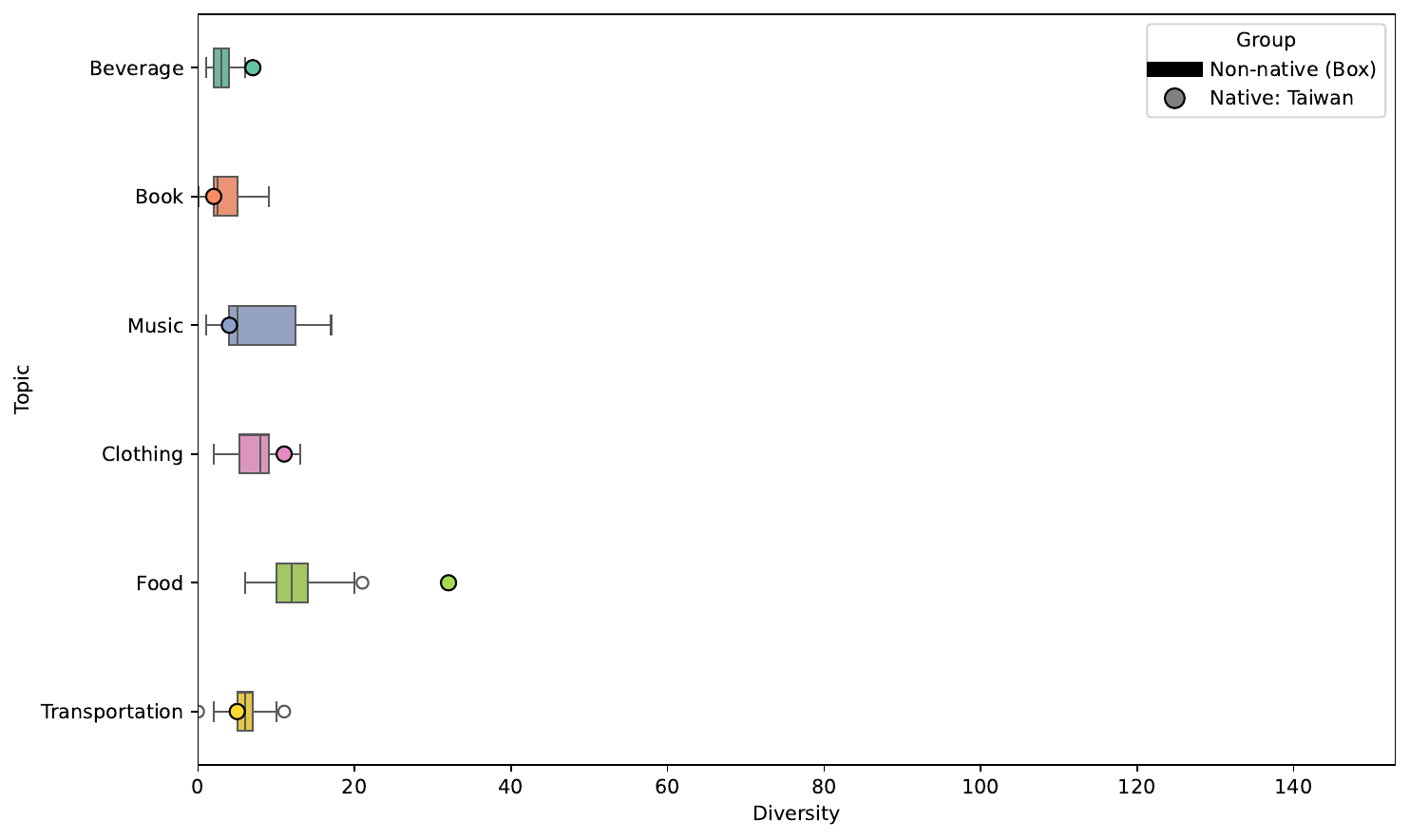} & & \\
Chinese (Traditional) & &
\end{tabular}
\caption{Box plots for diversity comparison between native and non-native languages for \textsc{DeepSeek} across 13 languages.}
\label{fig:box_deepseek}
\end{figure*}

\begin{figure*}[t]
\centering
\begin{tabular}{ccc}
\includegraphics[width=0.3\textwidth]{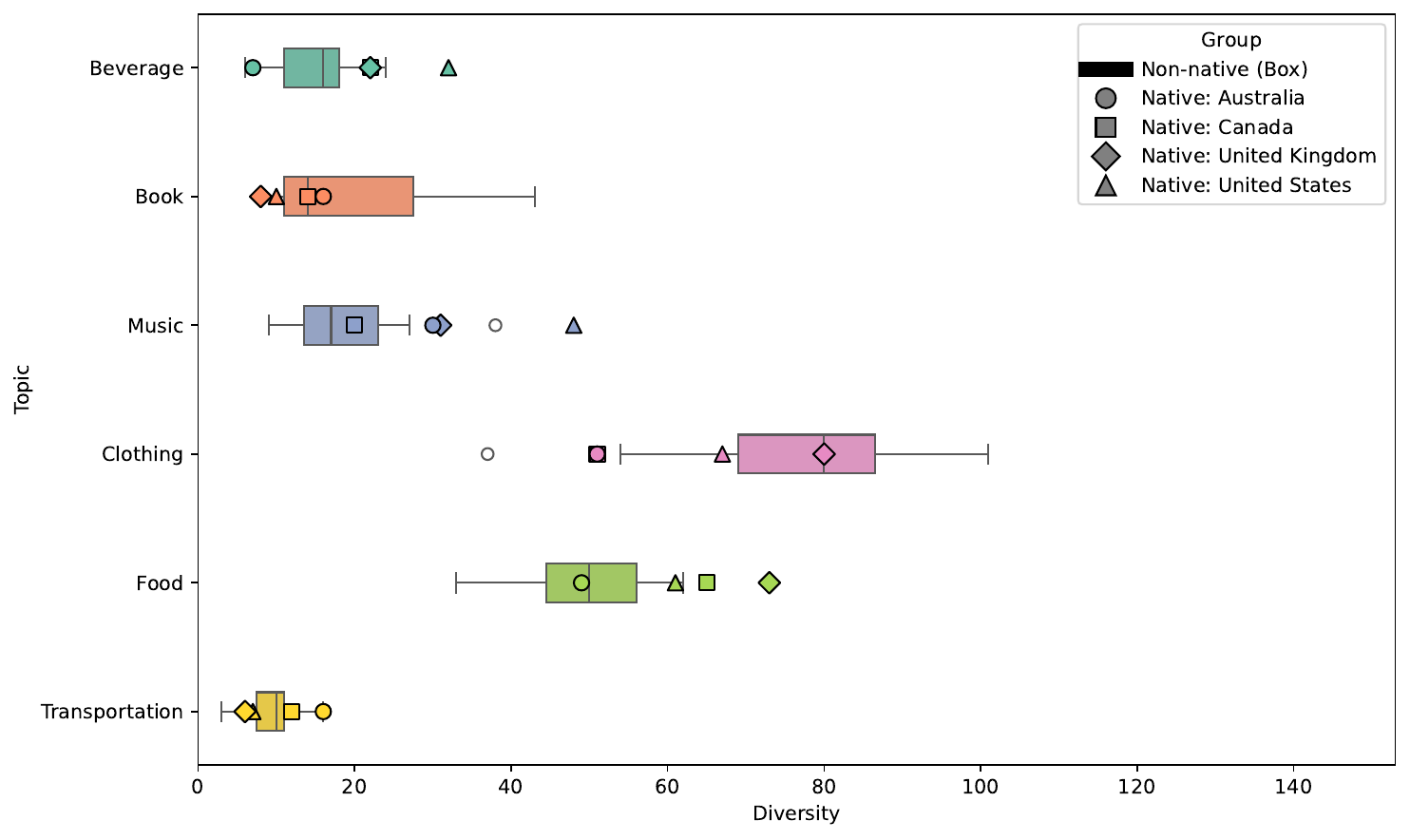} & 
\includegraphics[width=0.3\textwidth]{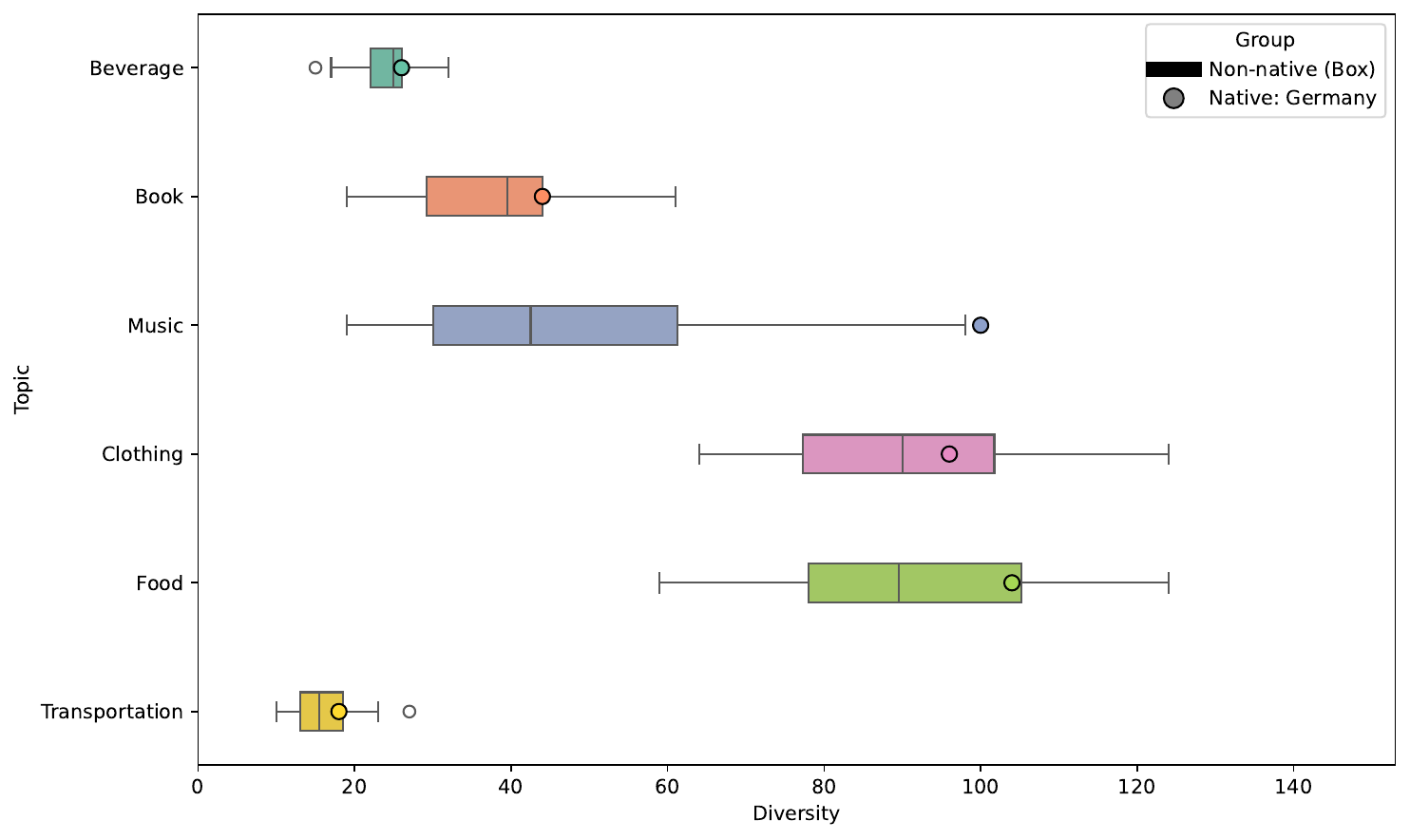} & 
\includegraphics[width=0.3\textwidth]{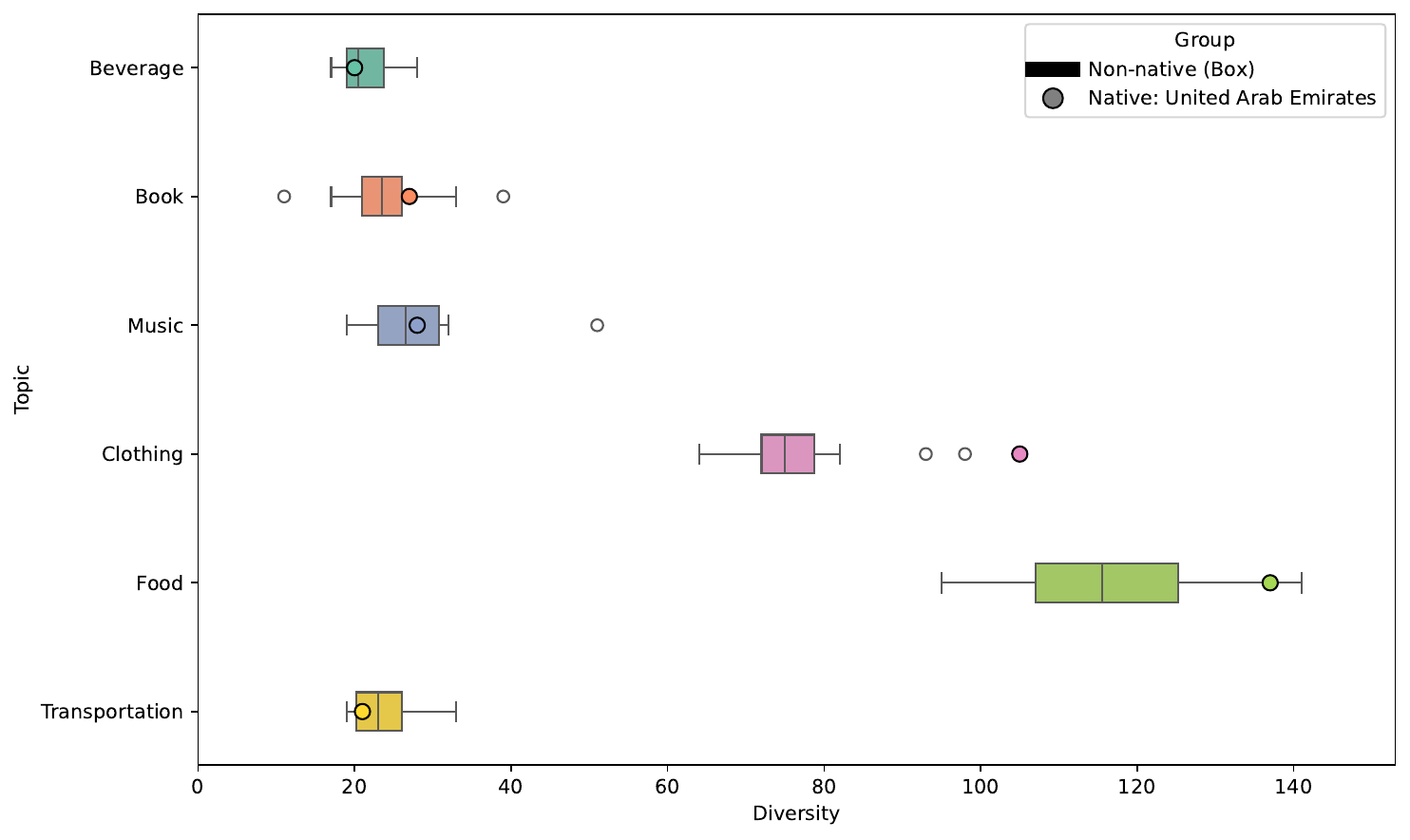} \\
English & German & Arabic \\
\includegraphics[width=0.3\textwidth]{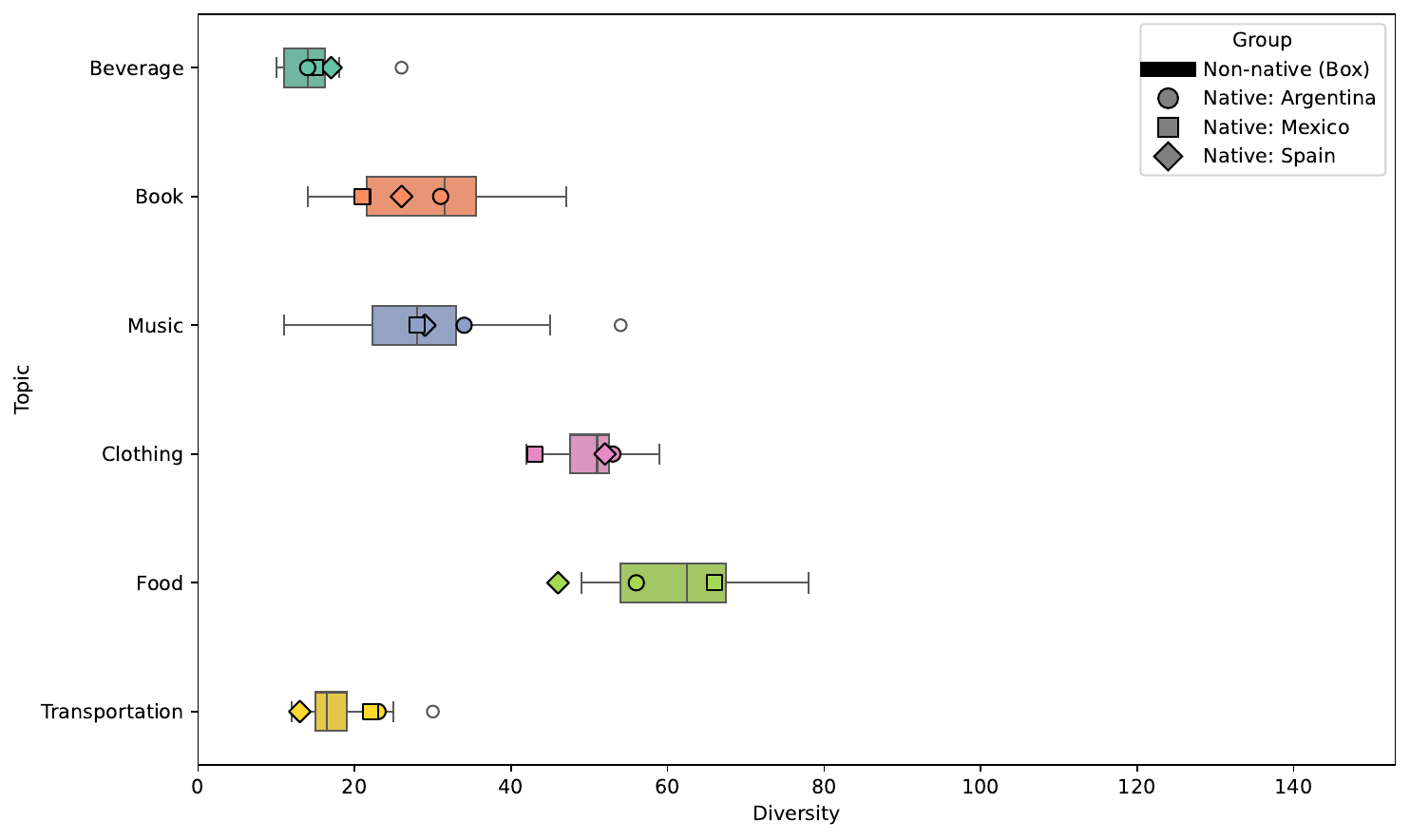} & 
\includegraphics[width=0.3\textwidth]{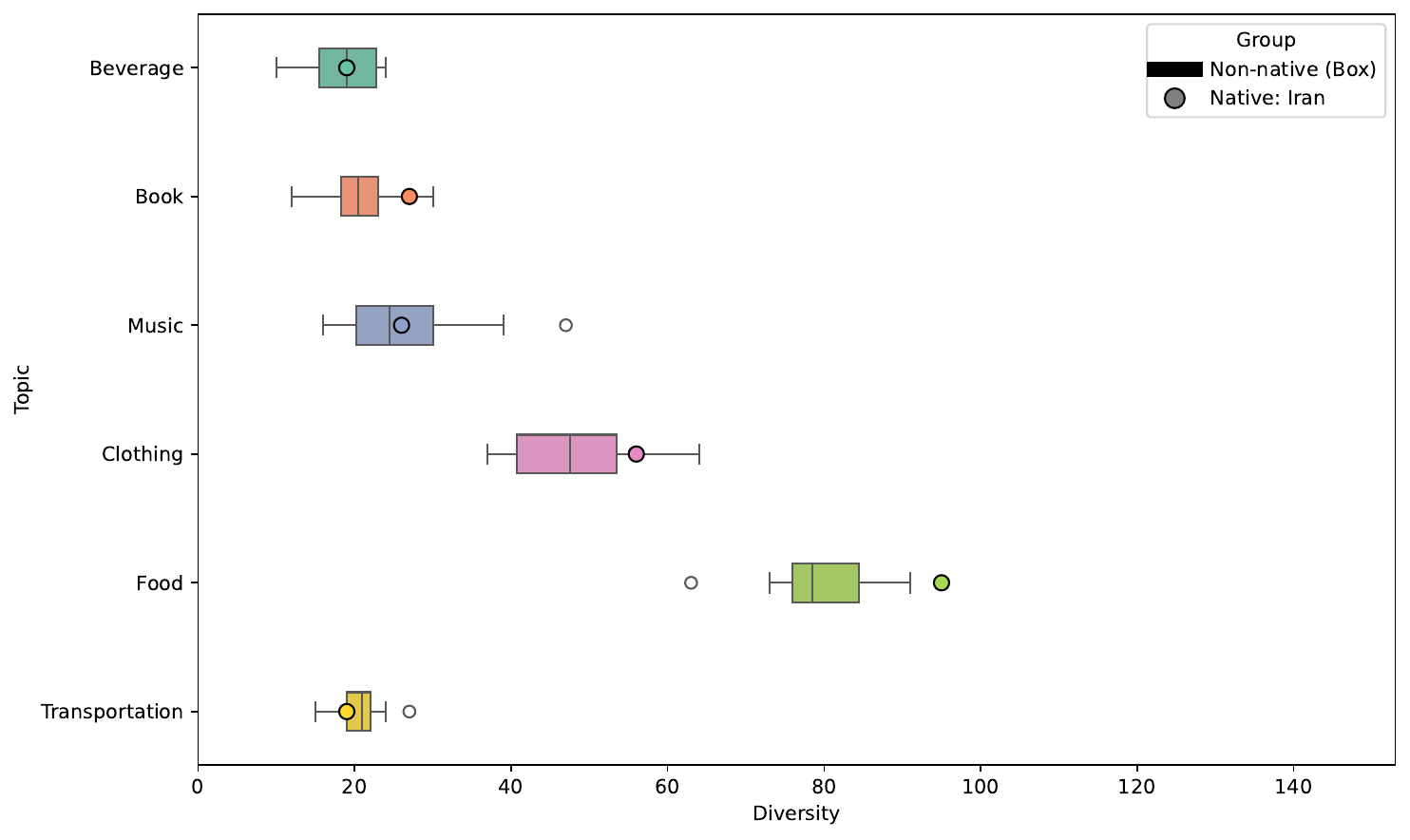} & 
\includegraphics[width=0.3\textwidth]{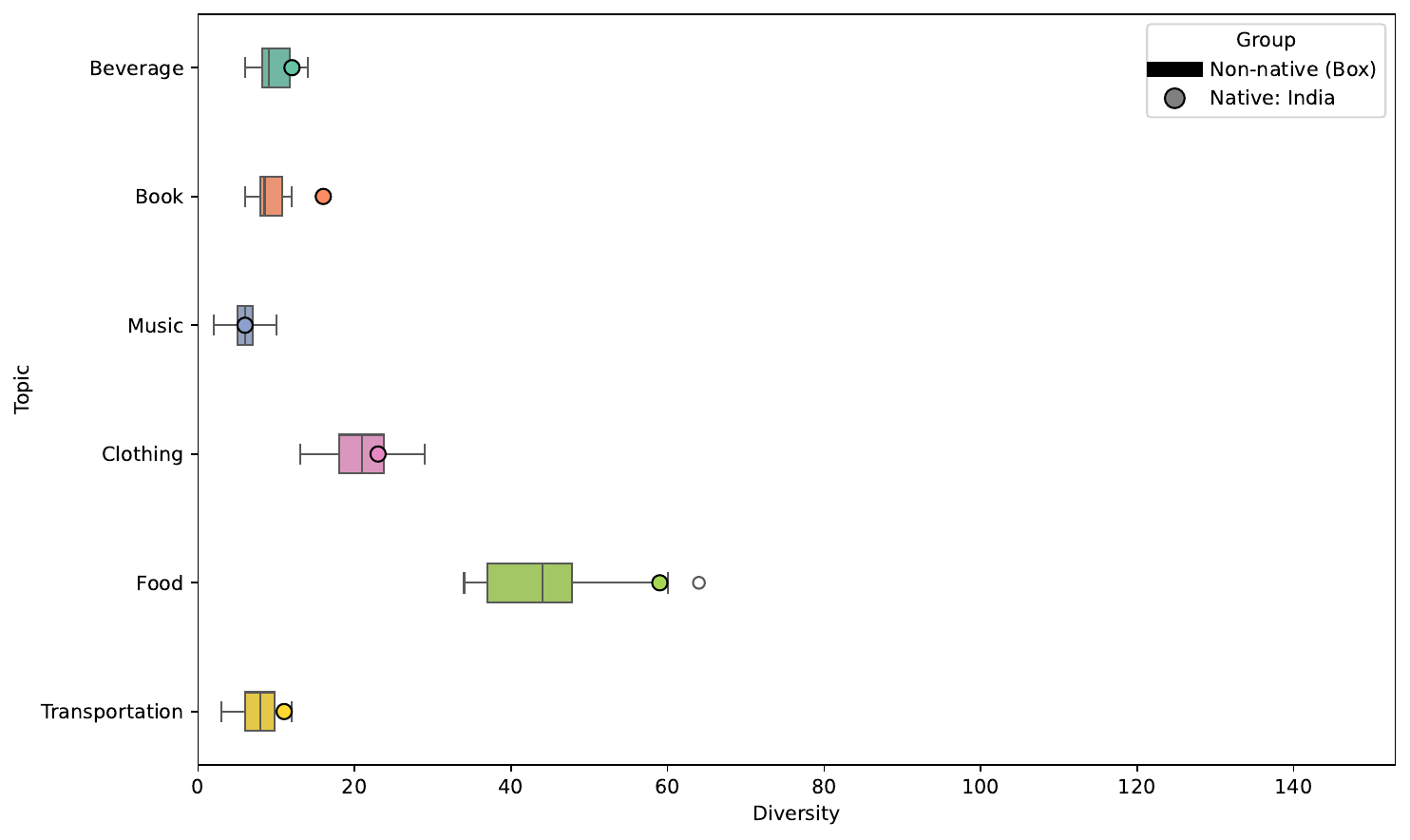} \\
Spanish & Persian & Hindi \\
\includegraphics[width=0.3\textwidth]{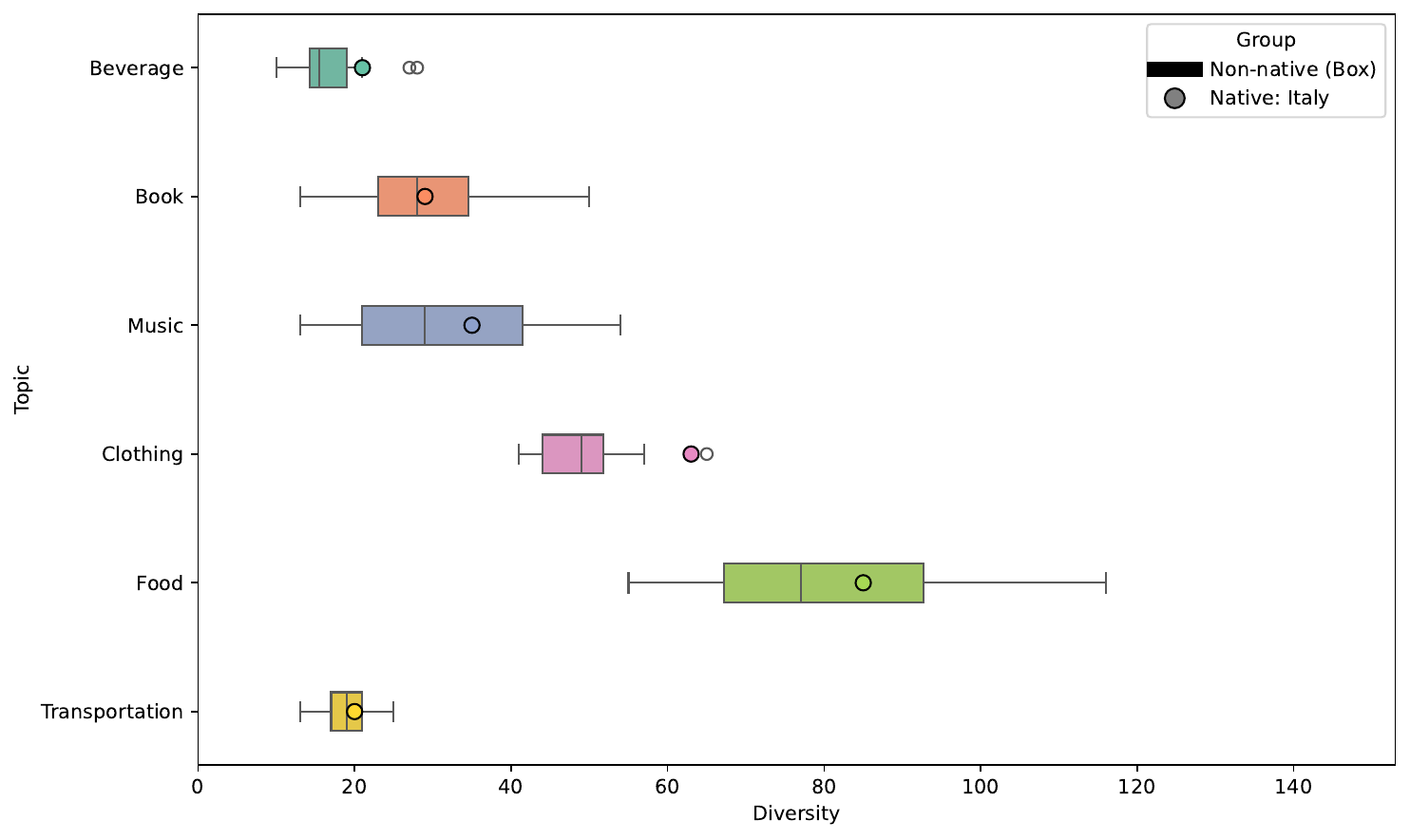} & 
\includegraphics[width=0.3\textwidth]{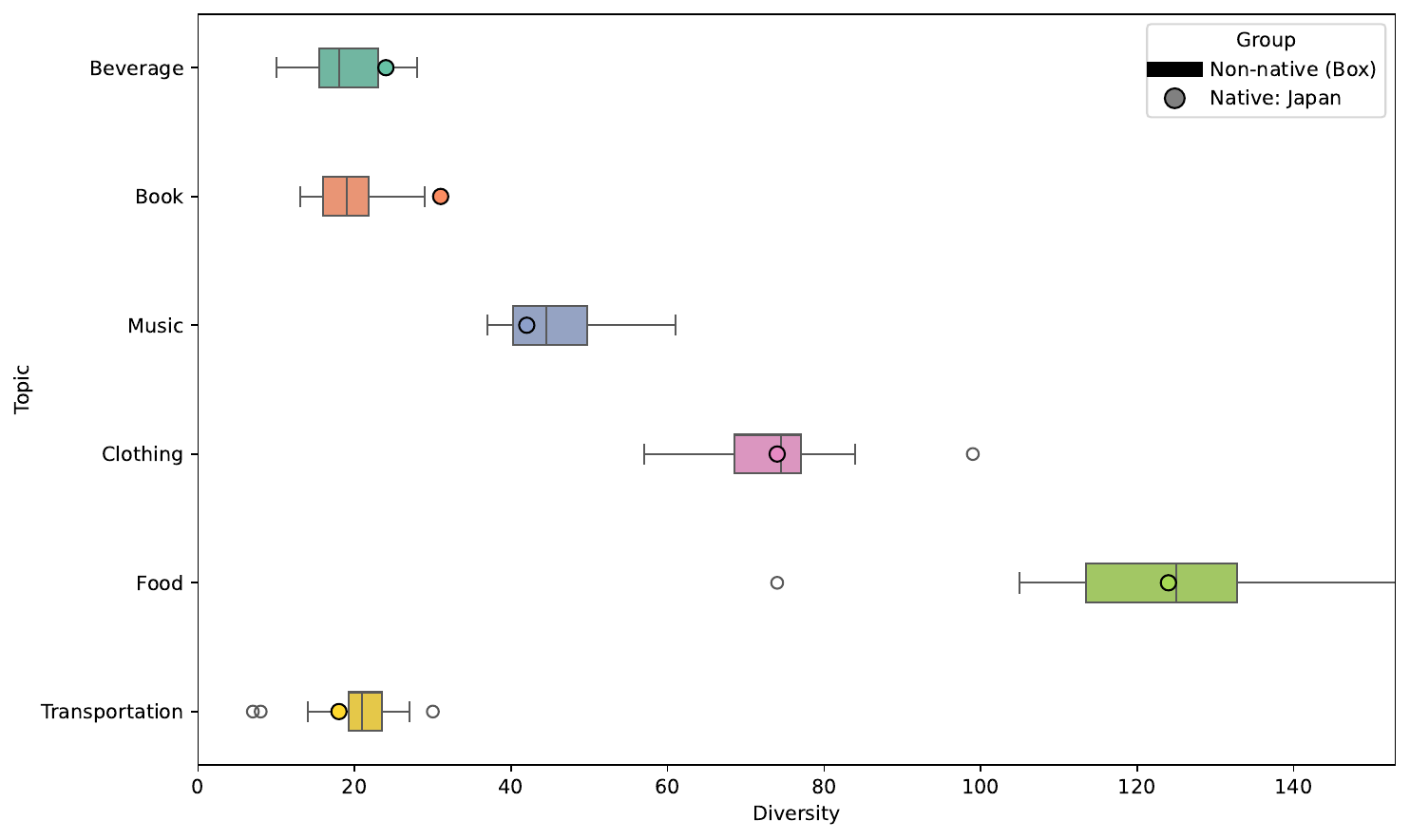} & 
\includegraphics[width=0.3\textwidth]{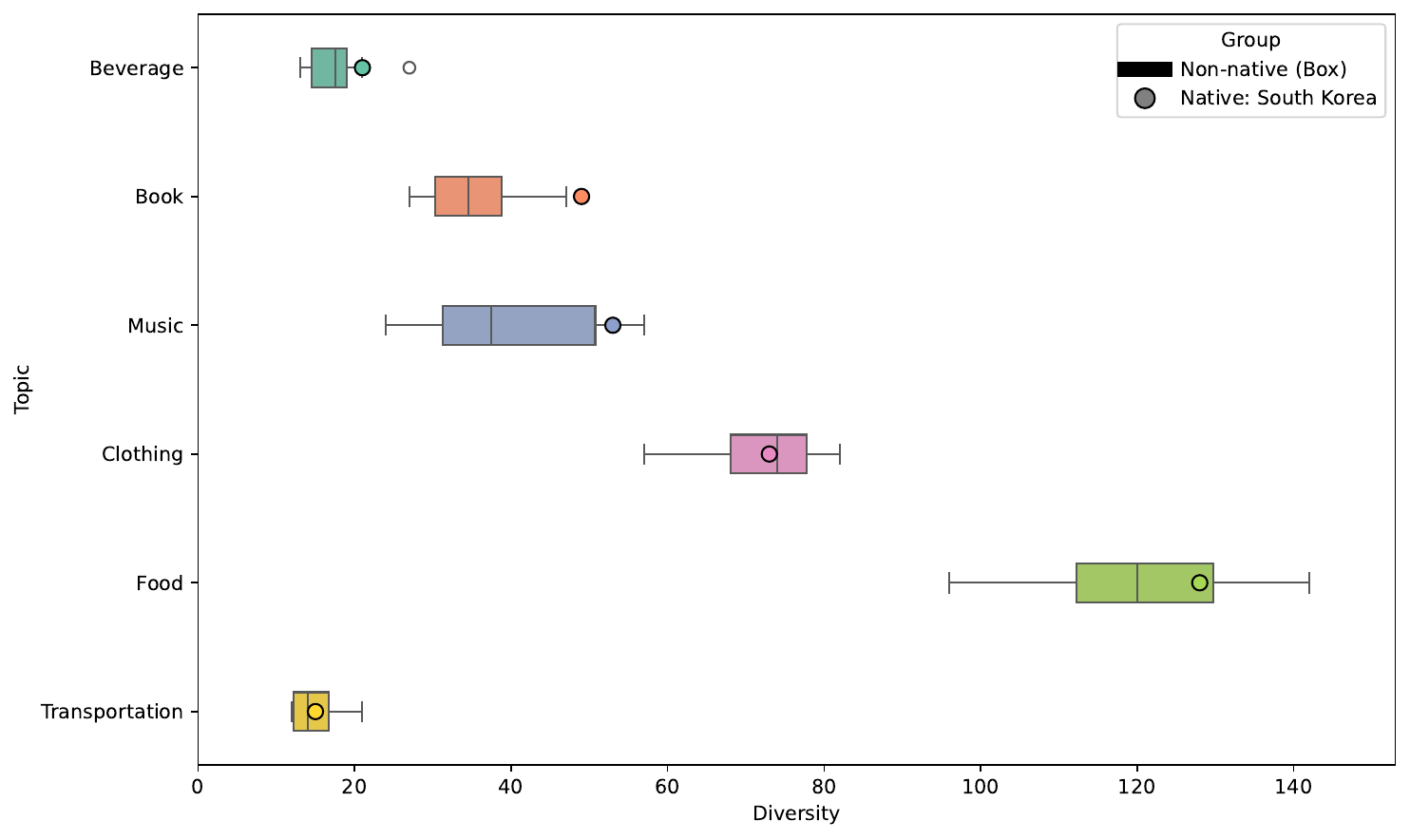} \\
Italian & Japanese & Korean \\
\includegraphics[width=0.3\textwidth]{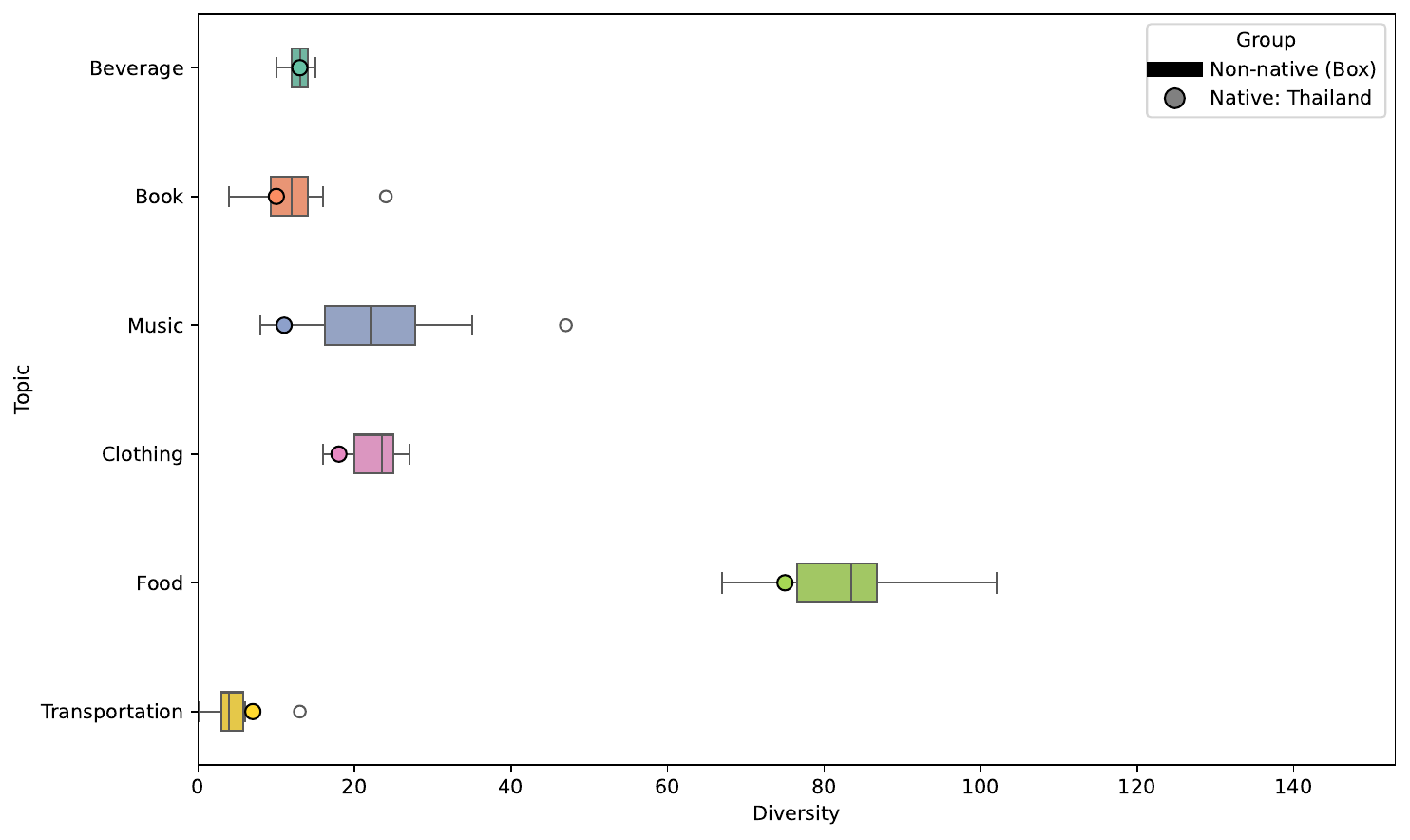} & 
\includegraphics[width=0.3\textwidth]{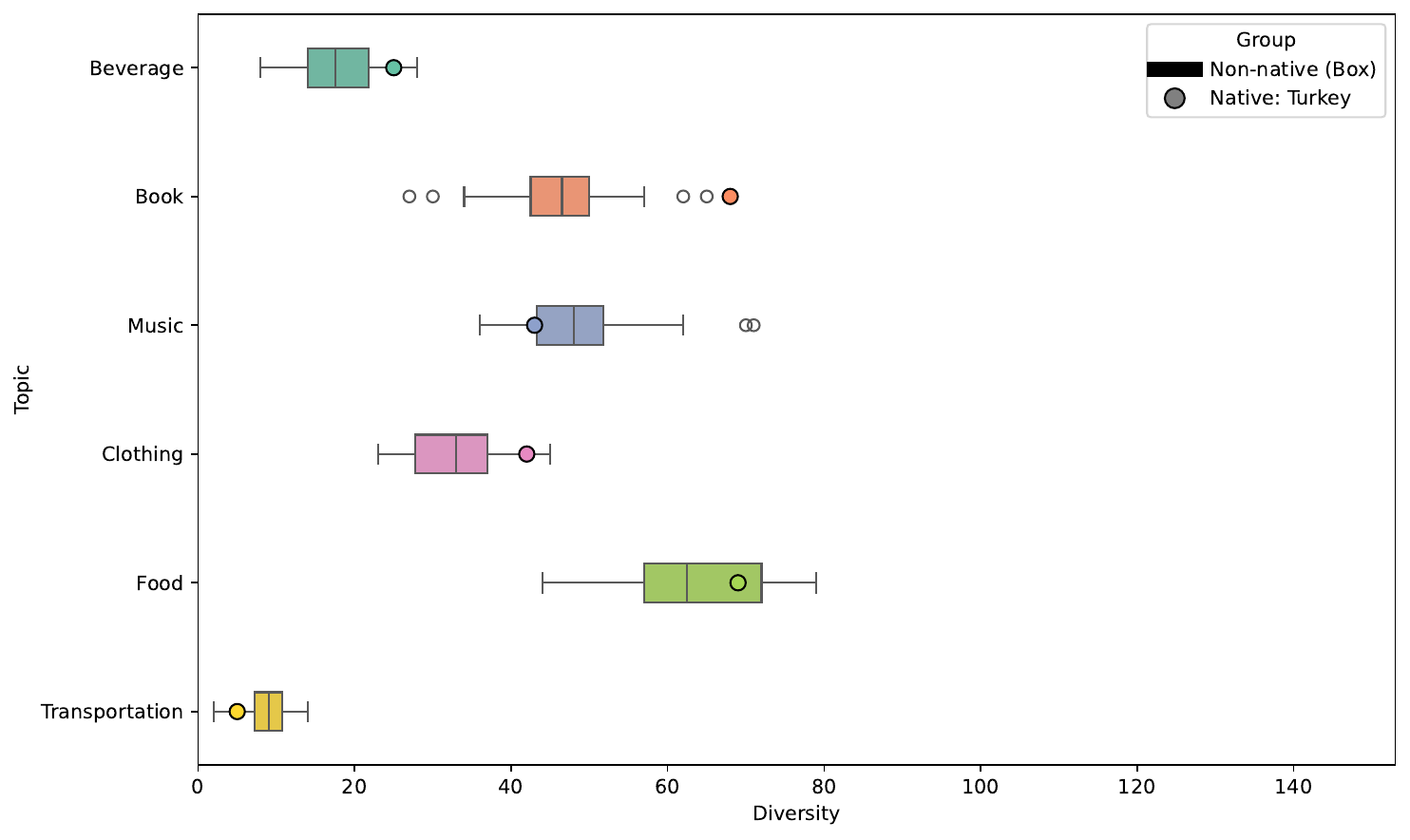} & 
\includegraphics[width=0.3\textwidth]{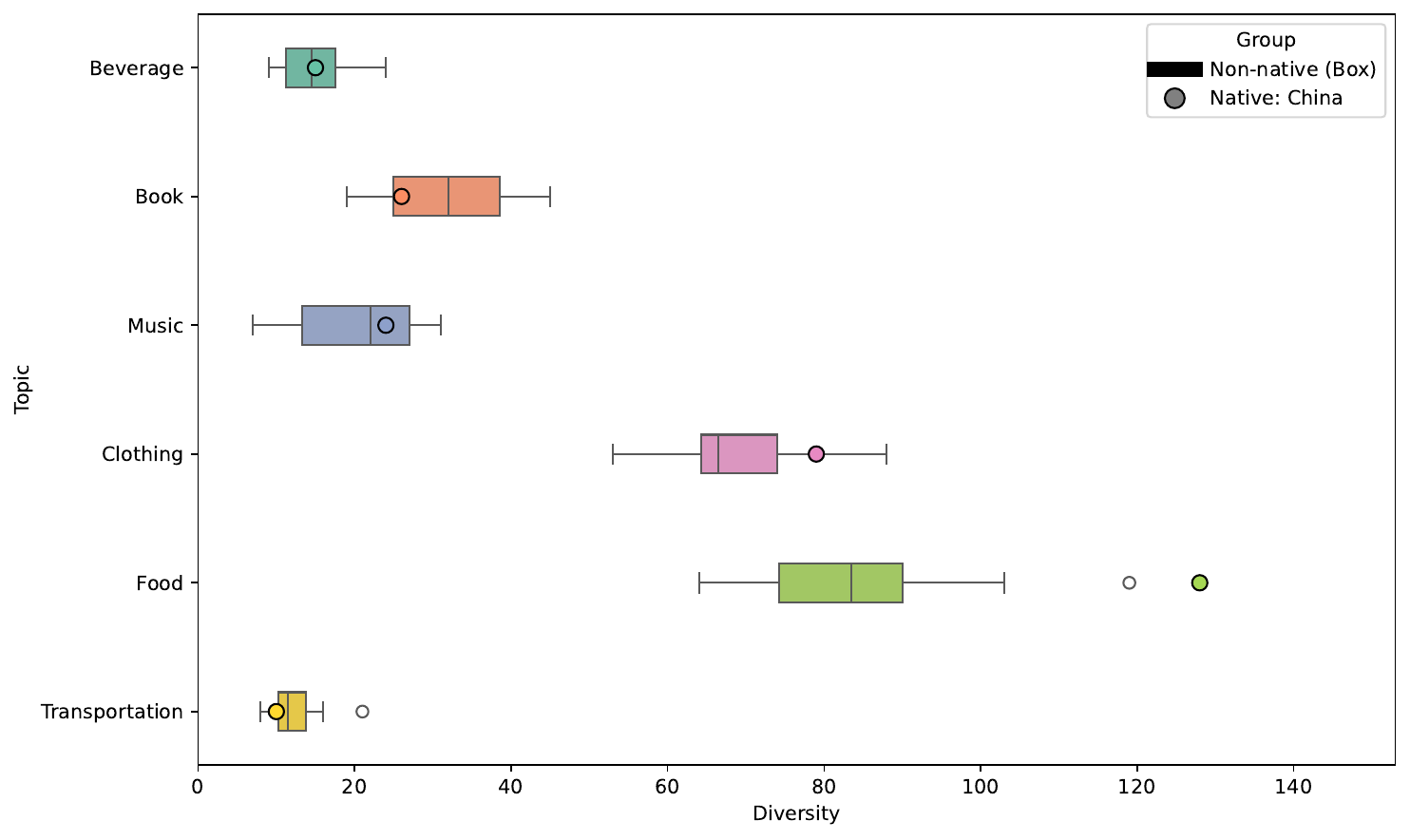} \\
Thai & Turkish & Chinese (Simplified) \\
\includegraphics[width=0.3\textwidth]{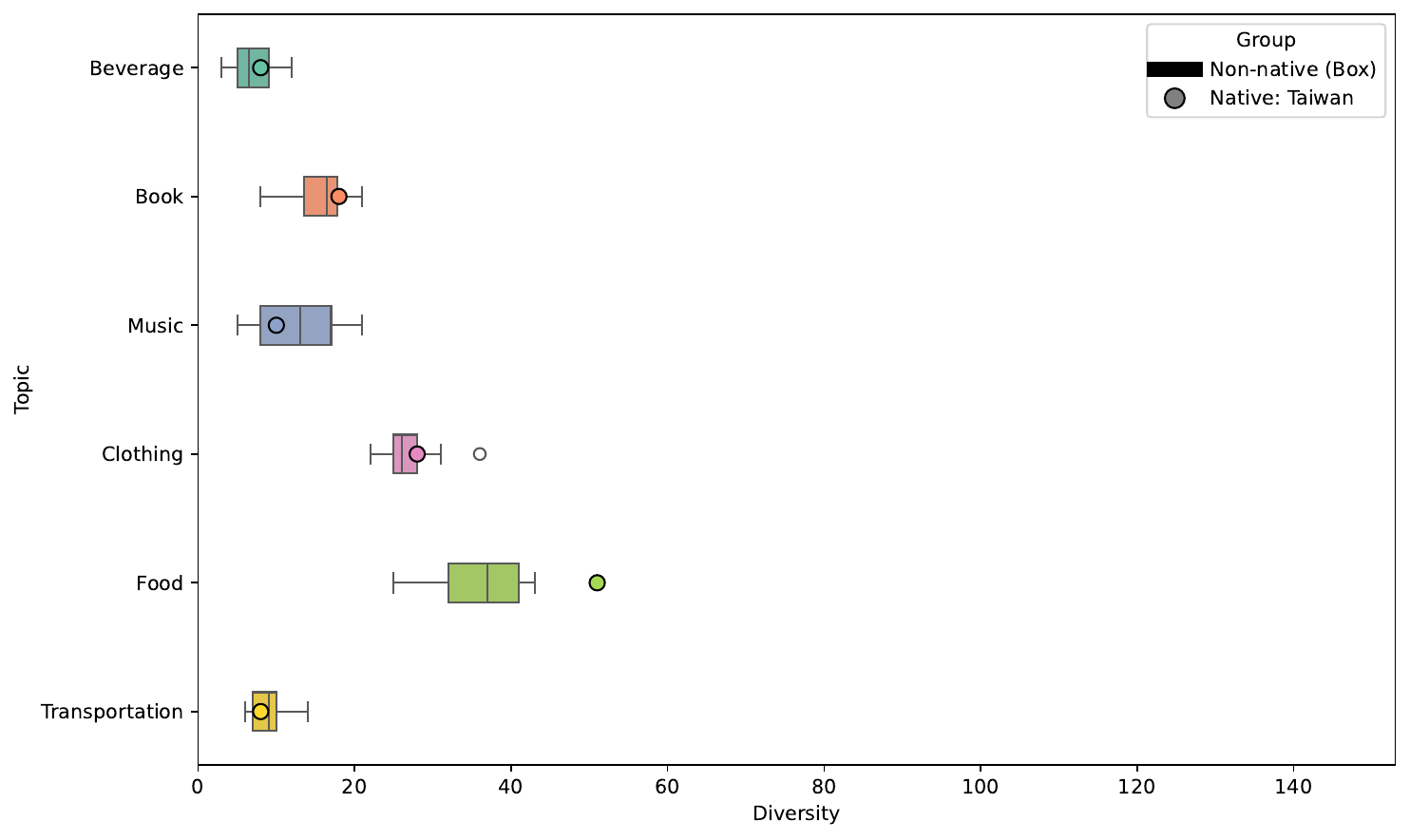} & & \\
Chinese (Traditional) & &
\end{tabular}
\caption{Box plots for diversity comparison between native and non-native languages for \textsc{Llama3-8B}}
\end{figure*}

\begin{figure*}[t]
\centering
\begin{tabular}{ccc}
\includegraphics[width=0.3\textwidth]{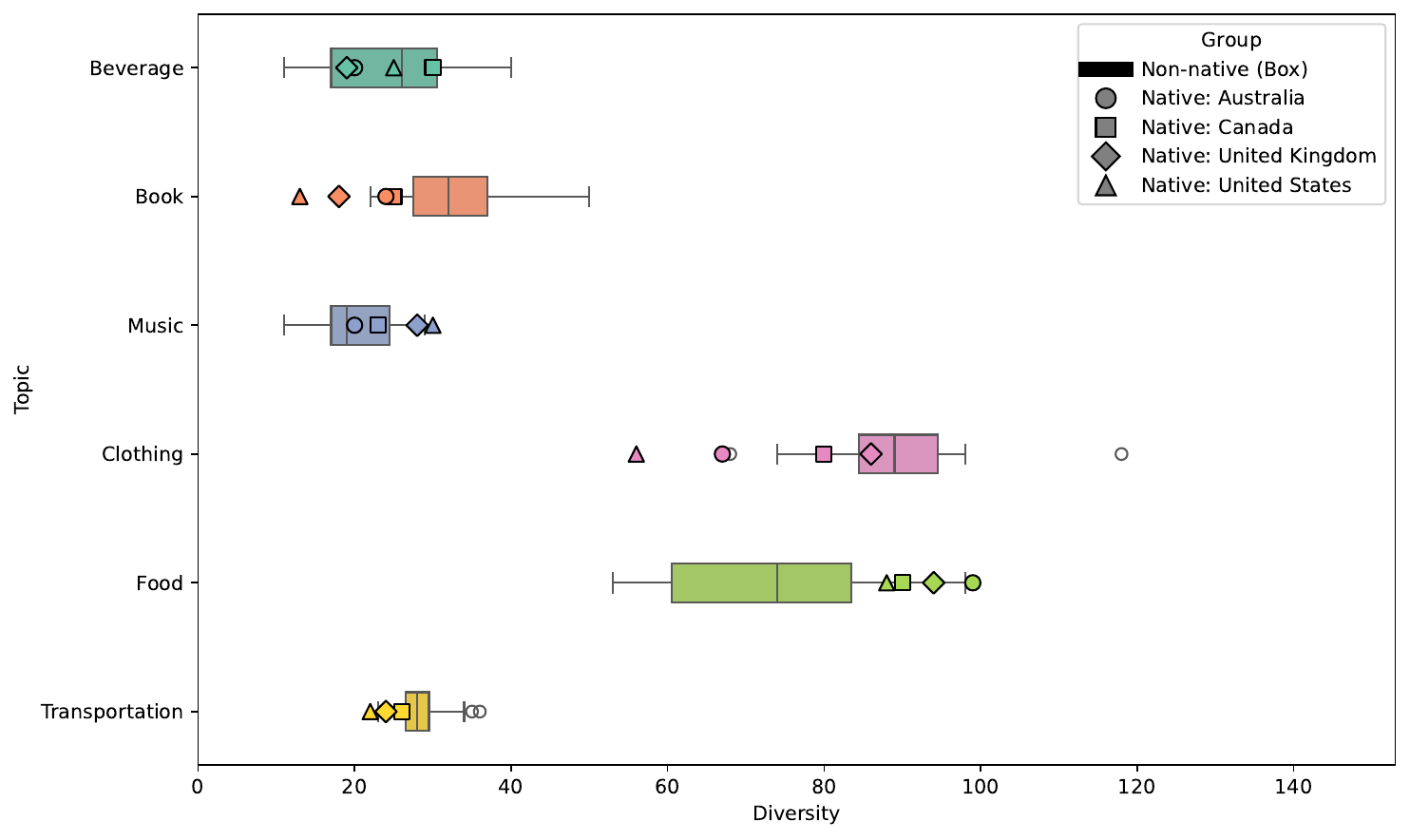} & 
\includegraphics[width=0.3\textwidth]{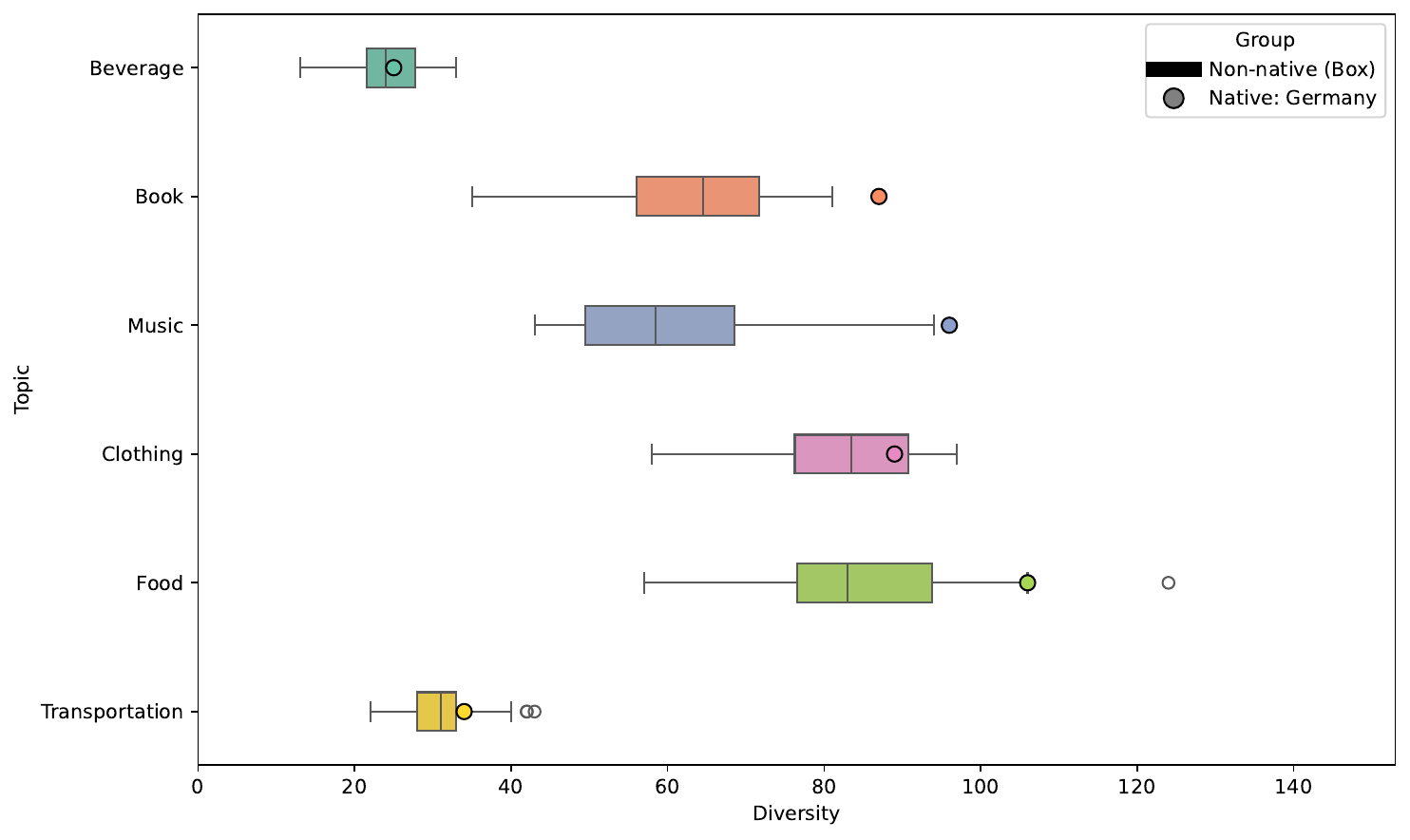} & 
\includegraphics[width=0.3\textwidth]{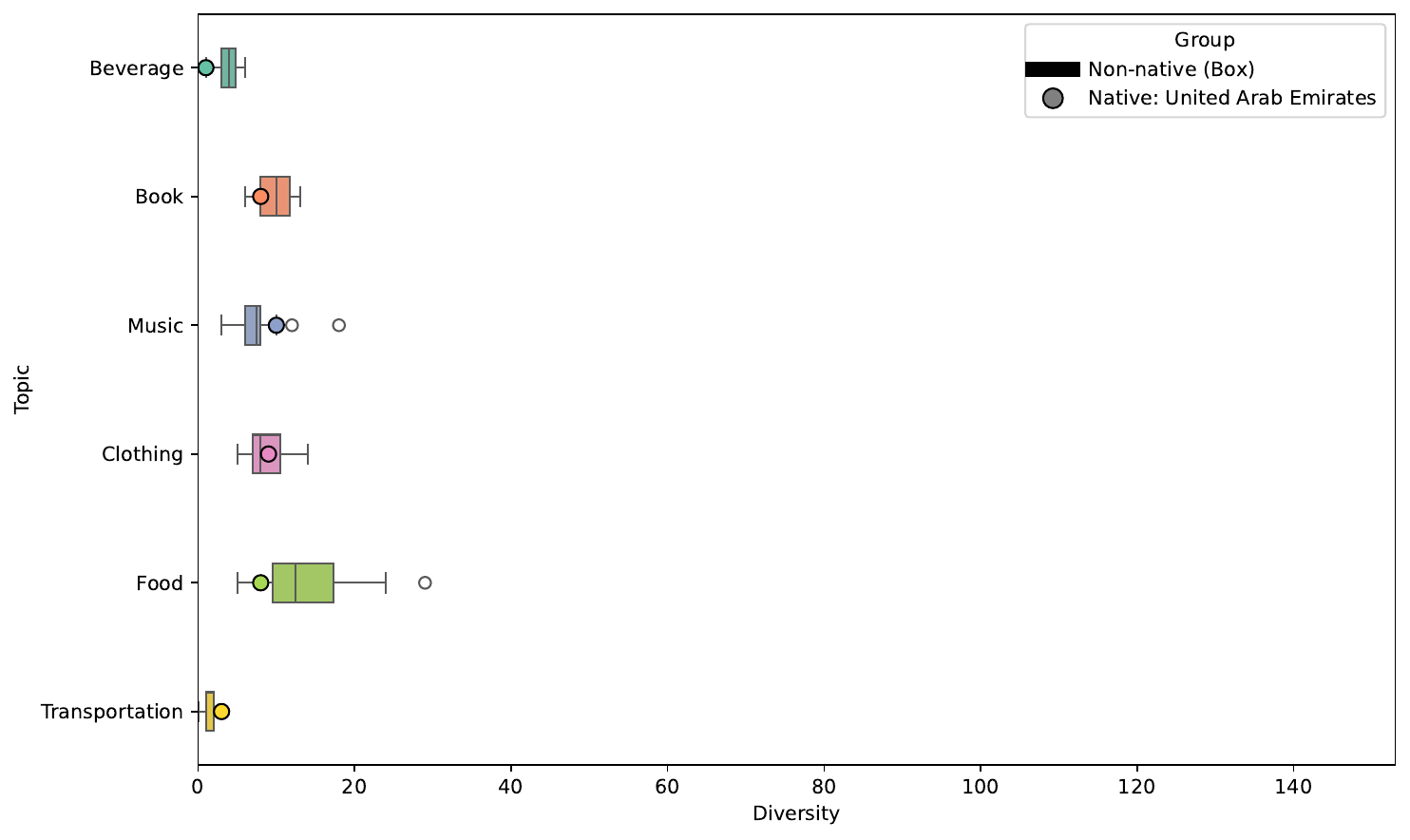} \\
English & German & Arabic \\
\includegraphics[width=0.3\textwidth]{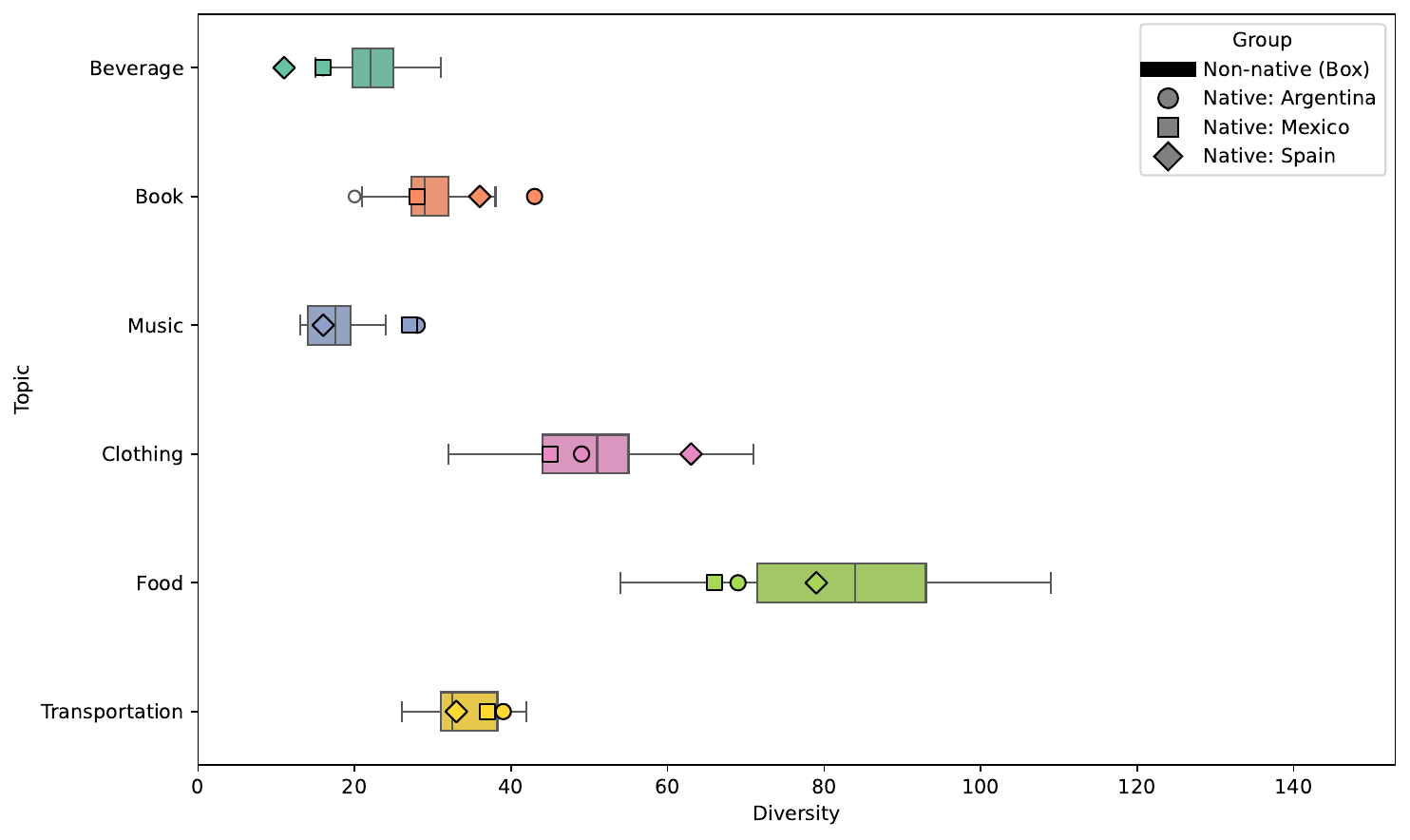} & 
\includegraphics[width=0.3\textwidth]{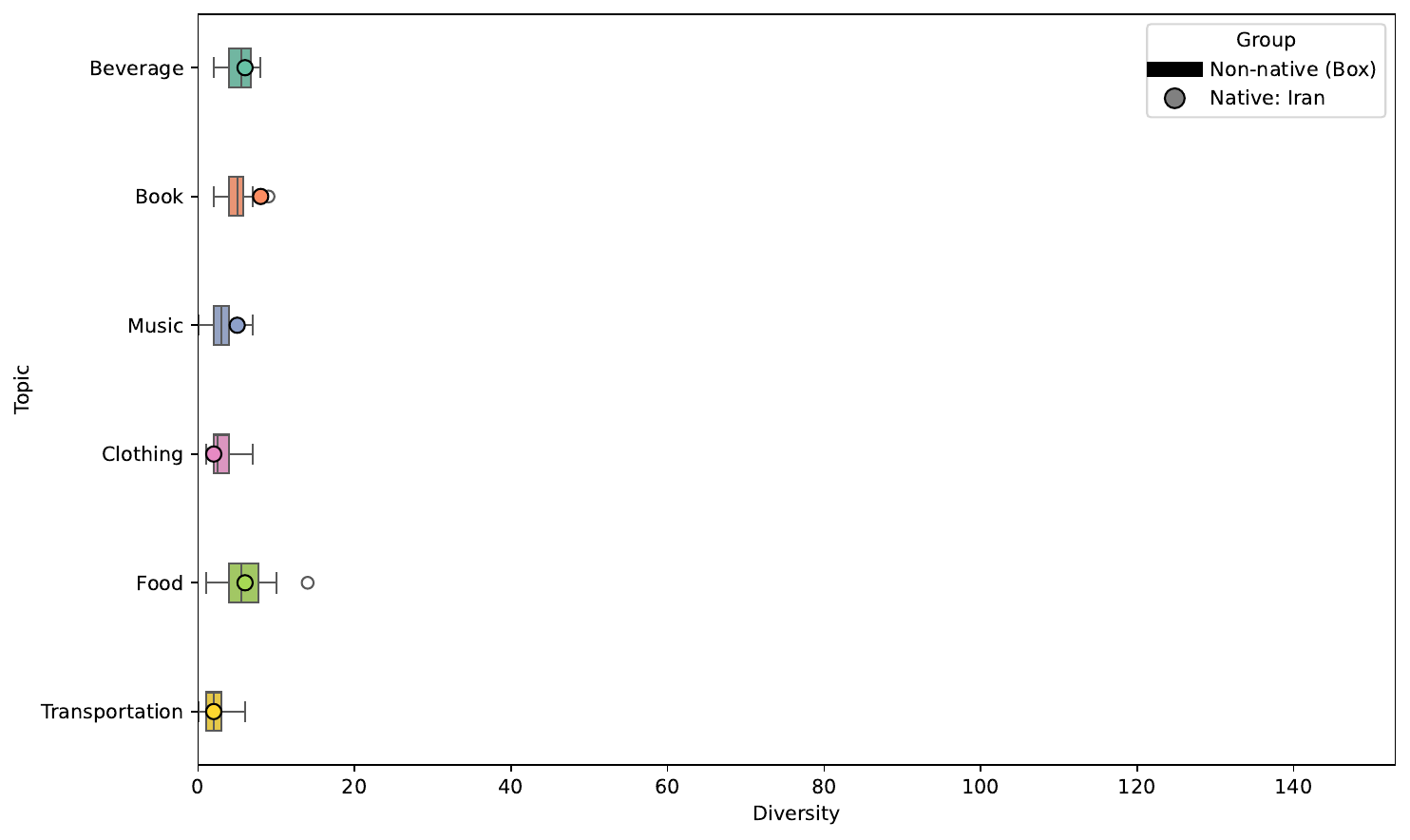} & 
\includegraphics[width=0.3\textwidth]{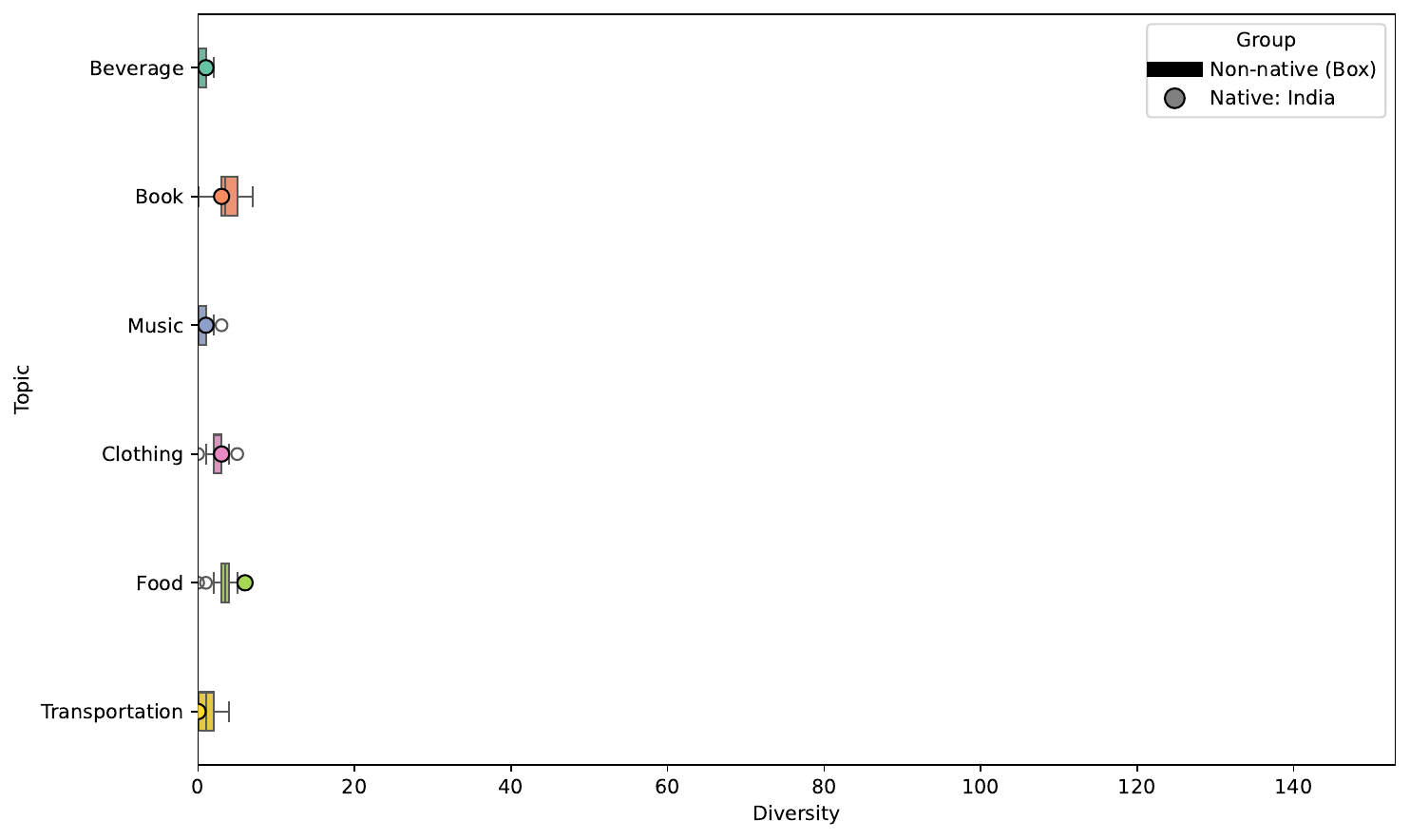} \\
Spanish & Persian & Hindi \\
\includegraphics[width=0.3\textwidth]{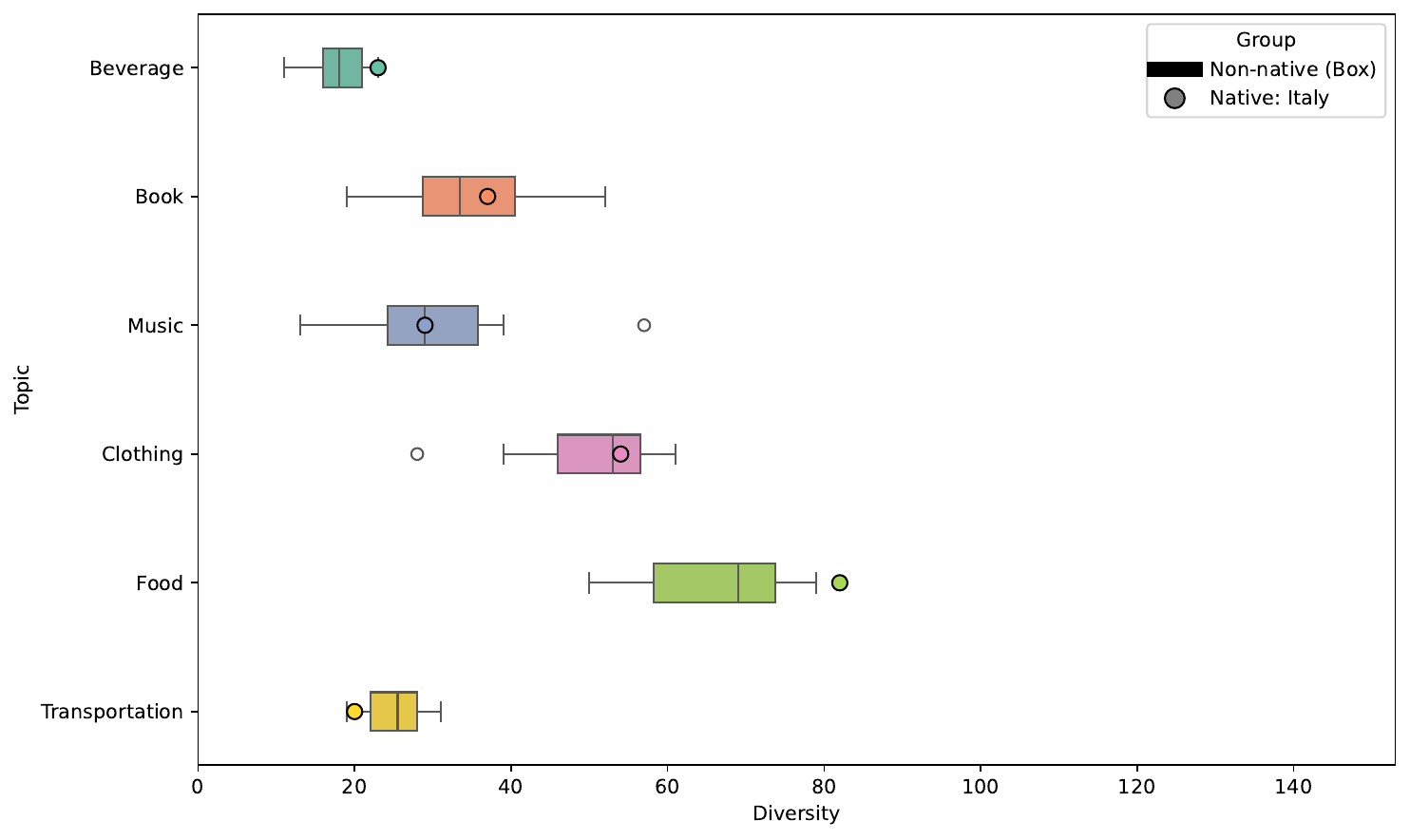} & 
\includegraphics[width=0.3\textwidth]{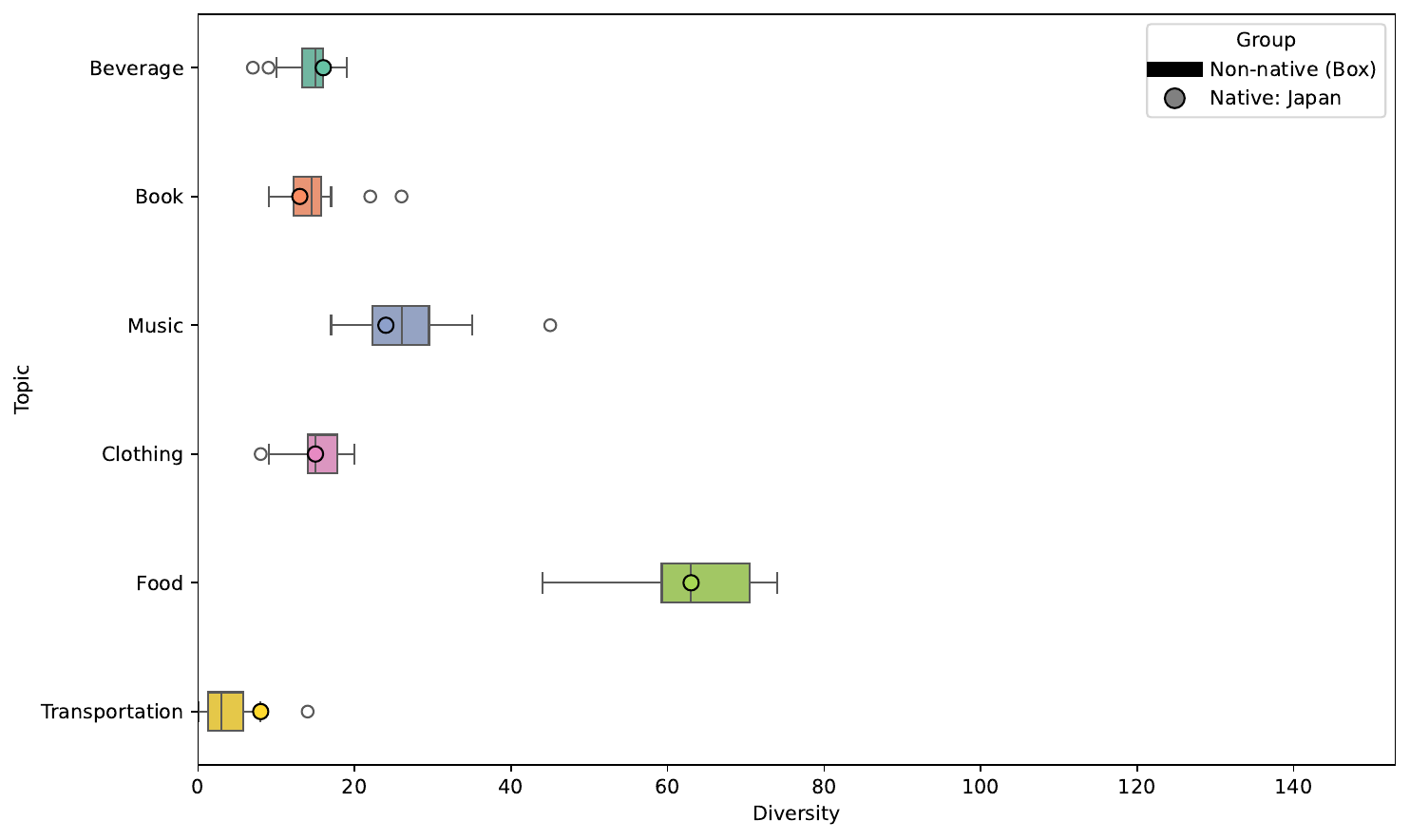} & 
\includegraphics[width=0.3\textwidth]{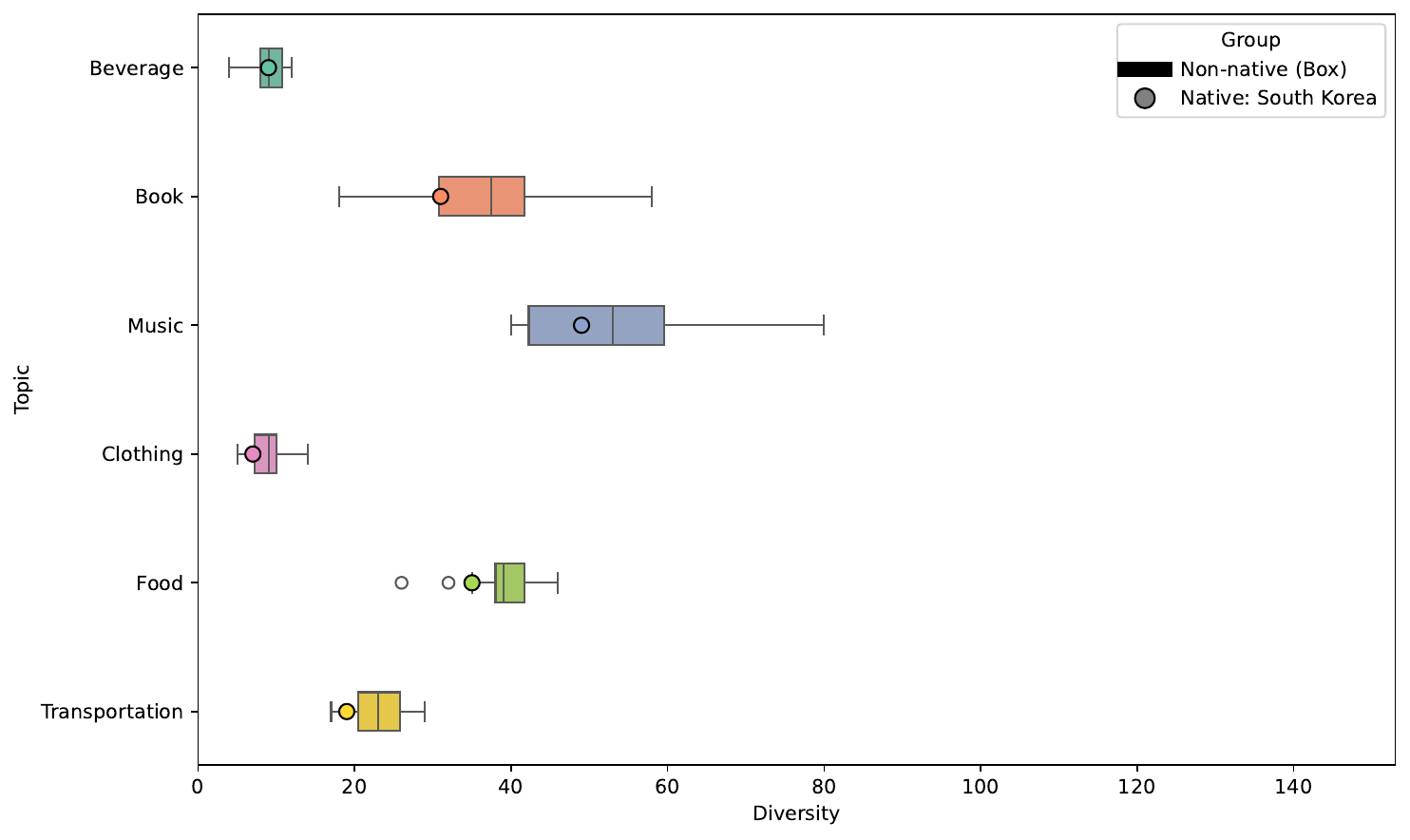} \\
Italian & Japanese & Korean \\
\includegraphics[width=0.3\textwidth]{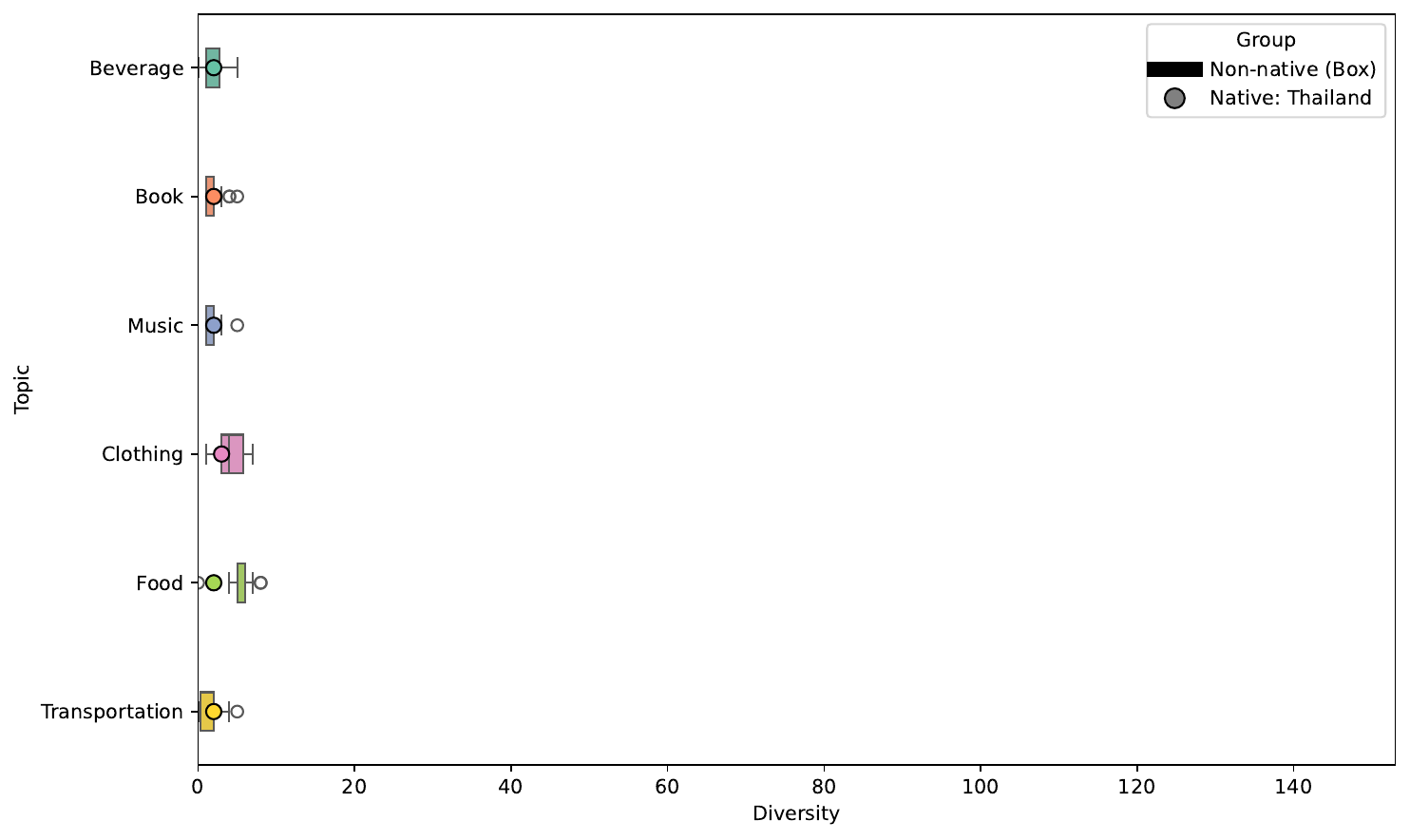} & 
\includegraphics[width=0.3\textwidth]{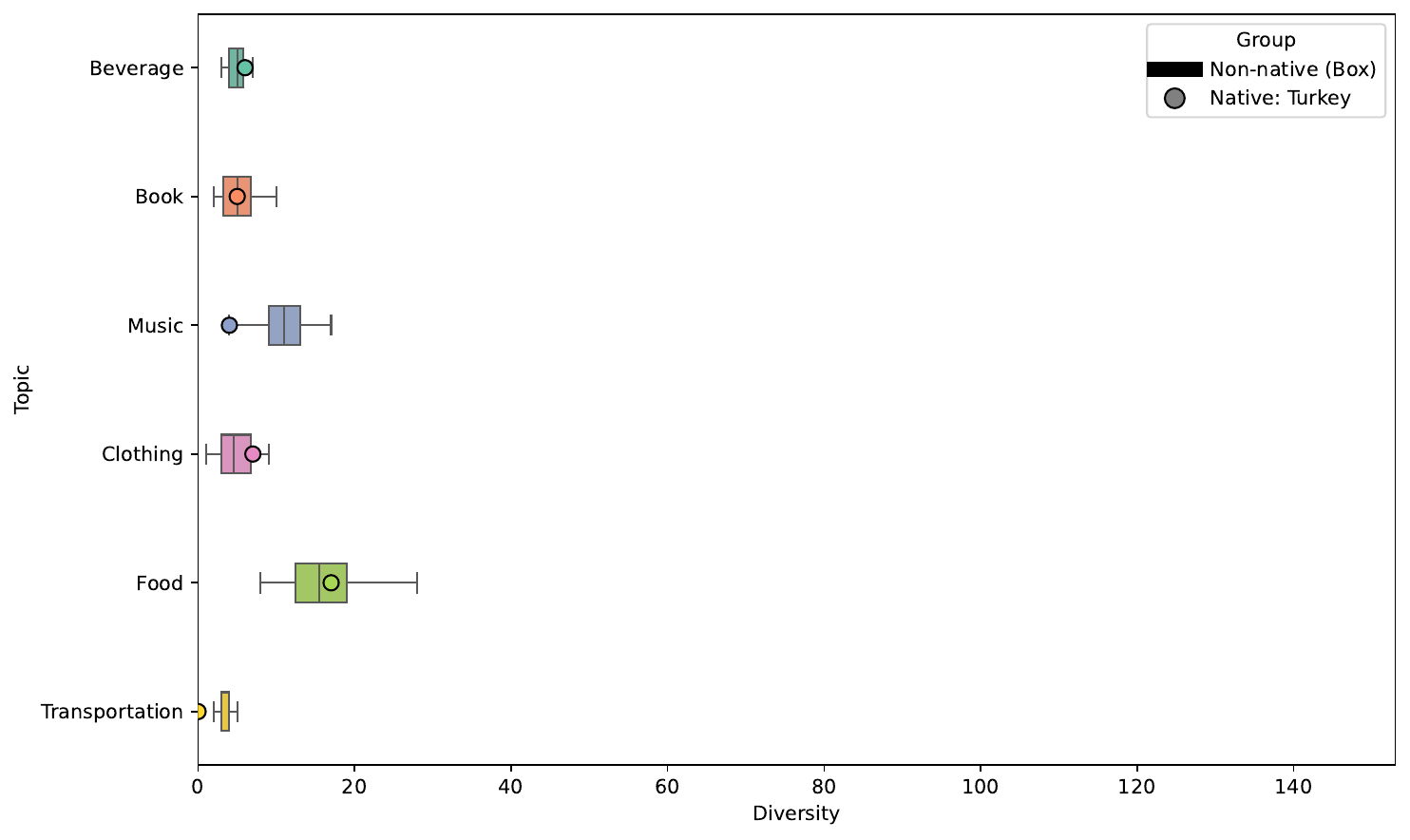} & 
\includegraphics[width=0.3\textwidth]{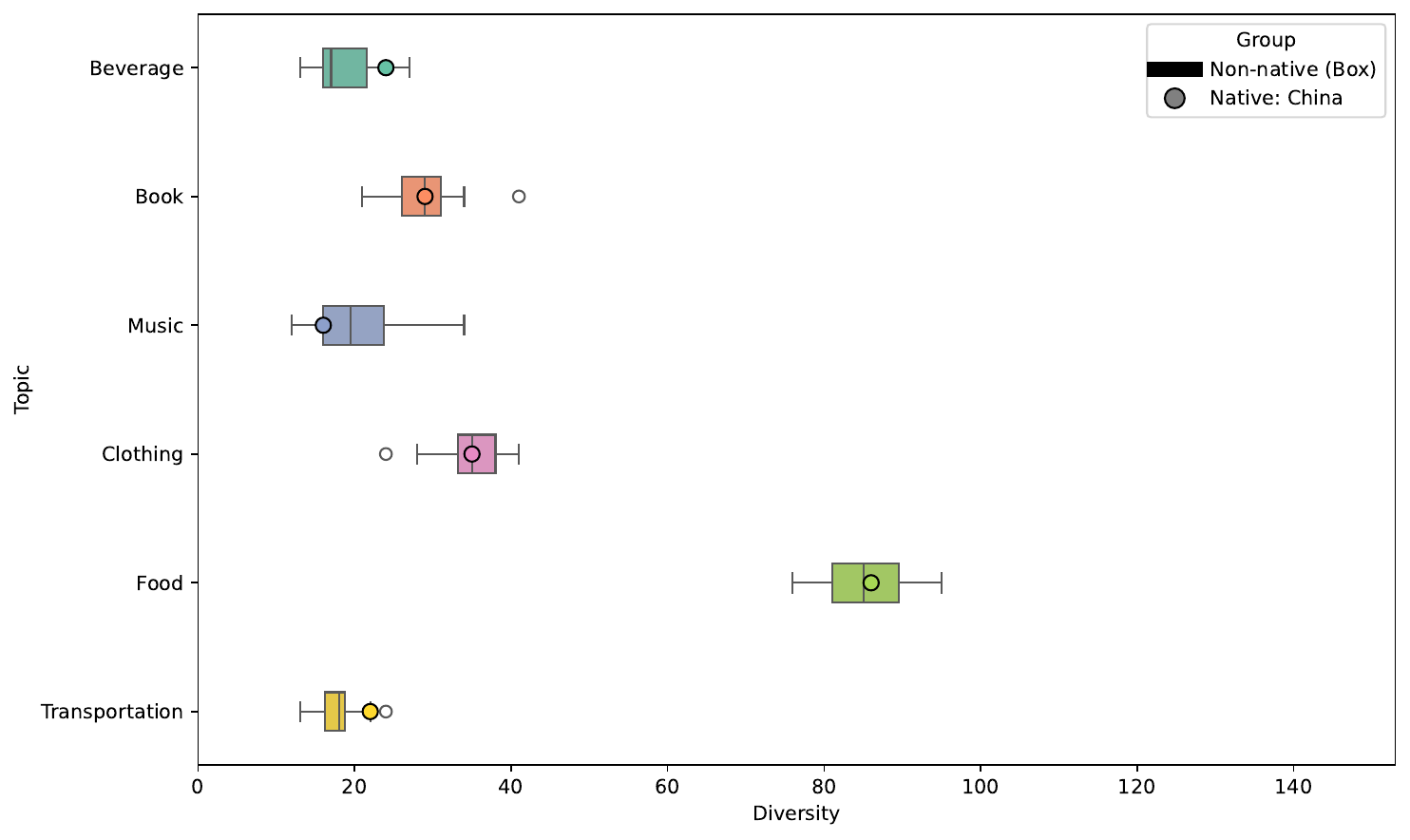} \\
Thai & Turkish & Chinese (Simplified) \\
\includegraphics[width=0.3\textwidth]{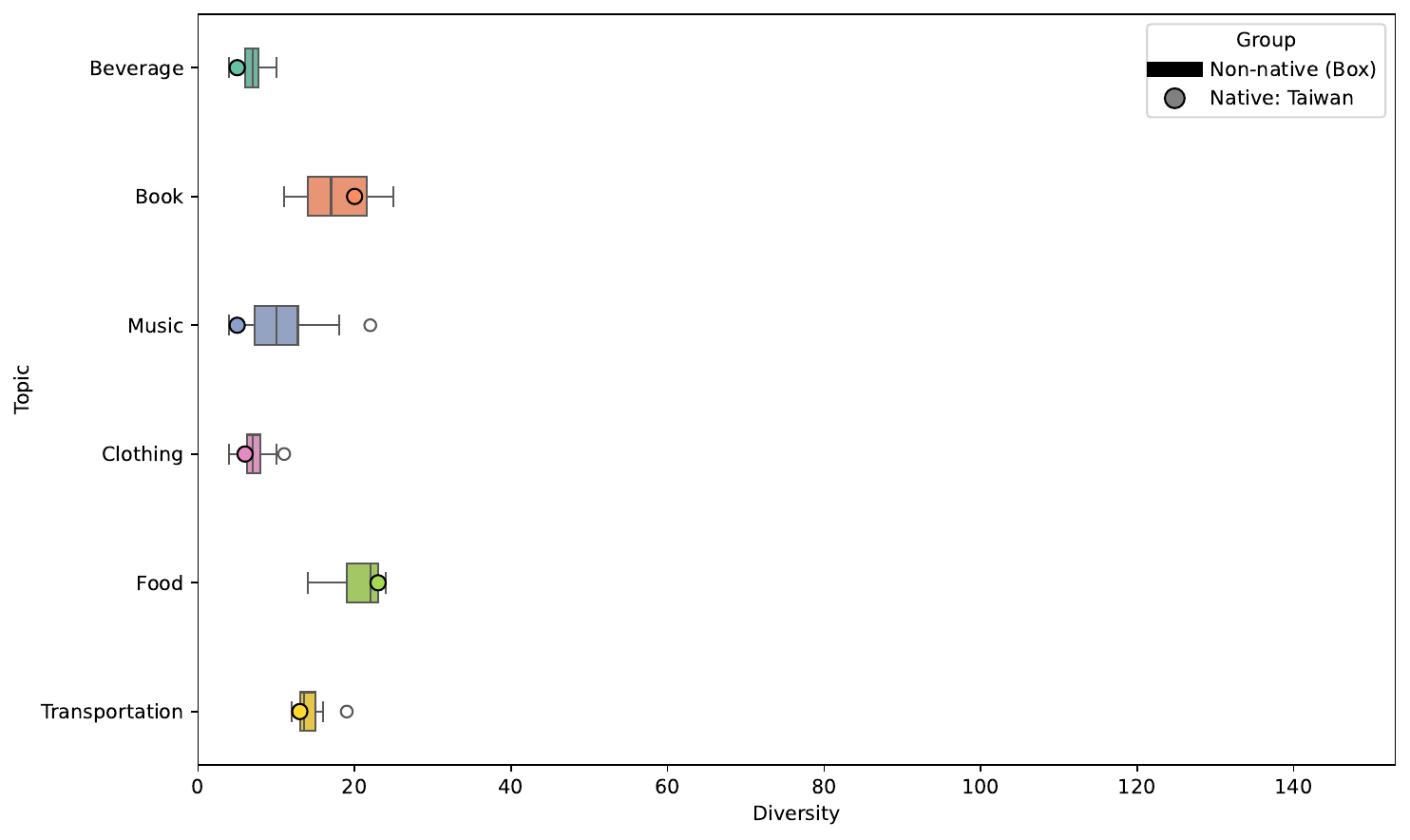} & & \\
Chinese (Traditional) & &
\end{tabular}
\caption{Box plots for diversity comparison between native and non-native languages for \textsc{Mistral} across 13 languages.}
\label{fig:box_mistral}
\end{figure*}

\begin{figure*}[t]
\centering
\resizebox{\textwidth}{!}{
\begin{tabular}{ccc}
\includegraphics[width=0.32\textwidth]{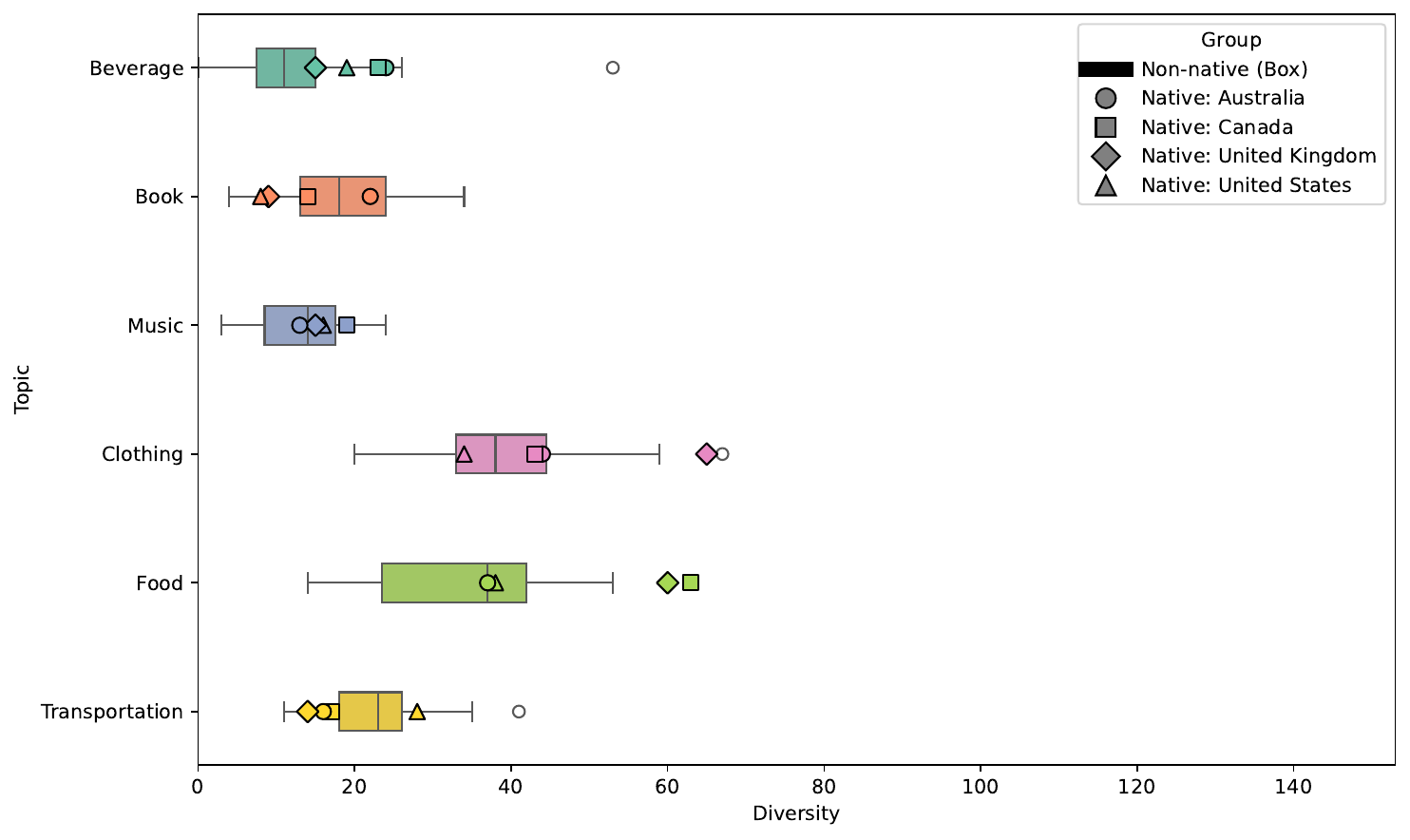} & 
\includegraphics[width=0.32\textwidth]{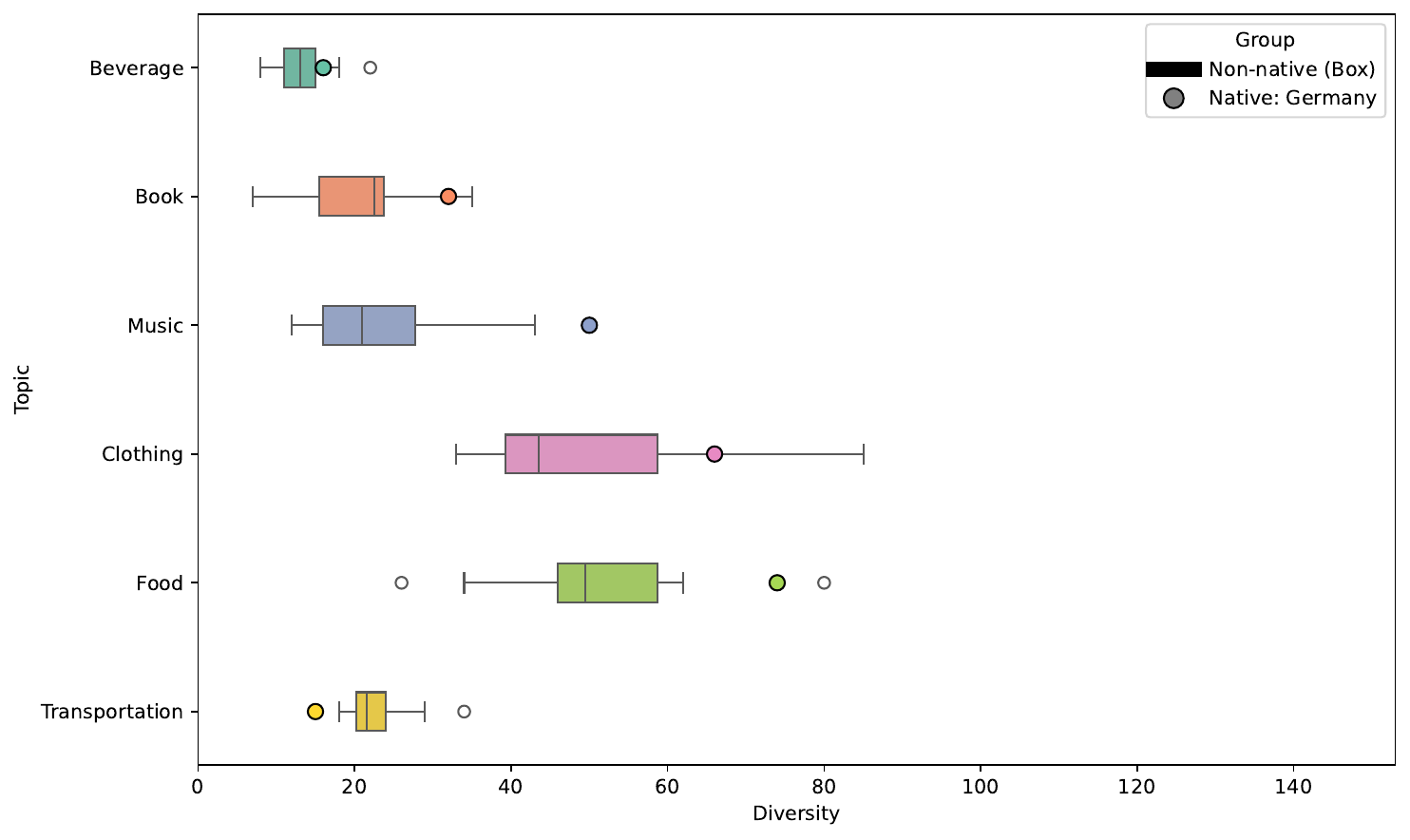} & 
\includegraphics[width=0.32\textwidth]{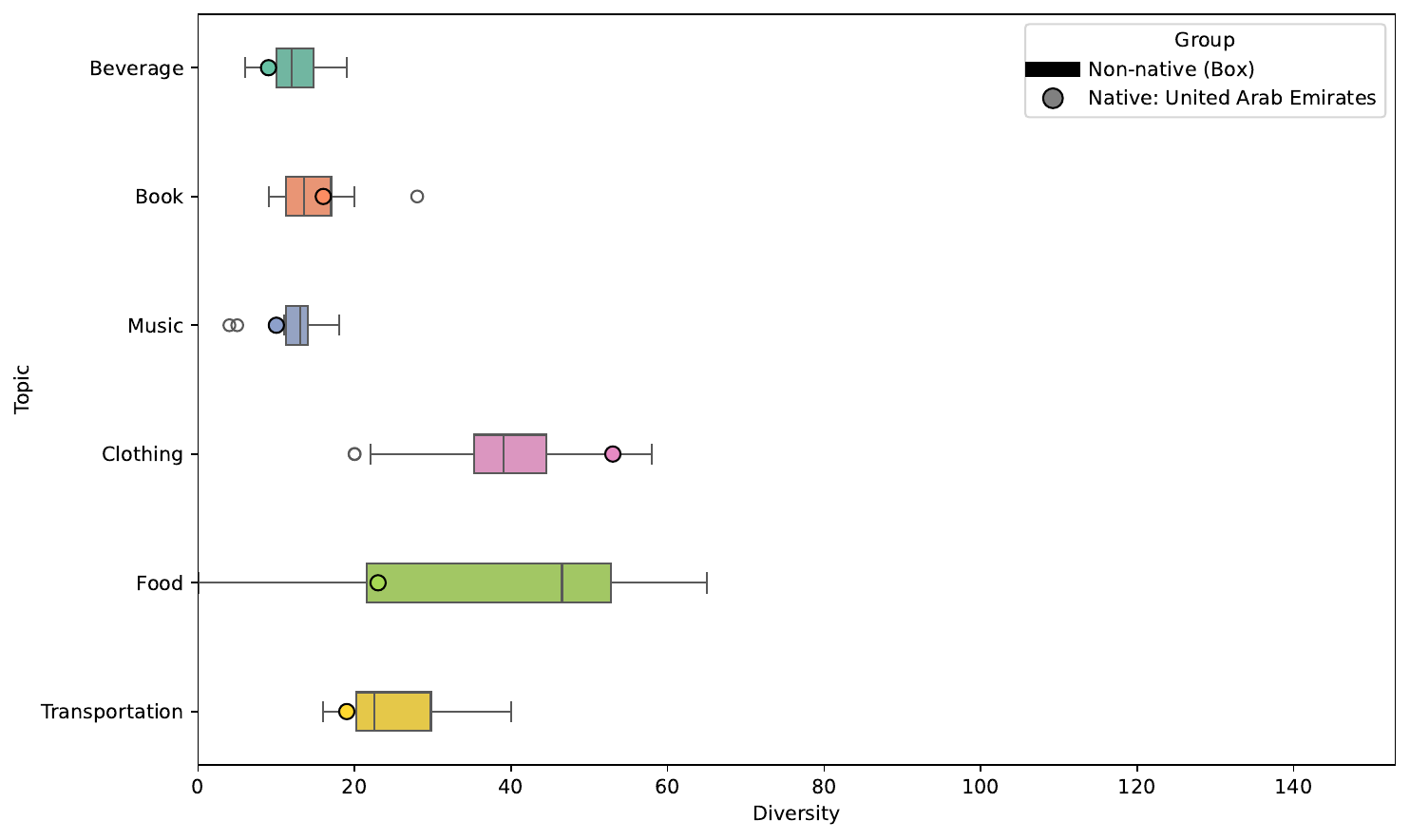} \\
English & German & Arabic \\
\includegraphics[width=0.3\textwidth]{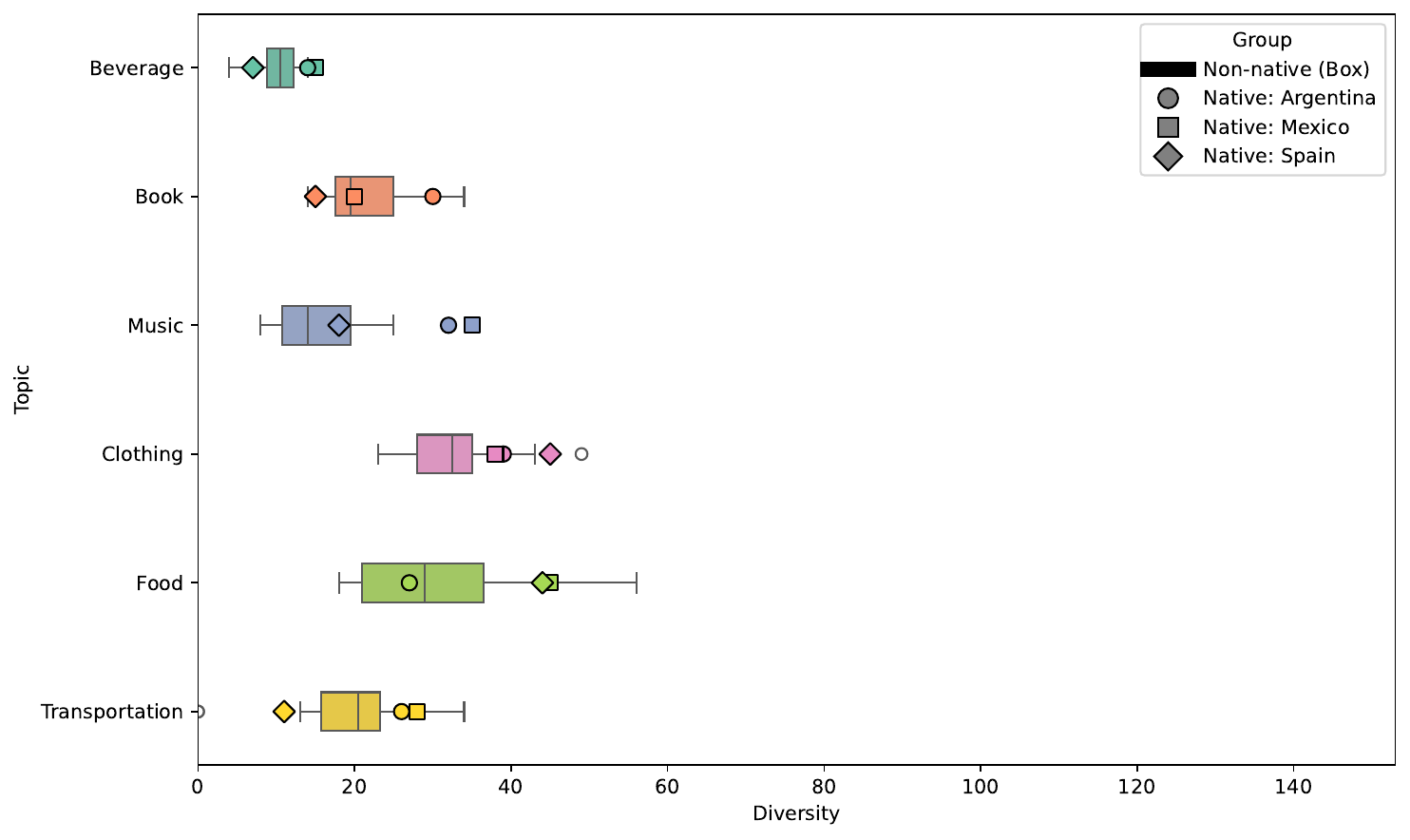} & 
\includegraphics[width=0.3\textwidth]{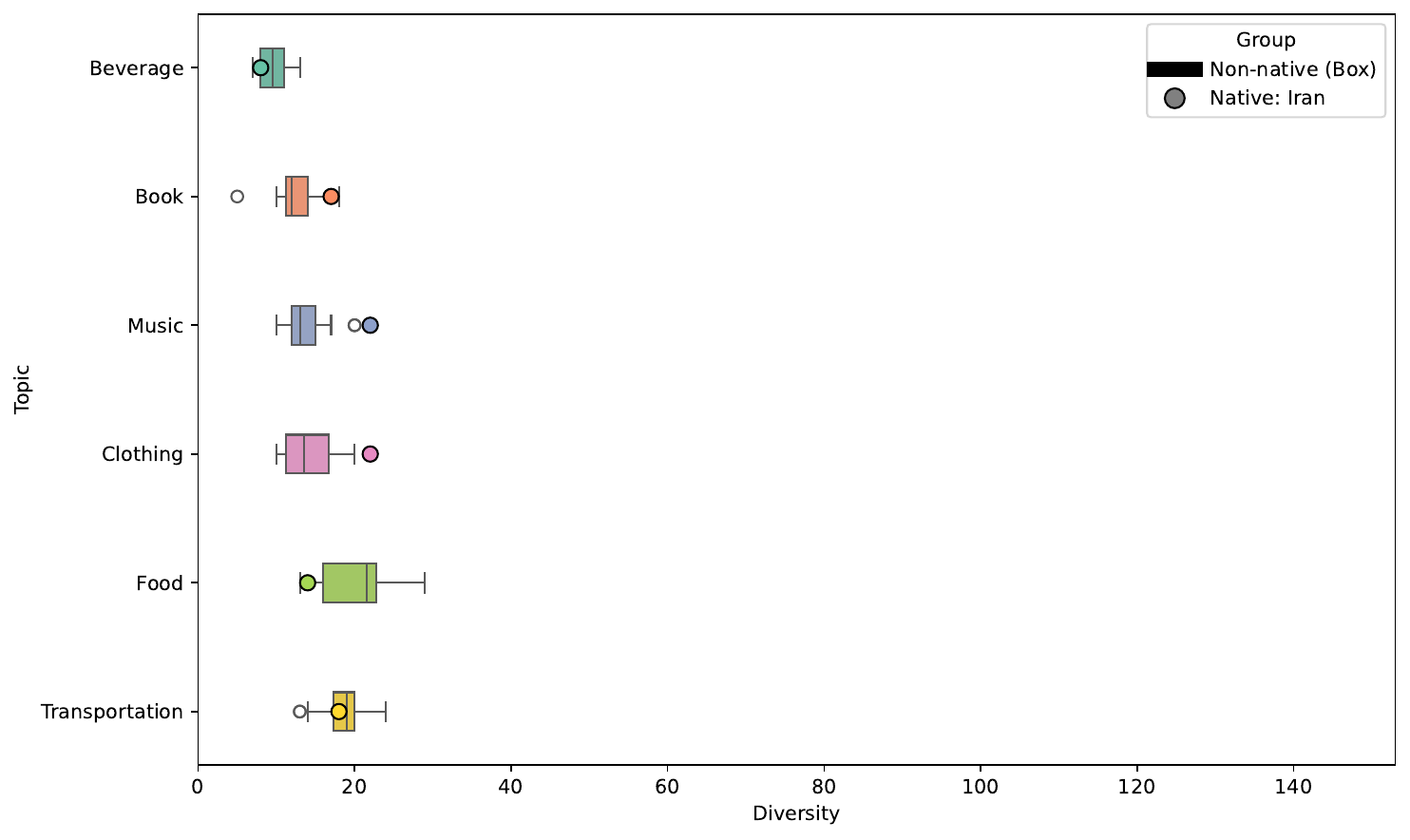} & 
\includegraphics[width=0.3\textwidth]{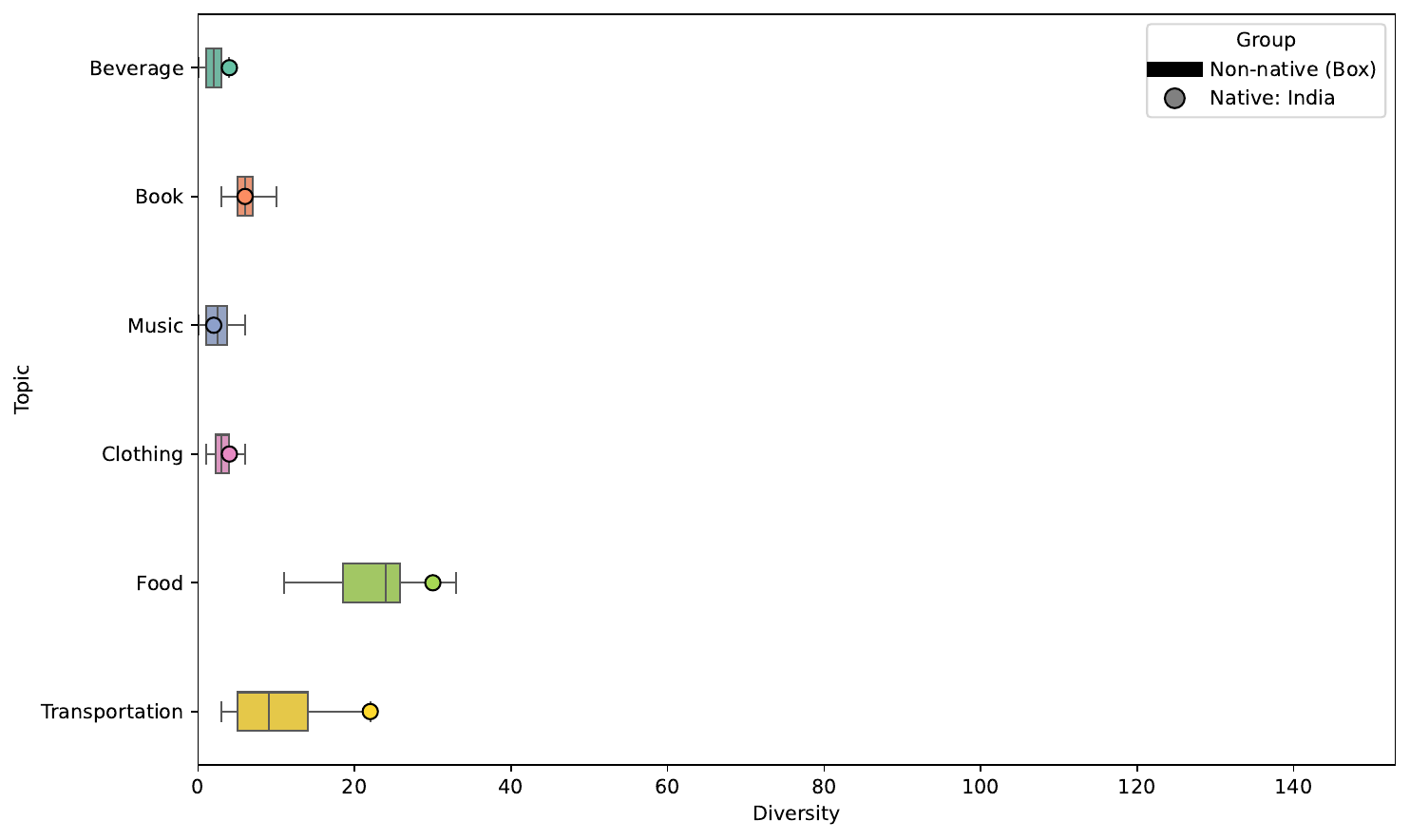} \\
Spanish & Persian & Hindi \\
\includegraphics[width=0.3\textwidth]{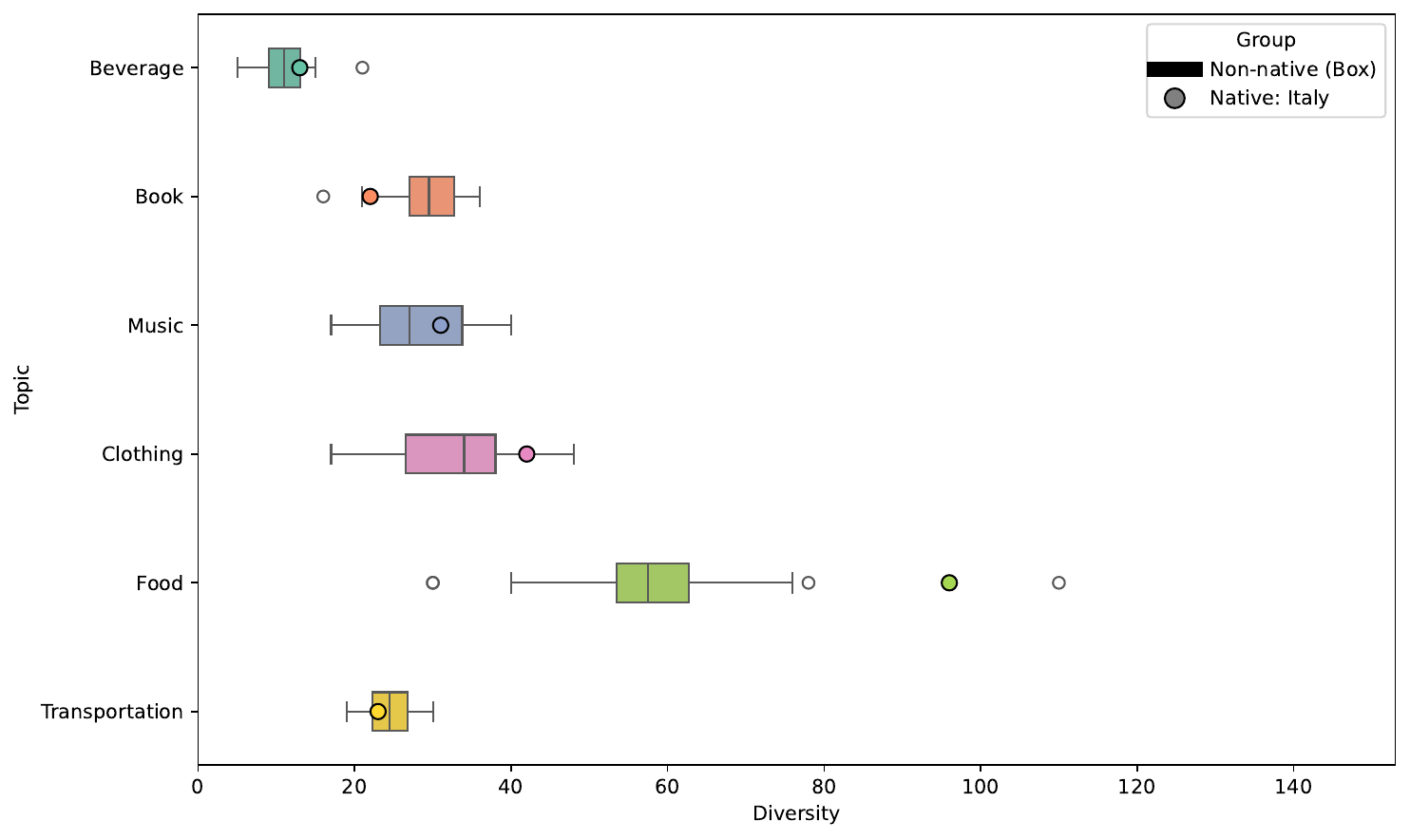} & 
\includegraphics[width=0.3\textwidth]{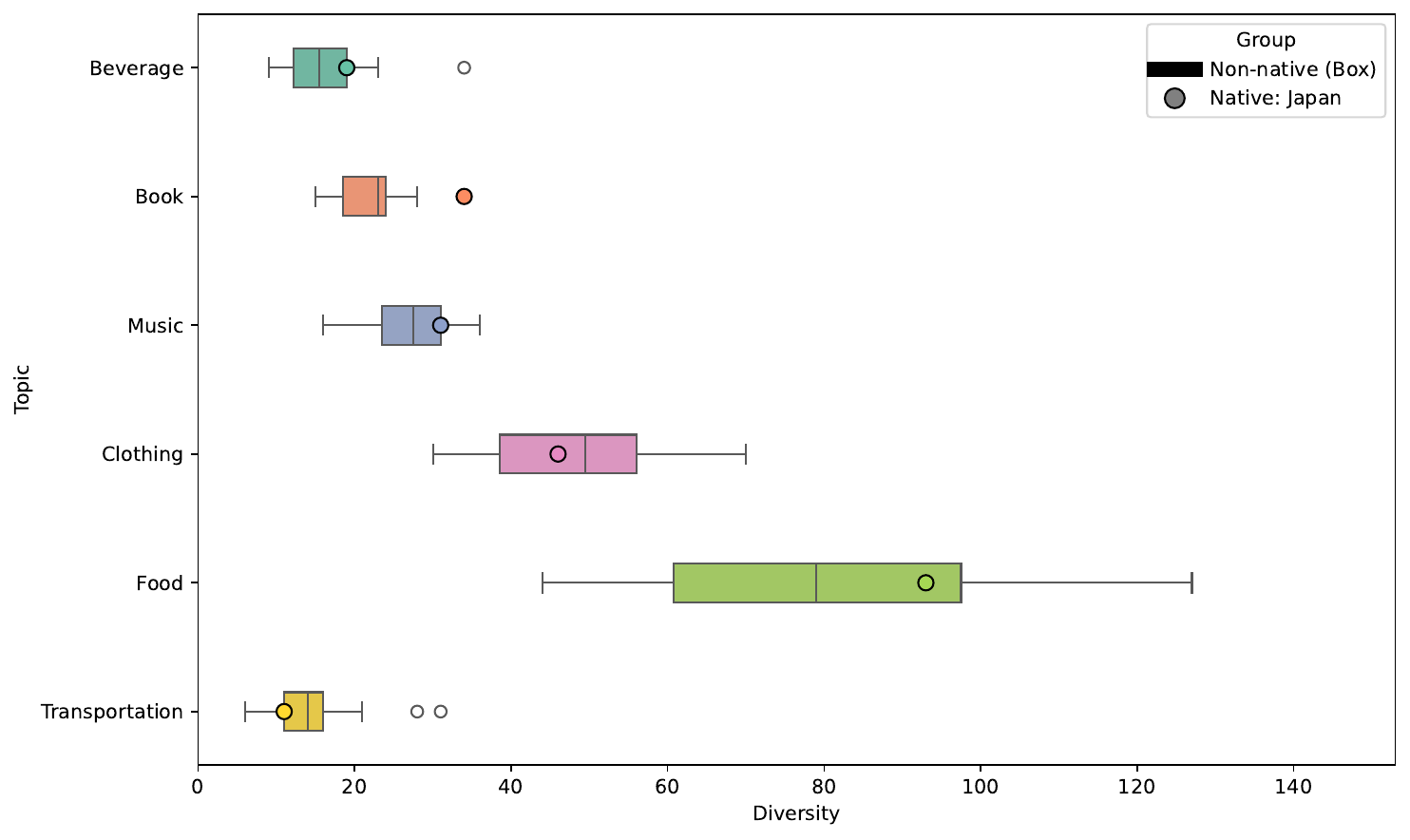} & 
\includegraphics[width=0.3\textwidth]{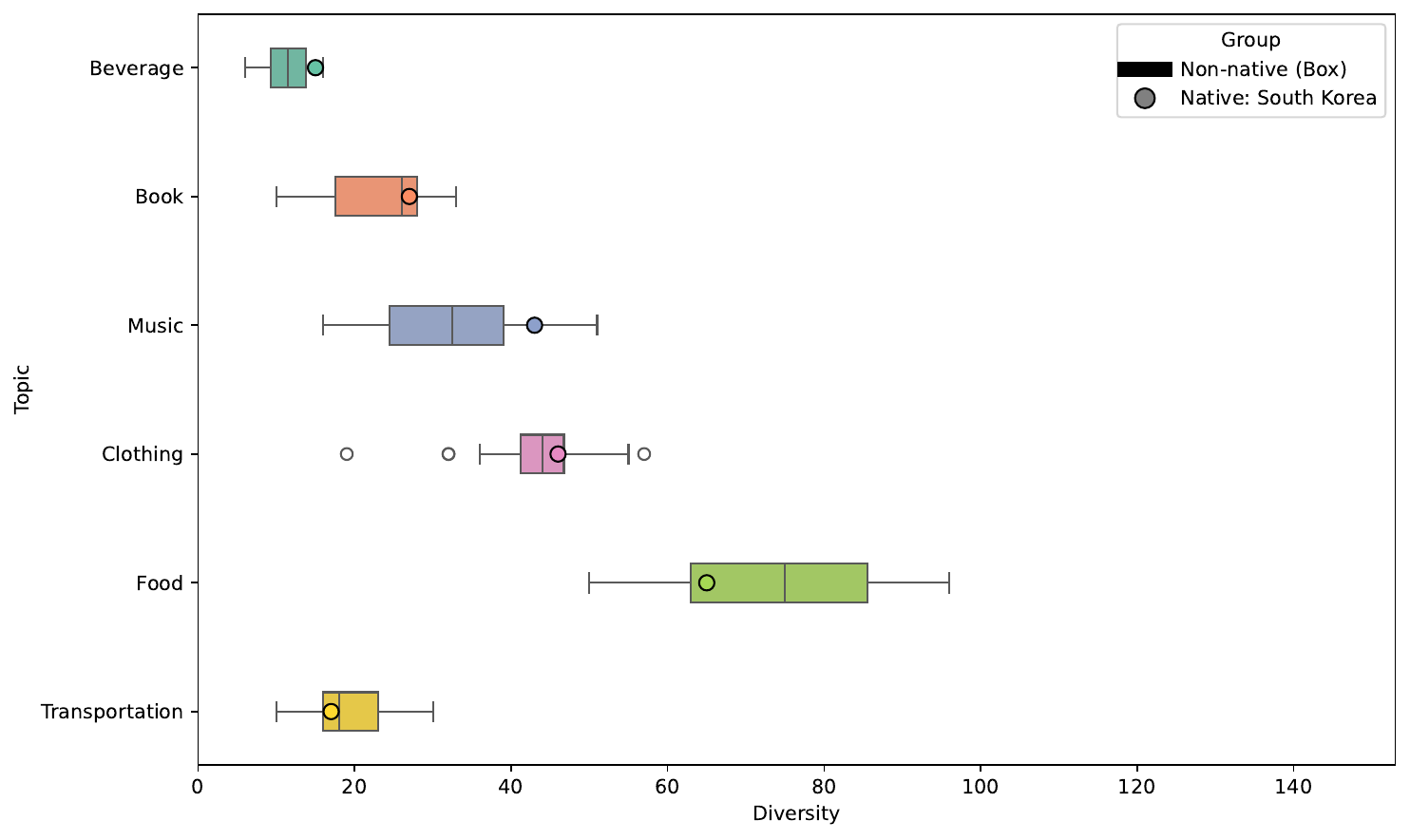} \\
Italian & Japanese & Korean \\
\includegraphics[width=0.3\textwidth]{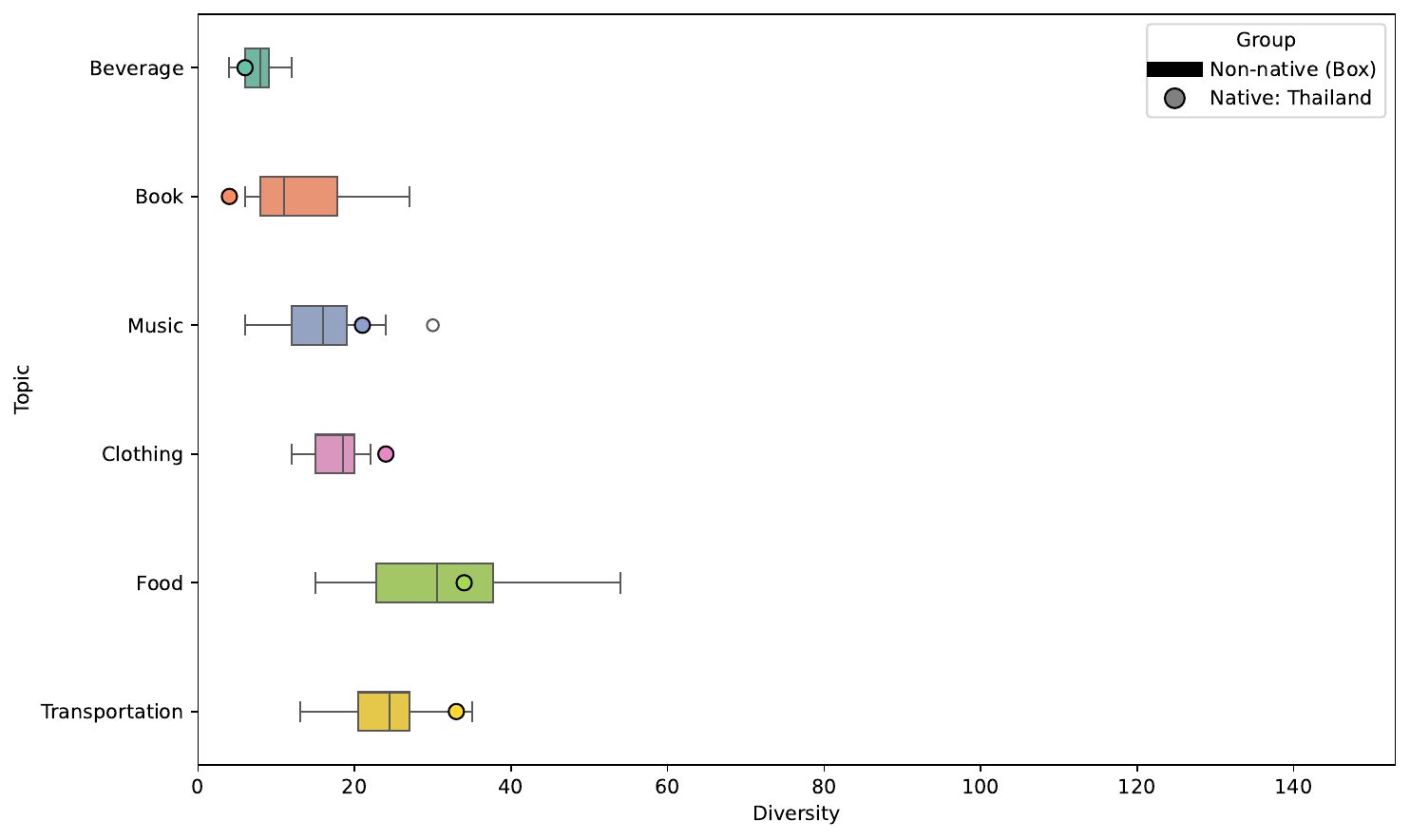} & 
\includegraphics[width=0.3\textwidth]{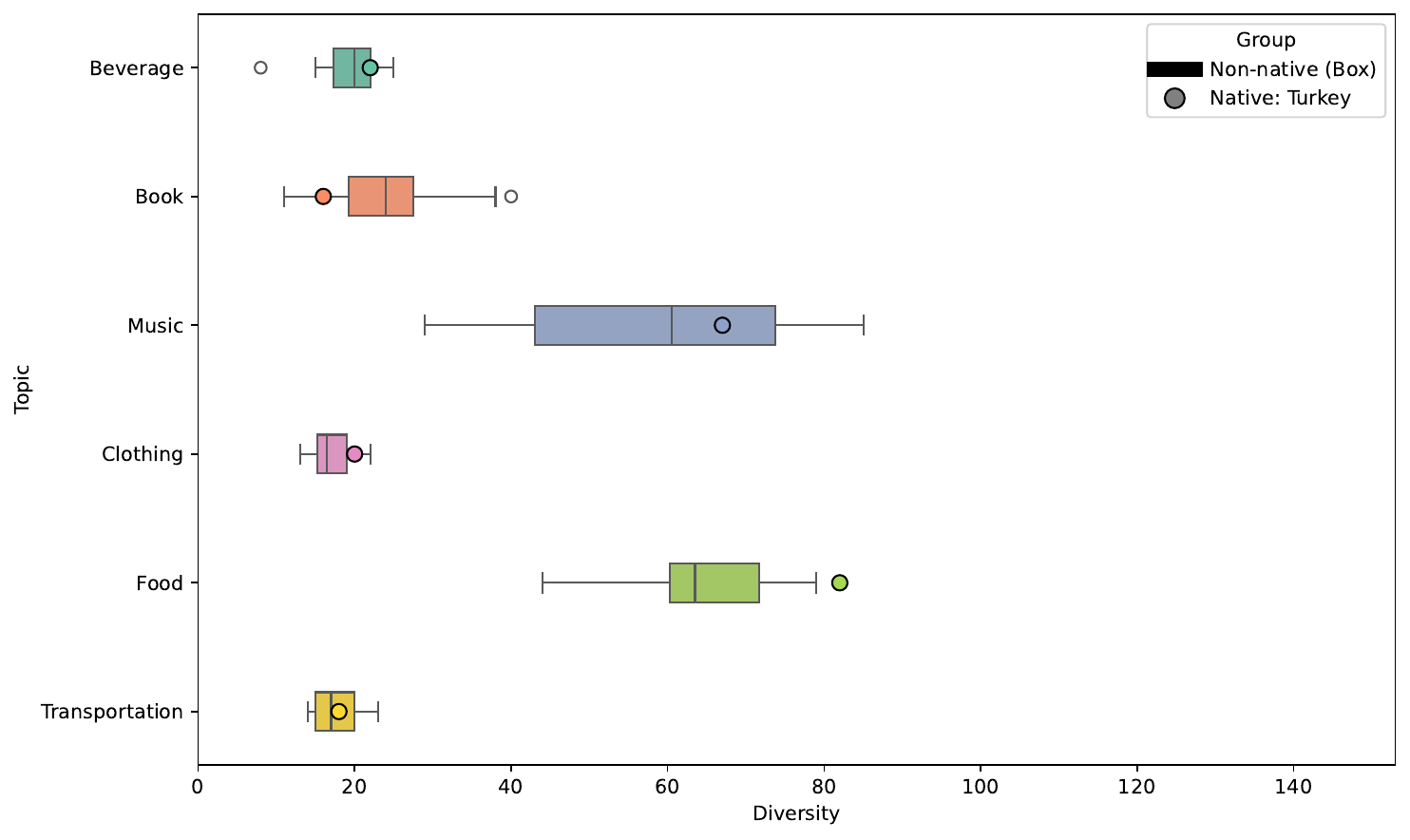} & 
\includegraphics[width=0.3\textwidth]{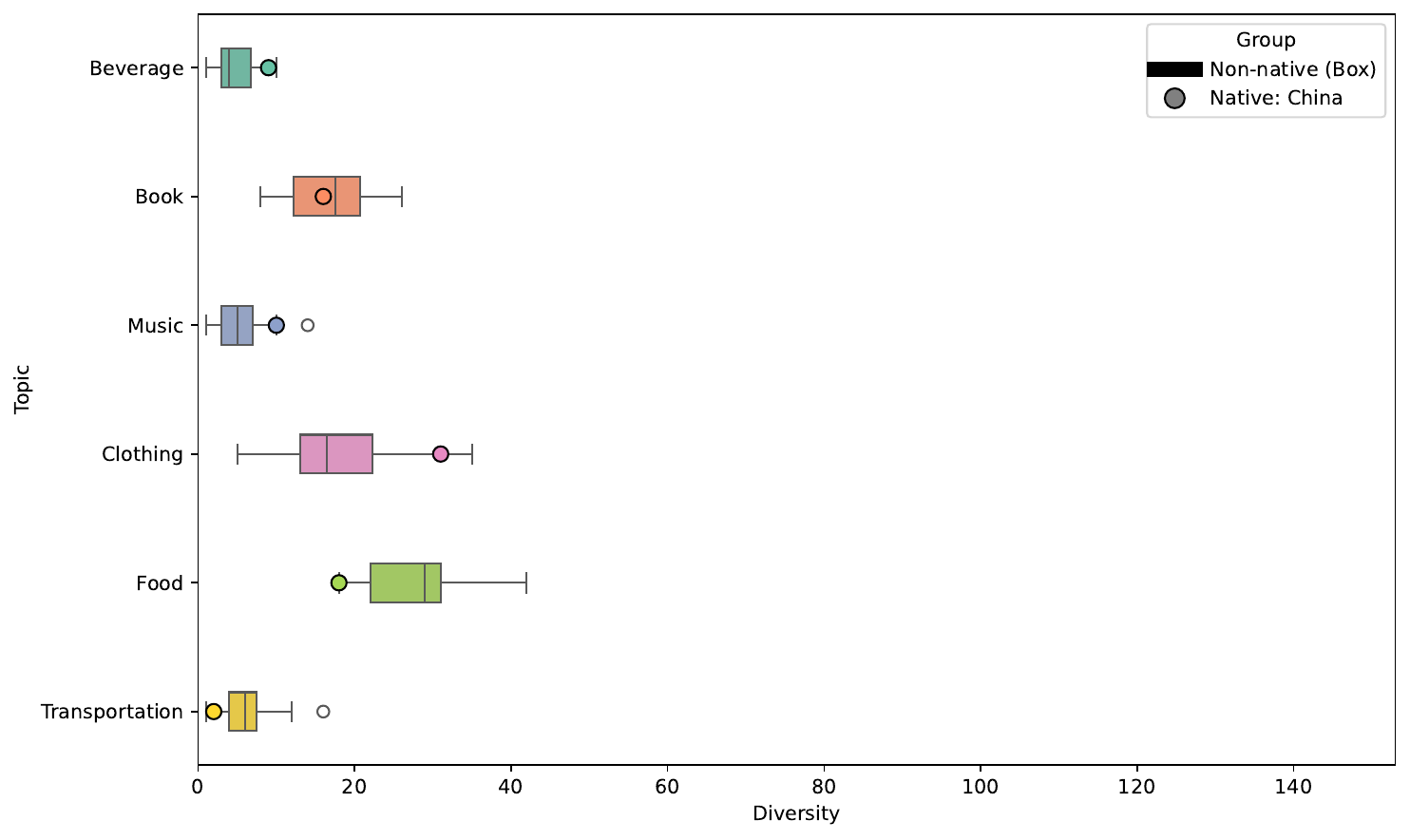} \\
Thai & Turkish & Chinese (Simplified) \\
\includegraphics[width=0.3\textwidth]{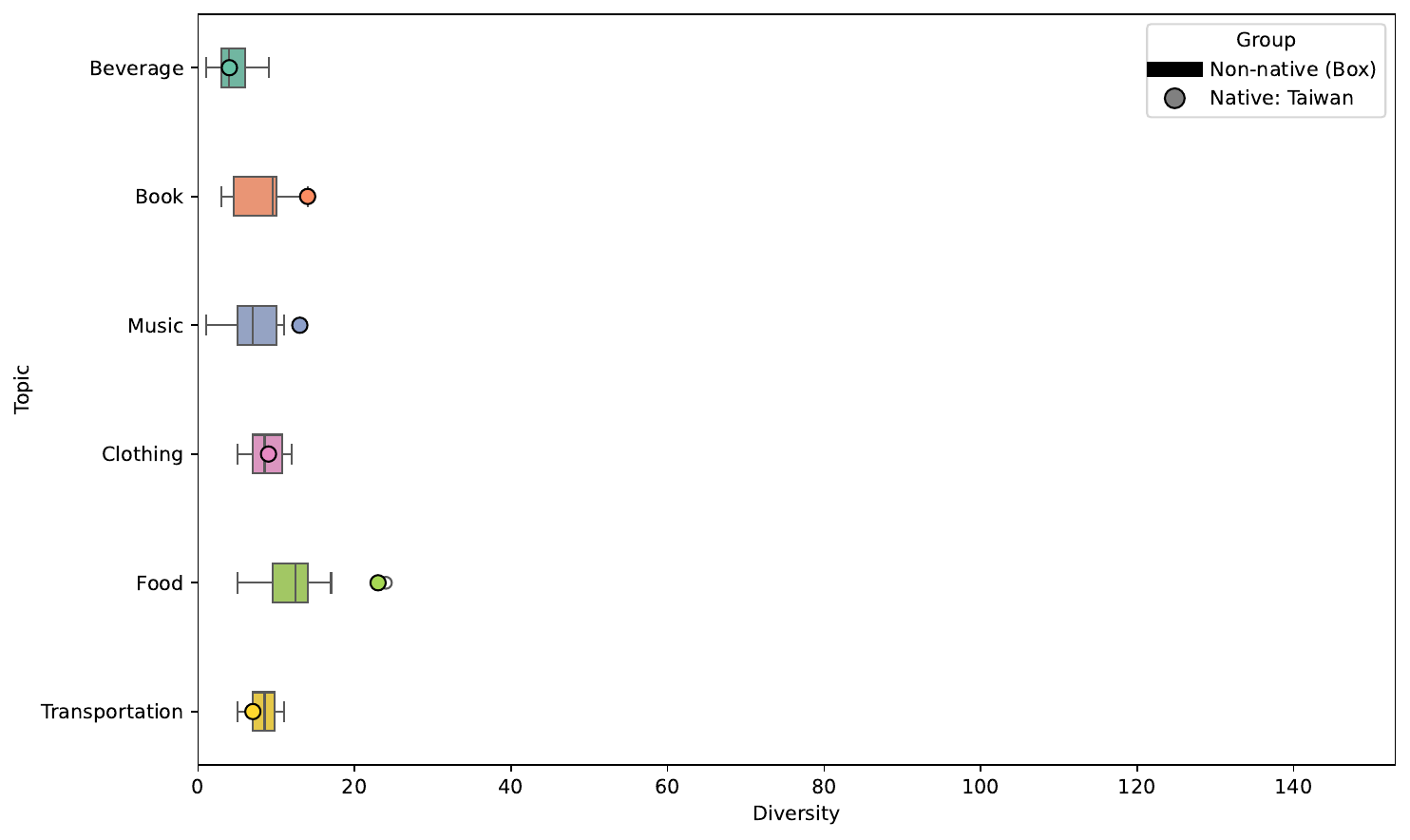} & & \\
Chinese (Traditional) & &
\end{tabular}}
\caption{Box plots for diversity comparison between native and non-native languages for \textsc{Qwen} across 13 languages.}
\label{fig:box_qwen}
\end{figure*}

\section{Results of Culture Specificity}
\label{sec:specificity}
Figures~\ref{fig:culturespec_aya}–\ref{fig:culturespec_qwen} illustrate the culture specificity of each model, computed by averaging results across all six topics. In this analysis, culture specificity is defined as the difference in entity representations between culturally contextualized prompts (i.e., those mentioning a specific country or region) and neutral prompts. A higher value indicates that the model adapts its output more strongly in response to the cultural cues embedded in the prompt.

In each figure, black-outlined cells highlight cases where the prompt language and the mentioned country form a native pairing. Interestingly, many of these native language–country pairs already exhibit relatively high specificity under the neutral prompt condition, suggesting that language itself can act as an implicit cultural signal, even in the absence of explicit country references.

Across all models, a consistent trend emerges: English exhibits the highest specificity gain, indicating that models tend to adjust their outputs most when prompts are in English. This may reflect English’s central role in pretraining corpora and its exposure to culturally diverse contexts. Additionally, as shown in Figure~\ref{fig:culturespec_qwen}, Chinese achieves a high specificity gain in the Qwen model, potentially reflecting stronger native-language adaptation capabilities.
\begin{figure*}[htbp]
  \centering
  \includegraphics[width=\columnwidth]{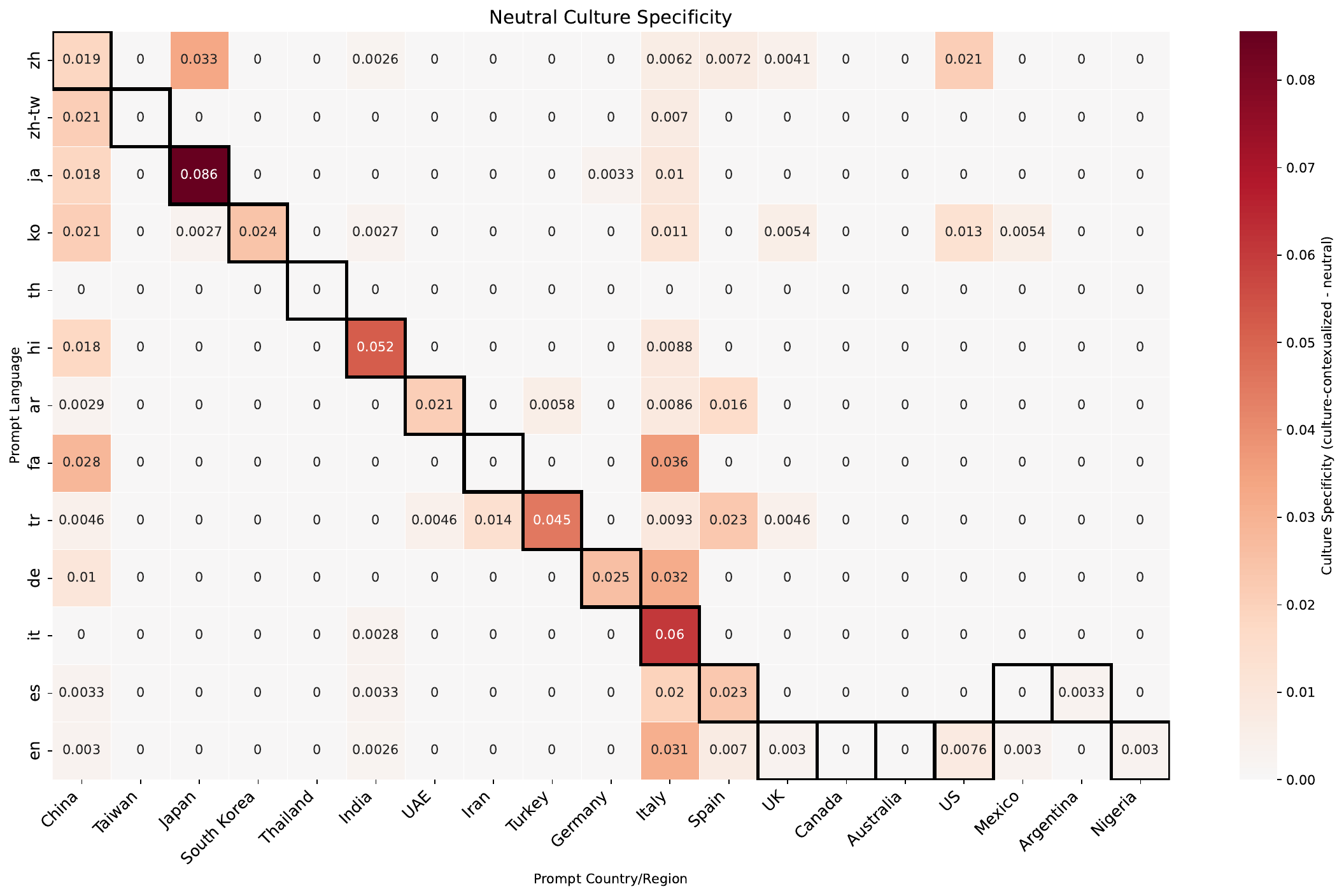}
  \vspace{0.8em}
  \includegraphics[width=\columnwidth]{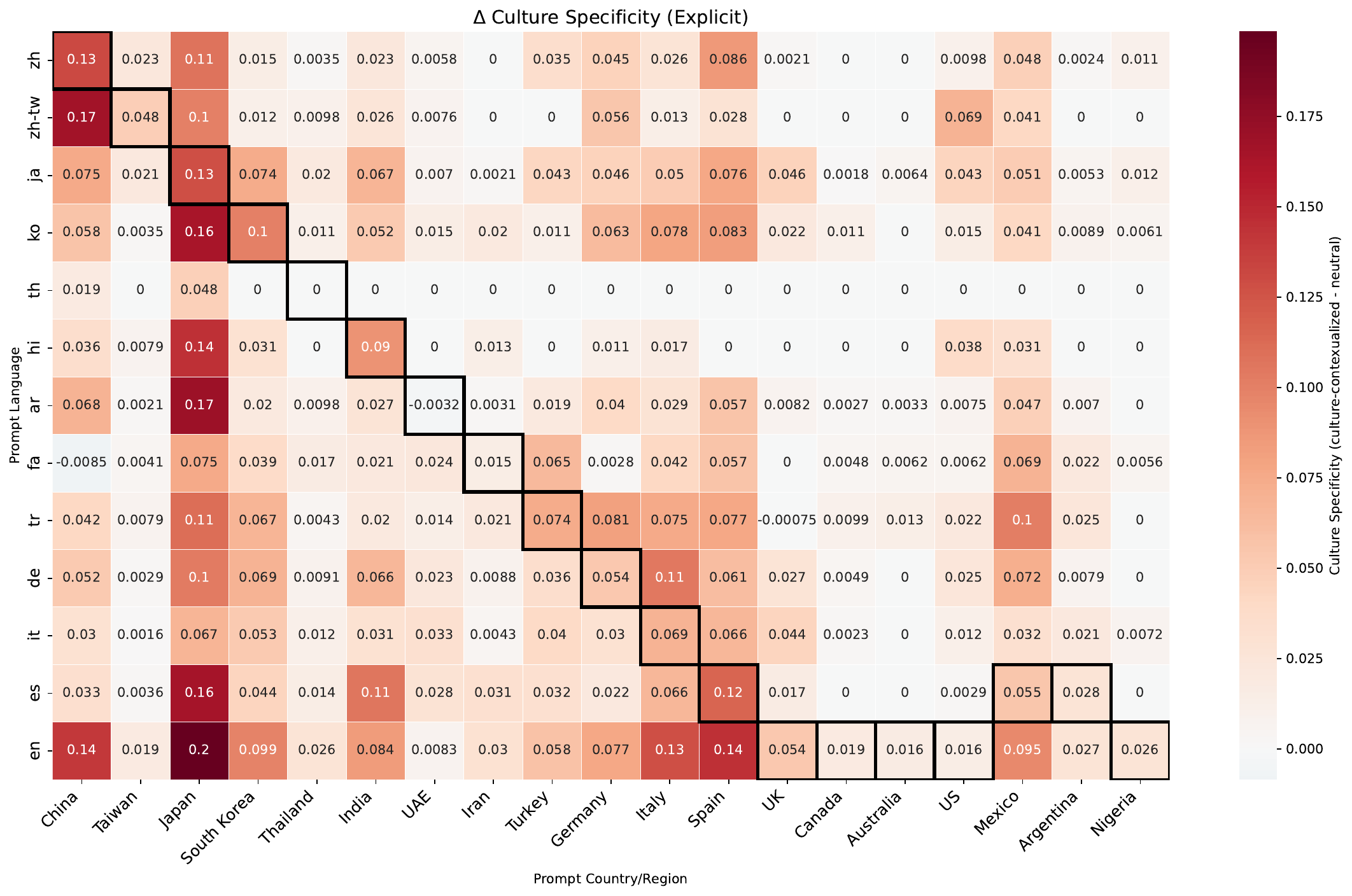}\\

  \caption{Culture specificity results for \textbf{Aya}  with a neutral prompt (left) and a prompt explicitly mentioning the country (right, showing delta to neutral).}
  \label{fig:culturespec_aya}
\end{figure*}

\begin{figure*}[htbp]
  \centering
  \includegraphics[width=\columnwidth]{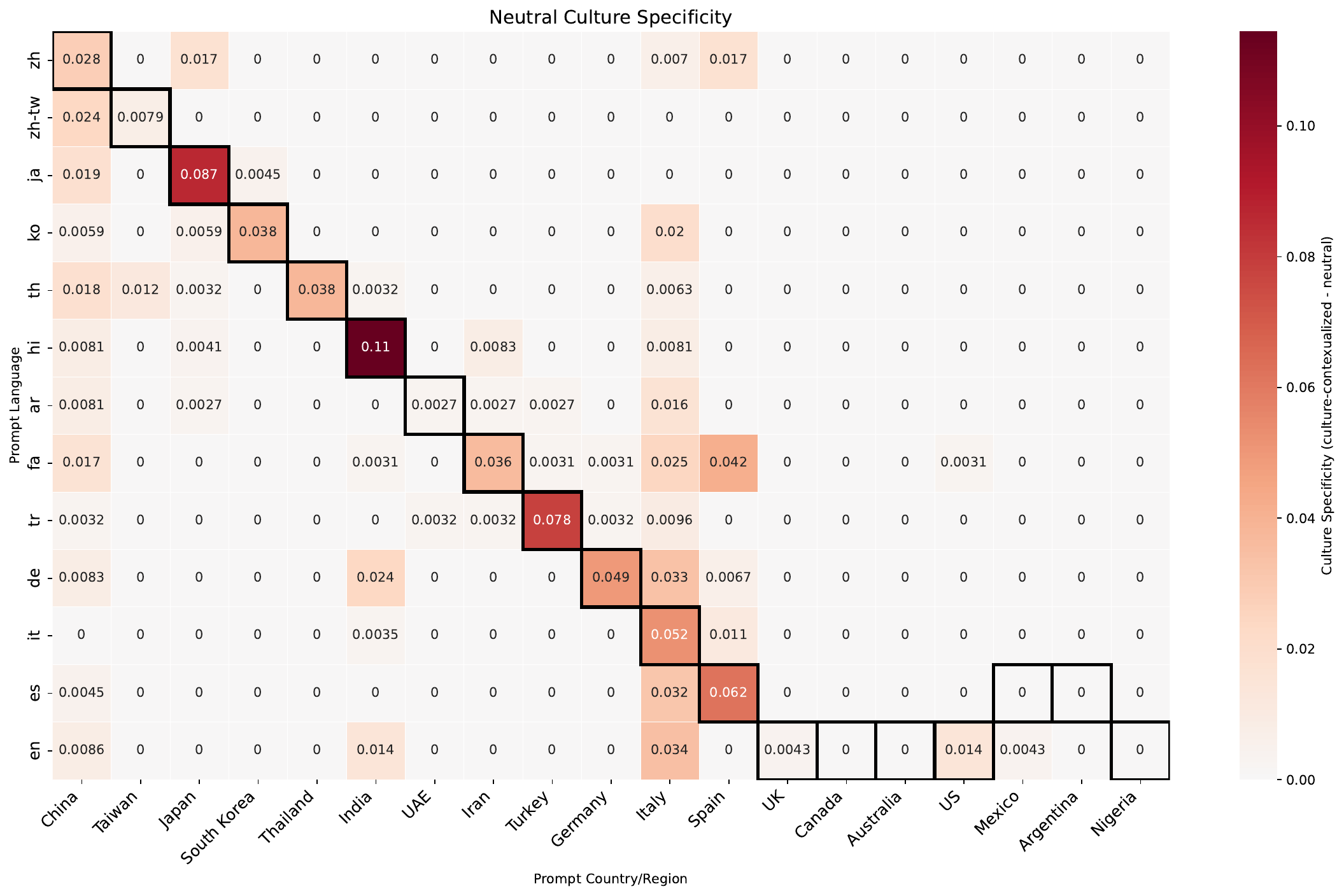}
  \vspace{0.8em}
  \includegraphics[width=\columnwidth]{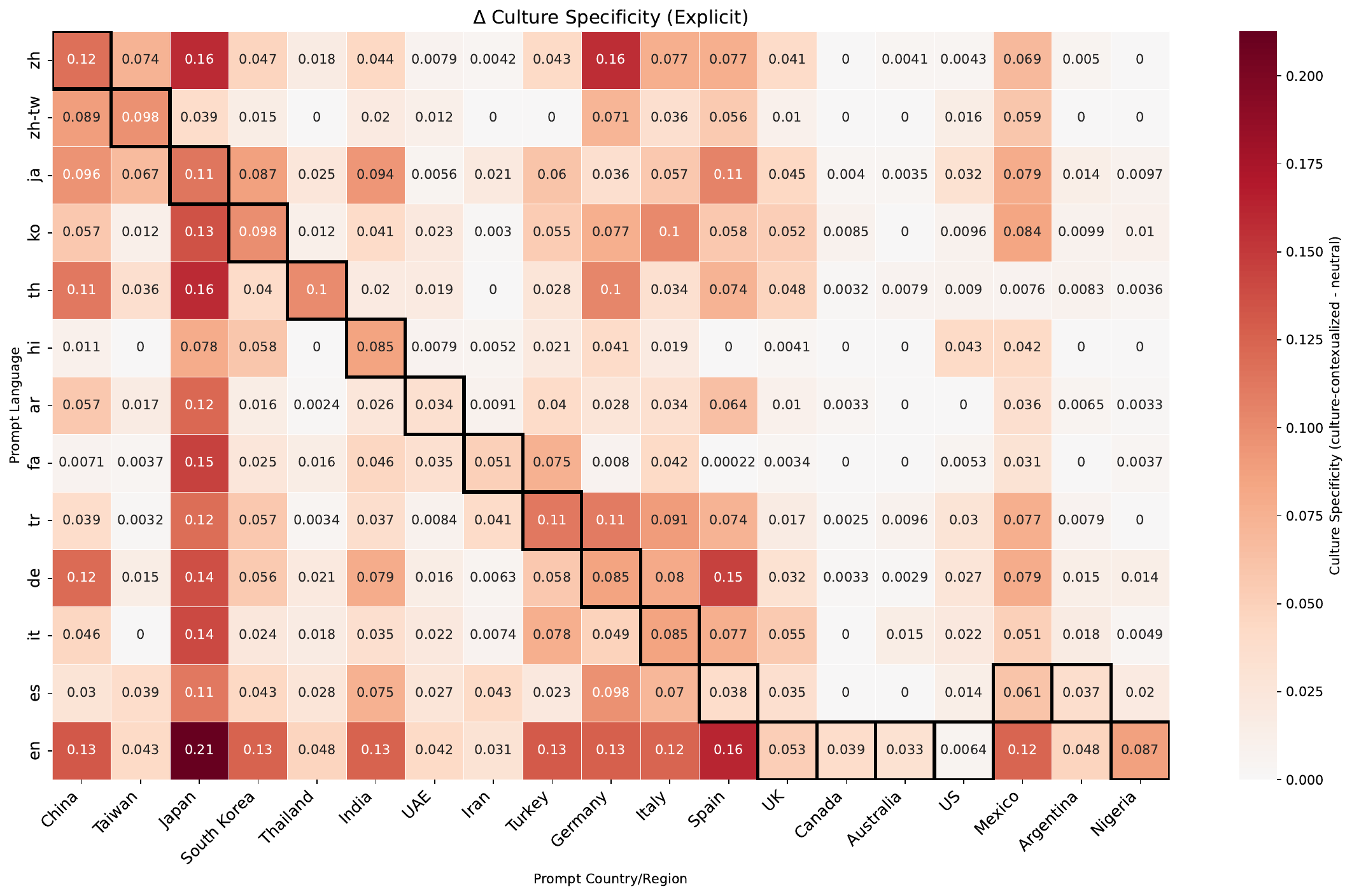}

  \caption{Culture specificity results for \textbf{ChatGPT}  with a neutral prompt (left) and a prompt explicitly mentioning the country (right, showing delta to neutral).}
  \label{fig:culturespec_chatgpt}
\end{figure*}

\begin{figure*}[htbp]
  \centering
  \includegraphics[width=\columnwidth]{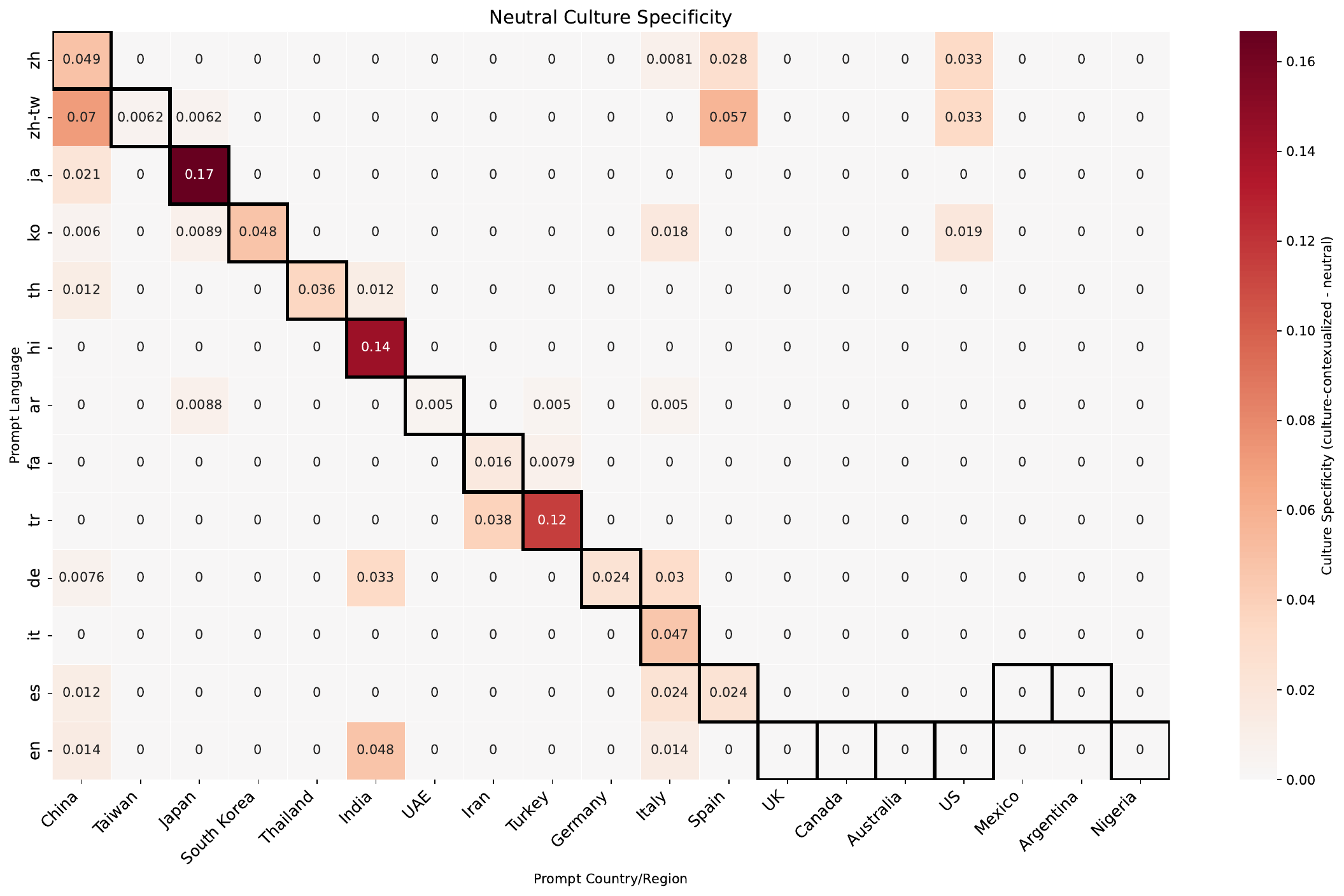}
  \vspace{0.8em}
  \includegraphics[width=\columnwidth]{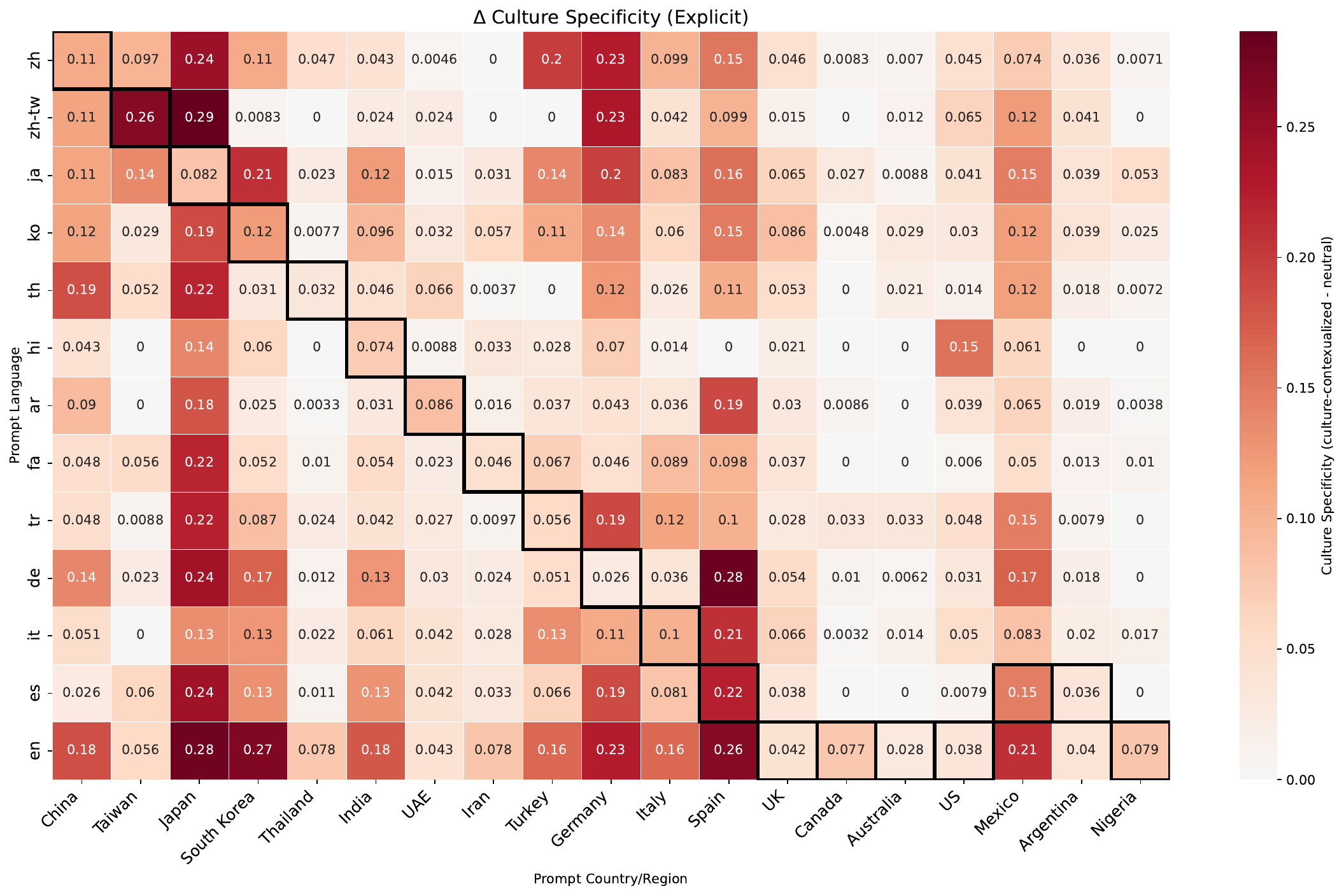}

  \caption{Culture specificity results for \textbf{DeepSeek}  with a neutral prompt (left) and a prompt explicitly mentioning the country (right, showing delta to neutral).}
  \label{fig:culturespec_deepseek}
\end{figure*}

\begin{figure*}[htbp]
  \centering
  \includegraphics[width=\columnwidth]{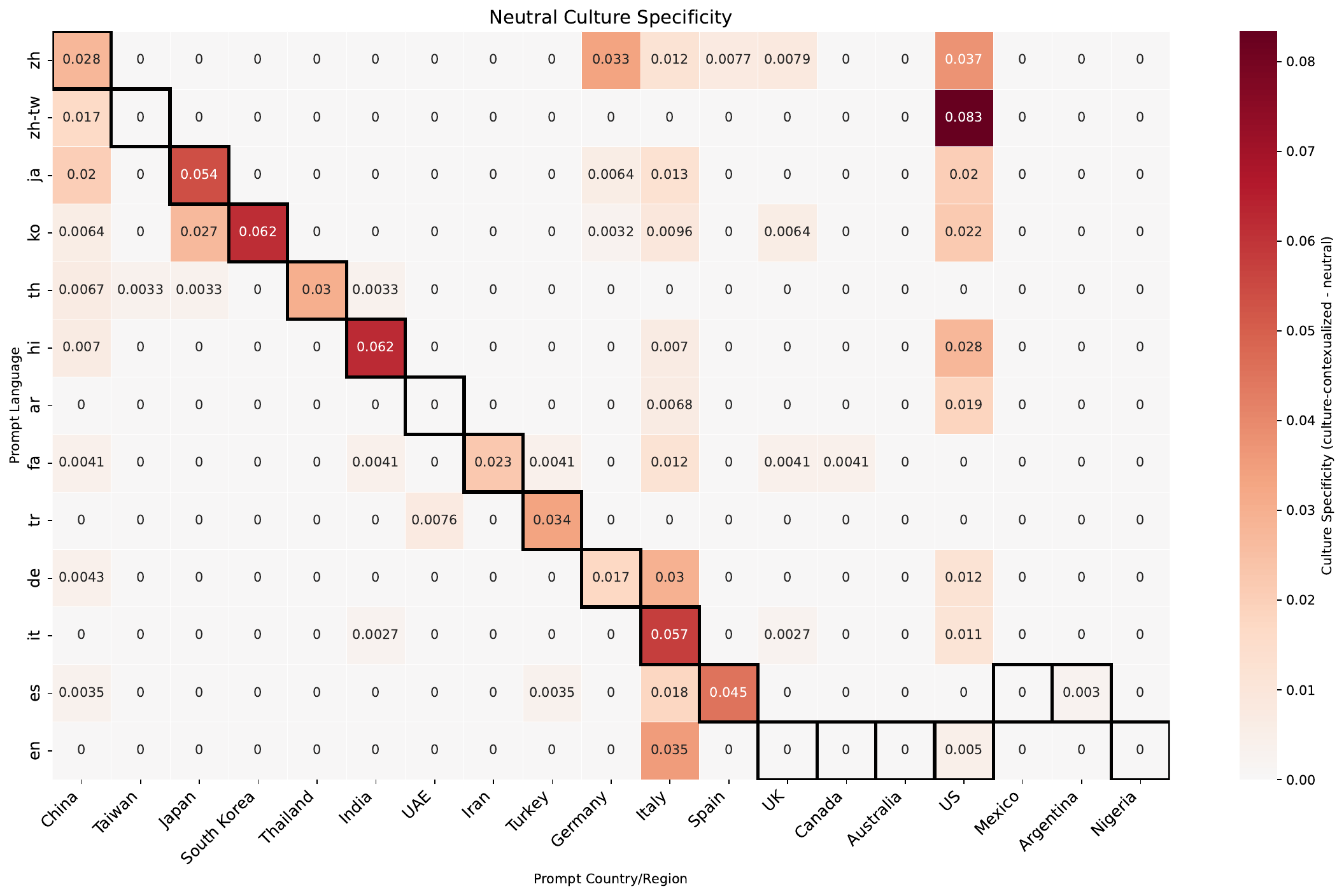}
  \vspace{0.8em}
  \includegraphics[width=\columnwidth]{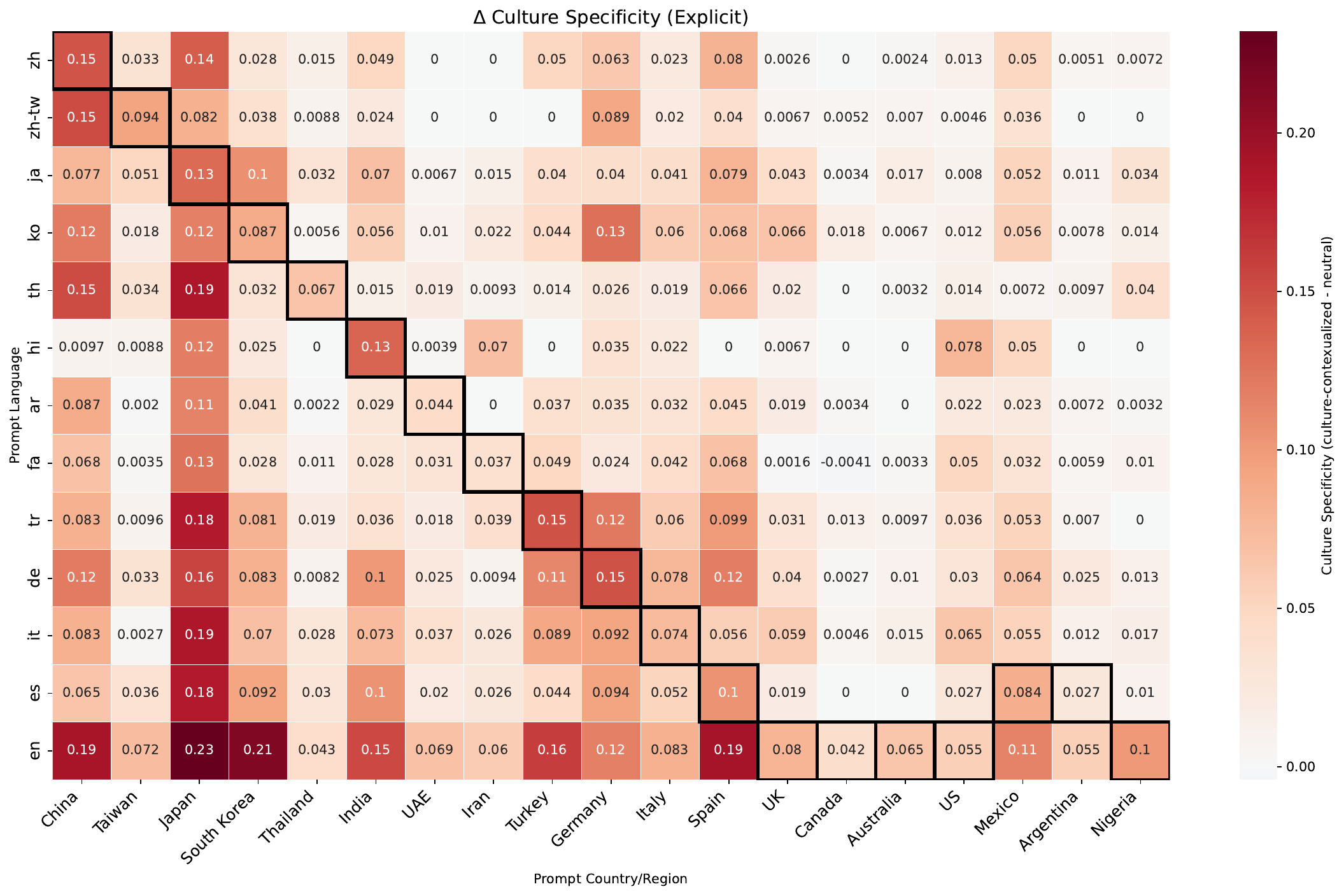}

  \caption{Culture specificity results for \textbf{Llama3-70B} with a neutral prompt (left) and a prompt explicitly mentioning the country (right, showing delta to neutral).}
  \label{fig:culturespec_llama3_70b}
\end{figure*}

\begin{figure*}[htbp]
  \centering
  \includegraphics[width=\columnwidth]{match/llama3_strict_nonbias.pdf}
  \vspace{0.8em}
  \includegraphics[width=\columnwidth]{match/llama3_strict.pdf}
  \vspace{0.8em}
  \includegraphics[width=\columnwidth]{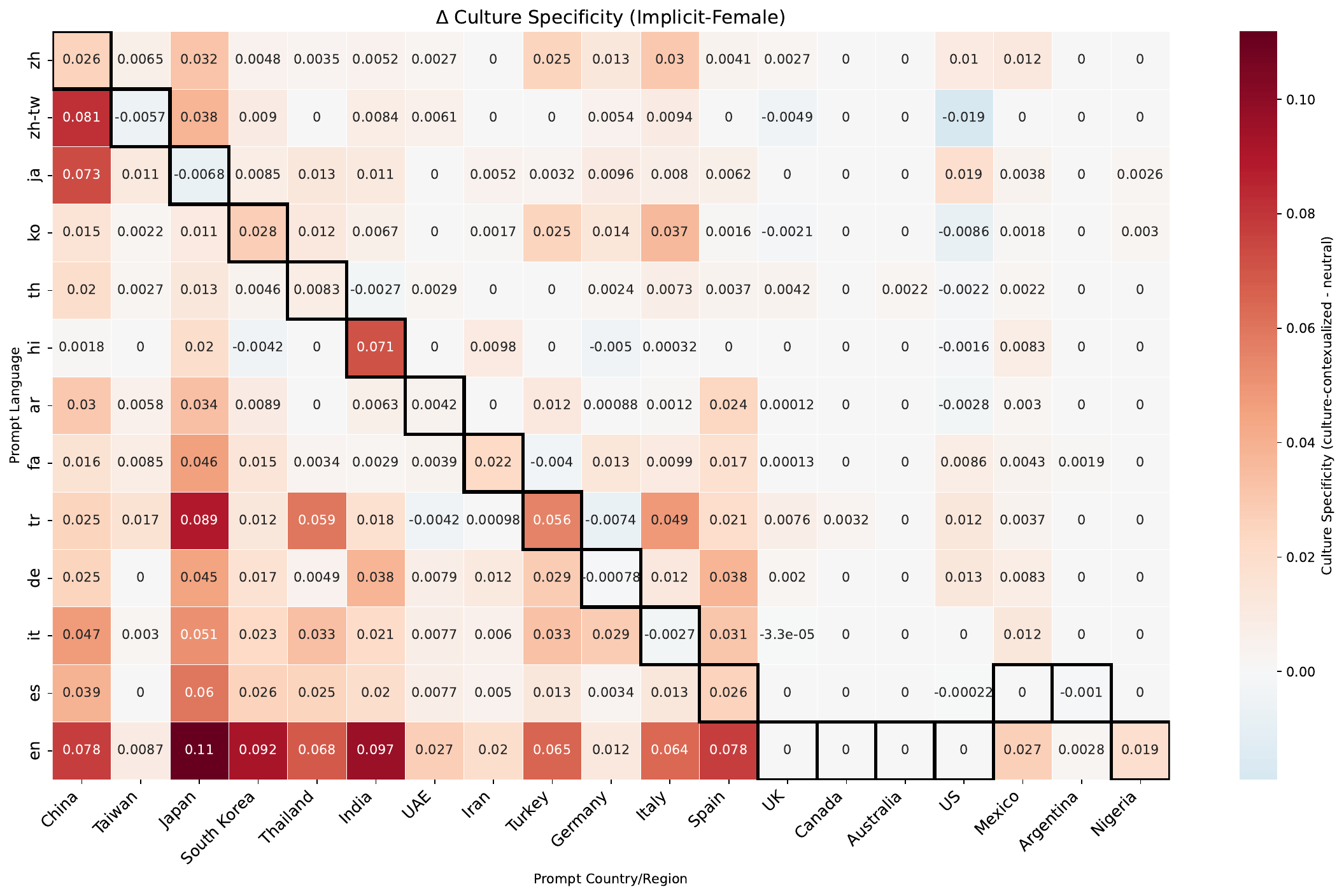}
  \vspace{0.8em}
  \includegraphics[width=\columnwidth]{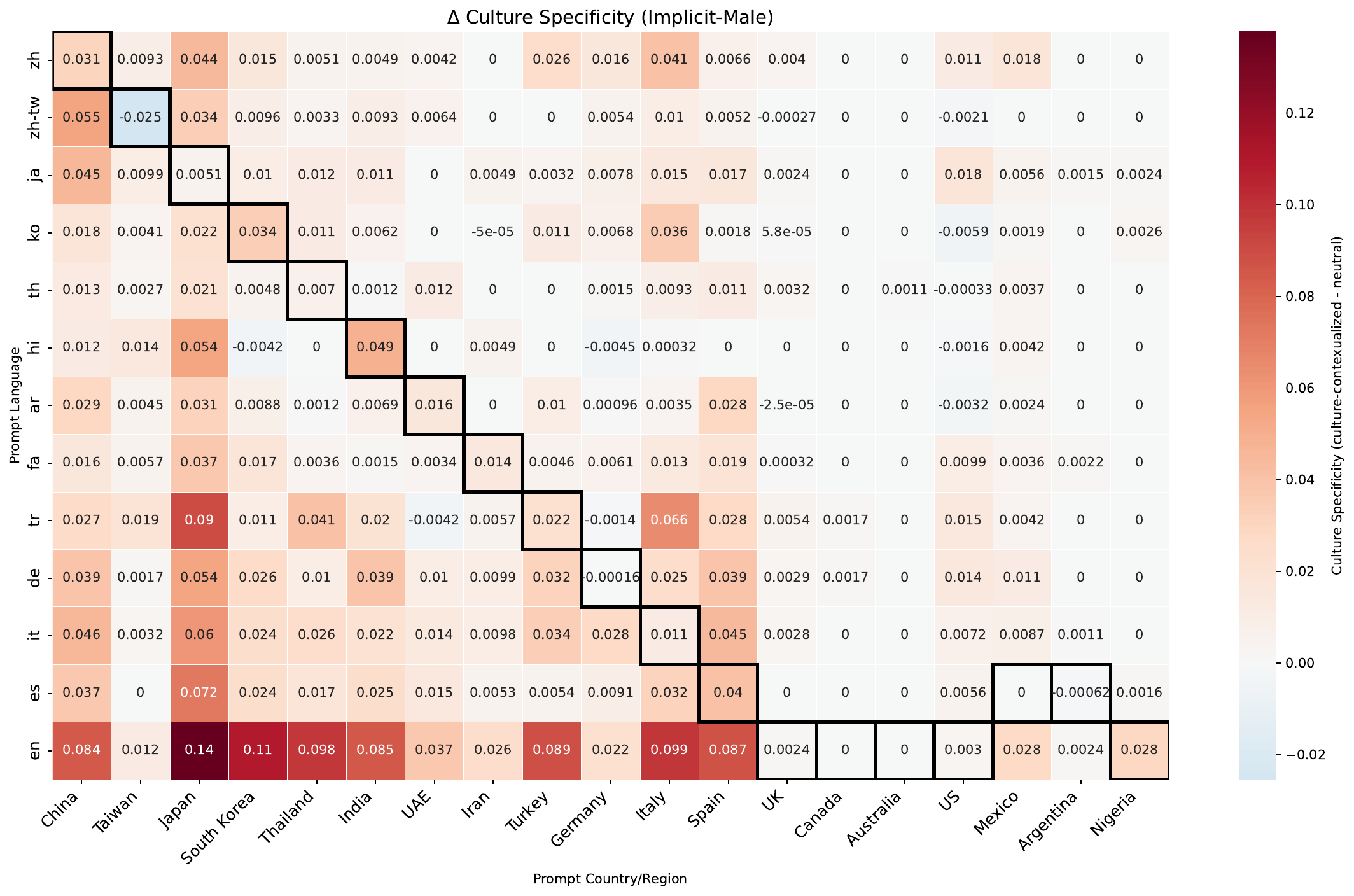}

  \caption{Culture specificity results for \textbf{Llama3} with a neutral prompt, a prompt explicitly mentioning the country (showing delta to neutral), and two prompts mentioning a popular female and male name as cultural context (showing delta to neutral).}
  \label{fig:culturespec_llama3}
\end{figure*}

\begin{figure*}[htbp]
  \centering
  \includegraphics[width=\columnwidth]{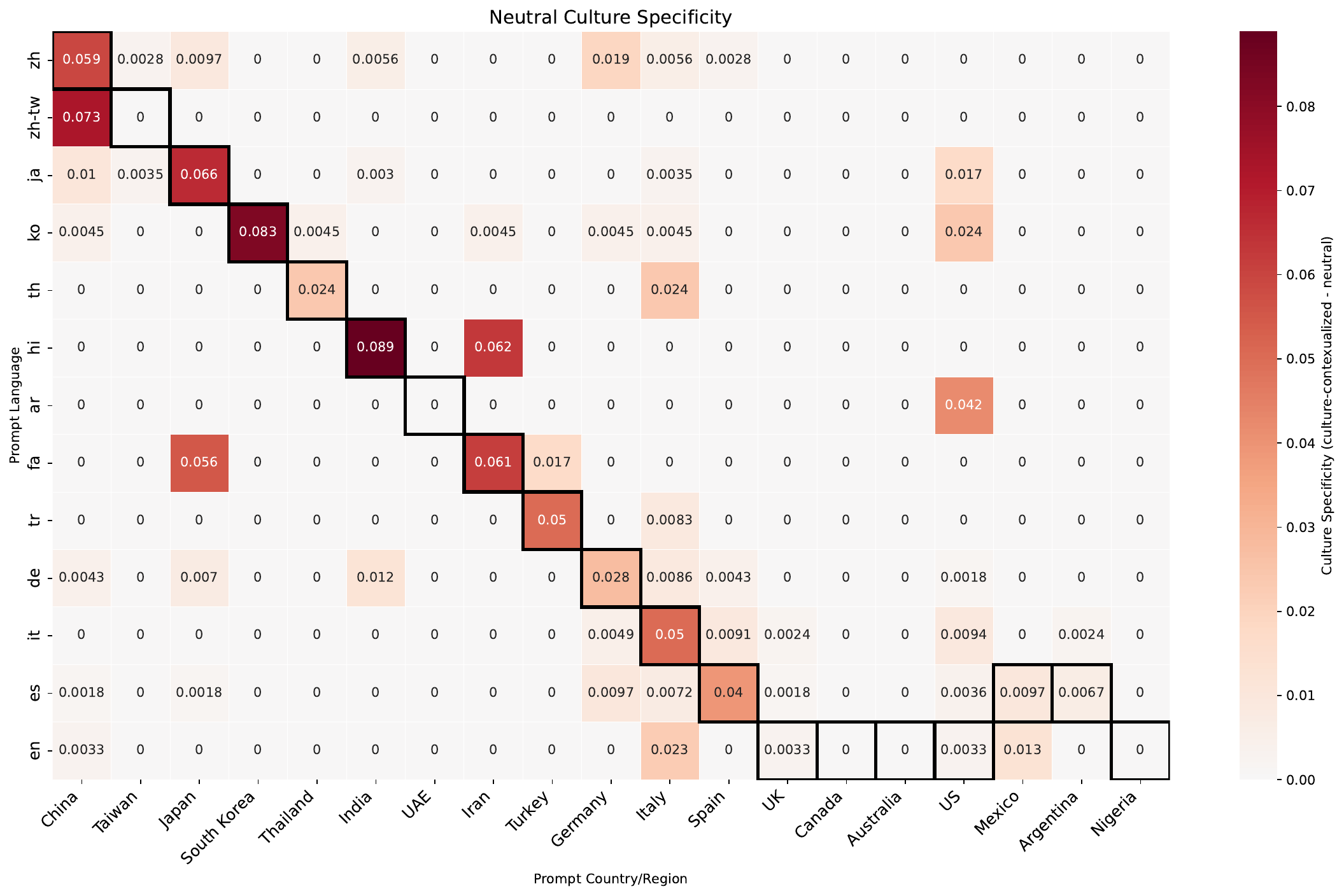}
  \vspace{0.8em}
  \includegraphics[width=\columnwidth]{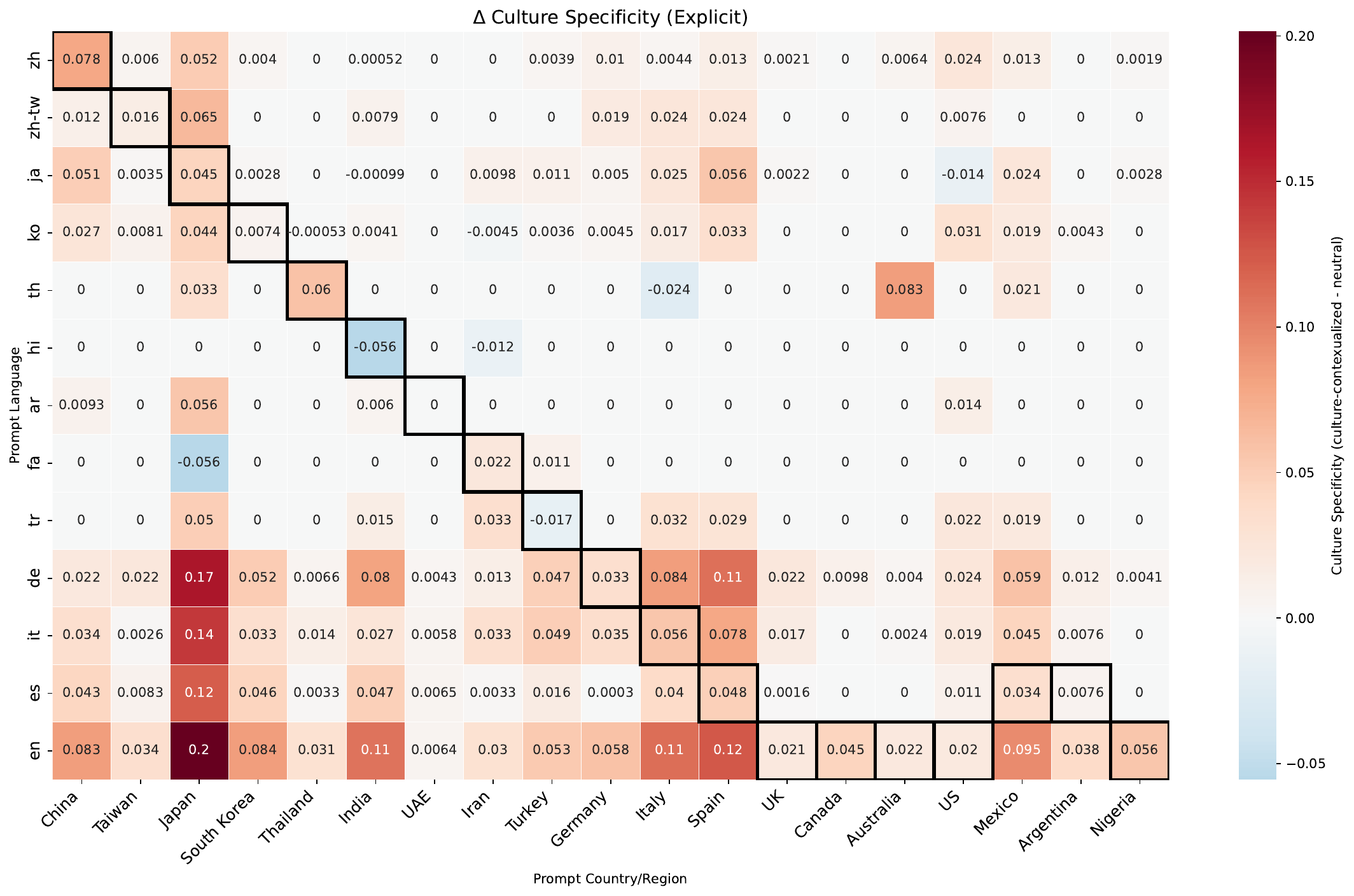}

  \caption{Culture specificity results for \textbf{Mistral} with a neutral prompt (left) and a prompt explicitly mentioning the country (right, showing delta to neutral).}
  \label{fig:culturespec_mistral}
\end{figure*}

\begin{figure*}[htbp]
  \centering
  \includegraphics[width=\columnwidth]{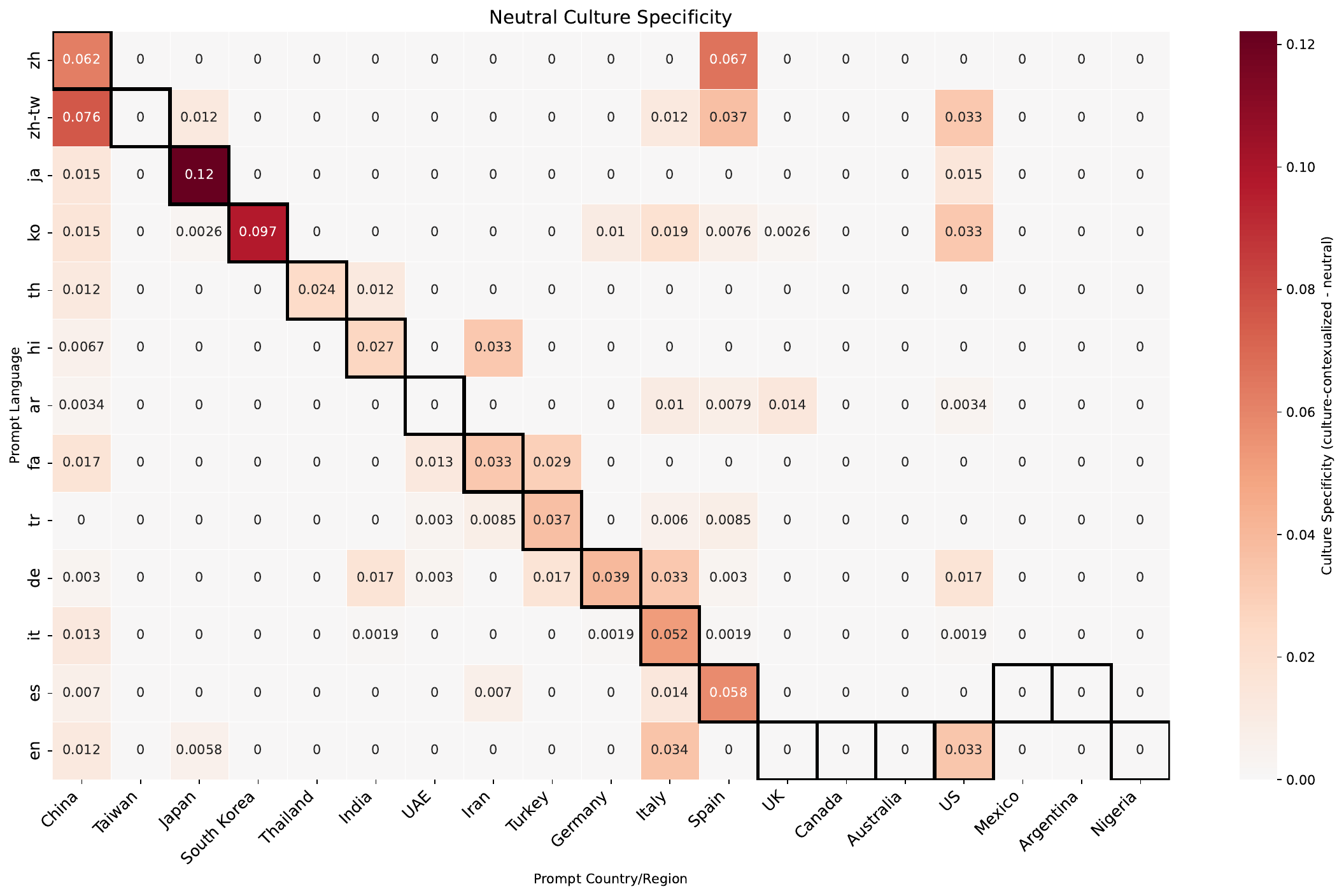}
  \vspace{0.8em}
  \includegraphics[width=\columnwidth]{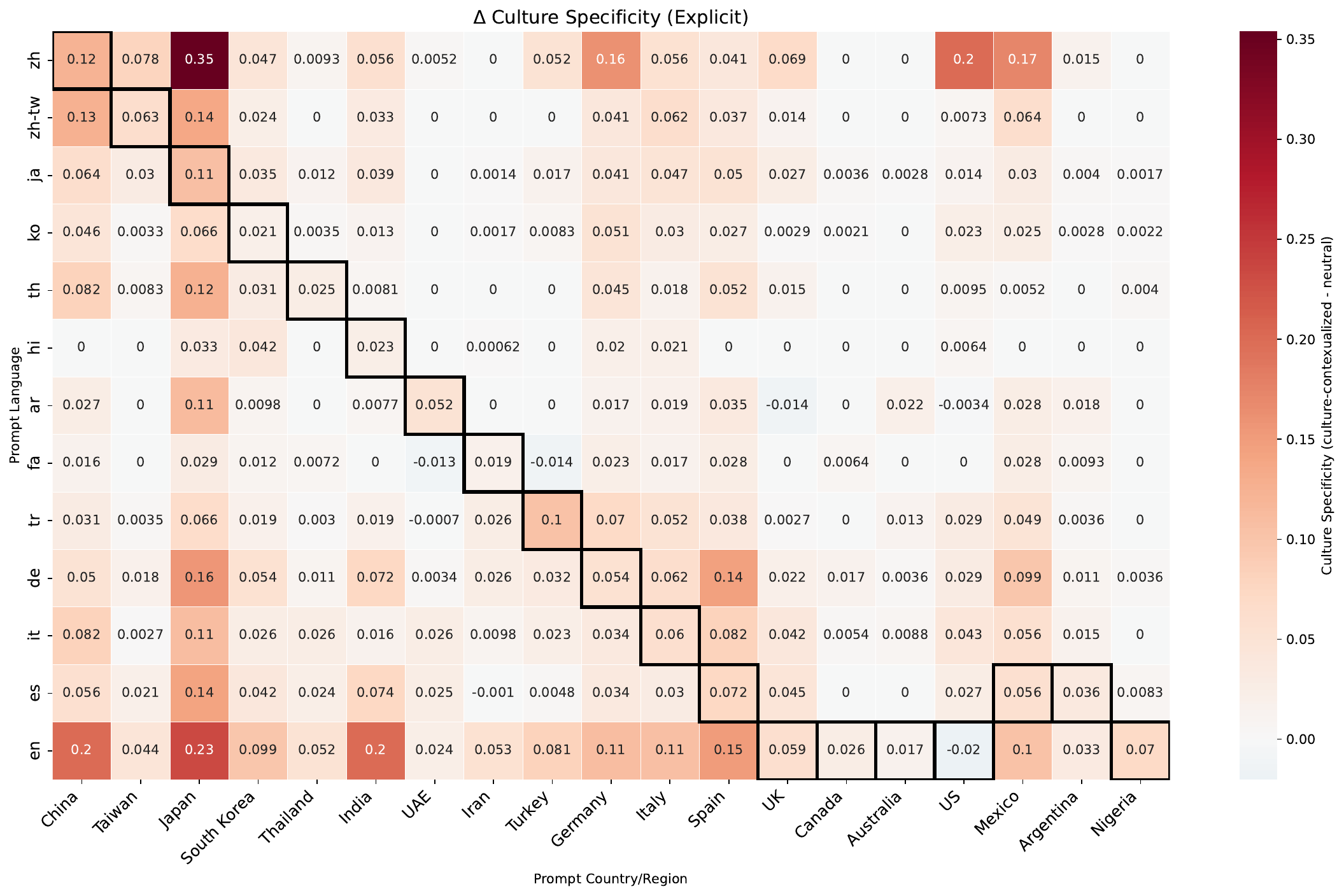}

  \caption{Culture specificity results for \textbf{Qwen}  with a neutral prompt (left) and a prompt explicitly mentioning the country (right, showing delta to neutral).}
  \label{fig:culturespec_qwen}
\end{figure*}

\vspace{-2pt}
\section{Results of Culture Consensus Results}
\label{sec:all_consensus}
To measure cultural consensus, we compute the pairwise overlap rate of entity QIDs across different languages when generating responses to semantically equivalent prompts. This overlap reflects the extent to which different languages share similar cultural interpretations of the same prompt.

In Figure~\ref{fig:consensus_chatgpt}-~\ref{fig:consensus_qwen}, we present the average consensus scores across all mentioned countries and six topics. To aid interpretation, we use yellow color coding to highlight the major Europe region. Notably, European languages consistently exhibit higher cultural consensus across models, suggesting a stronger alignment in entity representations within this region.
Our findings suggest that cultural signals interact with cross-lingual representations in complex ways, shaping both specificity and consensus across languages. Approaches such as contrastive learning~\cite{DBLP:conf/icassp/ChenGGLW22} offer a promising direction to more faithfully capturing culture-specific cues while maintaining stronger cross-lingual consensus.

\begin{figure*}[htbp]
  \centering
  \resizebox{\textwidth}{!}{
  \begin{tabular}{ccc}
    \includegraphics[width=0.3\textwidth]{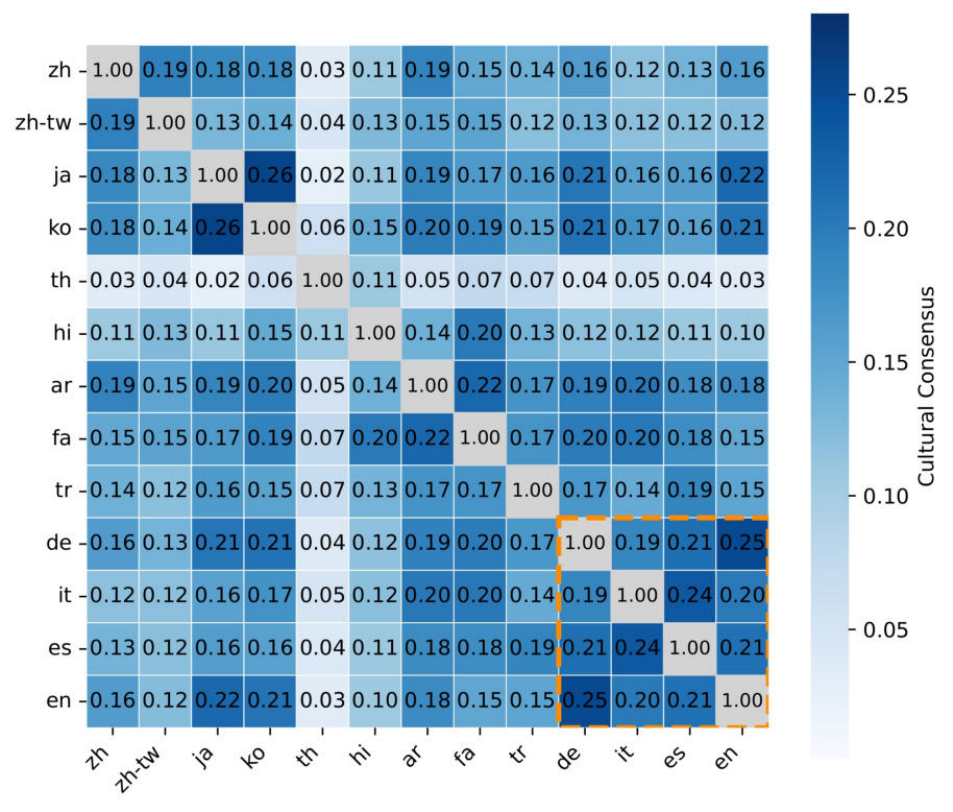} &
    \includegraphics[width=0.3\textwidth]{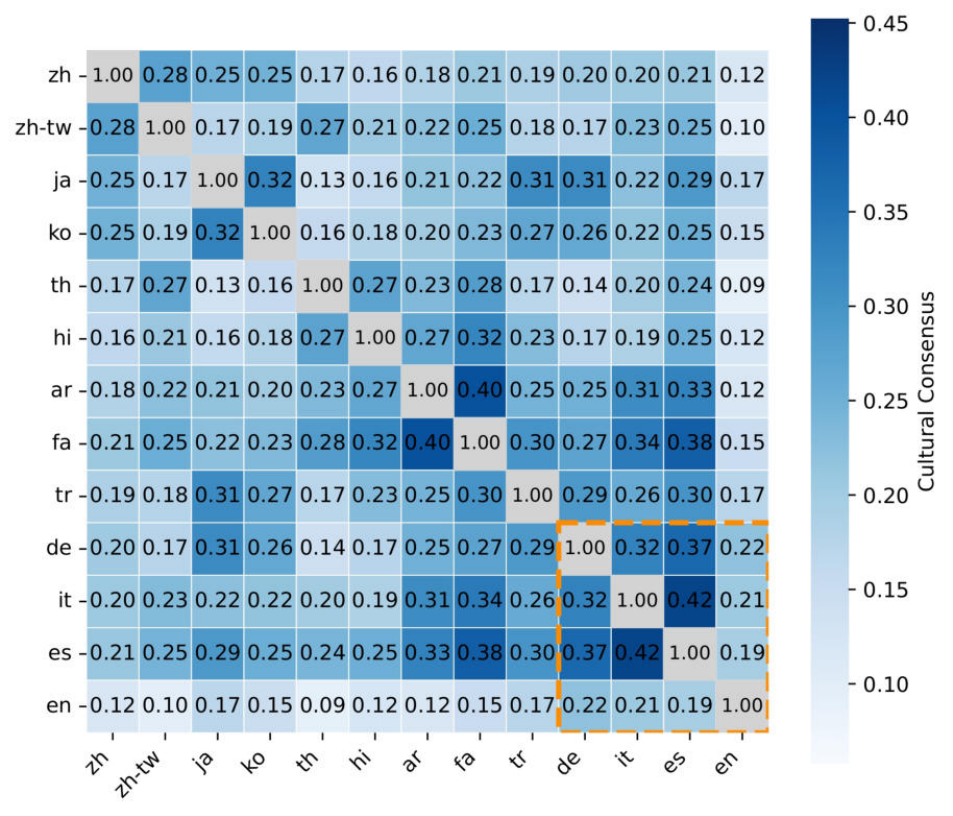} &
    \includegraphics[width=0.3\textwidth]{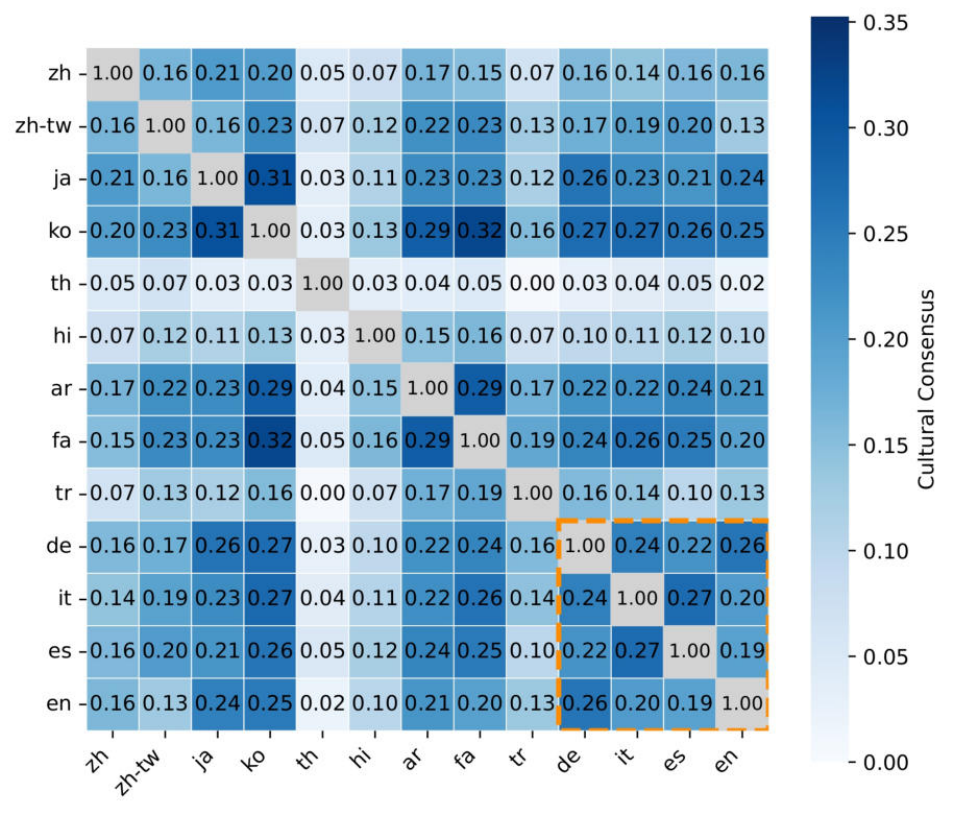} \\
    (a) Food & (b) Beverage & (c) Clothing \\[1.5ex]
    \includegraphics[width=0.3\textwidth]{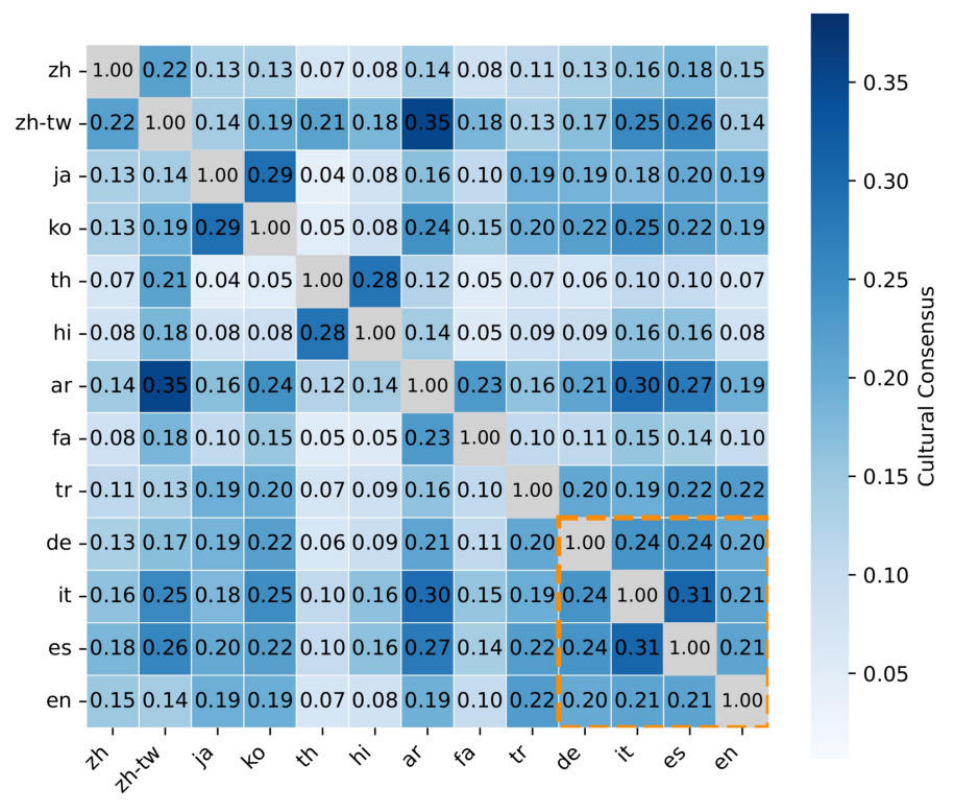} &
    \includegraphics[width=0.3\textwidth]{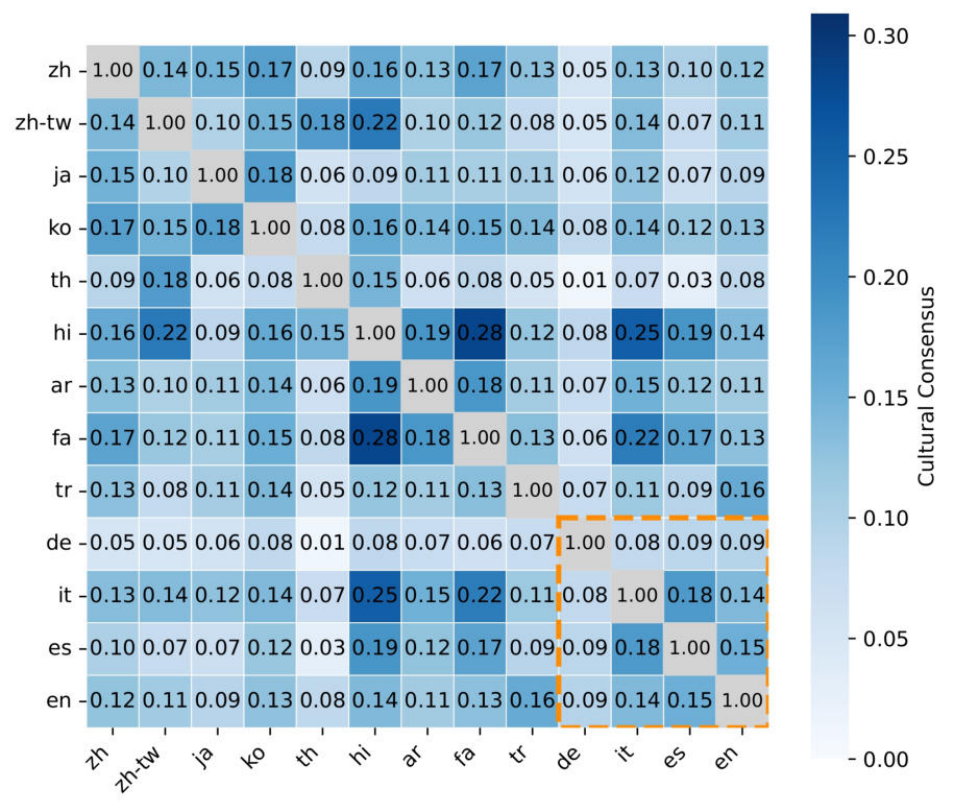} &
    \includegraphics[width=0.3\textwidth]{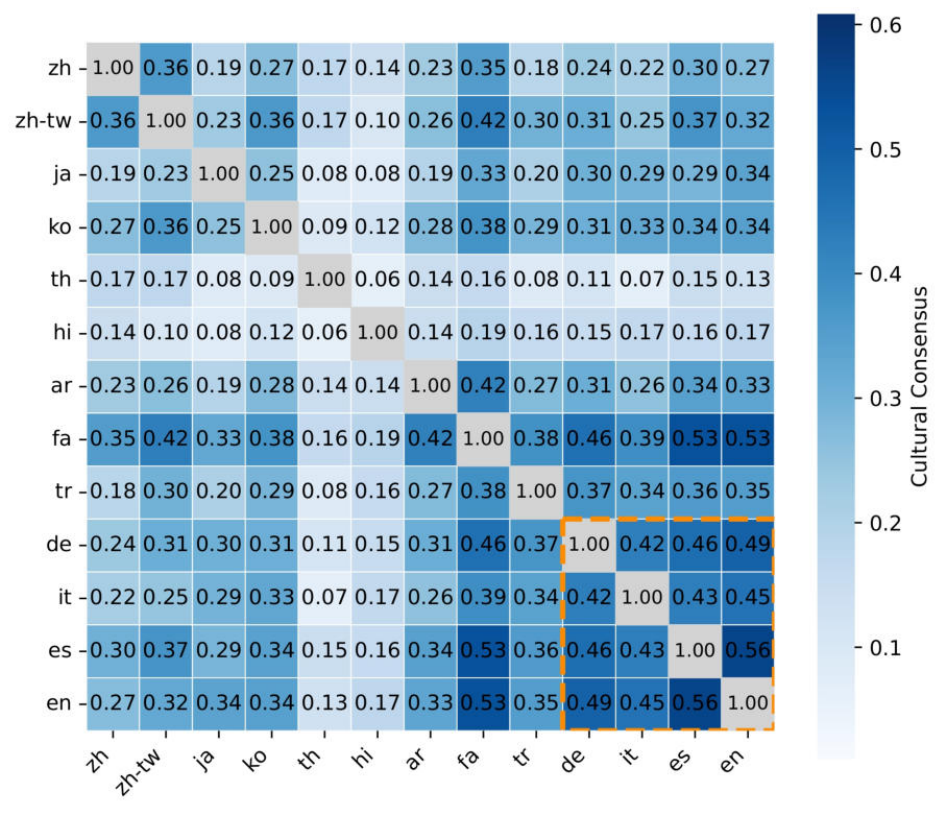} \\
    (d) Music & (e) Book & (f) Transportation
  \end{tabular}}
  \caption{Culture consensus results for \textbf{Aya} across six topics. Yellow colors are used to distinguish Europe language region.}
  \label{fig:consensus_aya}
\end{figure*}

\begin{figure*}[htbp]
  \centering
  \resizebox{\textwidth}{!}{
  \begin{tabular}{ccc}
    \includegraphics[width=0.3\textwidth]{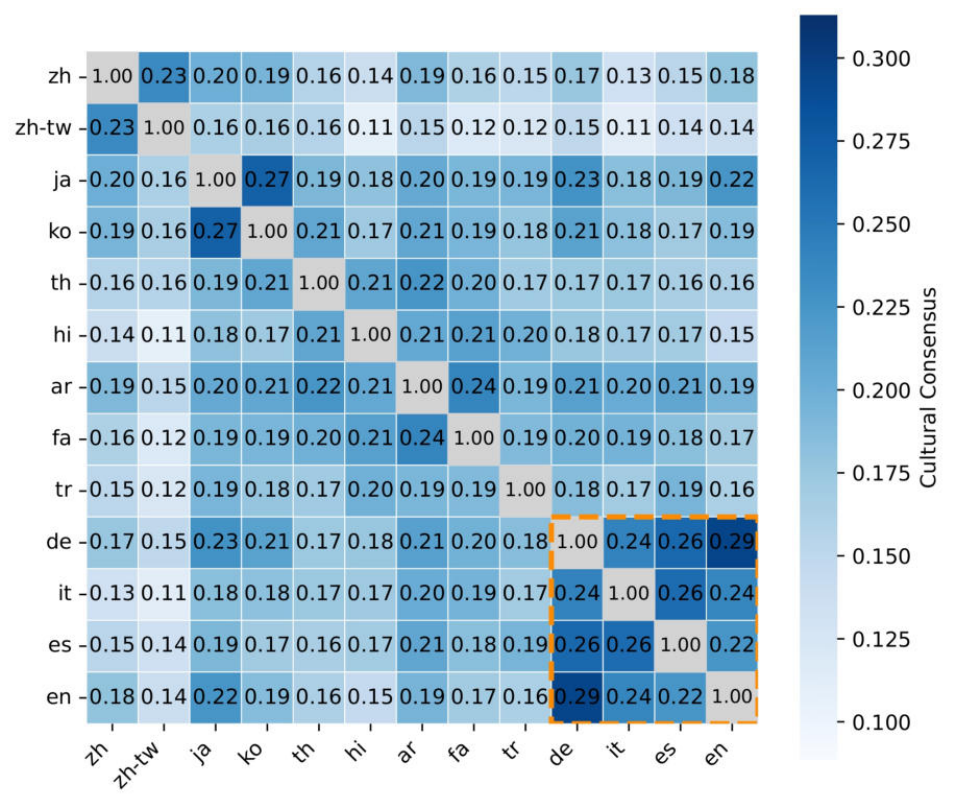} &
    \includegraphics[width=0.3\textwidth]{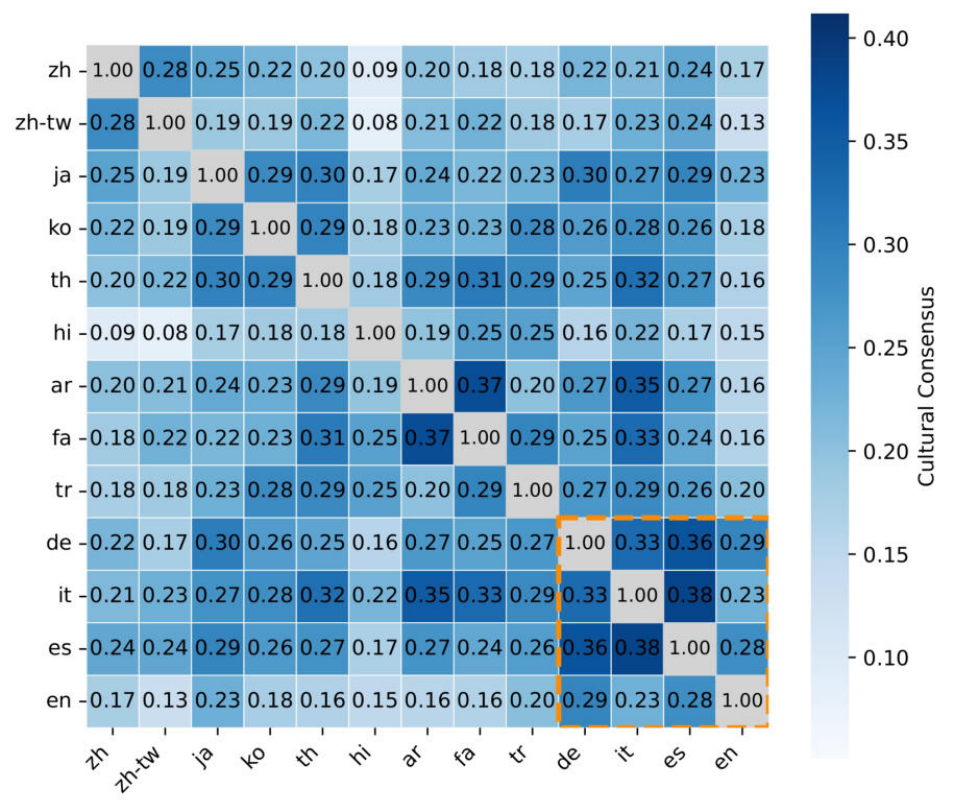} &
    \includegraphics[width=0.3\textwidth]{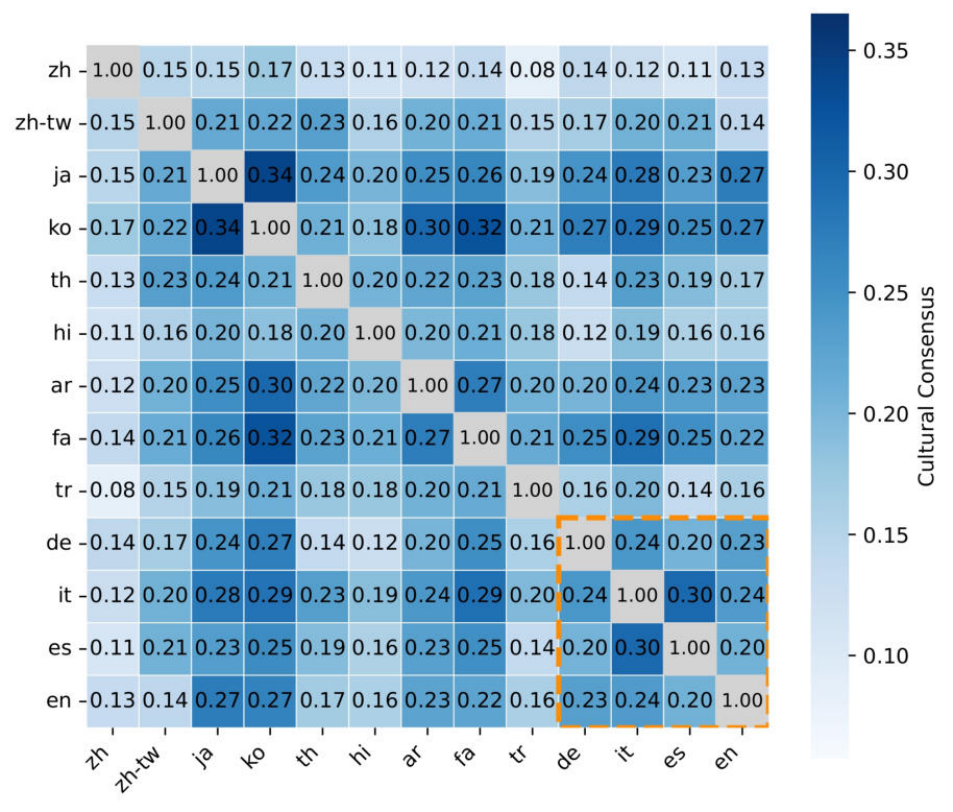} \\
    (a) Food & (b) Beverage & (c) Clothing \\[1.5ex]
    \includegraphics[width=0.3\textwidth]{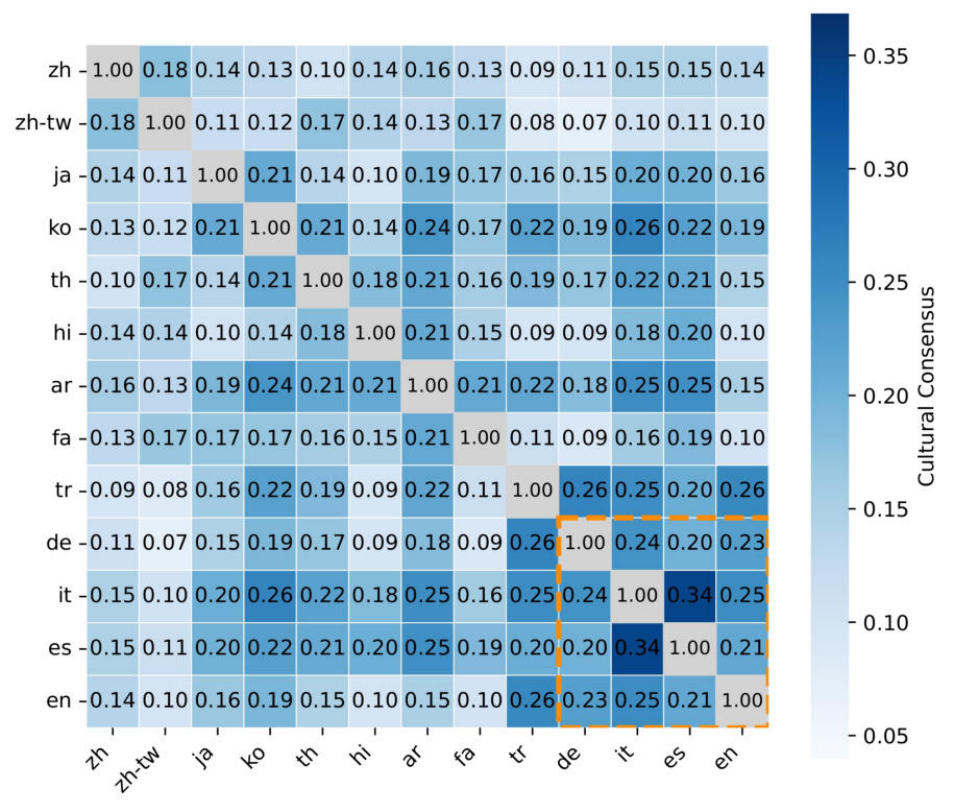} &
    \includegraphics[width=0.3\textwidth]{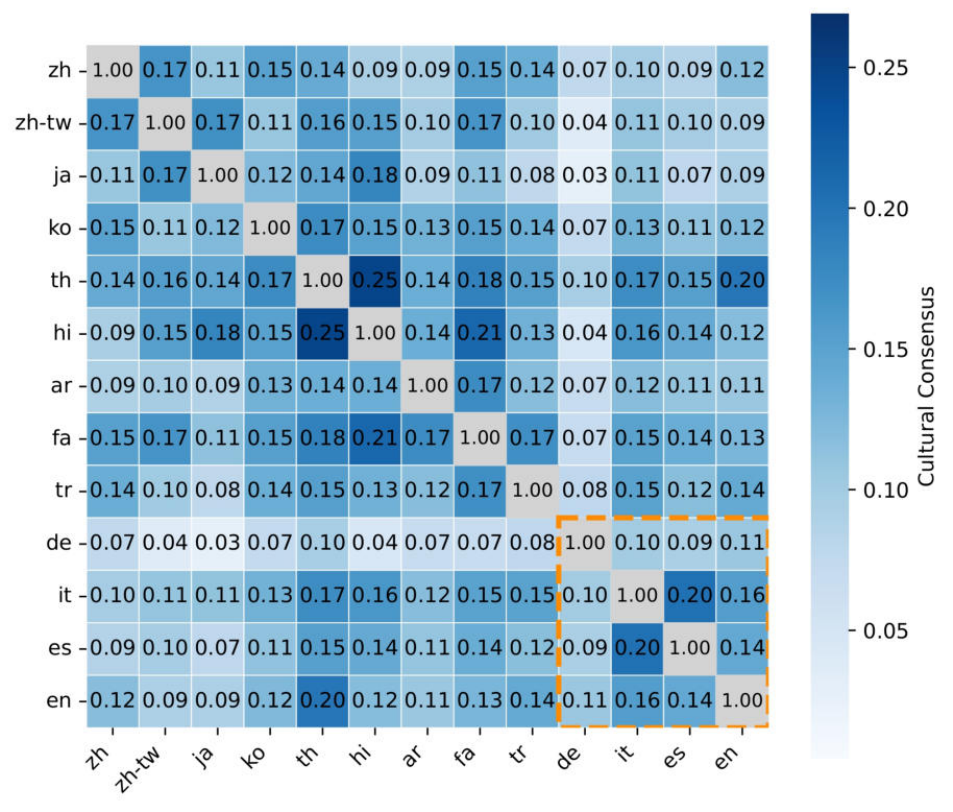} &
    \includegraphics[width=0.3\textwidth]{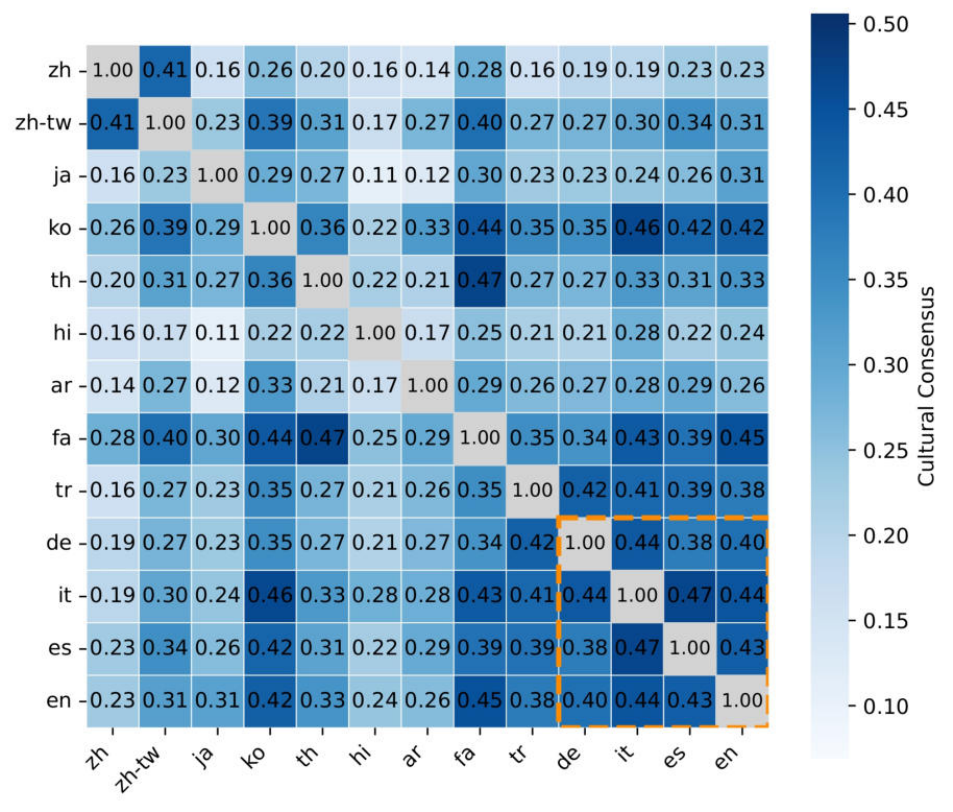} \\
    (d) Music & (e) Book & (f) Transportation
  \end{tabular}}
  \caption{Culture consensus results for \textbf{ChatGPT} across six topics. Yellow colors are used to distinguish Europe language region.}
  \label{fig:consensus_chatgpt}
\end{figure*}

\begin{figure*}[htbp]
  \centering
  \resizebox{\textwidth}{!}{
  \begin{tabular}{ccc}
    \includegraphics[width=0.3\textwidth]{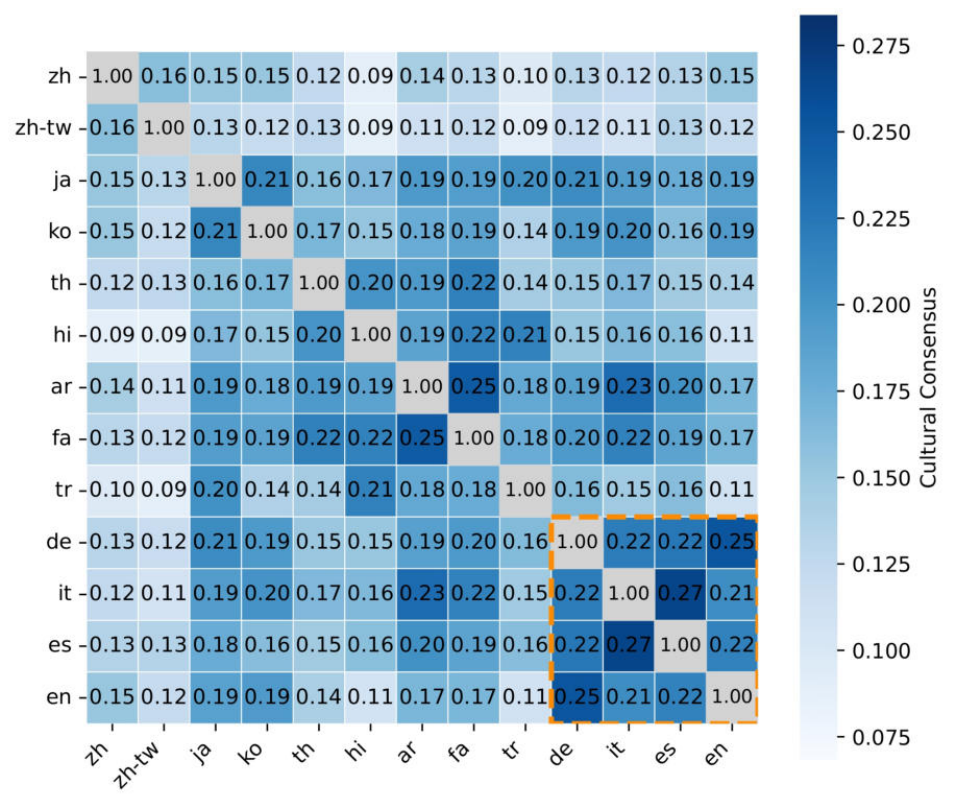} &
    \includegraphics[width=0.3\textwidth]{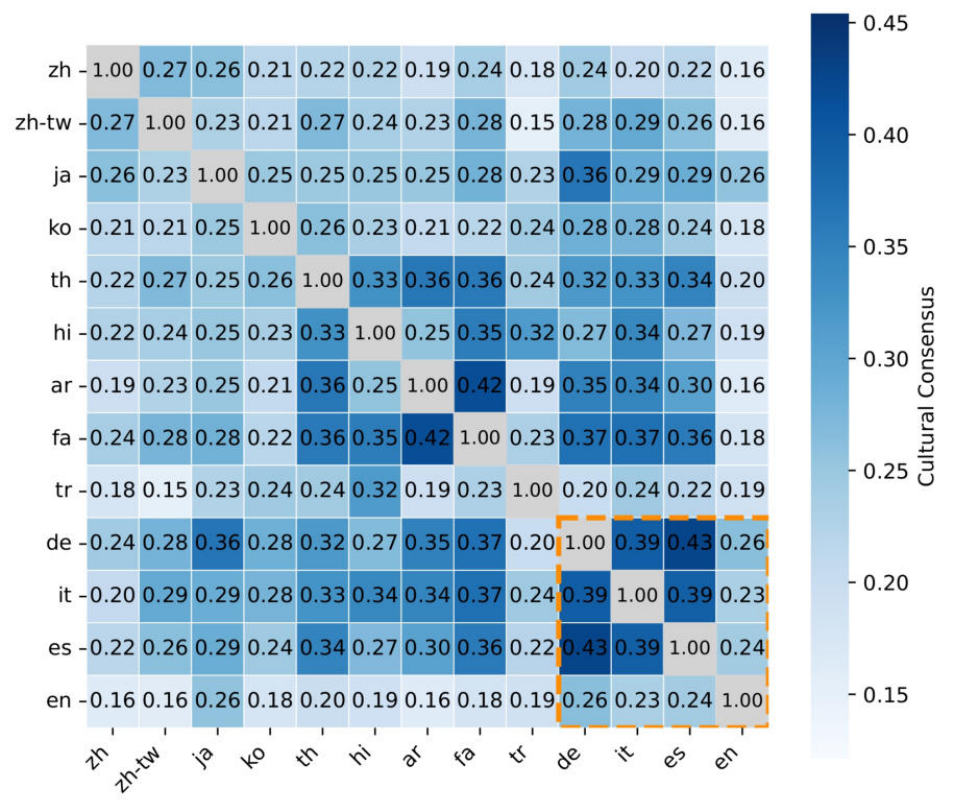} &
    \includegraphics[width=0.3\textwidth]{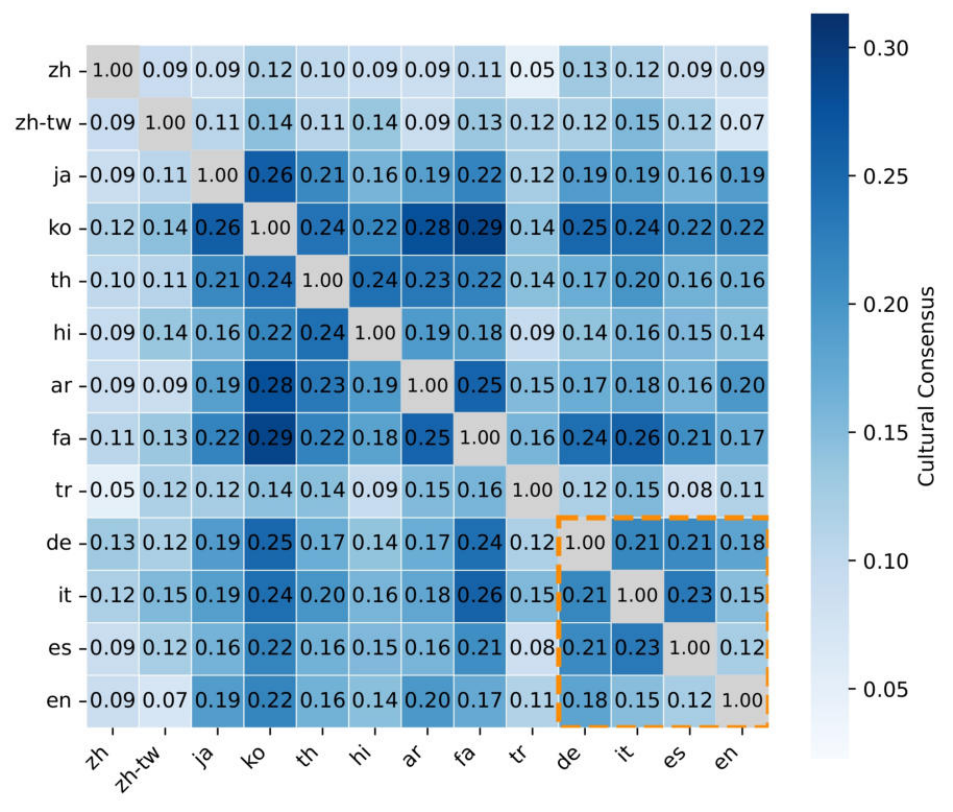} \\
    (a) Food & (b) Beverage & (c) Clothing \\[1.5ex]
    \includegraphics[width=0.3\textwidth]{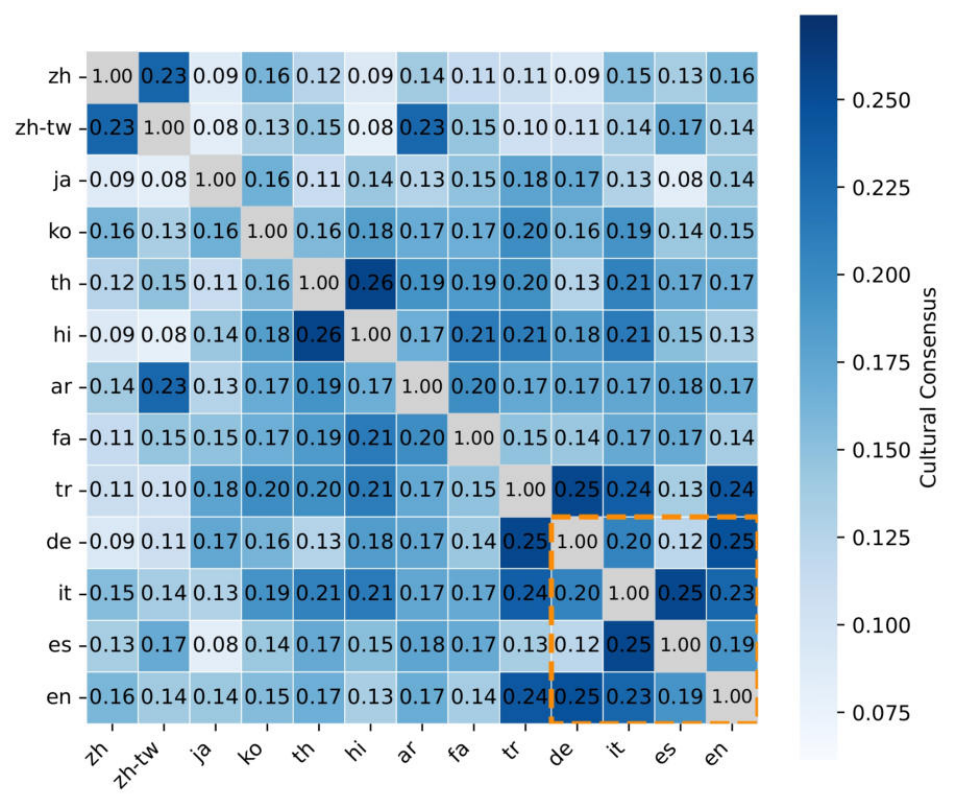} &
    \includegraphics[width=0.3\textwidth]{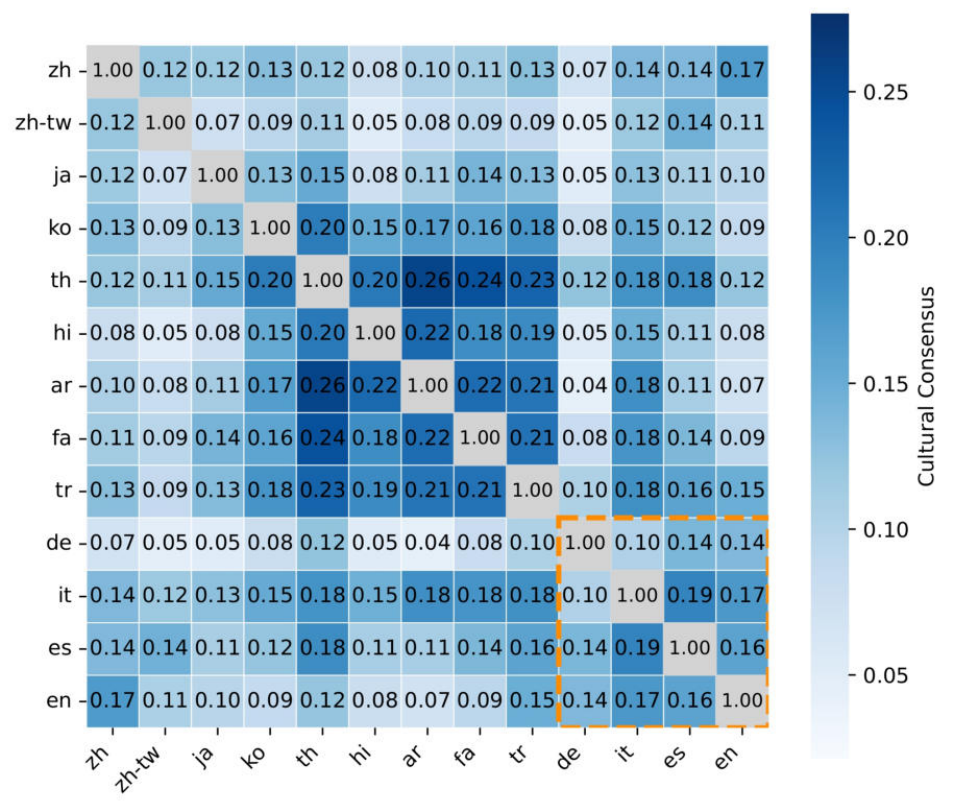} &
    \includegraphics[width=0.3\textwidth]{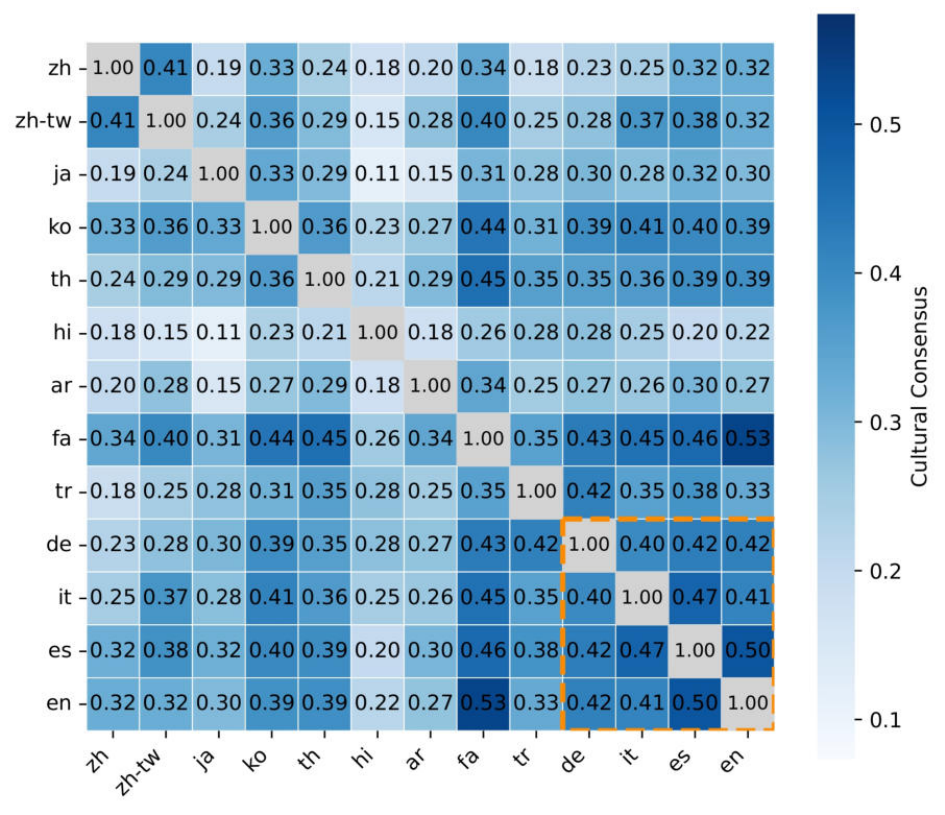} \\
    (d) Music & (e) Book & (f) Transportation
  \end{tabular}}
  \caption{Culture consensus results for \textbf{DeepSeek} across six topics. Yellow colors are used to distinguish Europe language region.}
  \label{fig:consensus_deepseek}
\end{figure*}

\begin{figure*}[htbp]
  \centering
  \resizebox{\textwidth}{!}{
  \begin{tabular}{ccc}
    \includegraphics[width=0.3\textwidth]{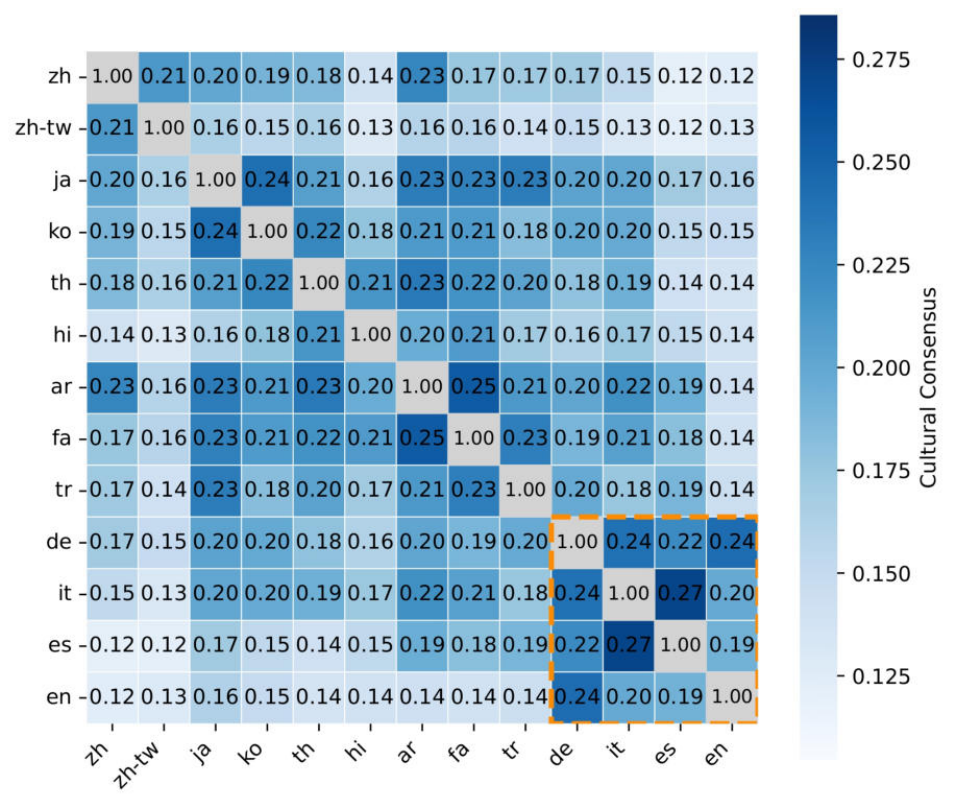} &
    \includegraphics[width=0.3\textwidth]{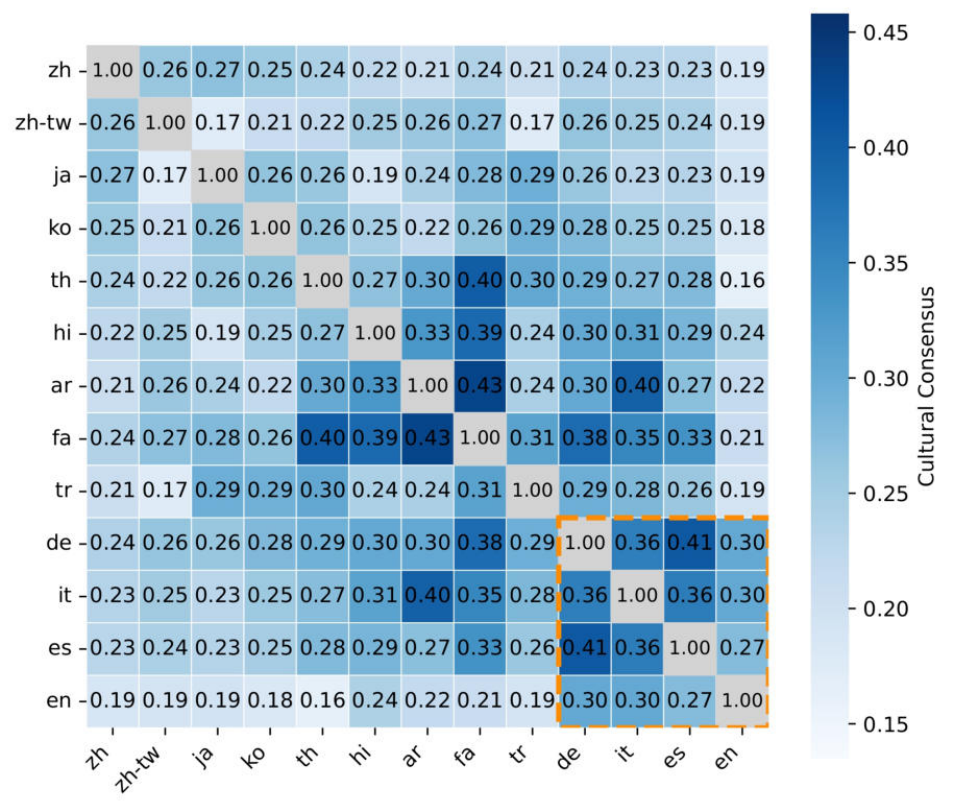} &
    \includegraphics[width=0.3\textwidth]{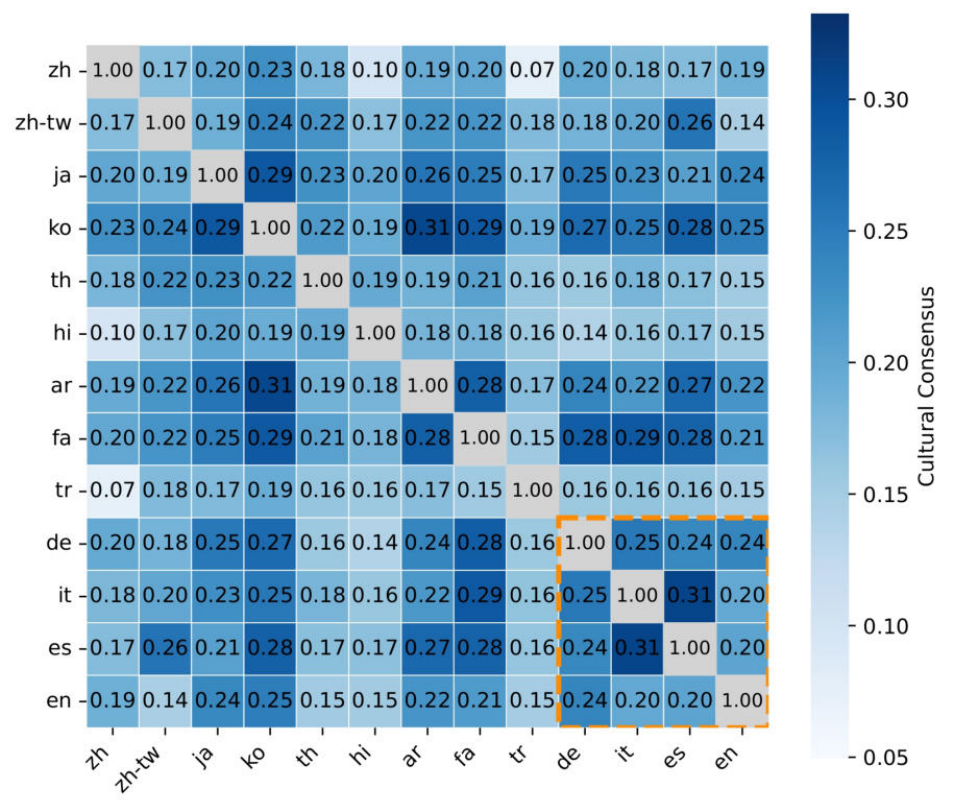} \\
    (a) Food & (b) Beverage & (c) Clothing \\[1.5ex]
    \includegraphics[width=0.3\textwidth]{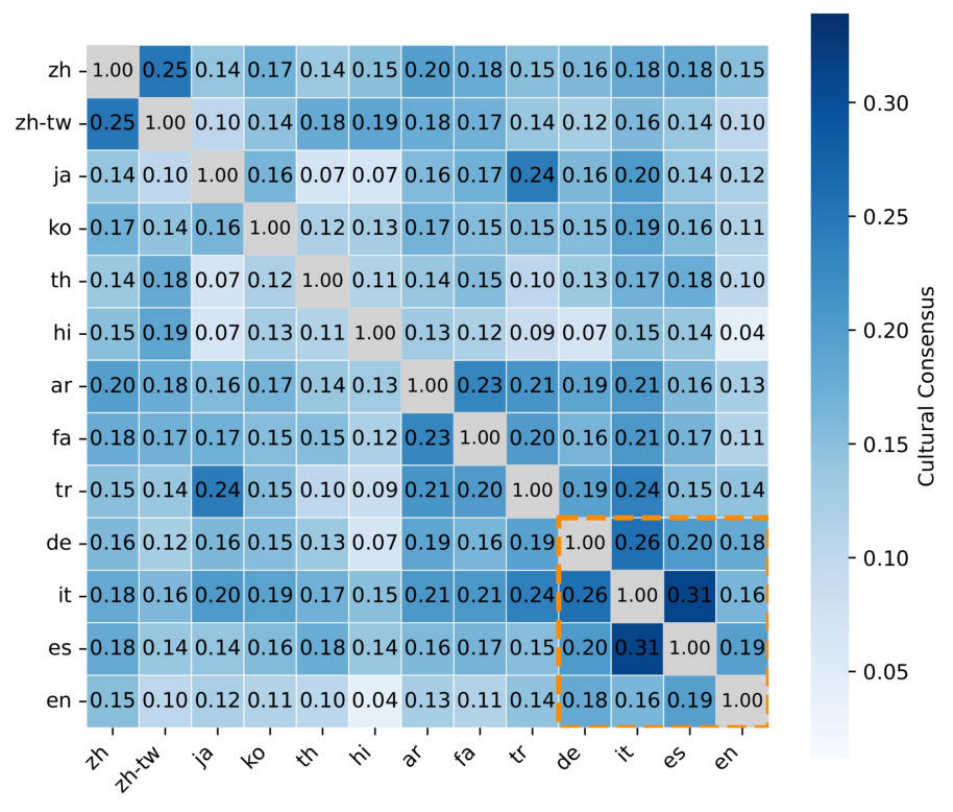} &
    \includegraphics[width=0.3\textwidth]{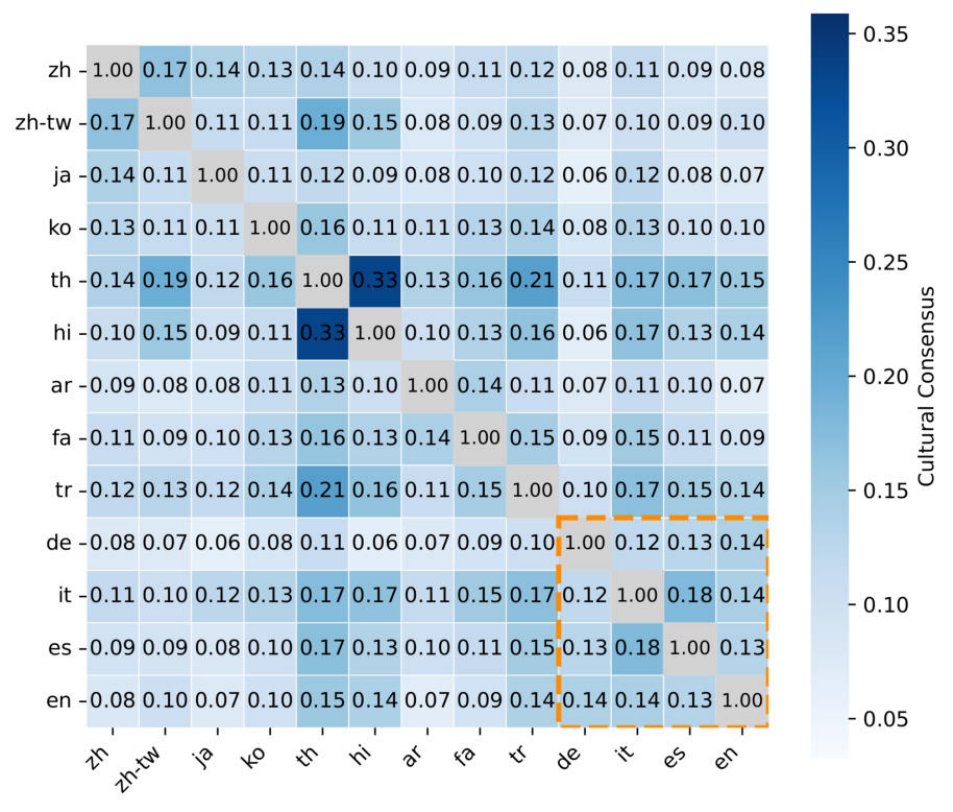} &
    \includegraphics[width=0.3\textwidth]{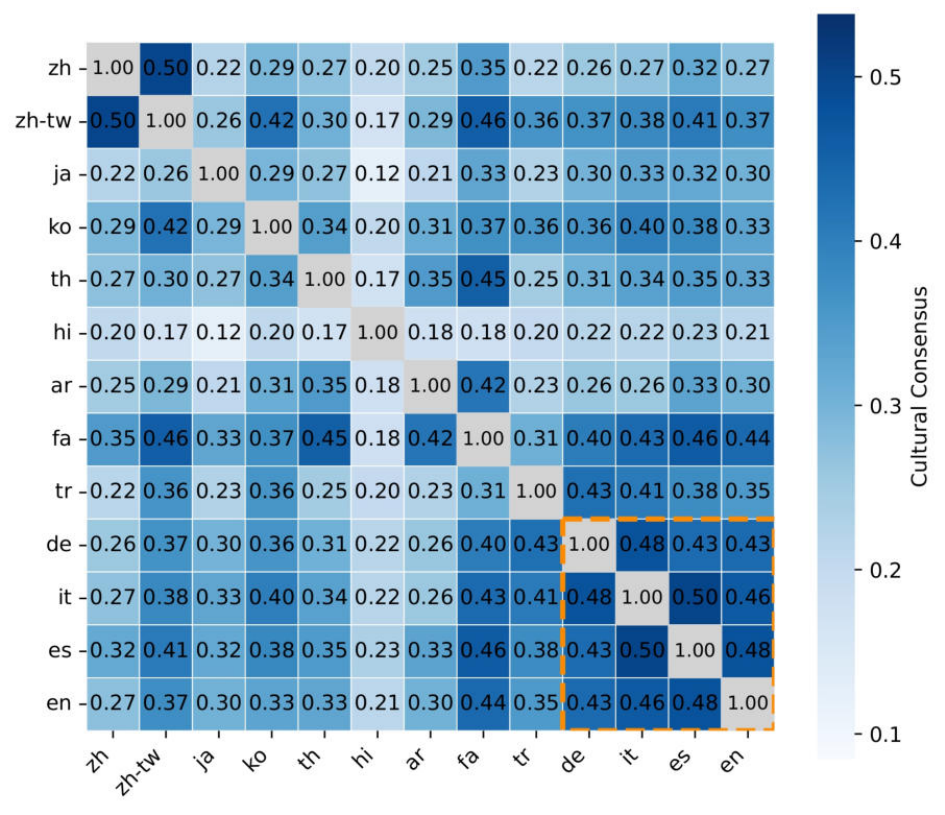} \\
    (d) Music & (e) Book & (f) Transportation
  \end{tabular}}
  \caption{Culture consensus results for \textbf{LLaMA3-70B} across six topics. Yellow colors are used to distinguish Europe language region.}
  \label{fig:consensus_llama3_70b}
\end{figure*}

\begin{figure*}[htbp]
  \centering
  \resizebox{\textwidth}{!}{
  \begin{tabular}{ccc}
    \includegraphics[width=0.3\textwidth]{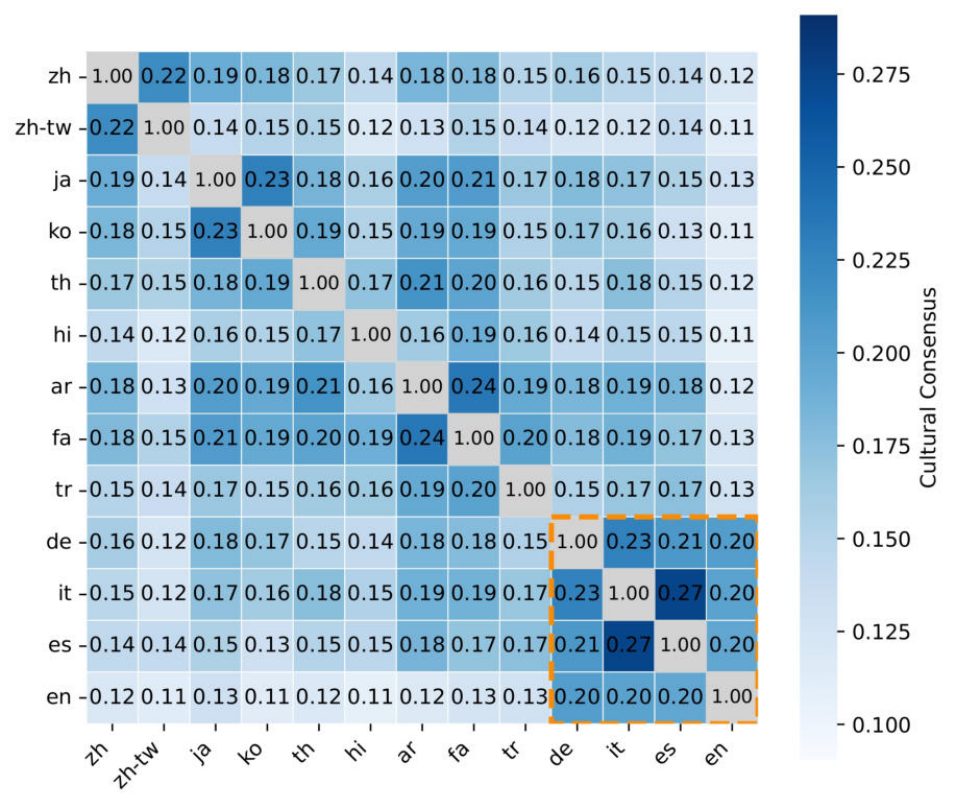} &
    \includegraphics[width=0.3\textwidth]{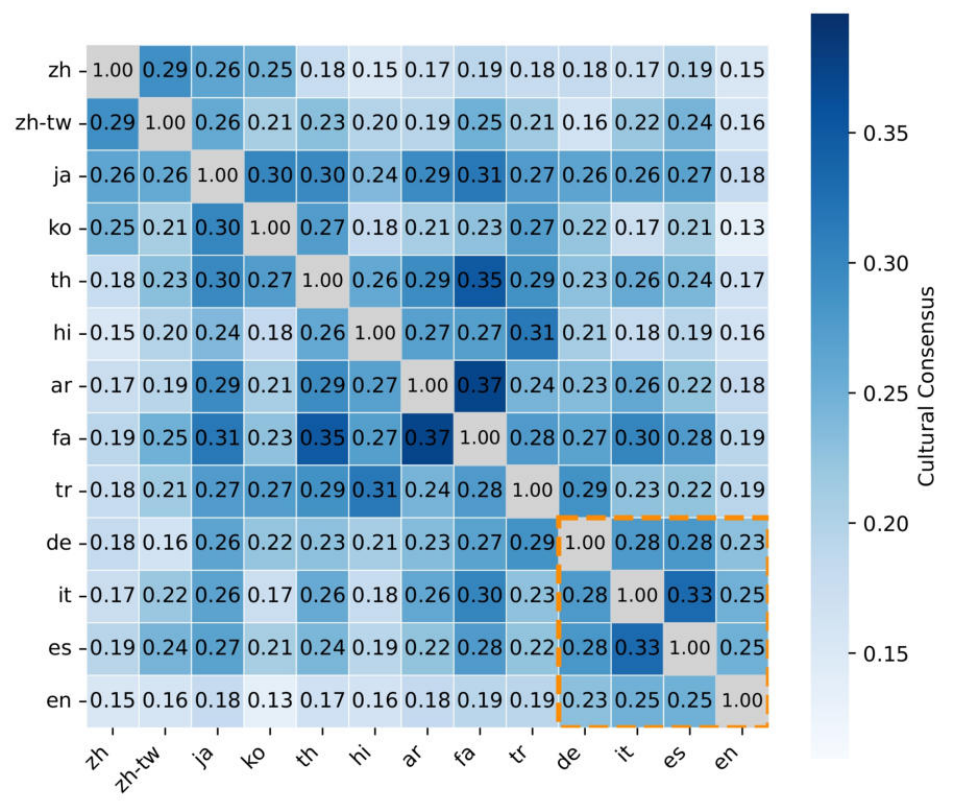} &
    \includegraphics[width=0.3\textwidth]{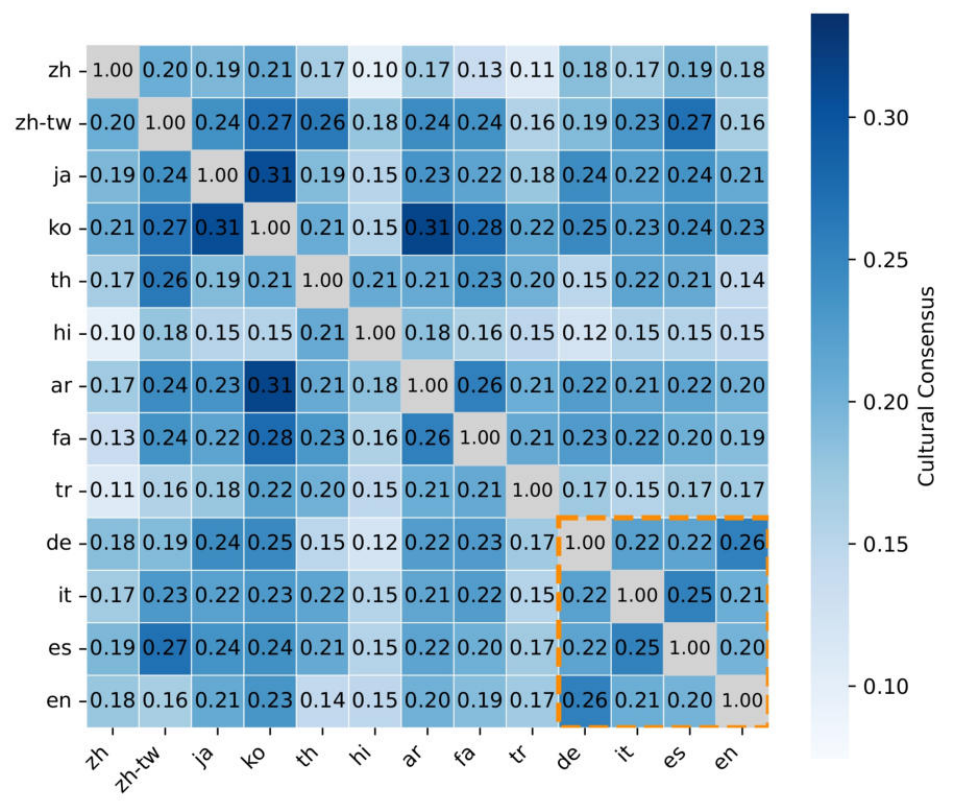} \\
    (a) Food & (b) Beverage & (c) Clothing \\[1.5ex]
    \includegraphics[width=0.3\textwidth]{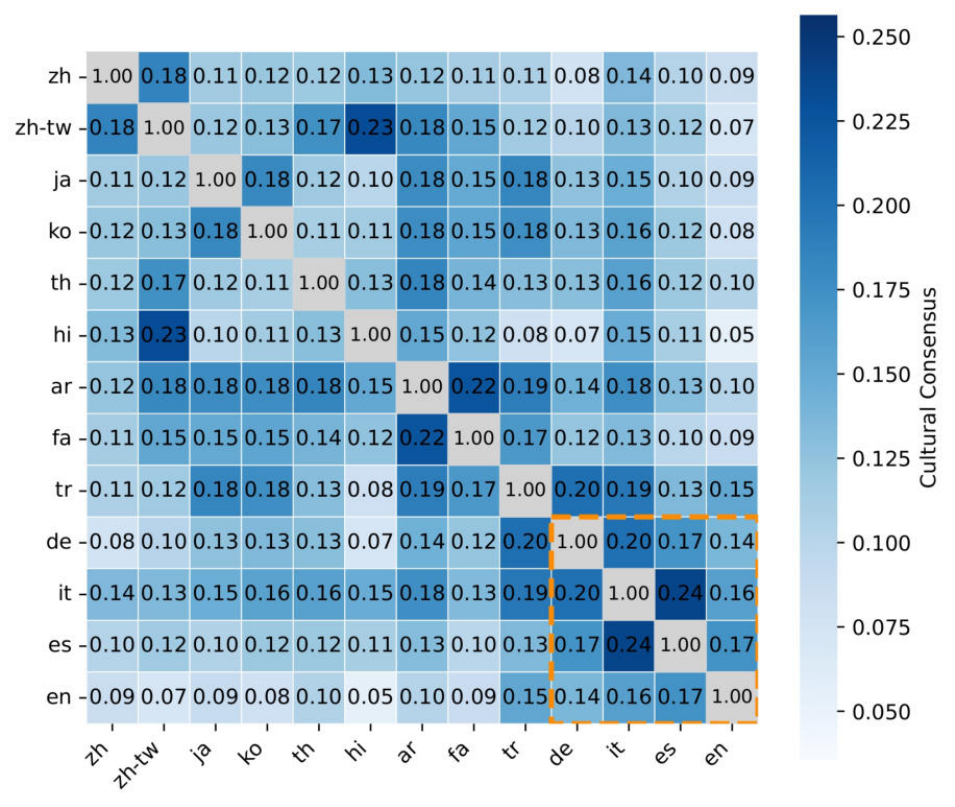} &
    \includegraphics[width=0.3\textwidth]{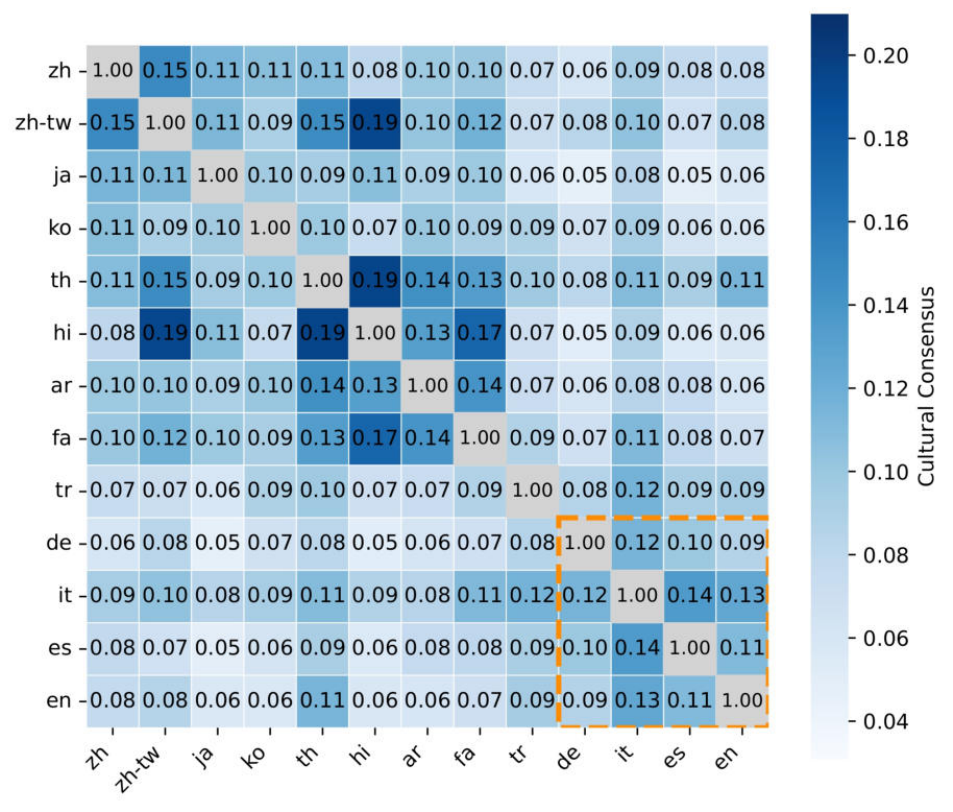} &
    \includegraphics[width=0.3\textwidth]{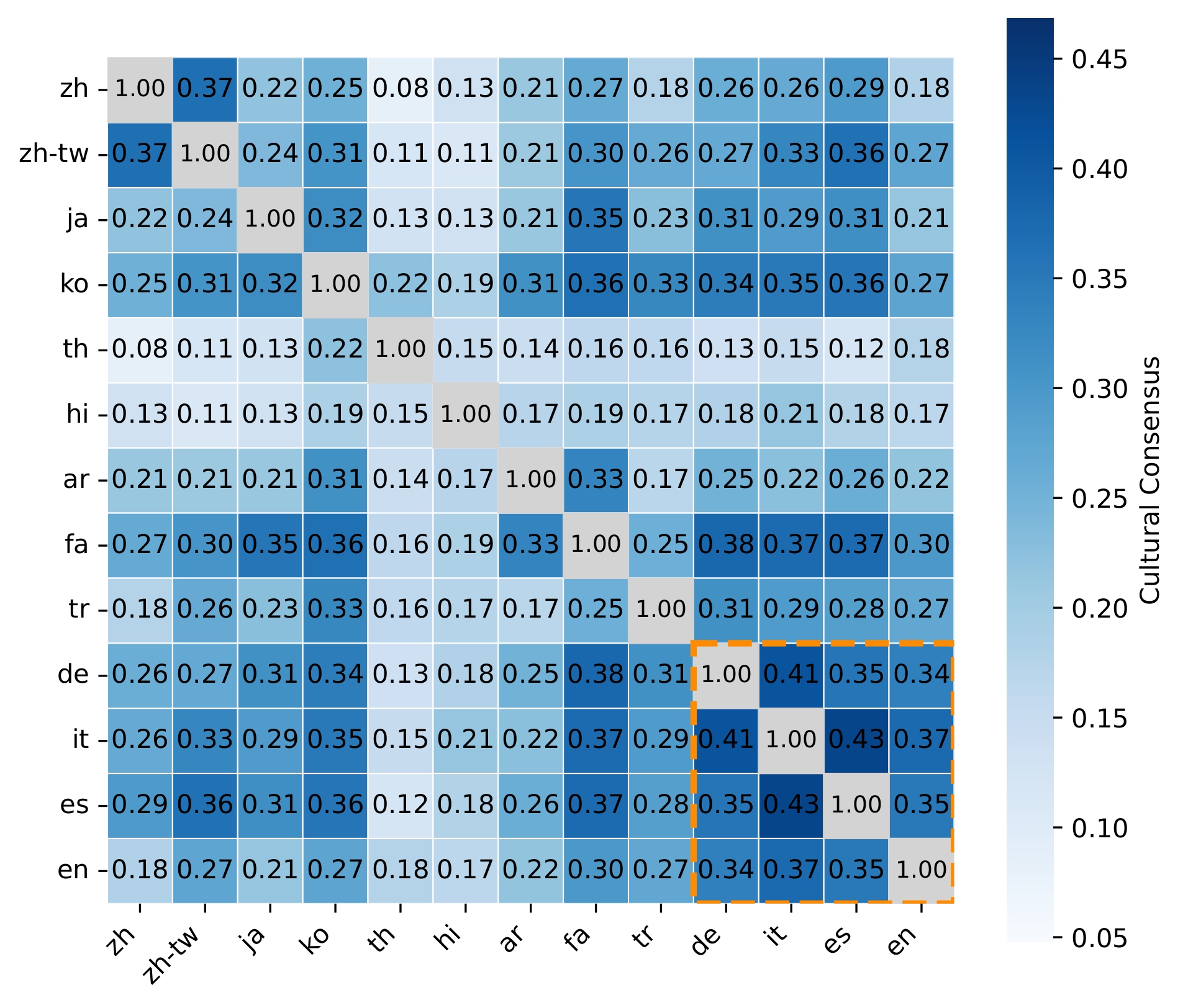} \\
    (d) Music & (e) Book & (f) Transportation
  \end{tabular}}
  \caption{Culture consensus results for \textbf{LLaMA3} across six topics. Yellow colors are used to distinguish Europe language region.}
  \label{fig:consensus_llama3}
\end{figure*}

\begin{figure*}[htbp]
  \centering
  \resizebox{\textwidth}{!}{
  \begin{tabular}{ccc}
    \includegraphics[width=0.3\textwidth]{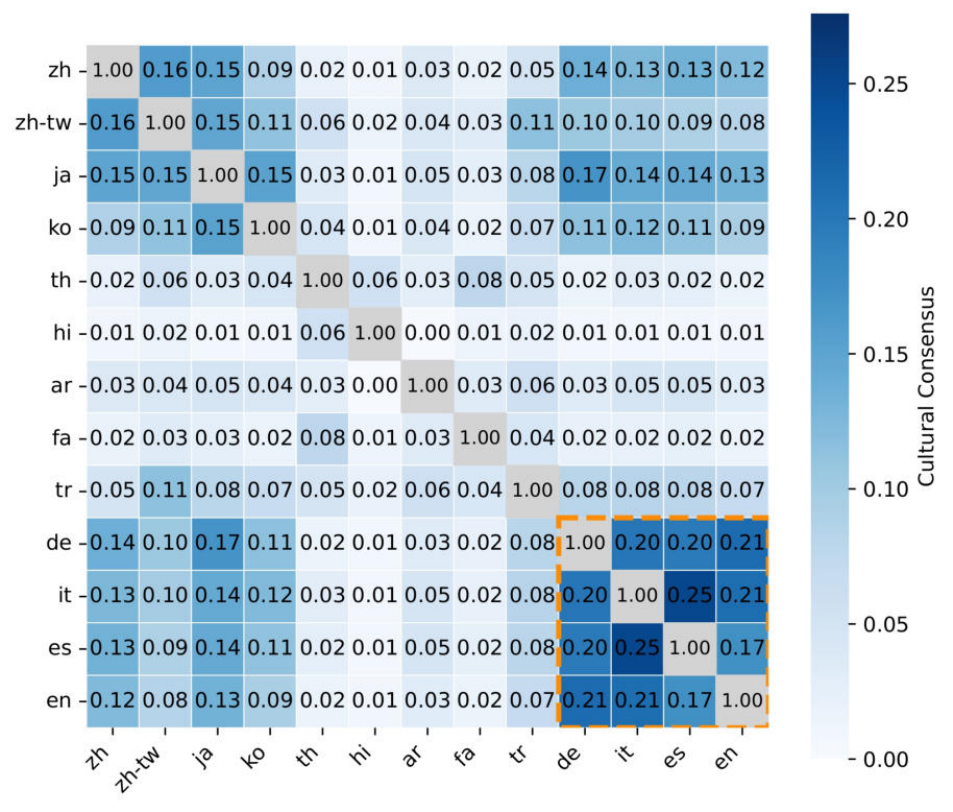} &
    \includegraphics[width=0.3\textwidth]{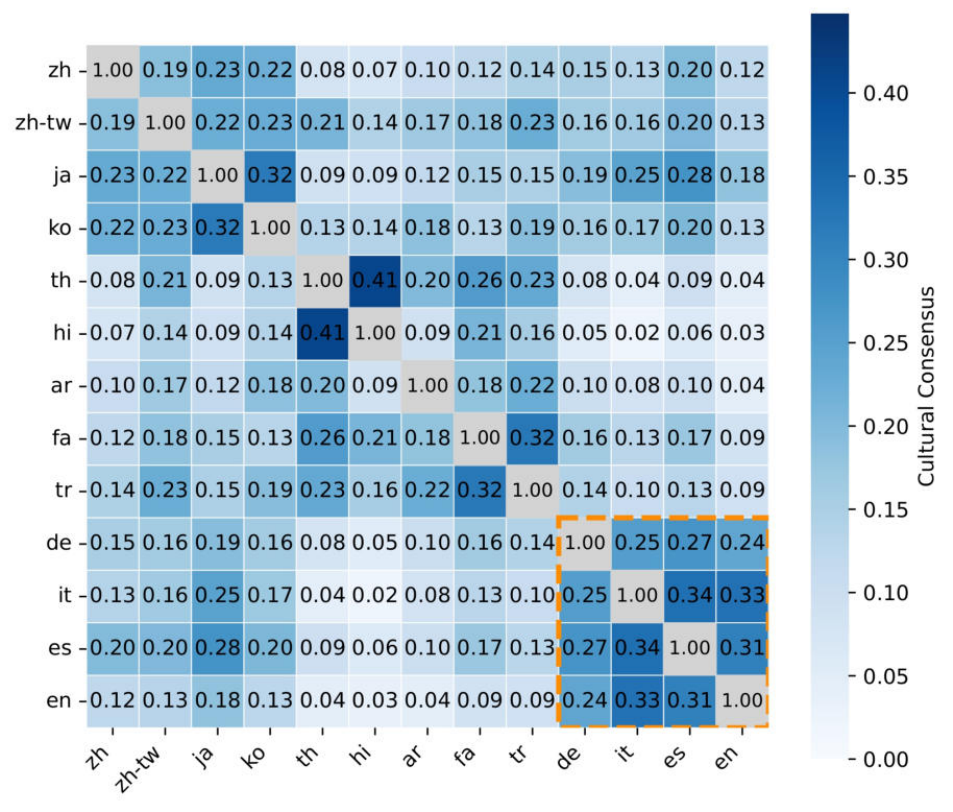} &
    \includegraphics[width=0.3\textwidth]{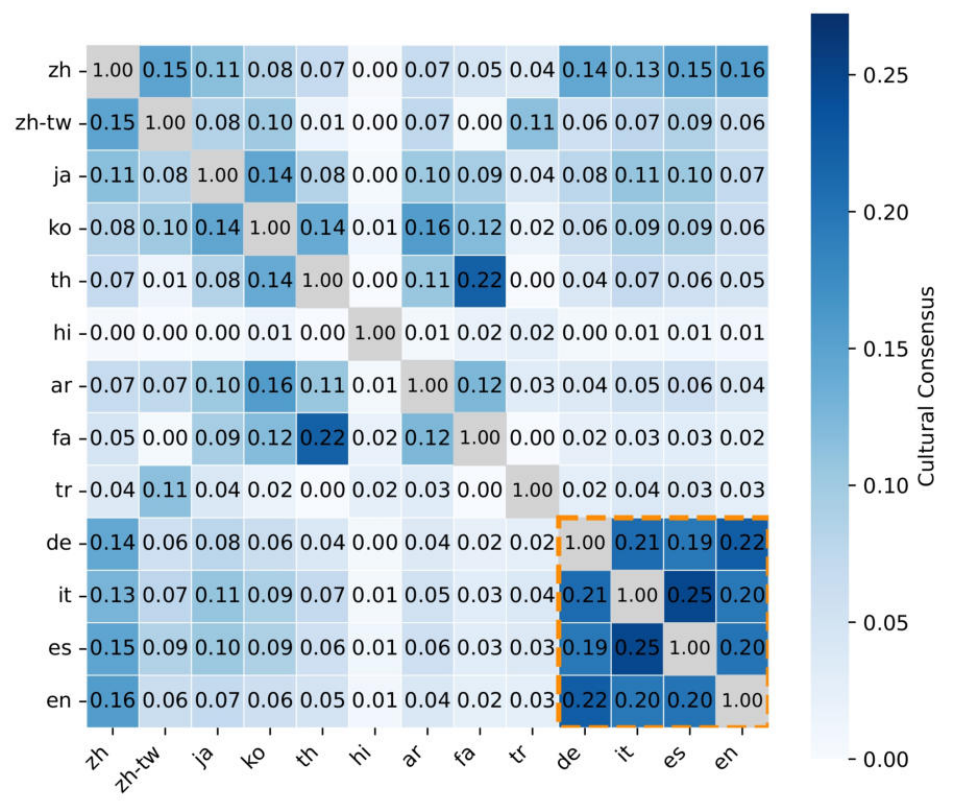} \\
    (a) Food & (b) Beverage & (c) Clothing \\[1.5ex]
    \includegraphics[width=0.3\textwidth]{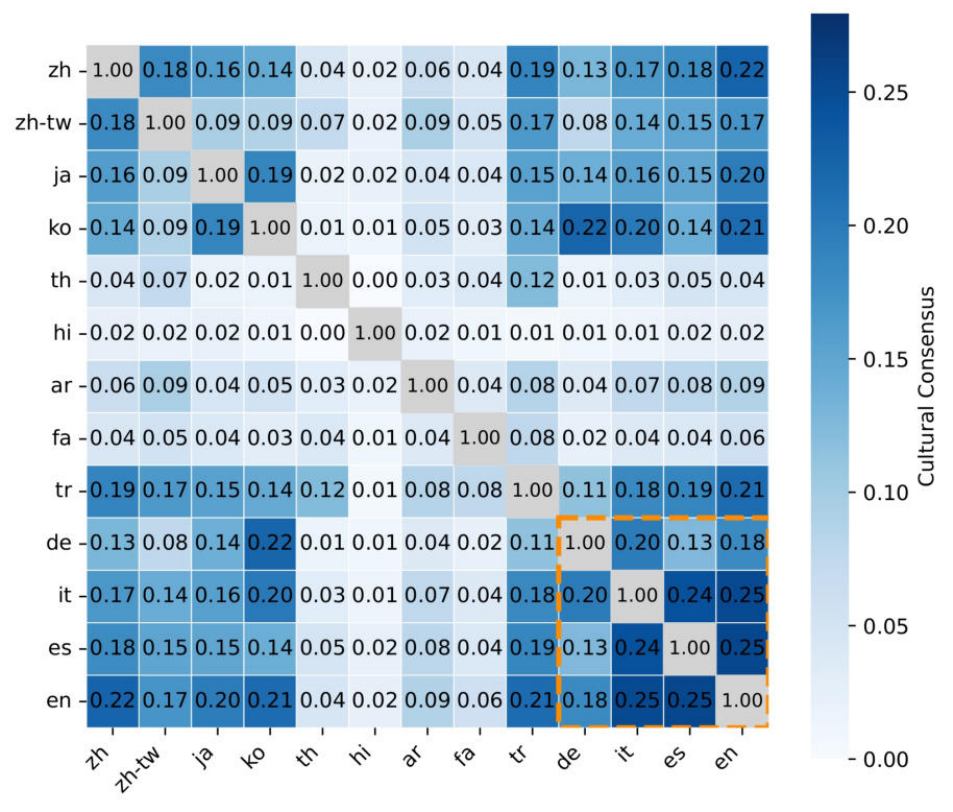} &
    \includegraphics[width=0.3\textwidth]{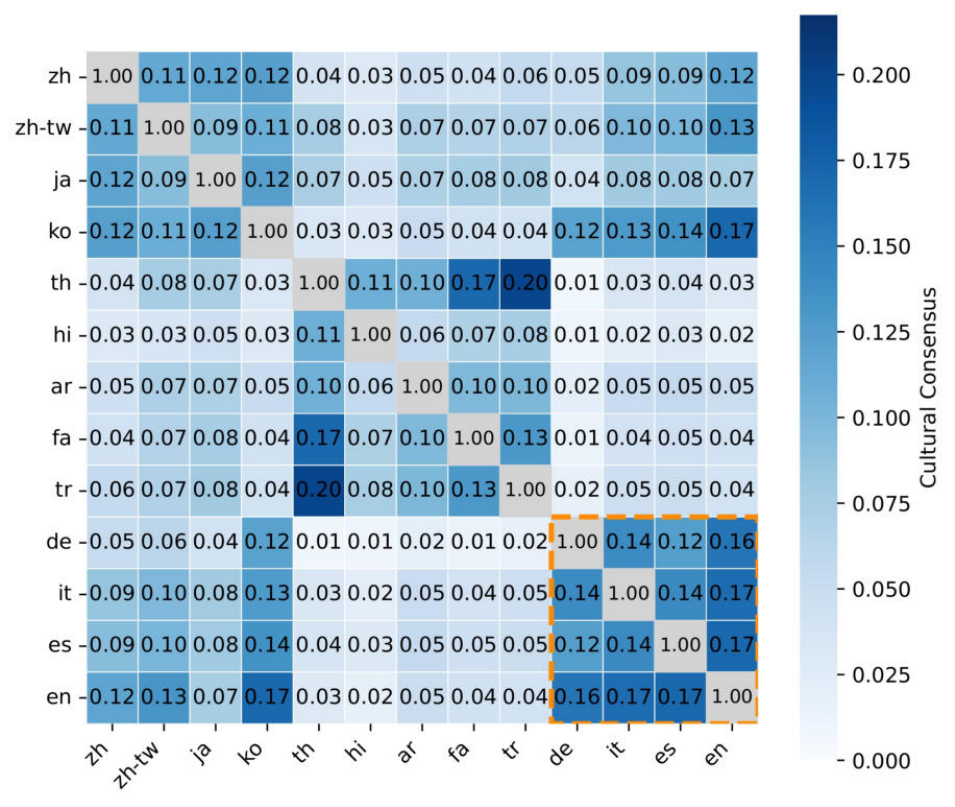} &
    \includegraphics[width=0.3\textwidth]{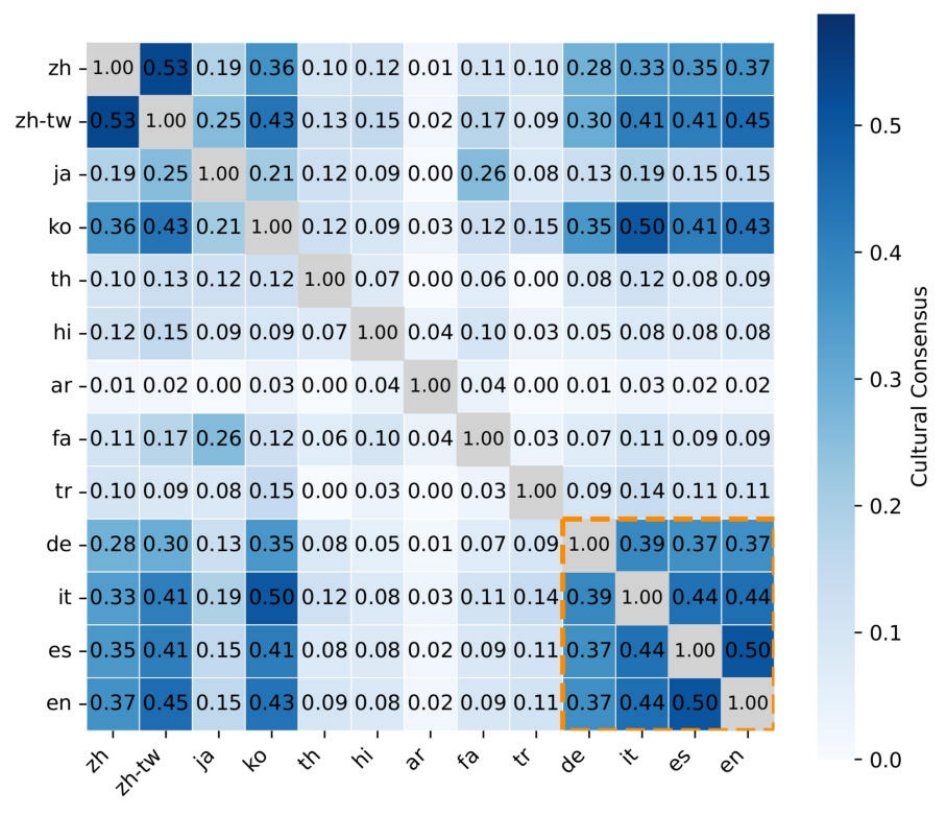} \\
    (d) Music & (e) Book & (f) Transportation
  \end{tabular}}
  \caption{Culture consensus results for \textbf{Mistral} across six topics. Yellow colors are used to distinguish Europe language region.}
  \label{fig:consensus_mistral}
\end{figure*}

\begin{figure*}[htbp]
  \centering
  \resizebox{\textwidth}{!}{
  \begin{tabular}{ccc}
    \includegraphics[width=0.3\textwidth]{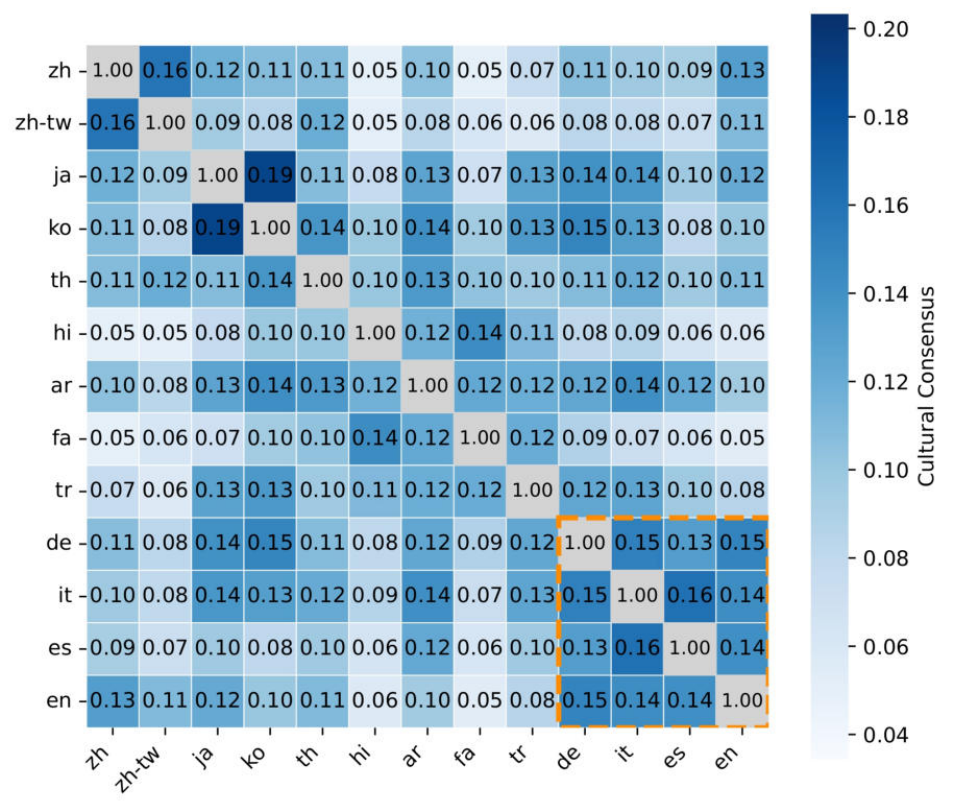} &
    \includegraphics[width=0.3\textwidth]{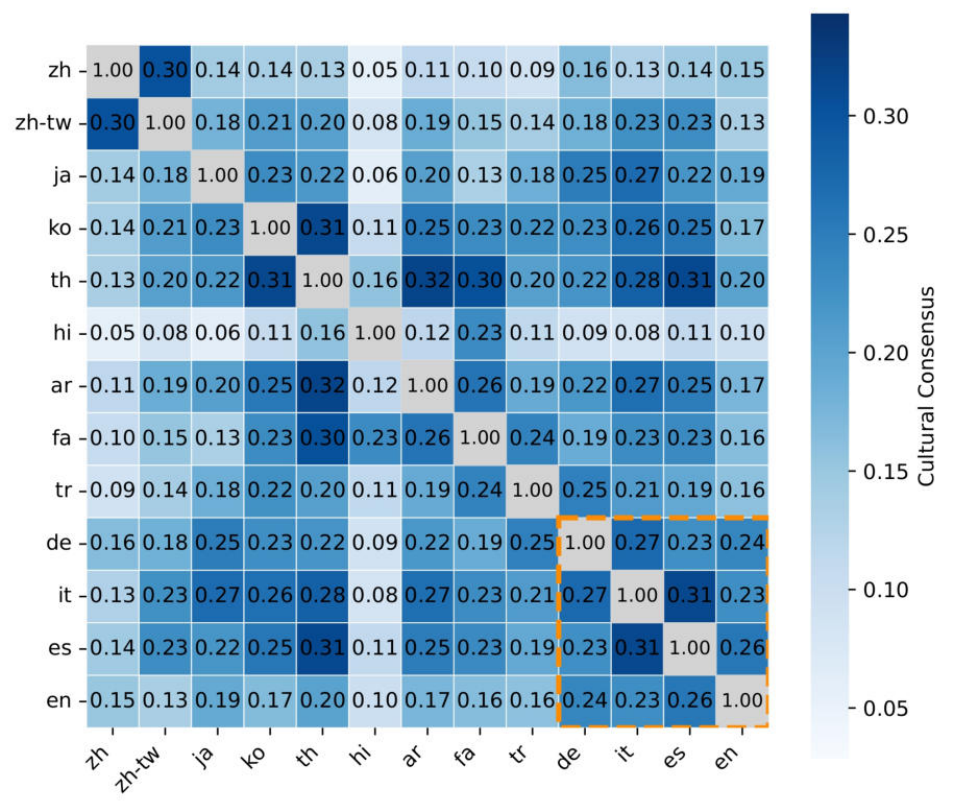} &
    \includegraphics[width=0.3\textwidth]{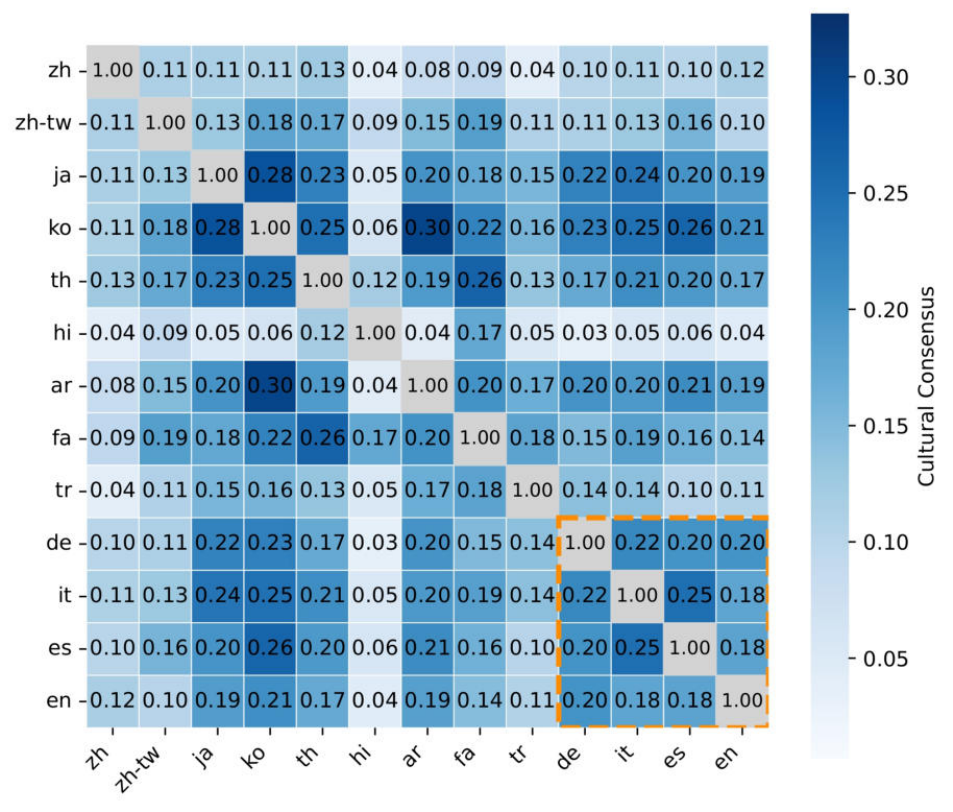} \\
    (a) Food & (b) Beverage & (c) Clothing \\[1.5ex]
    \includegraphics[width=0.3\textwidth]{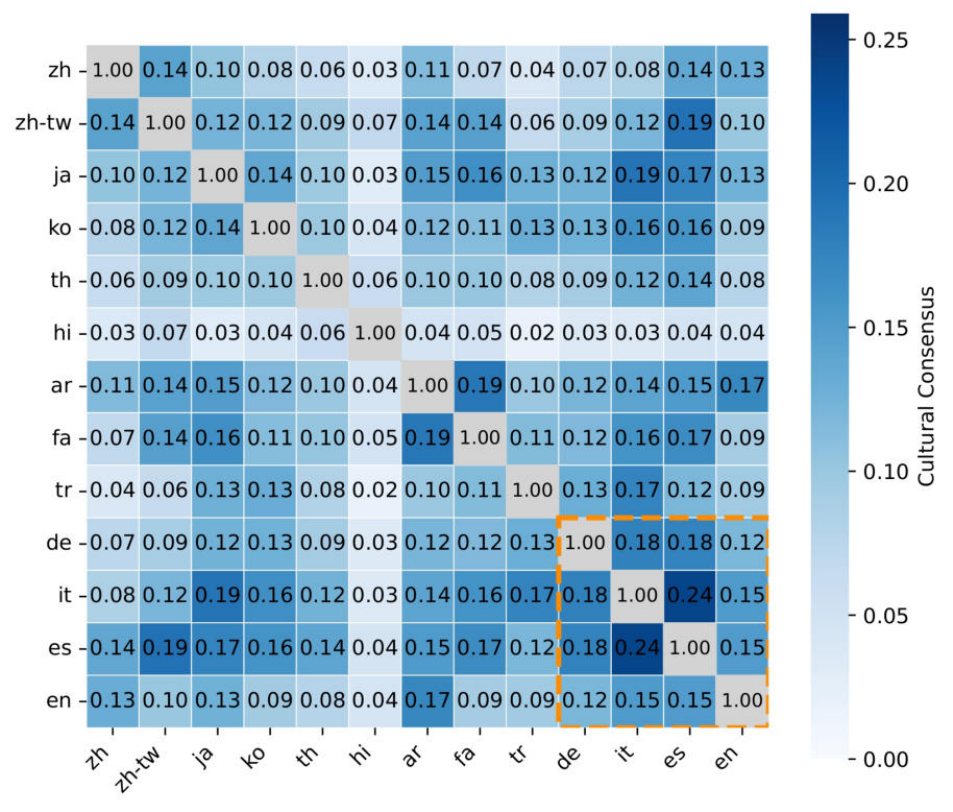} &
    \includegraphics[width=0.3\textwidth]{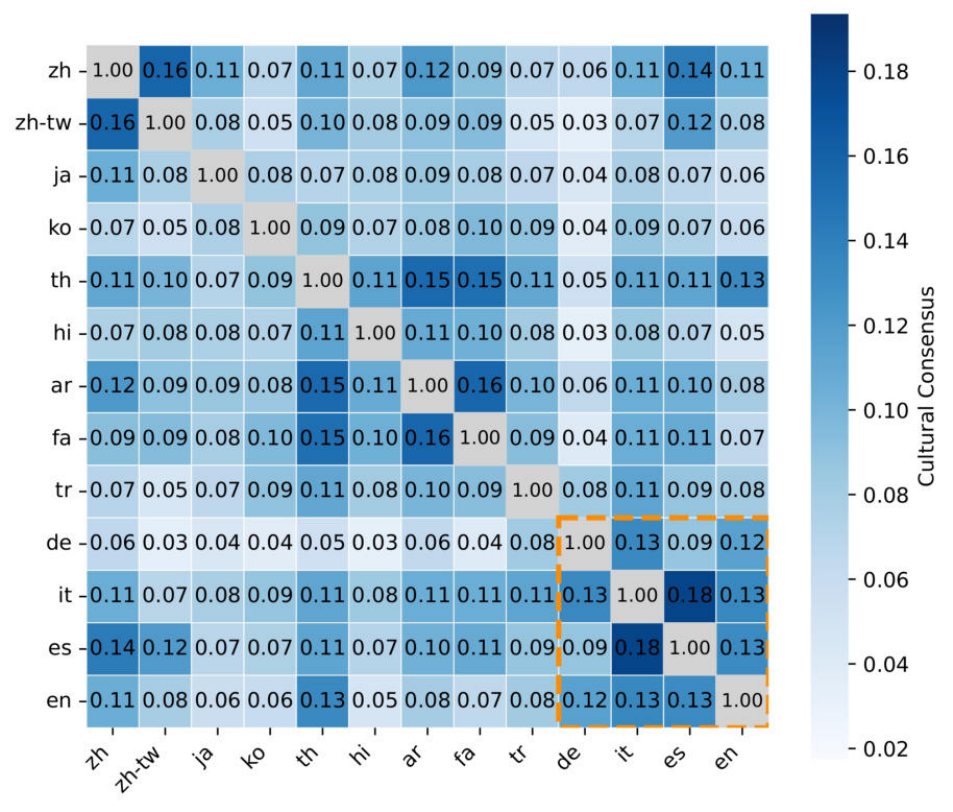} &
    \includegraphics[width=0.3\textwidth]{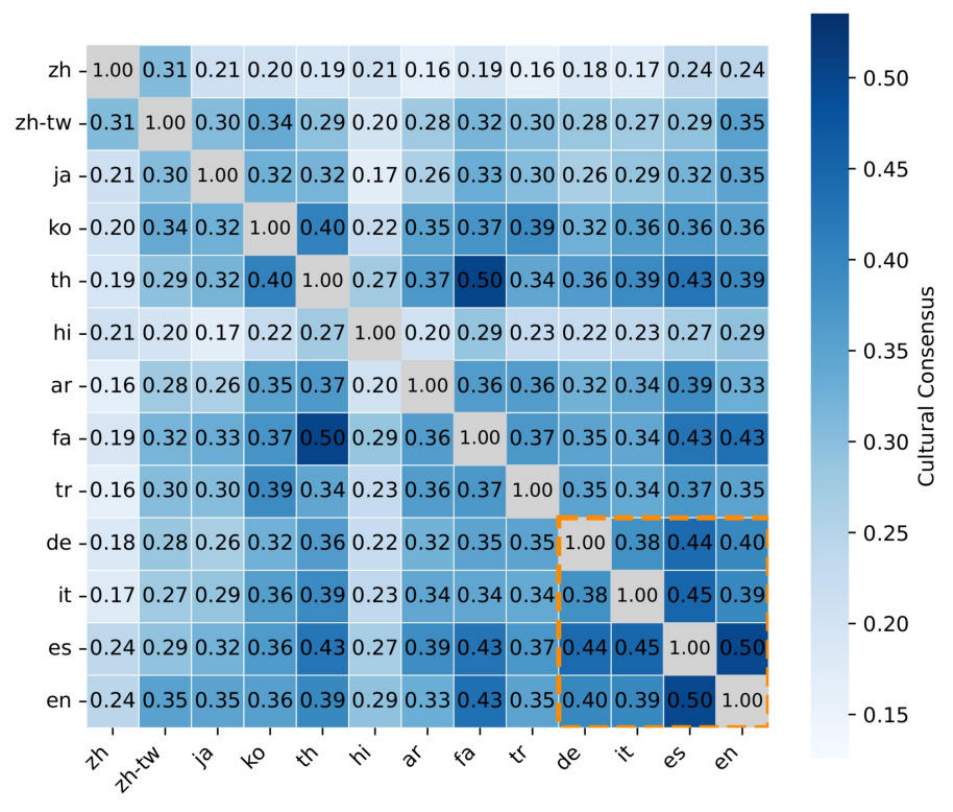} \\
    (d) Music & (e) Book & (f) Transportation
  \end{tabular}}
  \caption{Culture consensus results for \textbf{Qwen} across six topics. Yellow colors are used to distinguish Europe language region.}
  \label{fig:consensus_qwen}
\end{figure*}


\begin{thebibliography}{49}
\providecommand{\natexlab}[1]{#1}

\bibitem[{Agarwal et~al.(2024)Agarwal, Tanmay, Khandelwal, and Choudhury}]{agarwal-etal-2024-ethical}
Utkarsh Agarwal, Kumar Tanmay, Aditi Khandelwal, and Monojit Choudhury. 2024.
\newblock \href {https://aclanthology.org/2024.lrec-main.560/} {Ethical reasoning and moral value alignment of {LLM}s depend on the language we prompt them in}.
\newblock In \emph{Proceedings of the 2024 Joint International Conference on Computational Linguistics, Language Resources and Evaluation (LREC-COLING 2024)}, pages 6330--6340, Torino, Italia. ELRA and ICCL.

\bibitem[{Bang et~al.(2024)Bang, Chen, Lee, and Fung}]{bang-etal-2024-measuring}
Yejin Bang, Delong Chen, Nayeon Lee, and Pascale Fung. 2024.
\newblock \href {https://doi.org/10.18653/v1/2024.acl-long.600} {Measuring political bias in large language models: What is said and how it is said}.
\newblock In \emph{Proceedings of the 62nd Annual Meeting of the Association for Computational Linguistics (Volume 1: Long Papers)}, pages 11142--11159, Bangkok, Thailand. Association for Computational Linguistics.

\bibitem[{Bender et~al.(2021)Bender, Gebru, McMillan-Major, and Shmitchell}]{bender2021dangers}
Emily~M Bender, Timnit Gebru, Angelina McMillan-Major, and Shmargaret Shmitchell. 2021.
\newblock On the dangers of stochastic parrots: Can language models be too big?
\newblock In \emph{Proceedings of the 2021 ACM conference on fairness, accountability, and transparency}, pages 610--623.

\bibitem[{Berger and Ponti(2025)}]{berger-ponti-2025-cross}
Uri Berger and Edoardo Ponti. 2025.
\newblock \href {https://aclanthology.org/2025.naacl-long.478/} {Cross-lingual and cross-cultural variation in image descriptions}.
\newblock In \emph{Proceedings of the 2025 Conference of the Nations of the Americas Chapter of the Association for Computational Linguistics: Human Language Technologies (Volume 1: Long Papers)}, pages 9453--9465, Albuquerque, New Mexico. Association for Computational Linguistics.

\bibitem[{Bhatt and Diaz(2024)}]{bhatt-diaz-2024-extrinsic}
Shaily Bhatt and Fernando Diaz. 2024.
\newblock \href {https://doi.org/10.18653/v1/2024.findings-emnlp.942} {Extrinsic evaluation of cultural competence in large language models}.
\newblock In \emph{Findings of the Association for Computational Linguistics: EMNLP 2024}, pages 16055--16074, Miami, Florida, USA. Association for Computational Linguistics.

\bibitem[{Blodgett et~al.(2020)Blodgett, Barocas, Daum{\'e}~III, and Wallach}]{blodgett-etal-2020-language}
Su~Lin Blodgett, Solon Barocas, Hal Daum{\'e}~III, and Hanna Wallach. 2020.
\newblock \href {https://doi.org/10.18653/v1/2020.acl-main.485} {Language (technology) is power: A critical survey of {\textquotedblleft}bias{\textquotedblright} in {NLP}}.
\newblock In \emph{Proceedings of the 58th Annual Meeting of the Association for Computational Linguistics}, pages 5454--5476, Online. Association for Computational Linguistics.

\bibitem[{Chafe(1994)}]{chafe1994discourse}
Wallace Chafe. 1994.
\newblock \emph{Discourse, consciousness, and time: The flow and displacement of conscious experience in speaking and writing}.
\newblock University of Chicago Press.

\bibitem[{Chen et~al.(2022{\natexlab{a}})Chen, Guo, Gu, Liu, and Wang}]{DBLP:conf/icassp/ChenGGLW22}
Beiduo Chen, Wu~Guo, Bin Gu, Quan Liu, and Yongchao Wang. 2022{\natexlab{a}}.
\newblock \href {https://doi.org/10.1109/ICASSP43922.2022.9747720} {Multi-level contrastive learning for cross-lingual alignment}.
\newblock In \emph{{IEEE} International Conference on Acoustics, Speech and Signal Processing, {ICASSP} 2022, Virtual and Singapore, 23-27 May 2022}, pages 7947--7951. {IEEE}.

\bibitem[{Chen et~al.(2022{\natexlab{b}})Chen, Guo, Liu, and Tao}]{DBLP:conf/icpr/ChenGLT22}
Beiduo Chen, Wu~Guo, Quan Liu, and Kun Tao. 2022{\natexlab{b}}.
\newblock \href {https://doi.org/10.1109/ICPR56361.2022.9956721} {Feature aggregation in zero-shot cross-lingual transfer using multilingual {BERT}}.
\newblock In \emph{26th International Conference on Pattern Recognition, {ICPR} 2022, Montreal, QC, Canada, August 21-25, 2022}, pages 1428--1435. {IEEE}.

\bibitem[{Chen et~al.(2022{\natexlab{c}})Chen, Ma, Qi, Guo, Ling, and Liu}]{chen-etal-2022-ustc}
Beiduo Chen, Jun-Yu Ma, Jiajun Qi, Wu~Guo, Zhen-Hua Ling, and Quan Liu. 2022{\natexlab{c}}.
\newblock \href {https://doi.org/10.18653/v1/2022.semeval-1.223} {{USTC}-{NELSLIP} at {S}em{E}val-2022 task 11: Gazetteer-adapted integration network for multilingual complex named entity recognition}.
\newblock In \emph{Proceedings of the 16th International Workshop on Semantic Evaluation (SemEval-2022)}, pages 1613--1622, Seattle, United States. Association for Computational Linguistics.

\bibitem[{Cheng et~al.(2023)Cheng, Durmus, and Jurafsky}]{cheng-etal-2023-marked}
Myra Cheng, Esin Durmus, and Dan Jurafsky. 2023.
\newblock \href {https://doi.org/10.18653/v1/2023.acl-long.84} {Marked personas: Using natural language prompts to measure stereotypes in language models}.
\newblock In \emph{Proceedings of the 61st Annual Meeting of the Association for Computational Linguistics (Volume 1: Long Papers)}, pages 1504--1532, Toronto, Canada. Association for Computational Linguistics.

\bibitem[{Chiu et~al.(2024)Chiu, Jiang, Lin, Park, Li, Ravi, Bhatia, Antoniak, Tsvetkov, Shwartz, and Choi}]{chiu2024culturalbenchrobustdiversechallenging}
Yu~Ying Chiu, Liwei Jiang, Bill~Yuchen Lin, Chan~Young Park, Shuyue~Stella Li, Sahithya Ravi, Mehar Bhatia, Maria Antoniak, Yulia Tsvetkov, Vered Shwartz, and Yejin Choi. 2024.
\newblock \href {https://arxiv.org/abs/2410.02677} {Culturalbench: a robust, diverse and challenging benchmark on measuring the (lack of) cultural knowledge of llms}.
\newblock \emph{Preprint}, arXiv:2410.02677.

\bibitem[{Deshpande et~al.(2023)Deshpande, Murahari, Rajpurohit, Kalyan, and Narasimhan}]{deshpande-etal-2023-toxicity}
Ameet Deshpande, Vishvak Murahari, Tanmay Rajpurohit, Ashwin Kalyan, and Karthik Narasimhan. 2023.
\newblock \href {https://doi.org/10.18653/v1/2023.findings-emnlp.88} {Toxicity in chatgpt: Analyzing persona-assigned language models}.
\newblock In \emph{Findings of the Association for Computational Linguistics: EMNLP 2023}, pages 1236--1270, Singapore. Association for Computational Linguistics.

\bibitem[{Dewaele and Pavlenko(2003)}]{dewaele2003productivity}
Jean-Marc Dewaele and Aneta Pavlenko. 2003.
\newblock Productivity and lexical diversity in native and non-native speech: A study of cross-cultural effects.
\newblock \emph{Effects of the second language on the first}, 3:120.

\bibitem[{Ghaboura et~al.(2025)Ghaboura, More, Thawkar, Ghallabi, Thawakar, Khan, Cholakkal, Khan, and Anwer}]{ghaboura-etal-2025-time}
Sara Ghaboura, Ketan~Pravin More, Ritesh Thawkar, Wafa~Al Ghallabi, Omkar Thawakar, Fahad~Shahbaz Khan, Hisham Cholakkal, Salman Khan, and Rao~Muhammad Anwer. 2025.
\newblock \href {https://doi.org/10.18653/v1/2025.findings-acl.1211} {Time travel: A comprehensive benchmark to evaluate {LMM}s on historical and cultural artifacts}.
\newblock In \emph{Findings of the Association for Computational Linguistics: ACL 2025}, pages 23627--23641, Vienna, Austria. Association for Computational Linguistics.

\bibitem[{Grattafiori et~al.(2024)Grattafiori, Dubey, Jauhri, Pandey, Kadian, Al-Dahle, Letman, Mathur, Schelten, Vaughan et~al.}]{grattafiori2024llama}
Aaron Grattafiori, Abhimanyu Dubey, Abhinav Jauhri, Abhinav Pandey, Abhishek Kadian, Ahmad Al-Dahle, Aiesha Letman, Akhil Mathur, Alan Schelten, Alex Vaughan, and 1 others. 2024.
\newblock The llama 3 herd of models.
\newblock \emph{arXiv preprint arXiv:2407.21783}.

\bibitem[{Hedderich et~al.(2025)Hedderich, Wang, Zhao, Eichin, Fischer, and Plank}]{hedderich-etal-2025-whats}
Michael~A. Hedderich, Anyi Wang, Raoyuan Zhao, Florian Eichin, Jonas Fischer, and Barbara Plank. 2025.
\newblock \href {https://doi.org/10.18653/v1/2025.acl-long.985} {What{'}s the difference? supporting users in identifying the effects of prompt and model changes through token patterns}.
\newblock In \emph{Proceedings of the 63rd Annual Meeting of the Association for Computational Linguistics (Volume 1: Long Papers)}, pages 20093--20123, Vienna, Austria. Association for Computational Linguistics.

\bibitem[{Hu et~al.(2024)Hu, Maistro, and Hershcovich}]{hu-etal-2024-bridging}
Tianyi Hu, Maria Maistro, and Daniel Hershcovich. 2024.
\newblock \href {https://doi.org/10.18653/v1/2024.emnlp-main.61} {Bridging cultures in the kitchen: A framework and benchmark for cross-cultural recipe retrieval}.
\newblock In \emph{Proceedings of the 2024 Conference on Empirical Methods in Natural Language Processing}, pages 1068--1080, Miami, Florida, USA. Association for Computational Linguistics.

\bibitem[{Hutchinson et~al.(2020)Hutchinson, Prabhakaran, Denton, Webster, Zhong, and Denuyl}]{hutchinson-etal-2020-social}
Ben Hutchinson, Vinodkumar Prabhakaran, Emily Denton, Kellie Webster, Yu~Zhong, and Stephen Denuyl. 2020.
\newblock \href {https://doi.org/10.18653/v1/2020.acl-main.487} {Social biases in {NLP} models as barriers for persons with disabilities}.
\newblock In \emph{Proceedings of the 58th Annual Meeting of the Association for Computational Linguistics}, pages 5491--5501, Online. Association for Computational Linguistics.

\bibitem[{Jiang et~al.(2023)Jiang, Sablayrolles, Mensch, Bamford, Chaplot, Casas, Bressand, Lengyel, Lample, Saulnier et~al.}]{jiang2023mistral}
Albert~Q Jiang, Alexandre Sablayrolles, Arthur Mensch, Chris Bamford, Devendra~Singh Chaplot, Diego de~las Casas, Florian Bressand, Gianna Lengyel, Guillaume Lample, Lucile Saulnier, and 1 others. 2023.
\newblock Mistral 7b.
\newblock \emph{arXiv preprint arXiv:2310.06825}.

\bibitem[{Jiang and Joshi(2024)}]{jiang-joshi-2024-cpopqa}
Ming Jiang and Mansi Joshi. 2024.
\newblock \href {https://doi.org/10.18653/v1/2024.naacl-short.52} {{CP}op{QA}: Ranking cultural concept popularity by {LLM}s}.
\newblock In \emph{Proceedings of the 2024 Conference of the North American Chapter of the Association for Computational Linguistics: Human Language Technologies (Volume 2: Short Papers)}, pages 615--630, Mexico City, Mexico. Association for Computational Linguistics.

\bibitem[{Jin et~al.(2024)Jin, Kim, Lee, Yoo, Oh, and Lee}]{jin-etal-2024-kobbq}
Jiho Jin, Jiseon Kim, Nayeon Lee, Haneul Yoo, Alice Oh, and Hwaran Lee. 2024.
\newblock \href {https://doi.org/10.1162/tacl_a_00661} {{K}o{BBQ}: {K}orean bias benchmark for question answering}.
\newblock \emph{Transactions of the Association for Computational Linguistics}, 12:507--524.

\bibitem[{Jones and Galliers(1995)}]{jones1995evaluating}
Karen~Sparck Jones and Julia~R Galliers. 1995.
\newblock Evaluating natural language processing systems: An analysis and review.

\bibitem[{Liu et~al.(2024{\natexlab{a}})Liu, Feng, Xue, Wang, Wu, Lu, Zhao, Deng, Zhang, Ruan et~al.}]{liu2024deepseek}
Aixin Liu, Bei Feng, Bing Xue, Bingxuan Wang, Bochao Wu, Chengda Lu, Chenggang Zhao, Chengqi Deng, Chenyu Zhang, Chong Ruan, and 1 others. 2024{\natexlab{a}}.
\newblock Deepseek-v3 technical report.
\newblock \emph{arXiv preprint arXiv:2412.19437}.

\bibitem[{Liu et~al.(2024{\natexlab{b}})Liu, Gurevych, and Korhonen}]{liu2024culturally}
Chen~Cecilia Liu, Iryna Gurevych, and Anna Korhonen. 2024{\natexlab{b}}.
\newblock Culturally aware and adapted nlp: A taxonomy and a survey of the state of the art.
\newblock \emph{arXiv preprint arXiv:2406.03930}.

\bibitem[{Ma et~al.(2024)Ma, Wang, Hu, Haensch, Hedderich, Plank, and Kreuter}]{ma-etal-2024-potential}
Bolei Ma, Xinpeng Wang, Tiancheng Hu, Anna-Carolina Haensch, Michael~A. Hedderich, Barbara Plank, and Frauke Kreuter. 2024.
\newblock \href {https://doi.org/10.18653/v1/2024.findings-emnlp.513} {The potential and challenges of evaluating attitudes, opinions, and values in large language models}.
\newblock In \emph{Findings of the Association for Computational Linguistics: EMNLP 2024}, pages 8783--8805, Miami, Florida, USA. Association for Computational Linguistics.

\bibitem[{Myung et~al.(2024)Myung, Lee, Zhou, Jin, Putri, Antypas, Borkakoty, Kim, Perez-Almendros, Ayele et~al.}]{myung2024blend}
Junho Myung, Nayeon Lee, Yi~Zhou, Jiho Jin, Rifki Putri, Dimosthenis Antypas, Hsuvas Borkakoty, Eunsu Kim, Carla Perez-Almendros, Abinew~Ali Ayele, and 1 others. 2024.
\newblock Blend: A benchmark for llms on everyday knowledge in diverse cultures and languages.
\newblock \emph{Advances in Neural Information Processing Systems}, 37:78104--78146.

\bibitem[{Naous et~al.(2024)Naous, Ryan, Ritter, and Xu}]{naous-etal-2024-beer}
Tarek Naous, Michael~J Ryan, Alan Ritter, and Wei Xu. 2024.
\newblock \href {https://doi.org/10.18653/v1/2024.acl-long.862} {Having beer after prayer? measuring cultural bias in large language models}.
\newblock In \emph{Proceedings of the 62nd Annual Meeting of the Association for Computational Linguistics (Volume 1: Long Papers)}, pages 16366--16393, Bangkok, Thailand. Association for Computational Linguistics.

\bibitem[{Nayak et~al.(2024)Nayak, Jain, Awal, Reddy, Steenkiste, Hendricks, Stanczak, and Agrawal}]{nayak-etal-2024-benchmarking}
Shravan Nayak, Kanishk Jain, Rabiul Awal, Siva Reddy, Sjoerd~Van Steenkiste, Lisa~Anne Hendricks, Karolina Stanczak, and Aishwarya Agrawal. 2024.
\newblock \href {https://doi.org/10.18653/v1/2024.emnlp-main.329} {Benchmarking vision language models for cultural understanding}.
\newblock In \emph{Proceedings of the 2024 Conference on Empirical Methods in Natural Language Processing}, pages 5769--5790, Miami, Florida, USA. Association for Computational Linguistics.

\bibitem[{Nikandrou et~al.(2025)Nikandrou, Pantazopoulos, Vitsakis, Konstas, and Suglia}]{nikandrou-etal-2025-crope}
Malvina Nikandrou, Georgios Pantazopoulos, Nikolas Vitsakis, Ioannis Konstas, and Alessandro Suglia. 2025.
\newblock \href {https://aclanthology.org/2025.naacl-long.402/} {{CROPE}: Evaluating in-context adaptation of vision and language models to culture-specific concepts}.
\newblock In \emph{Proceedings of the 2025 Conference of the Nations of the Americas Chapter of the Association for Computational Linguistics: Human Language Technologies (Volume 1: Long Papers)}, pages 7917--7936, Albuquerque, New Mexico. Association for Computational Linguistics.

\bibitem[{Onohara et~al.(2025)Onohara, Miyai, Imajuku, Egashira, Baek, Yue, Neubig, and Aizawa}]{onohara-etal-2025-jmmmu}
Shota Onohara, Atsuyuki Miyai, Yuki Imajuku, Kazuki Egashira, Jeonghun Baek, Xiang Yue, Graham Neubig, and Kiyoharu Aizawa. 2025.
\newblock \href {https://aclanthology.org/2025.naacl-long.43/} {{JMMMU}: A {J}apanese massive multi-discipline multimodal understanding benchmark for culture-aware evaluation}.
\newblock In \emph{Proceedings of the 2025 Conference of the Nations of the Americas Chapter of the Association for Computational Linguistics: Human Language Technologies (Volume 1: Long Papers)}, pages 932--950, Albuquerque, New Mexico. Association for Computational Linguistics.

\bibitem[{OpenAI(2025)}]{openai2025gpt4o}
OpenAI. 2025.
\newblock Openai o3 and o4-mini.
\newblock \url{https://openai.com/index/introducing-o3-and-o4-mini/}.
\newblock Accessed: 2025-05-10.

\bibitem[{Pawar et~al.(2024)Pawar, Park, Jin, Arora, Myung, Yadav, Haznitrama, Song, Oh, and Augenstein}]{pawar2024surveyculturalawarenesslanguage}
Siddhesh Pawar, Junyeong Park, Jiho Jin, Arnav Arora, Junho Myung, Srishti Yadav, Faiz~Ghifari Haznitrama, Inhwa Song, Alice Oh, and Isabelle Augenstein. 2024.
\newblock \href {https://arxiv.org/abs/2411.00860} {Survey of cultural awareness in language models: Text and beyond}.
\newblock \emph{Preprint}, arXiv:2411.00860.

\bibitem[{Pawar et~al.(2025)Pawar, Park, Jin, Arora, Myung, Yadav, Haznitrama, Song, Oh, and Augenstein}]{10.1162/COLI.a.14}
Siddhesh Pawar, Junyeong Park, Jiho Jin, Arnav Arora, Junho Myung, Srishti Yadav, Faiz~Ghifari Haznitrama, Inhwa Song, Alice Oh, and Isabelle Augenstein. 2025.
\newblock \href {https://doi.org/10.1162/COLI.a.14} {Survey of cultural awareness in language models: Text and beyond}.
\newblock \emph{Computational Linguistics}, pages 1--96.

\bibitem[{Peters et~al.(2018)Peters, Gr{\"u}ter, and Borovsky}]{peters2018vocabulary}
Ryan~E Peters, Theres Gr{\"u}ter, and Arielle Borovsky. 2018.
\newblock Vocabulary size and native speaker self-identification influence flexibility in linguistic prediction among adult bilinguals.
\newblock \emph{Applied Psycholinguistics}, 39(6):1439--1469.

\bibitem[{Qi et~al.(2023)Qi, Fern{\'a}ndez, and Bisazza}]{qi-etal-2023-cross}
Jirui Qi, Raquel Fern{\'a}ndez, and Arianna Bisazza. 2023.
\newblock \href {https://doi.org/10.18653/v1/2023.emnlp-main.658} {Cross-lingual consistency of factual knowledge in multilingual language models}.
\newblock In \emph{Proceedings of the 2023 Conference on Empirical Methods in Natural Language Processing}, pages 10650--10666, Singapore. Association for Computational Linguistics.

\bibitem[{Schneider and Sitaram(2024)}]{schneider-sitaram-2024-m5}
Florian Schneider and Sunayana Sitaram. 2024.
\newblock \href {https://doi.org/10.18653/v1/2024.findings-emnlp.250} {M5 {--} a diverse benchmark to assess the performance of large multimodal models across multilingual and multicultural vision-language tasks}.
\newblock In \emph{Findings of the Association for Computational Linguistics: EMNLP 2024}, pages 4309--4345, Miami, Florida, USA. Association for Computational Linguistics.

\bibitem[{Seveso et~al.(2025)Seveso, Potert{\`i}, Federici, Mezzanzanica, and Mercorio}]{seveso-etal-2025-italic}
Andrea Seveso, Daniele Potert{\`i}, Edoardo Federici, Mario Mezzanzanica, and Fabio Mercorio. 2025.
\newblock \href {https://aclanthology.org/2025.naacl-long.68/} {{ITALIC}: An {I}talian culture-aware natural language benchmark}.
\newblock In \emph{Proceedings of the 2025 Conference of the Nations of the Americas Chapter of the Association for Computational Linguistics: Human Language Technologies (Volume 1: Long Papers)}, pages 1469--1478, Albuquerque, New Mexico. Association for Computational Linguistics.

\bibitem[{Shi et~al.(2024)Shi, Li, Zhang, Ziems, Yu, Horesh, Paula, and Yang}]{shi-etal-2024-culturebank}
Weiyan Shi, Ryan Li, Yutong Zhang, Caleb Ziems, Sunny Yu, Raya Horesh, Rog{\'e}rio Abreu~De Paula, and Diyi Yang. 2024.
\newblock \href {https://doi.org/10.18653/v1/2024.findings-emnlp.288} {{C}ulture{B}ank: An online community-driven knowledge base towards culturally aware language technologies}.
\newblock In \emph{Findings of the Association for Computational Linguistics: EMNLP 2024}, pages 4996--5025, Miami, Florida, USA. Association for Computational Linguistics.

\bibitem[{Singh et~al.(2025)Singh, Romanou, Fourrier, Adelani, Ngui, Vila-Suero, Limkonchotiwat, Marchisio, Leong, Susanto, Ng, Longpre, Ruder, Ko, Bosselut, Oh, Martins, Choshen, Ippolito, Ferrante, Fadaee, Ermis, and Hooker}]{singh-etal-2025-global}
Shivalika Singh, Angelika Romanou, Cl{\'e}mentine Fourrier, David~Ifeoluwa Adelani, Jian~Gang Ngui, Daniel Vila-Suero, Peerat Limkonchotiwat, Kelly Marchisio, Wei~Qi Leong, Yosephine Susanto, Raymond Ng, Shayne Longpre, Sebastian Ruder, Wei-Yin Ko, Antoine Bosselut, Alice Oh, Andre Martins, Leshem Choshen, Daphne Ippolito, and 4 others. 2025.
\newblock \href {https://doi.org/10.18653/v1/2025.acl-long.919} {Global {MMLU}: Understanding and addressing cultural and linguistic biases in multilingual evaluation}.
\newblock In \emph{Proceedings of the 63rd Annual Meeting of the Association for Computational Linguistics (Volume 1: Long Papers)}, pages 18761--18799, Vienna, Austria. Association for Computational Linguistics.

\bibitem[{{\"U}st{\"u}n et~al.(2024){\"U}st{\"u}n, Aryabumi, Yong, Ko, D{'}souza, Onilude, Bhandari, Singh, Ooi, Kayid, Vargus, Blunsom, Longpre, Muennighoff, Fadaee, Kreutzer, and Hooker}]{ustun-etal-2024-aya}
Ahmet {\"U}st{\"u}n, Viraat Aryabumi, Zheng Yong, Wei-Yin Ko, Daniel D{'}souza, Gbemileke Onilude, Neel Bhandari, Shivalika Singh, Hui-Lee Ooi, Amr Kayid, Freddie Vargus, Phil Blunsom, Shayne Longpre, Niklas Muennighoff, Marzieh Fadaee, Julia Kreutzer, and Sara Hooker. 2024.
\newblock \href {https://doi.org/10.18653/v1/2024.acl-long.845} {Aya model: An instruction finetuned open-access multilingual language model}.
\newblock In \emph{Proceedings of the 62nd Annual Meeting of the Association for Computational Linguistics (Volume 1: Long Papers)}, pages 15894--15939, Bangkok, Thailand. Association for Computational Linguistics.

\bibitem[{Vrande{\v{c}}i{\'c} and Kr{\"o}tzsch(2014)}]{vrandevcic2014wikidata}
Denny Vrande{\v{c}}i{\'c} and Markus Kr{\"o}tzsch. 2014.
\newblock Wikidata: a free collaborative knowledgebase.
\newblock \emph{Communications of the ACM}, 57(10):78--85.

\bibitem[{Wan et~al.(2023)Wan, Pu, Sun, Garimella, Chang, and Peng}]{wan-etal-2023-kelly}
Yixin Wan, George Pu, Jiao Sun, Aparna Garimella, Kai-Wei Chang, and Nanyun Peng. 2023.
\newblock \href {https://doi.org/10.18653/v1/2023.findings-emnlp.243} {{\textquotedblleft}kelly is a warm person, joseph is a role model{\textquotedblright}: Gender biases in {LLM}-generated reference letters}.
\newblock In \emph{Findings of the Association for Computational Linguistics: EMNLP 2023}, pages 3730--3748, Singapore. Association for Computational Linguistics.

\bibitem[{Wang et~al.(2025)Wang, Adel, Lange, Liu, Nie, Str{\"o}tgen, and Schuetze}]{wang-etal-2025-lost-multilinguality}
Mingyang Wang, Heike Adel, Lukas Lange, Yihong Liu, Ercong Nie, Jannik Str{\"o}tgen, and Hinrich Schuetze. 2025.
\newblock \href {https://doi.org/10.18653/v1/2025.acl-long.253} {Lost in multilinguality: Dissecting cross-lingual factual inconsistency in transformer language models}.
\newblock In \emph{Proceedings of the 63rd Annual Meeting of the Association for Computational Linguistics (Volume 1: Long Papers)}, pages 5075--5094, Vienna, Austria. Association for Computational Linguistics.

\bibitem[{Wang et~al.(2024)Wang, Jiao, Huang, Dai, Huang, Tu, and Lyu}]{wang-etal-2024-countries}
Wenxuan Wang, Wenxiang Jiao, Jingyuan Huang, Ruyi Dai, Jen-tse Huang, Zhaopeng Tu, and Michael Lyu. 2024.
\newblock \href {https://doi.org/10.18653/v1/2024.acl-long.345} {Not all countries celebrate thanksgiving: On the cultural dominance in large language models}.
\newblock In \emph{Proceedings of the 62nd Annual Meeting of the Association for Computational Linguistics (Volume 1: Long Papers)}, pages 6349--6384, Bangkok, Thailand. Association for Computational Linguistics.

\bibitem[{Yang et~al.(2024)Yang, Yang, Zhang, Hui, Zheng, Yu, Li, Liu, Huang, Wei et~al.}]{yang2024qwen2}
An~Yang, Baosong Yang, Beichen Zhang, Binyuan Hui, Bo~Zheng, Bowen Yu, Chengyuan Li, Dayiheng Liu, Fei Huang, Haoran Wei, and 1 others. 2024.
\newblock Qwen2. 5 technical report.
\newblock \emph{arXiv preprint arXiv:2412.15115}.

\bibitem[{Yao et~al.(2024)Yao, Jiang, Bobinac, Yang, and Hu}]{yao-etal-2024-benchmarking}
Binwei Yao, Ming Jiang, Tara Bobinac, Diyi Yang, and Junjie Hu. 2024.
\newblock \href {https://doi.org/10.18653/v1/2024.findings-emnlp.765} {Benchmarking machine translation with cultural awareness}.
\newblock In \emph{Findings of the Association for Computational Linguistics: EMNLP 2024}, pages 13078--13096, Miami, Florida, USA. Association for Computational Linguistics.

\bibitem[{Y{\"u}ksel et~al.(2024)Y{\"u}ksel, K{\"o}ksal, Senel, Korhonen, and Schuetze}]{yuksel-etal-2024-turkishmmlu}
Arda Y{\"u}ksel, Abdullatif K{\"o}ksal, L{\"u}tfi~Kerem Senel, Anna Korhonen, and Hinrich Schuetze. 2024.
\newblock \href {https://doi.org/10.18653/v1/2024.findings-emnlp.413} {{T}urkish{MMLU}: Measuring massive multitask language understanding in {T}urkish}.
\newblock In \emph{Findings of the Association for Computational Linguistics: EMNLP 2024}, pages 7035--7055, Miami, Florida, USA. Association for Computational Linguistics.

\bibitem[{Zhou et~al.(2025)Zhou, Karidi, Liu, Garneau, Cao, Chen, Li, and Hershcovich}]{zhou-etal-2025-mapo}
Li~Zhou, Taelin Karidi, Wanlong Liu, Nicolas Garneau, Yong Cao, Wenyu Chen, Haizhou Li, and Daniel Hershcovich. 2025.
\newblock \href {https://aclanthology.org/2025.naacl-long.496/} {Does mapo tofu contain coffee? probing {LLM}s for food-related cultural knowledge}.
\newblock In \emph{Proceedings of the 2025 Conference of the Nations of the Americas Chapter of the Association for Computational Linguistics: Human Language Technologies (Volume 1: Long Papers)}, pages 9840--9867, Albuquerque, New Mexico. Association for Computational Linguistics.

\end{thebibliography}
\end{document}